%% file: main.tex
\documentclass{article} 
\usepackage{iclr2027_conference,times}

\input{math_commands.tex}

\usepackage{hyperref}
\usepackage{url}
\usepackage{graphicx}
\usepackage{fontawesome5}
\usepackage{amsmath,amssymb,algorithm,algpseudocode}
\usepackage{booktabs}
\usepackage{float}
\usepackage[table]{xcolor}
\usepackage{colortbl}
\usepackage{xurl}

\definecolor{CMRule}{HTML}{C49A44}
\definecolor{AppInk}{HTML}{000000}
\definecolor{AppBlue}{HTML}{000000}
\definecolor{AppHeadInk}{HTML}{000000}
\definecolor{AppGold}{HTML}{FFFFFF}
\definecolor{AppRule}{HTML}{000000}
\definecolor{AppMuted}{gray}{0.55}
\definecolor{AppAccent}{HTML}{000000}
\newcommand{\AppHead}[1]{{\bfseries #1}}
\newcommand{\AppTableStyle}{%
  \footnotesize\rmfamily\arrayrulecolor{black}%
  \setlength{\tabcolsep}{3pt}\renewcommand{\arraystretch}{1.06}%
  \setlength{\aboverulesep}{0.3ex}\setlength{\belowrulesep}{0.45ex}%
  \setlength{\extrarowheight}{0.4pt}}
\newcommand{\AppAlgorithmStyle}{\footnotesize}

\newtheorem{appproposition}{Proposition}
\newtheorem{applemma}{Lemma}
\newtheorem{appdefinition}{Definition}
\newenvironment{appproof}{\par\noindent\textit{Proof.}\ \ignorespaces}{\hfill$\square$\par\smallskip}

\title{How Language Models \\ Organize Task Information}

\author{Kaiyang Wan \& Yuxia Wang \\
INSAIT, Sofia University “St. Kliment Ohridski”\\
Sofia, Bulgaria \\
\texttt{\{kaiyang.wan,yuxia.wang\}@insait.ai} 
}

\iclrfinalcopy 
\begin{document}

\maketitle

\begin{abstract}
    Dense linear probes assess task readability from language model (LM) representations, but do not directly reveal how many features suffice for prediction.
    We measure readout sparsity with $k^*$, the smallest feature count found by a fixed search protocol that recovers 95\% of dense-probe AUROC above chance.
    We evaluate 19 LMs from five labs on 14 core tasks.
    Ten base models give the nine reliably readable tasks similar feature counts and nearly the same order.
    This agreement extends to the selected features themselves, which on three tasks show cross-model response correspondence beyond class-mean differences and a label-correlation-matched null.
    Instruction-tuned deployment selectively reduces $k^*$, with both weights and input format contributing.
    In controlled synthetic experiments across 20 models, successfully learned targets admit single-feature readouts, whereas readable controls require tens of features.
    Reference-model $k^*$ is associated with label requirements across eight students, following an empirical scaling of $N^*\propto(k^*)^{0.65}$.
    After continual learning, saved heads can fail while linear readability remains high, and recovery requires fewer labels than initial task learning under the tested procedures.
    Pruning reduces linear readability, with wider pre-pruning readouts generally associated with larger losses.
    These findings distinguish task readability from readout sparsity and establish their complementary value for predicting learning requirements and diagnosing LM changes: \faGithub  \href{https://anonymous.4open.science/r/taskprobe_ICLR_anonymous-8368/}{k*}.
\end{abstract}

\input{sections/introduction}

\input{sections/method}

\input{sections/cross_model}

\input{sections/training}

\input{sections/application}

\input{sections/related_work}

\input{sections/conclusion}


\clearpage

\subsection*{AI use statement}

In this work, we used generative AI tools for providing feedback on the design of our experimental pipeline and for implementing methods using coding agents. We have not used generative AI tools for generating synthetic datasets, developing theoretical models or conceptual frameworks, formulating mathematical claims, providing ingredients for or writing proofs, proposing or refining hypotheses, translation, cleaning or reformatting datasets, qualitative or thematic data analysis, or interpreting results. Additionally, we used generative AI tools for creating and editing software code, summarizing existing literature, brainstorming, searching for information, and editing the paper to improve readability. We have reviewed all AI-assisted work. The research questions, hypotheses, and final experimental design were determined by the authors. AI tools were used as a discussion partner, and all suggestions were critically evaluated before adoption. All code produced or modified by coding agents was reviewed and tested by the authors to verify its correctness and that it faithfully implements the described methods. All literature summaries and information obtained via AI tools were verified against the original sources, and all cited references were manually added. AI-suggested text edits were reviewed to ensure they did not alter the technical content. We take responsibility for the final content of this work, including text, claims or artifacts.

\subsection*{Ethics Statement}
This work measures how language models represent task labels, and it involves no human subjects, no new data collection, and no annotation by the authors.
All 24 tasks draw on public labeled datasets under their published licenses (Table~\ref{tab:tasks}); we load each dataset from its public host, redistribute none of them, and reproduce no text from any pool.
Three sources contain harmful, toxic, or hateful language: WildGuardMix~\citep{han2024wildguard}, ToxicChat~\citep{lin2023toxicchat}, and the offensive and hate splits of TweetEval~\citep{barbieri2020tweeteval}; their texts enter only as probe inputs whose published labels the probes predict.
The PII detection task uses a synthetic corpus~\citep{ai4privacy2023pii} whose identifying spans are generated, so no real personal information enters the experiments.
All models are open-weight releases used under their respective licenses, and we release no model weights.

The behavioral interventions of Section~\ref{sec:behavior} include edits that lower refusal rates on instruction-tuned models.
These edits test whether selected features carry the behavior they predict, and they apply two published methods, SAE feature ablation and LEACE~\citep{belrose2023leace}, to open-weight models with full access to their activations.
The reported quantity is the gap between refusal rates on harmful and benign prompts under each edit and its controls, and the released measurements contain aggregate rates and feature indices rather than model completions to harmful prompts.
The edits need white-box access to the residual stream, and they add no capability beyond what these published methods already provide.

Several findings bear on safe deployment.
Harmfulness detection has a wider readout than language identification or intent classification and loses more linear readability under pruning (Section~\ref{sec:pruning}); saved harmfulness heads degrade after later training stages, while dense probes still read the label (Section~\ref{sec:continual}).
These measurements let practitioners audit a modified model for retained safety-relevant readouts before deployment, and the recovery experiments state how many labels restore a degraded head.
The labels for each task come from their source datasets and inherit those annotation conventions, so our measurements describe label readability and make no claim about whether a model's predictions are fair or accurate for any group of users.

Activation extraction runs on single GPUs and probe fitting runs on CPU, which keeps the study's compute footprint small.
The authors declare no competing interests.

\newcommand{\SuppURL}{\url{https://anonymous.4open.science/r/taskprobe_ICLR_anonymous-8368}}

\subsection*{Reproducibility Statement}

Appendix~\ref{app:measurement} specifies the measurement in enough detail to reimplement it, with the count estimator as Algorithms~\ref{alg:search} and~\ref{alg:aggregate} and a worked example that recomputes one reported cell from its probe outputs (Appendix~\ref{app:worked}).
The same appendix gives the construction of every task pool with its source, split, deduplication, and cue-removal steps (Appendix~\ref{app:tasks}), the probed models and layers (Appendix~\ref{app:models}), the SAE feature protocol (Appendix~\ref{app:sae}), the estimator variant used in each experiment (Appendix~\ref{app:protocol-registry}), and the seeds and compute (Appendix~\ref{app:compute}).
Appendix~\ref{app:reproduction} lists the procedure, the randomness, and the estimator behind each application result, and Appendix~\ref{app:interventions} does the same for the behavioral interventions.
Every proposition and lemma states its assumptions and carries a complete proof in the appendix where it appears, and Appendix~\ref{app:finite-sample} reports the measured precision of every ceiling against the threshold gap on which the count depends.
All datasets and model weights are public, and the pool construction is seeded, so the task pools can be rebuilt from their sources.
The supplementary material and the anonymized repository at \SuppURL{} contain code that implements the measurement pipeline, namely task-pool construction, activation extraction, SAE encoding, probe fitting, the count estimator, and its aggregation into the count matrix, together with the configuration of every model and a synthetic test of the probe stack.

\bibliography{iclr2027_conference}
\bibliographystyle{iclr2027_conference}

\appendix

\section{Appendix}
\input{appendix/appendix_guide}

\input{appendix/appendix_methods}

\input{appendix/appendix_cross_model}

\input{appendix/appendix_training}

\input{appendix/appendix_applications}

\end{document}

%% file: math_commands.tex
\usepackage{amsmath,amsfonts,bm}

\def\eqref#1{equation~\ref{#1}}

\def\1{\bm{1}}

\DeclareMathAlphabet{\mathsfit}{\encodingdefault}{\sfdefault}{m}{sl}
\SetMathAlphabet{\mathsfit}{bold}{\encodingdefault}{\sfdefault}{bx}{n}



%% file: sections/introduction.tex
\begin{figure*}[!t]
\centering
\includegraphics[width=14cm]{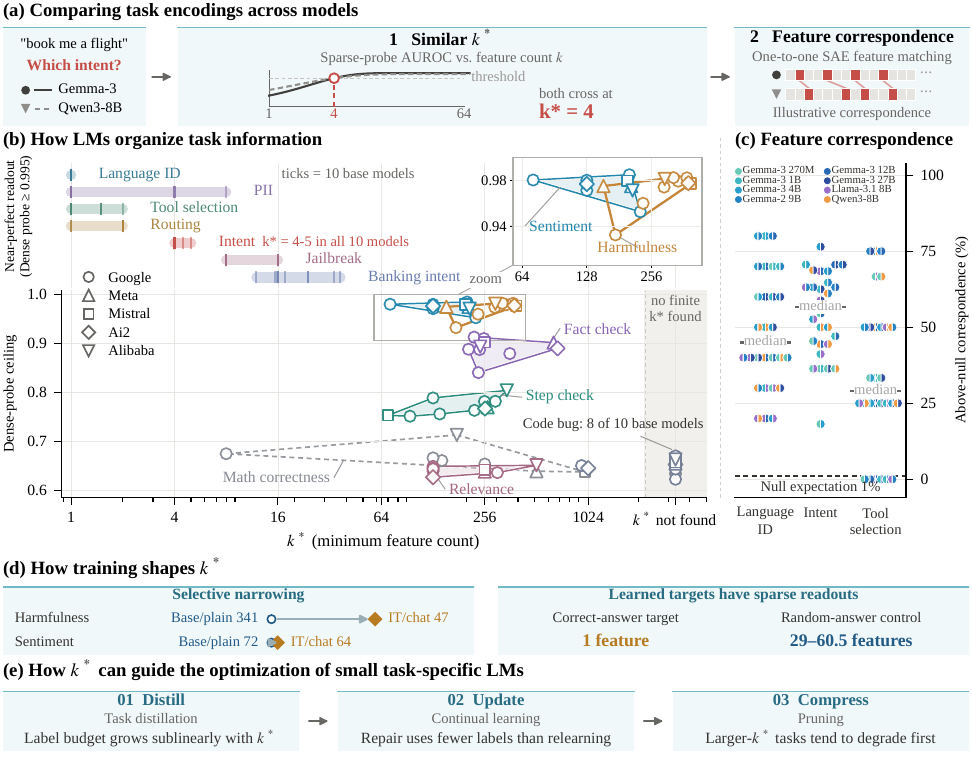}
\vspace{-20pt}
\caption{Task readout measurements reveal cross-model structure, its dependence on training, and implications for task learning and model modification.}
\label{fig:task-organization}
\vspace{-18pt}
\end{figure*}

\section{Introduction}
Language models (LMs) learn representations that support diverse tasks~\citep{devlin2019bert,brown2020language}, from assessing factual claims~\citep{thorne2018fever} to recognizing harmful requests~\citep{han2024wildguard}.
These representations can make task information accessible in different ways, with some tasks requiring only a few features for prediction and others requiring many~\citep{gurnee2023finding}.
Dense linear probes measure how accurately the full representation predicts task labels~\citep{alain2016understanding,belinkov2022probing}, but their performance alone does not distinguish these organizational patterns.
Building on prior work that examines probe accuracy alongside learning complexity~\citep{hewitt2019designing,voita2020information}, we focus on the number and identity of features supporting equally accurate predictions.
We study these differences to understand how LMs organize task information, how training shapes this organization, and what it reveals about subsequent learning and model changes.

To quantify these differences, we fit linear probes to selected feature subsets and compare their performance with a dense probe that uses the full representation.
We define $k^*$ as the smallest subset size our fixed search protocol identifies that recovers 95\% of the dense probe's AUROC above chance (Figure~\ref{fig:task-organization}a).
The dense-probe AUROC and $k^*$ jointly characterize how accurately and how sparsely a representation supports task prediction.
To examine feature correspondence across models, we additionally compare the responses of task-relevant sparse autoencoder (SAE) features~\citep{bricken2023monosemanticity,huben2024sparse} on shared inputs.
These measurements provide a common framework for characterizing cross-model readout structure, examining its dependence on training, and evaluating its predictive and diagnostic value for subsequent model changes.

Using this framework, we evaluate 19 LMs from five labs on 14 core tasks.
Across ten base models, we find that task readouts vary substantially in feature count across tasks but require similar counts within each task.
These readouts require only four to five residual coordinates for intent classification and yield task rankings with a median cross-lab Spearman correlation of 0.93 across nine tasks with consistently high dense-probe performance (Figure~\ref{fig:task-organization}b).
Their selected SAE features also show response correspondence across models on three tasks, exceeding that of unselected features with comparable label correlations after removing class-mean differences (Figure~\ref{fig:task-organization}c).
Selected language features further contribute to Gemma-2-9B-IT's adaptation of reply language to input language, as their ablation disrupts this behavior while matched control edits largely preserve it.
Together, these findings reveal shared structure in task readouts, both in the number of features that support prediction and in their responses to individual inputs.

This shared readout structure motivates examining how training shapes access to task information.
We compare nine base models with their instruction-tuned counterparts and find task-selective reductions in $k^*$.
By independently varying weights and input format, we find that both affect readout sparsity, with task-dependent interactions.
To isolate the contribution of target supervision, we train 20 controlled models on binary factors, pairing target aggregates with correct answers and control aggregates with independent random answers.
Successfully learned targets admit single-feature readouts, while readable controls require tens of features, even when a control shares half its factors with a trained target (Figure~\ref{fig:task-organization}d).
These results show that training can selectively concentrate access to supervised targets within representations that also support broader readouts of related information.

The dependence of readout structure on training motivates evaluating its predictive and diagnostic value for subsequent learning and model modification (Figure~\ref{fig:task-organization}e).
Across eight student models, reference-model $k^*$ predicts label requirements for supervised adaptation, following an empirical relation $N^*\propto(k^*)^{0.65}$.
After continual learning, dense probes retain near-baseline performance despite failures of earlier classification heads, and recovery requires fewer labels than initial learning under the tested procedures.
Under pruning, dense-probe performance declines, with wider pre-pruning readouts generally associated with greater losses.
Yet allocating labels by feature count and protecting parameters associated with selected readout features do not consistently outperform the tested baselines.
These results establish the value of readout measurements for predicting supervision requirements and distinguishing failures of classification heads from losses of linear readability.

Our contributions are as follows:
\begin{enumerate}
    \item In Section~\ref{sec:measurement}, we introduce $k^*$ and specify a protocol for measuring linear readability and readout sparsity, with explicit procedures for feature search, confirmation, and aggregation.
    \item In Section~\ref{sec:agreement}, we identify similar feature counts and consistent task rankings across base models, together with cross-model response correspondence among selected SAE features.
    \item In Section~\ref{sec:training}, we show that weights and input format shape readout sparsity, and that successful learning promotes single-feature readouts of targets in controlled experiments.
    \item In Section~\ref{sec:applications}, we demonstrate the predictive and diagnostic value of readout measurements for supervision requirements, forgetting, and pruning, while identifying limitations of the tested readout-based interventions.
\end{enumerate}

%% file: sections/method.tex
\section{Task Encoding Sparsity Measurement}
\label{sec:measurement}

In this work, we measure the linear readability of task information and the number of representation coordinates needed to recover that readability.

\textbf{Readout sparsity.}
We formulate each task as one or more binary classification problems, using one-versus-rest labels for multiclass tasks, and treat each coordinate of a frozen LM representation as a feature.
For a fixed class decision, layer, and probe seed, let $A(k)$ denote the highest test AUROC among candidate probes using $k$ features.
Let $C>0.5$ denote the dense-probe reference specified below.
We define normalized performance $s(k)$ and the smallest qualifying feature count $k^*$ as
\begin{equation}
    s(k)=\frac{A(k)-0.5}{C-0.5},
    \qquad
    k^*=\min\{k:s(k)\geq 0.95\},
    \label{eq:kstar}
\end{equation}
where the minimum ranges over feature counts evaluated by our search.
We set $k^*=\infty$ if no candidate meets the threshold and aggregate these local counts as described below.

\textbf{Interpretation.}
The dense-probe ceiling $C$ measures linear readability, while $k^*$ measures how many features suffice to recover 95\% of its performance above chance under the specified search protocol.
We interpret them jointly and treat a task--model pair as reliably readable when its task-level readability ceiling is above 0.90.
A small count at a low ceiling need not support accurate prediction, and the ceiling serves as a reference for linear readout of the current LM representation.
Counts depend on the representation basis and evaluation protocol; they do not specify intrinsic task complexity or a unique set of necessary features.

\textbf{Searching for sparse readouts.}
For the main count comparisons, we average residual activations over input tokens.
A preliminary single-seed sweep across the model's depth identifies a task-specific window for repeated measurement, typically comprising four consecutive entries in the probed layer grid (Appendix~\ref{app:aggregation}).
Within this window, we construct candidate probes using two search procedures.
One ranks features by their absolute two-sample $t$-scores on the training data and fits L2-regularized logistic probes to the top-$k$ subsets~\citep{gurnee2023finding}.
The other fits L1-regularized logistic probes along a regularization path, with nonzero coefficients defining each candidate support.
At each layer and seed, the local $k^*$ is the smallest feature count meeting the performance threshold across both procedures (Appendix~\ref{app:measurement}).

\textbf{Confirmation and aggregation.}
For each of the five seeds, we draw a stratified subsample containing 85\% of the training data and repeat the measurement on the same test set.
For each class decision, we set $C$ to the highest dense-probe test AUROC across the measured layers and seeds, providing a shared reference for all local counts.
To avoid relying on a single layer, we take the larger local count in each adjacent pair of probed layers, then the minimum across pairs, separately for each decision and seed.
We obtain the task-level count by taking the median across class decisions within each seed, followed by the median across seeds.
The task-level dense-probe reference is the median of the class-decision references.
This aggregation summarizes feature counts across potentially different supports at different layers, decisions, and seeds (Appendix~\ref{app:aggregation}).

\textbf{Models and tasks.}
We evaluate 14 tasks on 19 LMs from five labs, comprising ten base models and nine instruction-tuned counterparts.
The models range from 270M to 27B parameters and span the Gemma~\citep{team2024gemma,team2025gemma}, Llama~\citep{grattafiori2024llama}, Mistral~\citep{jiang2023mistral7b}, OLMo~\citep{olmo20242}, and Qwen~\citep{yang2025qwen3} families.
Five dispatch tasks cover language identification~\citep{keung2020multilingual,conneau2018xnli,cer2017semeval}, tool selection~\citep{glaiveai2023glaive}, routing~\citep{cobbe2021training,hendrycks2021measuring}, and intent classification~\citep{larson2019evaluation,casanueva2020efficient}.
The remaining tasks comprise three safety judgments~\citep{ai4privacy2023pii,han2024wildguard}, three text judgments~\citep{socher2013recursive,bajaj2016ms,thorne2018fever}, and three correctness checks on code and mathematics~\citep{zhou2019devign,zheng2025processbench}.
Ten additional tasks on Gemma-2-9B support the label-budget and input-baseline analyses.
We provide more details in Appendix~\ref{app:measurement}.

\textbf{Experiment-specific readout protocols.}
To compare feature responses across models, we select task-relevant SAE features at the last input token in the eight base models with available SAEs.
We select these features using the preliminary sweep results, following the protocol in Appendix~\ref{app:sae}.
For the controlled synthetic experiments, we probe representations at the query position using the protocol specified in Section~\ref{sec:toy} and Appendix~\ref{app:toy-results}.
In each analysis, we compare feature counts under a common representation and measurement protocol.

%% file: sections/cross_model.tex
\section{Shared Structure in Task Readouts Across Models}

\label{sec:agreement}
Models differ in scale and training, yet may organize task information into readouts with common properties.
In this section, we first compare feature counts identified by the fixed search protocol, then examine whether selected SAE features respond similarly to shared inputs.
We further test the behavioral relevance of selected features through targeted interventions.

\begin{figure*}[!h]
    \centering
    \vspace{-1mm}
    \includegraphics[width=\linewidth]{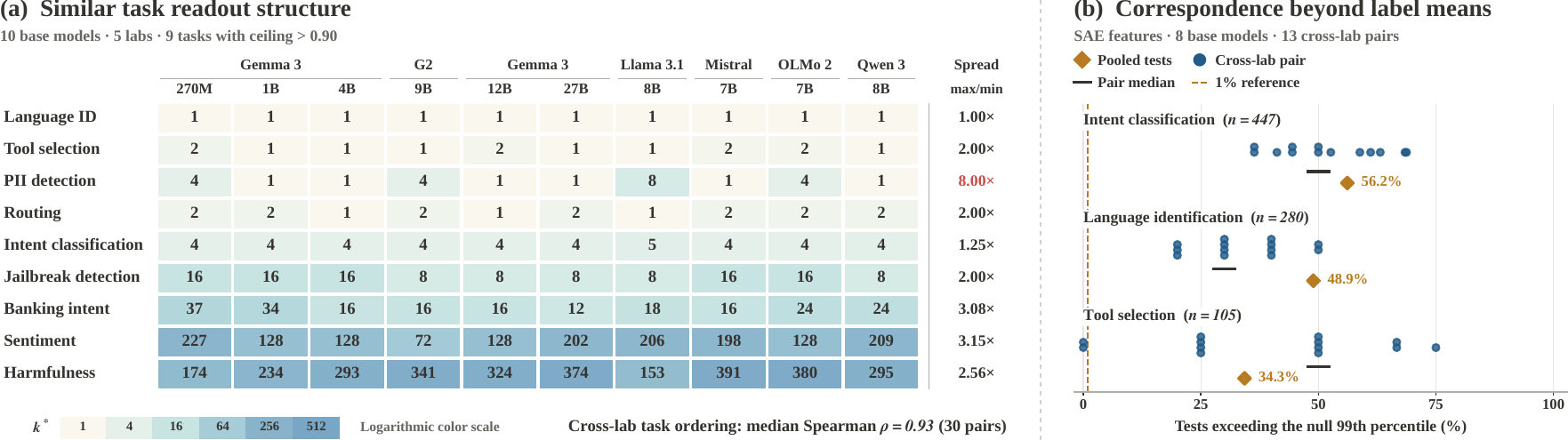}
    \vspace{-5mm}
    \caption{(a) Task readout counts and rankings are similar across base models, and (b) selected SAE features exhibit cross-model response correspondence beyond class-mean differences.}
    \label{fig:cross-model-story}
\end{figure*}

\subsection{Task Readout Structure Is Similar Across Models}
\label{sec:counts}

\textbf{Motivation and experimental setup.}
To determine whether models share task readout structure despite differences in scale and training, we compare residual-coordinate counts $k^*$ across ten base models under the fixed protocol of Section~\ref{sec:measurement}.
We retain the nine tasks that are reliably readable in every model, with task-level dense-probe AUROC above 0.90.
We measure each task's cross-model spread as its maximum count divided by its minimum, and compare task rankings using Spearman correlations over 30 cross-lab model pairs.
Both summaries use unrounded counts, and Appendix~\ref{app:ordering} provides aggregation details and the full 14-task comparisons.

\textbf{Similar feature counts and consistent task rankings across LMs.}
Feature counts vary substantially across tasks but remain comparatively similar across LMs within each task (Figure~\ref{fig:cross-model-story}a).
Language identification requires one feature in every LM, intent classification requires four to five, and harmfulness detection requires 153--391.
Eight of the nine tasks have a cross-model spread below 3.2. PII detection is the exception, ranging from one to eight features.
The LMs also rank the nine tasks similarly, with a median Spearman correlation of 0.93 across cross-lab pairs.
This agreement remains at 0.86--0.92 after controlling for how sparsely a bag-of-words probe reads each task from the raw text (Appendix~\ref{app:ordering}).
Together, these comparisons reveal a common ordering of tasks by readout sparsity across the tested base LMs.

\textbf{Task rankings are robust to estimation choices.}
Varying the performance threshold, feature-search path, or layer-selection rule yields task rankings closely aligned with the main protocol, with Spearman correlations of 0.97--0.99.
By contrast, changing the representation basis or input conditions can shift absolute feature counts (Appendix~\ref{app:protocol-controls}).
These results support a stable ordering across the tested estimation procedures while showing that the number of features supporting a readout depends on how task information is represented and presented.

\subsection{Feature Correspondence and Behavioral Relevance}

\label{sec:correspondence}

\label{sec:behavior}

\textbf{Motivation and experimental setup.}
Similar feature counts leave open whether selected features exhibit corresponding input responses, contribute to behavior, or provide the only available readout.
We compare selected SAE features at the last input token on shared inputs across 28 pairs of eight base models, including 13 cross-lab pairs.
For each model pair and class decision, we subtract class-specific mean activations and match the centered responses one to one by absolute correlation.
We compare each score with a label-correlation-matched null constructed from unselected features (Appendix~\ref{app:matching}).
To assess behavioral relevance~\citep{elazar2021amnesic}, we ablate selected language SAE features during generation, using unselected features with similar activation frequencies and random-direction edits of equal norm as controls.
In a separate harmfulness experiment, we compare readouts from highly ranked residual coordinates with refitted dense probes after fixing those coordinates to their training means (Appendix~\ref{app:interventions}).
 
\textbf{Selected features exhibit correspondence beyond class-mean differences.}
Across the three tasks with sufficient SAE coverage, matching scores exceed the null's 99th percentile in 56.2\% of tests for intent classification, 48.9\% for language identification, and 34.3\% for tool selection (Figure~\ref{fig:cross-model-story}b).
For intent classification, the median fraction of class decisions exceeding this threshold is 50\% across the 13 cross-lab pairs.
Binary task labels do not specify how feature responses should vary among inputs within each class, yet the selected features exhibit partial correspondence in this variation across models.
This finding identifies a shared aspect of task readout structure that feature counts and label correlations alone do not characterize.
 
\textbf{Language-identification features contribute to reply-language adaptation.}
On Gemma-2-9B-IT, ablating one to four selected SAE features for Japanese, Greek, or Hindi reduces the gap in target-language reply probability between target-language and other-language inputs from 0.88--0.99 to at most 0.015.
Ablations of unselected features with similar activation frequencies and random-direction edits of equal norm largely preserve this gap (Appendix~\ref{app:interventions}).
This contrast supports a behavioral role for the features under the interventions.
Features selected for identifying input language thus also contribute to the model's adaptation of reply language to that input.
 
\textbf{Selected coordinates support sufficient but nonexclusive readouts.}
In a separate harmfulness experiment, probes using only the 256 highest-ranked residual coordinates achieve AUROCs of 0.963--0.982 across three instruction-tuned models.
Fixing these coordinates to their training means leaves refitted dense-probe AUROC within 0.0011 of the full-representation baseline (Appendix~\ref{app:interventions}).
Thus, both the selected coordinates and their complements support accurate prediction.
This coexistence shows why identifying a small predictive subset does not determine which coordinates to remove to eliminate linear readability.

%% file: sections/training.tex
\section{Training Shapes Task Readout Sparsity}
\label{sec:training}

The shared readout structure across base models motivates examining how training changes readout sparsity.
In this section, we first compare base models with their instruction-tuned (IT) counterparts, distinguishing the contributions of model weights and input format.
We then use controlled synthetic experiments to test whether supervision for a target promotes a sparse readout of that target.

\begin{figure*}[!ht]
    \centering
    \includegraphics[width=\linewidth]{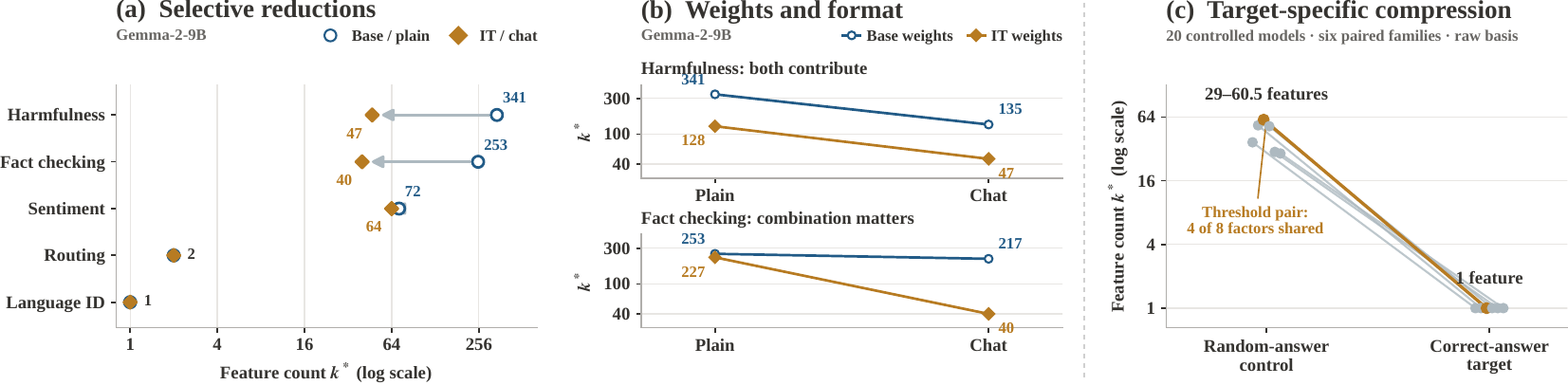}
    \vspace{-5mm}
    \caption{Instruction-tuned deployment reduces feature counts unevenly across tasks through weight and format effects, while supervised targets admit single-feature readouts in controlled experiments.}
    \vspace{-5mm}
    \label{fig:training-story}
\end{figure*}

\subsection{Instruction-Tuned Readouts Depend on Weights and Input Format}

\textbf{Motivation and experimental setup.}
Instruction tuning uses instruction--response supervision to shape how models answer user requests.
This motivates testing whether readout sparsity changes on tasks relevant to safe and accurate responses, including harmfulness detection and fact checking.
We compare nine base models with their IT counterparts under their respective deployment formats, retaining tasks that are reliably readable in both settings and have finite feature counts (Appendix~\ref{app:training-pairs}).
This comparison combines weight changes and input format changes.
To distinguish their contributions and test their interaction, we cross base and IT weights with plain-text and chat-formatted inputs on Gemma-2-9B (Appendix~\ref{app:factorial}).

\textbf{Deployment reductions are task-selective.}
Instruction-tuned deployment reduces readout counts unevenly across tasks (Figure~\ref{fig:training-story}a).
On Gemma-2-9B, harmfulness detection falls from $k^*=341$ to $47$ and fact checking from $253$ to $40$, while sentiment changes from $72$ to $64$.
Already sparse readouts, including language identification, tool selection, and routing, show little change.
Harmfulness reductions also vary across models, ranging from approximately $1.1$-fold on Gemma-3-270M to $8.5$-fold on Gemma-3-12B (Appendix~\ref{app:training-pairs}).
The deployment change therefore reshapes the sparsity profile across tasks, with its magnitude depending on the model.

\textbf{Similar deployment reductions reflect different weight--format interactions.}
Harmfulness detection and fact checking both show approximately six- to seven-fold deployment reductions, but differ in their dependence on weights and input format (Figure~\ref{fig:training-story}b).
For harmfulness detection, changing only the format reduces $k^*$ from $341$ to $135$, changing only the weights reduces it to $128$, and changing both reduces it to $47$.
Each factor produces a similar reduction under either setting of the other, consistent with approximately separable effects on the multiplicative scale.
For fact checking, either change alone yields a modest reduction from $253$ to $217$ or $227$, respectively, whereas their combination lowers the count to $40$.
Restricting all four conditions to shared probed layers preserves these medians (Appendix~\ref{app:factorial}).
These results show that the effect of weight updates on readout sparsity can depend strongly on the input format used to access task information.

\subsection{Target Supervision Promotes Sparse Readouts}
\label{sec:toy}

\begin{figure}[!ht]
    \centering
    \includegraphics[width=\linewidth]{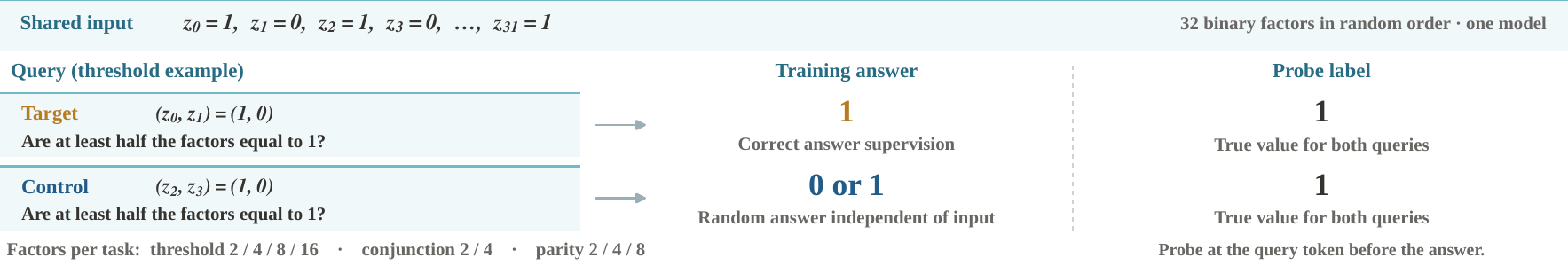}
    \vspace{-5mm}
    \caption{Target queries receive correct training answers and control queries receive independent random answers, while probes predict the true aggregate values for both.}
    \label{fig:toy-design}
\end{figure}

\textbf{Motivation and experimental setup.}
To examine whether target supervision promotes sparse readouts, we construct tasks whose answers are fully determined by the input but differ in whether training rewards their prediction.
We train 18 transformers and two GRUs on binary-factor sequences using the target--control supervision in Figure~\ref{fig:toy-design}, which illustrates the tokenized inputs and queries in readable notation.
Within each function family, target and control aggregates have matching true-label distributions.
An overlapping threshold pair tests whether learning a related aggregate suffices for sparse readout formation.
We measure readouts of the true aggregate values for both targets and controls on fresh sequences, extracting representations at the query position before the answer appears.
Appendices~\ref{app:toy-design} and~\ref{app:toy-results} detail the training and probe protocols.

\textbf{Sparse readouts are specific to supervised targets.}
Across all 20 models, every reliably readable supervised aggregate admits a one-feature readout, whereas readable controls require tens of features, with family-level medians of $29$--$60.5$ in the original basis (Figure~\ref{fig:training-story}c).
The controls remain linearly readable within the same models, demonstrating that linear readability alone does not determine readout sparsity.
Even the threshold control, which shares four of its eight factors with a supervised target, requires a median of $60.5$ features, showing that learning a related aggregate does not suffice to produce an equally sparse readout.
This target--control separation persists in the tested PCA basis (Appendix~\ref{app:toy-results}).
Four- and eight-factor parity targets fail the readability criterion, limiting the single-feature result to targets that acquire reliably readable representations.
Together, these findings support a target-specific role for supervision in organizing task information into sparse readouts, beyond the availability of that information or partial overlap in its input dependencies.

\textbf{Single-feature readouts summarize multiple input dependencies.}
Supervised threshold targets over $2$, $4$, $8$, and $16$ input factors all admit one-feature readouts under the tested protocol (Appendix~\ref{app:toy-results}).
Thus, increasing the number of factors that determine the target need not increase the number of coordinates required to read it out after training.
A single coordinate can summarize predictive evidence from multiple factors, while the computation that produces this coordinate may involve many features.
This distinction places $k^*$ at the level of access to a learned representation as it measures the coordinates needed for prediction under a specified basis and probe protocol.
The observed sparsity therefore concerns the organization of target information and does not reduce the information retained by the full representation.

%% file: sections/application.tex
\section{Implications for Learning and Maintaining Task-Specific LMs}
\label{sec:applications}

Building on the training-induced changes in readout structure, we explore the predictive and diagnostic value of readout measurements for learning and model modification.
We examine whether feature counts anticipate label requirements, whether linear readability persists after continual learning disrupts earlier classification heads, and whether pre-pruning counts anticipate its loss.

\subsection{Feature Counts Anticipate Label Requirements}
\label{sec:distillation}

\textbf{Motivation and experimental setup.}
To test whether readout structure in one model predicts supervision requirements in others, we measure $k^*$ in a frozen Gemma-2-9B reference model and evaluate label requirements across eight students from four model families.
We adapt each student at multiple label budgets using LoRA~\citep{hu2021lora}, a classification head, and ground-truth labels.
For each student--task pair, we define $N^*$ as the label count needed to reach 95\% of the AUROC gain above chance attained at the largest tested budget.
We fit a power-law relation between reference-model $k^*$ and student $N^*$ with a shared exponent and student-specific intercepts, and assess prediction by leaving out each task in turn (Appendices~\ref{app:student-training} and~\ref{app:label-inference}).

\begin{figure*}[!ht]
    \centering
    \includegraphics[width=\linewidth]{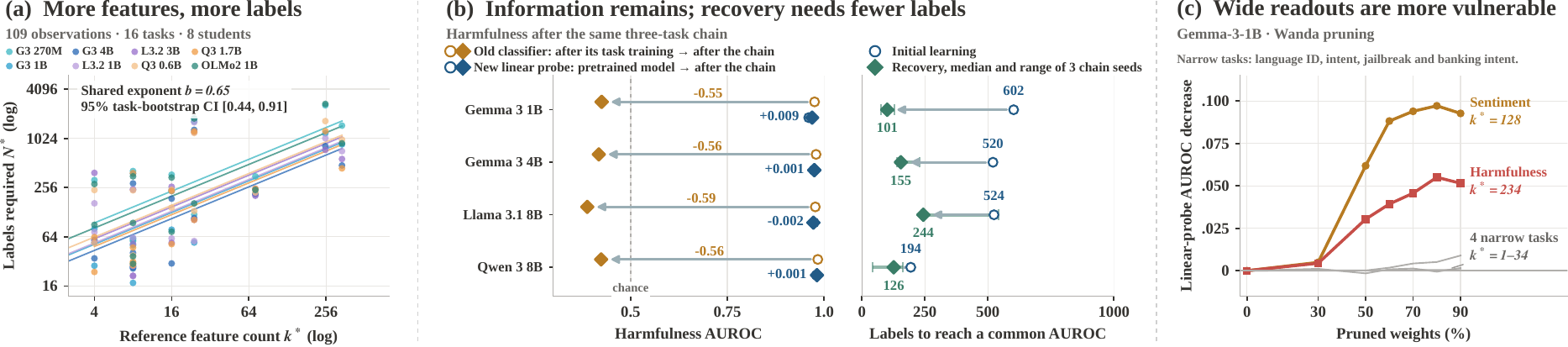}
    \vspace{-5mm}
    \caption{Feature counts anticipate label requirements and pruning vulnerability, while dense probes reveal retained linear readability after saved-head failure.}
    \vspace{-5mm}
    \label{fig:applications-story}
\end{figure*}

\textbf{Reference-model readouts anticipate label requirements across students.}
Tasks with wider readouts in the reference model generally require more labeled examples for student adaptation.
Across 109 uncensored student--task observations spanning eight students and 16 tasks reliably readable in the reference model, the fitted relation is $N^*\propto(k^*)^{0.65}$, with a task-bootstrap 95\% interval of $[0.44,0.91]$ for the exponent (Figure~\ref{fig:applications-story}a).
Leaving out each task in turn yields a 90th-percentile multiplicative prediction error of $5.3$, indicating substantial uncertainty in individual budget estimates.
The association thus links readout sparsity in a frozen reference representation to supervision requirements for parameter adaptation in other models.
This link remains after controlling for transferability scores or for how well a bag-of-words probe reads the raw text, as Appendix~\ref{app:learning-controls} reports.
It supports a predictive role for readout structure across the tested students and raises the question of why sparser readouts cost fewer labels to learn.

\textbf{Allocation by predicted label requirements does not consistently save labels.}
To test whether the predictive relationship improves budget allocation, we replay learning curves for nine tasks and allocate labels in proportion to predicted $N^*$.
Reaching the same joint utility targets requires a median of $1.03$ times the uniform-allocation budget (Appendix~\ref{app:allocation}).
A rule based on a 16-label pilot requires $0.66$ times that budget under the replay accounting, which excludes separate pilot costs.
The tested proportional rule therefore yields no consistent savings despite the association between feature counts and label requirements.
This outcome highlights a distinction between predicting where a learning curve reaches a performance criterion and determining where additional labels most improve a joint objective.

\subsection{Linear Readability Can Persist after Behavioral Forgetting}
\label{sec:continual}

\textbf{Motivation and experimental setup.}
An earlier task's classification head may fail after subsequent training even if the updated representation retains linearly readable task information.
To examine this possibility, we sequentially fine-tune four models on classification tasks and compare saved-head performance with newly fitted dense probes on the same frozen backbones.
To assess the supervision needed to relearn an earlier task, we compare recovery using LoRA and a new classification head with initial learning from the pretrained model.
We measure label requirements in both settings on a shared grid of 16 to 1024 labels, against the AUROC at which initial learning attains 95\% of its gain above chance at 1024 labels (Appendices~\ref{app:student-training} and~\ref{app:continual}).

\textbf{Saved-head failure coexists with retained linear readability.}
After the harmfulness--sentiment--jailbreak sequence, saved harmfulness heads lose a median of $0.55$--$0.59$ AUROC across the four models relative to their performance immediately after harmfulness training.
Yet median dense-probe ceilings remain within $0.009$ of their pretrained baselines (Figure~\ref{fig:applications-story}b, left).
The updated representations thus support accurate linear readouts even when the previously trained heads no longer predict reliably.
These results distinguish the persistence of an existing readout from the continued availability of linearly readable task information, motivating separate evaluation of both after sequential training (Appendix~\ref{app:continual}).

\textbf{Relearning requires fewer labels than initial adaptation.}
Across all four models, harmfulness recovery requires fewer labels than initial learning, with medians of $101$--$244$ and $194$--$602$ labels, respectively (Figure~\ref{fig:applications-story}b, right).
Against each curve's own largest-budget performance instead, the advantage holds in three models and nearly vanishes on Qwen3-8B, and the two procedures also differ in input-length limits (Appendix~\ref{app:continual}).
The recovery advantage therefore holds under the tested procedures.
Together with the probe results, this finding shows that substantial saved-head failure can coexist with retained linear readability and reduced supervision requirements for relearning.

\subsection{Readout Sparsity Anticipates Pruning Vulnerability}
\label{sec:pruning}

\textbf{Motivation and experimental setup.}
The retained linear readability after sequential training raises the question of how task readouts withstand weight pruning.
We apply magnitude pruning and Wanda~\citep{sun2024simple} to four models from three labs across six tasks and test whether pre-pruning $k^*$ anticipates losses of linear readability.
To examine whether selected readout coordinates can also guide task preservation, we protect their associated weight rows and columns during pruning.
We compare this scheme with random-coordinate protection and ordinary pruning at matched actual sparsity.
We provide more details in Appendix~\ref{app:pruning} and Appendix~\ref{app:pruning-recovery}.

\textbf{Wider readouts are associated with greater pruning losses.}
Language identification and intent classification retain high dense-probe performance under pruning, whereas sentiment and harmfulness detection, which have wider readouts, incur larger ceiling losses (Figure~\ref{fig:applications-story}c).
At sparsities of at least 60\%, pre-pruning $k^*$ correlates with ceiling loss at approximately $0.89$ or higher in 30 of 32 model--method--sparsity conditions, with adjusted permutation $q<0.05$ (Appendix~\ref{app:pruning}).
The association broadly distinguishes stable from vulnerable tasks, but sentiment loses more than harmfulness detection in all four models, despite requiring fewer features in three.
Readout sparsity thus indicates pruning vulnerability but does not reliably order losses among the wider-readout tasks.
The reduced ceilings also distinguish these pruning outcomes from the retained linear readability observed after sequential training.

\textbf{Readout-guided parameter protection does not consistently preserve task performance.}
On Gemma-2-9B, protecting weight rows and columns associated with selected residual coordinates does not consistently outperform protection based on random coordinates.
At matched actual sparsity, it performs worse than ordinary Wanda pruning (Appendix~\ref{app:pruning-recovery}).
Preserving these weights need not preserve the selected activations, which also depend on computations elsewhere in the model.
The result illustrates a gap between identifying coordinates sufficient for task readout and specifying parameter constraints sufficient for task preservation.

Readout measurements anticipate supervision requirements, reveal retained linear readability after saved-head failure, and identify broad differences in pruning vulnerability.
Their predictive and diagnostic value does not automatically translate into effective intervention, as the tested allocation and parameter-protection schemes yield no consistent advantage over their baselines.

%% file: sections/related_work.tex
\section{Related Work}
\label{sec:related}

\textbf{Probing and cross-model readout structure.}
Sparse probing studies how many neurons support prediction~\citep{gurnee2023finding}, while minimum-description-length probing measures the coding cost of labels given a representation~\citep{voita2020information}.
Building on sparse probing, we compare the feature counts needed to recover a prescribed fraction of dense-probe performance across tasks and models.
Convergent learning examines correspondence between independently trained networks~\citep{li2015convergent}, and SAE-based analyses extend these comparisons to LM feature spaces~\citep{lan2024quantifying}.
Our correspondence analysis focuses on task-selected features and tests whether their responses agree within classes beyond a label-correlation-matched null.

\textbf{Training and behavioral relevance.}
Feature-drift analysis studies how fine-tuning repurposes LM features~\citep{galichin2026feature}.
We examine how these changes affect task readout sparsity, separate weight and input-format effects, and test target supervision in controlled experiments.
Amnesic probing distinguishes readable information from its contribution to behavior~\citep{elazar2021amnesic}, while concept-erasure methods seek to remove linear access to specified concepts~\citep{belrose2023leace}.
Our interventions examine both the behavioral contribution of task-selected language features and the persistence of harmfulness readouts outside selected coordinates, separating predictive sufficiency from behavioral contribution and readout necessity.

\textbf{Learning requirements and representation forgetting.}
LogME and H-score assess the transferability of pretrained representations~\citep{you2021logme,bao2019information}.
We investigate whether task readout sparsity in one reference model predicts the labels required for adaptation across student models.
Prior work uses refitted linear probes to distinguish representation forgetting from task-performance degradation~\citep{davari2022probing}.
We build on this distinction by jointly evaluating saved heads, linear readability, and relearning requirements across LM families, and contrast retained readability after sequential training with its decline under pruning.

%% file: sections/conclusion.tex
\section{Conclusion and Future Work}
\label{sec:discussion}

In this work, we studied how LMs organize task information by jointly measuring linear readability and readout sparsity.
Under a fixed protocol, base models exhibit similar task-level feature counts and partial correspondence among selected SAE features.
Training and input format reshape this structure, while controlled experiments show that successfully learned targets can admit single-feature readouts alongside wider readouts of related controls.
Readout measurements also anticipate supervision requirements and broad pruning vulnerability, while dense probes distinguish saved-head failure from losses of linear readability.
Together, these findings connect the organization of task readouts with how models learn and change.

Beyond feature counts, future work could characterize readout directions and activation magnitudes along them under controlled representation scaling.
Tracking these properties during training could distinguish changes in readout support, orientation, and the strength of task-relevant activations.
Combining these measurements with causal interventions would test when sparse readouts provide effective directions for activation steering and whether alternative readouts predict recovery after model modification.
These studies would connect the information accessible to a probe with the computations that govern model behavior.

%% file: appendix/appendix_guide.tex
\label{app:guide}
\begingroup
\makeatletter
\c@tocdepth=2\relax
\newcommand{\AppGuideGroup}[3]{%
  \par\addvspace{1em}%
  \@dottedtocline{1}{0em}{2.8em}%
    {\bfseries\numberline{#1}#2}{\bfseries#3}%
  \nopagebreak[4]}
\newcommand{\AppGuideEntry}[3]{%
  \@dottedtocline{2}{1em}{3.6em}%
    {\numberline{#1}#2}{#3}}
\makeatother

\noindent{\large\bfseries Appendix contents}\par
\smallskip
\noindent Measurement definitions, experimental details, and supporting evidence.\par
\medskip
\noindent\rule{\linewidth}{0.4pt}\par
\small
\noindent\hfill\textit{Page}\par

\AppGuideGroup{\ref{app:measurement}}{Measurement protocol}{\pageref{app:measurement}}
\AppGuideEntry{\ref{app:notation}}{Notation and Formal Definition}{\pageref{app:notation}}
\AppGuideEntry{\ref{app:inputs}}{Input Representations}{\pageref{app:inputs}}
\AppGuideEntry{\ref{app:tasks}}{Task Pools, Splits, and Class Decisions}{\pageref{app:tasks}}
\AppGuideEntry{\ref{app:probes}}{Probe Fitting}{\pageref{app:probes}}
\AppGuideEntry{\ref{app:search}}{Support Search at One Layer}{\pageref{app:search}}
\AppGuideEntry{\ref{app:aggregation}}{Layer windows, ceilings, and aggregation}{\pageref{app:aggregation}}
\AppGuideEntry{\ref{app:worked}}{Worked Example on One Cell}{\pageref{app:worked}}
\AppGuideEntry{\ref{app:finite-sample}}{Finite-Sample Behavior of the Count}{\pageref{app:finite-sample}}
\AppGuideEntry{\ref{app:models}}{Models and Probed Layers}{\pageref{app:models}}
\AppGuideEntry{\ref{app:sae}}{SAE Feature Protocol}{\pageref{app:sae}}
\AppGuideEntry{\ref{app:design}}{Design Choices and Their Measured Effects}{\pageref{app:design}}
\AppGuideEntry{\ref{app:selection}}{Selection Effects and Statistical Units}{\pageref{app:selection}}
\AppGuideEntry{\ref{app:protocol-registry}}{Estimator Variants Across Experiments}{\pageref{app:protocol-registry}}
\AppGuideEntry{\ref{app:compute}}{Software and Compute}{\pageref{app:compute}}
\AppGuideEntry{\ref{app:scope}}{Scope of the Measurement}{\pageref{app:scope}}
\AppGuideEntry{\ref{app:recommendations}}{Recommendations for measuring sparsity}{\pageref{app:recommendations}}

\AppGuideGroup{\ref{app:agreement}}{Cross-model agreement}{\pageref{app:agreement}}
\AppGuideEntry{\ref{app:ordering}}{Task Ordering Across Models}{\pageref{app:ordering}}
\AppGuideEntry{\ref{app:protocol-controls}}{Protocol and Estimator Controls}{\pageref{app:protocol-controls}}
\AppGuideEntry{\ref{app:inputbase}}{Bag-of-Words Input Baseline}{\pageref{app:inputbase}}
\AppGuideEntry{\ref{app:matching}}{Cross-Model Feature Matching}{\pageref{app:matching}}
\AppGuideEntry{\ref{app:dictionary}}{Leave-One-Out Dictionary Test}{\pageref{app:dictionary}}
\AppGuideEntry{\ref{app:interventions}}{Behavioral Interventions}{\pageref{app:interventions}}
\AppGuideEntry{\ref{app:agreement-summary}}{Findings and open questions}{\pageref{app:agreement-summary}}

\AppGuideGroup{\ref{app:training}}{Training experiments}{\pageref{app:training}}
\AppGuideEntry{\ref{app:training-pairs}}{Paired Deployment Comparisons}{\pageref{app:training-pairs}}
\AppGuideEntry{\ref{app:factorial}}{Weights and Format as a Factorial Design}{\pageref{app:factorial}}
\AppGuideEntry{\ref{app:toy-design}}{Synthetic World and Training Objective}{\pageref{app:toy-design}}
\AppGuideEntry{\ref{app:toy-results}}{Synthetic Readout, Coverage, and Controls}{\pageref{app:toy-results}}
\AppGuideEntry{\ref{app:training-summary}}{Findings and open questions}{\pageref{app:training-summary}}

\AppGuideGroup{\ref{app:applications}}{Learning and maintenance}{\pageref{app:applications}}
\AppGuideEntry{\ref{app:app-figure}}{Construction of the application figure}{\pageref{app:app-figure}}
\AppGuideEntry{\ref{app:student-training}}{Student Adaptation and Label Sources}{\pageref{app:student-training}}
\AppGuideEntry{\ref{app:label-count}}{Label requirements and censoring}{\pageref{app:label-count}}
\AppGuideEntry{\ref{app:label-inference}}{Uncertainty and held-out prediction}{\pageref{app:label-inference}}
\AppGuideEntry{\ref{app:learning-controls}}{Predictive value and teacher controls}{\pageref{app:learning-controls}}
\AppGuideEntry{\ref{app:allocation}}{Multi-Task Budget Replay}{\pageref{app:allocation}}
\AppGuideEntry{\ref{app:continual}}{Continual learning and recovery}{\pageref{app:continual}}
\AppGuideEntry{\ref{app:pruning}}{Pruning operators and statistical tests}{\pageref{app:pruning}}
\AppGuideEntry{\ref{app:pruning-recovery}}{Pruning Recovery and Support Protection}{\pageref{app:pruning-recovery}}
\AppGuideEntry{\ref{app:reproduction}}{Reproducibility of the Application Results}{\pageref{app:reproduction}}
\AppGuideEntry{\ref{app:applications-summary}}{Findings, recommendations, and open questions}{\pageref{app:applications-summary}}

\endgroup

%% file: appendix/appendix_methods.tex
\subsection{Details of the Task Encoding Sparsity Measurement}
\label{app:measurement}

This appendix specifies the measurement of Section~\ref{sec:measurement} in enough detail to recompute every $k^*$ and every ceiling reported for the 14 core tasks on the 19 models and for the ten additional tasks on Gemma-2-9B.
We call this set of task--model readings the task matrix.
The appendix proceeds from definitions to estimates and then to guarantees.
Section~\ref{app:notation} states the definition formally, and Sections~\ref{app:inputs} to~\ref{app:probes} specify the representations, the task pools, and the probe fits that enter it.
Sections~\ref{app:search} and~\ref{app:aggregation} give the search and aggregation procedures as algorithms, with the order-theoretic properties the paper relies on.
Section~\ref{app:worked} traces one task--model cell through the procedure with its actual numbers.
Section~\ref{app:finite-sample} measures how precisely the test split determines each ceiling and turns test-split reuse into an explicit finite-sample statement.
Sections~\ref{app:models} and~\ref{app:sae} list the models and describe the SAE protocol.
Section~\ref{app:design} justifies each design choice and points to the control that measures its effect, and Sections~\ref{app:selection} to~\ref{app:scope} state the statistical units, estimator variants, compute, and conditions under which each claim holds.

Table~\ref{tab:claim-evidence-measurement} maps every statement of Section~\ref{sec:measurement} to the place where this appendix defines or supports it.
Table~\ref{tab:protocol-glance} collects every setting of the main protocol in one place.

\begin{table}[ht]
\centering\AppTableStyle
\caption{\textbf{Claim-to-evidence map for Section~\ref{sec:measurement}.}
Each row pairs one statement of the main text with its formal object in this appendix and with the quantity that supports it.}
\label{tab:claim-evidence-measurement}
\begin{tabular}{@{}p{0.34\linewidth}p{0.25\linewidth}p{0.33\linewidth}@{}}
\toprule
\rowcolor{AppGold} \AppHead{Statement in Section~\ref{sec:measurement}} & \AppHead{Formal object} & \AppHead{Supporting quantity} \\
\midrule
$k^*$ is the smallest searched count recovering 95\% of the dense AUROC above chance & Equations~\ref{eq:threshold} and~\ref{eq:layer-reading} & Definition, with the threshold example in Section~\ref{app:notation} \\
A finite $k^*$ certifies a qualifying readout, and $\infty$ records that no candidate qualified & Lemma~\ref{lem:basic}(i) & Exact statement about the searched family \\
Raising the threshold can only raise $k^*$ & Lemma~\ref{lem:basic}(ii) and (iv) & Holds for local, confirmed, and task-level counts \\
$k^*$ refers to the model's own coordinates & Proposition~\ref{prop:affine} & Random-rotation control in Table~\ref{tab:rotation} \\
Two search procedures supply candidates & Algorithm~\ref{alg:search} & Ranked path alone keeps the task order at 0.99 (Table~\ref{tab:estimator}) \\
A single-seed sweep places a window of four grid layers & Section~\ref{app:aggregation} & 240 of 266 core cells use the four-layer window \\
Five seeds with 85\% stratified subsamples on a fixed test split & Section~\ref{app:aggregation} & Seed range reported for every cell \\
Neighbor confirmation, then medians over class decisions and seeds & Equation~\ref{eq:nc}, Algorithm~\ref{alg:aggregate}, Lemma~\ref{lem:basic}(iii) & Best single layer keeps the task order at 0.98 (Table~\ref{tab:estimator}), worked example in Table~\ref{tab:worked} \\
Reliable readability means a task ceiling above 0.90 & Readability flag in Section~\ref{app:aggregation} & Flag shown in every count table \\
Reuse of the test split has a measured and bounded effect on the count & Section~\ref{app:finite-sample}, Proposition~\ref{prop:sandwich} & Ceiling standard errors below the threshold gap in 99.4\% of readable class decisions (Table~\ref{tab:ceiling-precision}) \\
14 core tasks on 19 models, ten additional tasks on Gemma-2-9B & Tables~\ref{tab:tasks} and~\ref{tab:models} & Pool sizes, positive rates, and layer grids \\
SAE supports follow a separate sweep-based protocol & Section~\ref{app:sae} & Counts compared only within one basis \\
\bottomrule
\end{tabular}
\end{table}

\begin{table}[ht]
\centering\AppTableStyle
\caption{\textbf{The main protocol at a glance.}
Each setting holds for all 19 models and 24 tasks unless the middle column states otherwise, and the last column names the section that defines and justifies it.}
\label{tab:protocol-glance}
\begin{tabular}{@{}p{0.27\linewidth}p{0.50\linewidth}p{0.15\linewidth}@{}}
\toprule
\rowcolor{AppGold} \AppHead{Component} & \AppHead{Setting} & \AppHead{Defined in} \\
\midrule
Input length & 768 tokens, 1024 for mathematical-step correctness & \ref{app:inputs} \\
Input format & Raw text for base models, one chat-template user turn for instruction-tuned models & \ref{app:inputs} \\
Representation & Residual stream after a block, mean over all attended positions & \ref{app:notation} \\
Numerics & Forward pass in bfloat16, stored features in float32 & \ref{app:inputs} \\
Standardization & Training-split mean and standard deviation per feature & \ref{app:probes} \\
Probe & Logistic regression with intercept, L2 penalty, class-balanced weights & \ref{app:probes} \\
Regularization & Inverse strength $\gamma\in\{0.01,0.1,1,10\}$ chosen on a stratified 20\% validation portion & \ref{app:probes} \\
Feature ranking & Absolute two-sample Welch $t$-statistic on training data & \ref{app:search} \\
Candidate sizes & Prefixes of size $1,2,4,\dots,1024$, plus L1 supports at ten strengths from $10^{-3}$ to $10^{1.5}$ & \ref{app:search} \\
Score & Test AUROC of the probe's decision function & \ref{app:notation} \\
Threshold & 95\% of the ceiling's AUROC above chance & \ref{app:notation} \\
Layers & Four adjacent layers of the probed grid around the sweep's best layer & \ref{app:aggregation} \\
Seeds & Five seeds, each with a stratified 85\% training subsample and a fixed test split & \ref{app:aggregation} \\
Aggregation & Neighbor confirmation, then median over class decisions, then median over seeds & \ref{app:aggregation} \\
Readability flag & A task ceiling at or below 0.90 marks a count that we do not interpret as a task size & \ref{app:aggregation} \\
\bottomrule
\end{tabular}
\end{table}

\subsubsection{Notation and Formal Definition}
\label{app:notation}

\textbf{Representations.}
Let $f$ be a frozen LM with $L$ transformer blocks and residual width $d$.
For an input text $x$ tokenized into positions $1,\dots,T_x$, let $h_\ell(x,t)\in\mathbb{R}^d$ denote the residual stream after block $\ell$ at position $t$, with blocks indexed from $0$.
We form the representation of $x$ at layer $\ell$ by mean pooling over attended positions,
\begin{equation}
    \phi_\ell(x)=\frac{1}{T_x}\sum_{t=1}^{T_x} h_\ell(x,t)\ \in\mathbb{R}^{d},
    \label{eq:pool}
\end{equation}
and each of the $d$ coordinates of $\phi_\ell(x)$ is one feature.
The SAE protocol of Section~\ref{app:sae} instead starts from the last position $h_\ell(x,T_x)$ and maps it into a sparse dictionary.

\textbf{Tasks and class decisions.}
A task provides a training split $\mathcal{D}_{\mathrm{tr}}$, a test split $\mathcal{D}_{\mathrm{te}}$, and a finite set of class decisions $\mathcal{C}$.
Each class decision $c\in\mathcal{C}$ assigns a binary label $y^{c}(x)\in\{0,1\}$ to every text $x$ in both splits.
Binary tasks have a single class decision.
Multiclass tasks have one class decision per evaluated class, which marks that class as positive and every other text as negative.
We write $\mathcal{P}_c=\{x\in\mathcal{D}_{\mathrm{te}}:y^c(x)=1\}$ and $\mathcal{N}_c=\{x\in\mathcal{D}_{\mathrm{te}}:y^c(x)=0\}$ for the positive and negative test texts of class decision $c$, with sizes $n_1^{\mathrm{te}}$ and $n_0^{\mathrm{te}}$.

\textbf{Probes and scores.}
For a feature subset $S\subseteq[d]$ with $[d]=\{1,\dots,d\}$, we write $\phi_{\ell,S}(x)\in\mathbb{R}^{|S|}$ for the restriction of $\phi_\ell(x)$ to the features in $S$.
A probe on $S$ is a linear classifier fitted to $\{(\phi_{\ell,S}(x),y^c(x)):x\in\mathcal{D}_{\mathrm{tr}}\}$ by the procedure of Section~\ref{app:probes}, and it returns a real-valued decision function $g_S$.
The score of the probe is its AUROC on the test split, which equals the Mann--Whitney statistic of the decision values~\citep{mann1947test},
\begin{equation}
    A_{c,\ell}(S)=\frac{1}{n_1^{\mathrm{te}}\,n_0^{\mathrm{te}}}\sum_{x\in\mathcal{P}_c}\sum_{x'\in\mathcal{N}_c}\Bigl(\mathbf{1}\bigl[g_S(x)>g_S(x')\bigr]+\tfrac12\,\mathbf{1}\bigl[g_S(x)=g_S(x')\bigr]\Bigr),
    \label{eq:auroc}
\end{equation}
where $g_S(x)$ abbreviates $g_S(\phi_{\ell,S}(x))$.
The dense probe uses the full set $[d]$.
Equation~\ref{eq:auroc} has two properties that the protocol uses.
It depends on the decision values only through their order, so any strictly increasing transformation of $g_S$ leaves the score unchanged and no calibration enters the measurement.
It averages over positive and negative texts separately, so the class prior does not enter the score, which keeps one-vs-rest decisions with positive rates near 1\% on the same scale as balanced tasks.
A score of $0.5$ is the chance level of any scorer that ignores the input.

\textbf{Ceiling and threshold.}
Let $\Lambda$ be the set of layers probed in the five-seed measurement of Section~\ref{app:aggregation}, let $R$ be its set of seeds, and let a superscript $r$ mark the seed of a fit.
The dense-probe reference $C$ of Equation~\ref{eq:kstar}, which we call the ceiling, is the largest dense-probe AUROC of a class decision over probed layers and seeds.
The threshold follows from it,
\begin{equation}
    C_c=\max_{\ell\in\Lambda,\,r\in R} A^{(r)}_{c,\ell}([d]),
    \qquad
    \tau_c = 0.5 + 0.95\,(C_c-0.5).
    \label{eq:threshold}
\end{equation}
The threshold $\tau_c$ is the AUROC at which the normalized score $s(k)$ of Equation~\ref{eq:kstar} equals $0.95$, so a probe qualifies exactly when its AUROC reaches $\tau_c$.
For example, a ceiling of $0.98$ gives $\tau_c=0.956$, and a ceiling of $0.90$ gives $\tau_c=0.88$.
The normalization measures performance above chance, so tasks with different ceilings face the same relative requirement.
The factor $0.95$ is a convention rather than a derived constant, and Section~\ref{app:design} reports the effect of replacing it by $0.90$ or $0.975$.

\textbf{Design rationale.}
The threshold is relative to the ceiling of the same class decision, so a count never rewards a task merely for being easy to read densely.
An absolute AUROC threshold would conflate readability with sparsity, because a task with a ceiling of $0.91$ and a task with a ceiling of $1.00$ would face requirements of different relative difficulty.

\textbf{Local count.}
At one layer and one seed, the search of Section~\ref{app:search} evaluates a finite family $\mathcal{S}^{(r)}_{c,\ell}$ of candidate subsets.
The local count of Section~\ref{sec:measurement} is the size of the smallest qualifying candidate,
\begin{equation}
    k^*_{c,\ell,r}=\min\bigl\{|S| : S\in\mathcal{S}^{(r)}_{c,\ell},\ A^{(r)}_{c,\ell}(S)\ge\tau_c\bigr\},
    \label{eq:layer-reading}
\end{equation}
and it equals $\infty$ when no candidate qualifies.
We call the candidate that attains this minimum the support of the class decision at that layer and seed.
All counts take values in the extended set $\overline{\mathbb{N}}=\{1,2,\dots\}\cup\{\infty\}$, ordered so that $\infty$ exceeds every finite value.
Section~\ref{app:aggregation} turns the local counts into the task-level $k^*$ reported for each task--model cell.

\textbf{Takeaway for practitioners.}
Dense-probe AUROC alone does not reveal how many coordinates carry a task.
In the probe-family control of Table~\ref{tab:probefamily} on Gemma-2-9B, intent classification reaches a ceiling of $1.0000$ with a count of 4, while harmfulness detection reaches $0.9785$ with a count of 256.
Two tasks that a dense probe ranks as almost equally readable thus differ by a factor of 64 in the size of their readout, and reporting the ceiling together with $k^*$ separates the two quantities.

\subsubsection{Input Representations}
\label{app:inputs}

\textbf{Tokenization.}
We tokenize each text with the probed model's own tokenizer and truncate it to at most 768 tokens.
Mathematical-step correctness uses 1024 tokens because each input includes a problem and a partial solution ending at the evaluated step.

\textbf{Input format.}
Base models receive the raw text with the tokenizer's default special tokens and no chat template.
Instruction-tuned models receive the same text as a single user turn rendered through the model's chat template, followed by the template's generation prompt, so the last position is the one at which the model would begin its reply.

\textbf{Numerics.}
We run one forward pass per text in bfloat16 precision, read the hidden states after every probed block of Table~\ref{tab:models}, and store the pooled vectors of Equation~\ref{eq:pool} in float32, because the residual activations of the Gemma-3 models exceed the float16 range.

\textbf{Pooling.}
The mean in Equation~\ref{eq:pool} runs over all attended positions, including special and template tokens, and the last position is the final non-padding token.
Template tokens therefore enter the mean-pooled representation of instruction-tuned models, and the input-format factorial of Section~\ref{sec:training} separates their contribution from that of the tuned weights.

\textbf{Layer convention.}
Layer indices are block indices starting at $0$, and layer $\ell$ denotes the residual stream after block $\ell$.
The SAE suites in Section~\ref{app:sae} use the same layer naming.

\subsubsection{Task Pools, Splits, and Class Decisions}
\label{app:tasks}

\textbf{Question.}
The measurement compares models on identical inputs, so every task must be one fixed pool of texts whose labels cannot be predicted from leakage or from trivial surface statistics.
This section specifies how we build, split, and balance such pools.

\textbf{Data.}
Each task is a pool of texts per split together with one binary label vector per class decision.
We build every pool once with a fixed seed and reuse it for all 19 models, so differences between models never come from differences in data.
Table~\ref{tab:tasks} lists all 24 pools with their sources, numbers of class decisions, sizes, positive rates, and test splits.

\textbf{Splits.}
Where the source provides labeled training and test splits, we keep them, and where its test split is unlabeled we use the development split.
Otherwise we hold out 20\% of the examples for testing, and we choose the held-out unit to prevent leakage.
Mathematical-answer and mathematical-step correctness split by problem, since ProcessBench contains several solutions of the same problem.
PII detection, which flags personally identifiable information, splits by the masked template, since the source re-instantiates one template with many identity values.
Tool selection splits by the request text after replacing every digit string with a placeholder, since the requests are templated.
Harmfulness and jailbreak detection deduplicate prompts and then split stratified by label.
The ten additional tasks on Gemma-2-9B keep their native splits, and we remove duplicate texts within each split and drop test texts that also occur in the training split.

\textbf{Balance and caps.}
Sentiment, relevance, fact checking, routing, and the additional binary tasks are balanced exactly before capping.
Harmfulness detection and code-bug detection keep their natural class ratios.
Mathematical-answer correctness also keeps its natural ratio, which is close to one half because ProcessBench pairs correct and incorrect solutions of the same problem.
PII detection assigns each text at random to the positive or the negative construction described below.
Mathematical-step correctness keeps every erroneous step and at most two correct steps per solution, then subsamples correct steps to twice the number of erroneous ones.
One-vs-rest tasks whose evaluated classes form a subset of the source classes, namely CLINC150, Banking77, tool selection, MASSIVE, and 20 Newsgroups, keep every text of the evaluated classes and fill the remainder of the cap with texts of other classes, so each class decision has a small positive rate.
Language identification, AG News, and Emotion evaluate every class of their pools, so the cap subsamples all classes alike.
For Banking77 and tool selection, whose class frequencies are unequal, we cap the pool so that the rarest evaluated class keeps a positive rate of at least 5\%.
We evaluate a class decision only when it has at least five positives in each split for the core tasks and at least 20 for the additional tasks.
Section~\ref{app:finite-sample} shows that the smaller class count of the test split sets the precision of every AUROC, which is the reason for this minimum, and in the five-seed measurement the rarer test class holds at least 30 texts in every class decision of the 14 core tasks.

\textbf{Class decisions.}
CLINC150 uses 20 intents sampled uniformly from its 150 in-scope intents.
Banking77 uses its ten most frequent intents.
Tool selection uses the four most frequent functions among conversations whose first assistant turn is a function call, and the input is the user request alone without the tool schema.
Language identification draws on a public 20-language identification corpus assembled from multilingual product reviews, XNLI, and translated STS benchmark sentences.
We sample ten of its languages with English forced into the sample, namely Arabic, Bulgarian, English, French, Greek, Hindi, Japanese, Turkish, Urdu, and Vietnamese.
Among the additional tasks, AG News uses all four topics, Emotion all six emotions, 20 Newsgroups ten sampled groups, and MASSIVE 20 sampled intents of which 15 meet the minimum positive count.

\textbf{Positive classes.}
Routing marks problems drawn from MATH rather than GSM8K, after removing MATH problems with figure code and dropping problems longer than 1200 characters from both sources.
PII detection marks the original texts of the ai4privacy corpus, and its negatives are the same texts with every annotated span replaced by a generic filler, so each text serves as either a positive or a negative and the two classes share domain and register.
Harmfulness detection marks prompts labeled harmful in WildGuardMix.
Jailbreak detection also draws on WildGuardMix but uses only harmful prompts, and it marks the adversarially framed ones against the plainly stated ones, so the label tracks adversarial framing rather than harmful content.
Toxicity marks user turns labeled toxic in ToxicChat.
Sentiment tasks mark positive reviews, subjectivity marks subjective sentences, and the two TweetEval tasks mark offensive and hateful tweets.
Relevance marks the passage selected for a query in MS~MARCO, with one selected and one unselected passage per query, each concatenated to the query.
Fact checking marks refuted claims against supported ones, and the probe sees only the claim text without evidence.
Code-bug detection marks defective C functions.
Mathematical-answer correctness marks complete solutions whose final answer is wrong.
Mathematical-step correctness marks the first erroneous step of a ProcessBench solution against correct steps, each rendered together with its problem and the preceding steps.

\textbf{Control for length.}
Raw adversarial prompts in WildGuardMix are several times longer than plainly stated ones, so a probe could separate the jailbreak classes by length alone.
We therefore length-match the jailbreak pool before splitting.
We bin prompts by character length on quantiles of the pooled distribution and pair each positive with a negative of nearly equal length within its bin, keeping only paired prompts.
If length still carried the label after matching, we would expect a classifier that reads only text length to reach an AUROC well away from $0.5$.
In the matched pool, that length-only classifier reaches an AUROC of 0.482 on the training split and 0.468 on the test split, and the median character lengths of the two classes differ by less than 10\%.

\textbf{Truncation.}
We truncate language-identification texts to 256 characters, relevance passages to 800, code functions to 1024, and jailbreak prompts to 2000.
We drop PII texts longer than 400 characters and tool requests longer than 512, so that a positive can never lose its annotated spans to truncation.

\textbf{Design rationale.}
Each construction above removes one alternative explanation of a readable label.
First, fixed pools shared by all 19 models ensure that a difference between two models cannot come from a difference in their data.
Second, the grouped hold-outs split by problem, by masked template, and by digit-normalized request, so a probe cannot score well by recognizing a near-duplicate of a training text.
Third, the PII negatives are the positive texts themselves with the annotated spans replaced, so the two classes share domain, register, and length distribution, and a probe must read the presence of identifying spans rather than the source of the text.
Fourth, the length-matched jailbreak pool removes the strongest surface cue of that task, and the length-only AUROC of 0.482 and 0.468 confirms its removal.
Fifth, jailbreak detection contrasts two groups of harmful prompts, so harmful content cannot explain the label.

\textbf{Scope.}
Routing measures source classification as a proxy for difficulty, since MATH problems use LaTeX notation and GSM8K problems are plain prose, and we report it as such.
Fact checking without evidence can rest on the wording of the claim, and Appendix~\ref{app:inputbase} measures how far a bag-of-words baseline on the raw text reaches on every task of Table~\ref{tab:tasks}.

\begin{table}[t]
\AppTableStyle
    \centering
    \caption{\textbf{Task statistics.}
    Each row is one task, and each class decision is probed separately.
    Train and test give the sizes of the pools shared by every model, and the positive rate is the fraction of positives in the training pool, as a range for tasks with several class decisions.
    The split column states whether the test pool is the source's test split, its development split, or a 20\% hold-out formed as in Section~\ref{app:tasks}.
    The lower block lists the ten additional tasks evaluated only on Gemma-2-9B.}
    \label{tab:tasks}
    \resizebox{\linewidth}{!}{%
    \begin{tabular}{llrrrrl}
        \toprule
\rowcolor{AppGold}
        \AppHead{Task} & \AppHead{Source} & \AppHead{Decisions} & \AppHead{Train} & \AppHead{Test} & \AppHead{Positive rate} & \AppHead{Split} \\
        \midrule
        Language identification & \citep{keung2020multilingual,conneau2018xnli,cer2017semeval} & 10 & 8000 & 3000 & 0.09 to 0.10 & test \\
        Tool selection & \citep{glaiveai2023glaive} & 4 & 3300 & 840 & 0.05 to 0.12 & hold-out \\
        Routing & GSM8K vs.\ MATH~\citep{cobbe2021training,hendrycks2021measuring} & 1 & 6000 & 2000 & 0.50 & test \\
        Intent (CLINC150) & \citep{larson2019evaluation} & 20 & 8000 & 3000 & 0.01 & test \\
        Intent (Banking77) & \citep{casanueva2020efficient} & 10 & 3420 & 780 & 0.05 & test \\
        PII detection & \citep{ai4privacy2023pii} & 1 & 6000 & 2000 & 0.49 & hold-out \\
        Jailbreak detection & WildGuardMix~\citep{han2024wildguard} & 1 & 1904 & 474 & 0.50 & hold-out \\
        Harmfulness detection & WildGuardMix~\citep{han2024wildguard} & 1 & 6000 & 2000 & 0.51 & hold-out \\
        Sentiment & SST-2~\citep{socher2013recursive} & 1 & 8000 & 856 & 0.50 & dev \\
        Relevance & MS MARCO~\citep{bajaj2016ms} & 1 & 6000 & 2000 & 0.50 & dev \\
        Fact checking & FEVER~\citep{thorne2018fever,atanasova2020generating} & 1 & 8000 & 3000 & 0.50 & dev \\
        Code-bug detection & Devign via CodeXGLUE~\citep{zhou2019devign,lu2021codexglue} & 1 & 6000 & 2000 & 0.46 & test \\
        Mathematical-answer correctness & ProcessBench~\citep{zheng2025processbench} & 1 & 2697 & 703 & 0.50 & hold-out \\
        Mathematical-step correctness & ProcessBench~\citep{zheng2025processbench} & 1 & 5334 & 1329 & 0.33 & hold-out \\
        \midrule
        News topic & AG News~\citep{zhang2015character} & 4 & 8000 & 3000 & 0.25 & test \\
        Movie-review sentiment & IMDB~\citep{maas2011learning} & 1 & 8000 & 3000 & 0.50 & test \\
        Business-review sentiment & Yelp~\citep{zhang2015character} & 1 & 8000 & 3000 & 0.50 & test \\
        Subjectivity & SUBJ~\citep{pang2004sentimental} & 1 & 7934 & 1934 & 0.50 & test \\
        Offensive language & TweetEval~\citep{barbieri2020tweeteval} & 1 & 7870 & 472 & 0.50 & test \\
        Hate speech & TweetEval~\citep{barbieri2020tweeteval} & 1 & 7562 & 2230 & 0.50 & test \\
        Emotion & \citep{saravia2018carer} & 6 & 8000 & 1989 & 0.04 to 0.34 & test \\
        Intent (MASSIVE) & \citep{fitzgerald2023massive} & 15 & 8000 & 2949 & 0.01 to 0.08 & test \\
        Newsgroup topic & 20 Newsgroups~\citep{lang1995newsweeder} & 10 & 8000 & 3000 & 0.06 to 0.07 & test \\
        Toxicity & ToxicChat~\citep{lin2023toxicchat} & 1 & 750 & 646 & 0.50 & test \\
        \bottomrule
    \end{tabular}}
\end{table}

\subsubsection{Probe Fitting}
\label{app:probes}

All probes share one fitting procedure, which we denote $\textsc{FitScore}(Z_{\mathrm{tr}},y,Z_{\mathrm{te}},S)$ and use as a subroutine in Algorithm~\ref{alg:search}, and which returns the test AUROC of a probe on the features in $S$.

\textbf{Standardization.}
Let $X_{\mathrm{tr}}\in\mathbb{R}^{n\times d}$ hold the training representations of one class decision at one layer, after the subsampling of Section~\ref{app:aggregation}.
We standardize every feature $j$ with the mean $\mu_j$ and the standard deviation $\sigma_j$ of its column over the $n$ training texts,
\begin{equation}
    z_j=\frac{x_j-\mu_j}{\sigma_j},
    \qquad
    \mu_j=\frac1n\sum_{i=1}^n x_{ij},
    \qquad
    \sigma_j^2=\frac1n\sum_{i=1}^n (x_{ij}-\mu_j)^2,
    \label{eq:standardize}
\end{equation}
and set $\sigma_j=1$ where the column is constant.
The same $\mu$ and $\sigma$ transform the test representations, so no test statistic enters the fit.
For the sparse SAE features of Section~\ref{app:sae}, we divide by $\sigma_j$ without centering, which preserves the sparsity of the features.

\textbf{Objective.}
Given a feature subset $S$ with $|S|=k$, the probe is a logistic regression with an intercept, an L2 penalty, and class-balanced example weights $\omega_{y}$,
\begin{equation}
    \min_{w\in\mathbb{R}^{k},\,b\in\mathbb{R}}\ \tfrac{1}{2}\lVert w\rVert_2^2 + \gamma\sum_{i=1}^{n}\omega_{y_i}\log\bigl(1+\exp(-\tilde y_i(w^\top z_{i,S}+b))\bigr),
    \qquad
    \omega_{y}=\frac{n}{2\,n_{y}},
    \label{eq:probe}
\end{equation}
where $\tilde y_i=2y_i-1\in\{-1,+1\}$ is the signed label, $n_y$ is the number of training examples with label $y$, and $\gamma>0$ is the inverse regularization strength.
We write $\gamma$ rather than the customary $C$ to keep $C$ for the ceiling.
The weights satisfy $n_1\omega_1=n_0\omega_0=n/2$, so both classes carry equal total weight, which matters for the one-vs-rest decisions with positive rates near 1\%.
The decision function of the fitted probe is $g_S(z)=w^\top z_S+b$.
We solve Equation~\ref{eq:probe} with the primal trust-region Newton solver of LIBLINEAR~\citep{fan2008liblinear} as wrapped by scikit-learn~\citep{pedregosa2011scikit}, with at most 2000 iterations.
LIBLINEAR treats the intercept as an additional constant feature, so the penalty in practice also includes $b^2/2$.
This term is identical for every subset and every model, and it leaves the comparisons unchanged.
With the intercept penalized, the objective is strictly convex in $(w,b)$, so each L2 fit has a unique minimizer.

\textbf{Regularization selection.}
We choose $\gamma$ from $\Gamma_2=\{0.01,0.1,1,10\}$ on a validation portion of the training split.
The validation portion holds 20\% of each class, rounded down, drawn with the seed of the fit.
For each $\gamma$ we fit Equation~\ref{eq:probe} on the remaining 80\% and compute the AUROC of the decision function on the validation portion, which we write $\mathrm{AUROC}_{\mathrm{val}}(\gamma)$.
The selected strength is
\begin{equation}
    \hat\gamma=\min\Bigl\{\gamma\in\Gamma_2 : \mathrm{AUROC}_{\mathrm{val}}(\gamma)=\max_{\gamma'\in\Gamma_2}\mathrm{AUROC}_{\mathrm{val}}(\gamma')\Bigr\},
    \label{eq:gamma-select}
\end{equation}
so ties resolve toward stronger regularization.
We refit on the full training split with $\hat\gamma$, and the returned score is the AUROC of the refitted decision function on the test split.
The dense probe that supplies the ceiling follows the same procedure with $S=[d]$.

\textbf{L1 variant.}
The L1 path of Algorithm~\ref{alg:search} replaces $\tfrac12\lVert w\rVert_2^2$ in Equation~\ref{eq:probe} by $\lVert w\rVert_1$ and fits a probe on all $d$ standardized features jointly at each of the ten strengths
\begin{equation}
    \Gamma_1=\bigl\{10^{-3+4.5\,i/9} : i=0,1,\dots,9\bigr\},
    \label{eq:l1-grid}
\end{equation}
which are spaced logarithmically from $10^{-3}$ to $10^{1.5}$.
LIBLINEAR solves this problem by coordinate descent, again with the intercept as a constant feature.
Each fit yields a support $T(\gamma)=\{j:w_j(\gamma)\neq0\}$ and a test AUROC of that fit's own decision function.
We drop fits with an empty support.
These fits do not select $\gamma$ on validation data, since each value of $\gamma$ is a separate candidate.

\textbf{Standardization and the L1 penalty.}
Standardization makes the L1 selection independent of the units of each coordinate.
Writing the fitted probe in raw coordinates as $w^\top z+b=v^\top x+b'$ gives $v_j=w_j/\sigma_j$ and $b'=b-\sum_j w_j\mu_j/\sigma_j$, and therefore
\begin{equation}
    \lVert w\rVert_1=\sum_{j=1}^{d}\sigma_j\,\lvert v_j\rvert.
    \label{eq:l1-weighted}
\end{equation}
The L1 path on standardized features is thus a weighted lasso on raw features whose weights equal the training standard deviations, apart from the penalty on the intercept, which in raw coordinates applies to $b'+\sum_j v_j\mu_j$.
A coordinate with a large raw scale receives no advantage in selection, and a coordinate with a small raw scale receives no penalty for it.

\textbf{Design rationale.}
Three choices keep the probe fit from favoring particular features or tasks.
Training-split standardization with the weighting of Equation~\ref{eq:l1-weighted} prevents coordinates with large activations, which occur in some residual dimensions, from dominating the selection.
Class-balanced weights prevent rare positives from being ignored by the loss, and the class-prior invariance of Equation~\ref{eq:auroc} prevents the positive rate from entering the score.
Validation-based choice of $\gamma$ on training data prevents the regularization strength from being tuned on the test split.

\subsubsection{Support Search at One Layer}
\label{app:search}

\begin{figure}[t]
\centering
\includegraphics[width=\linewidth]{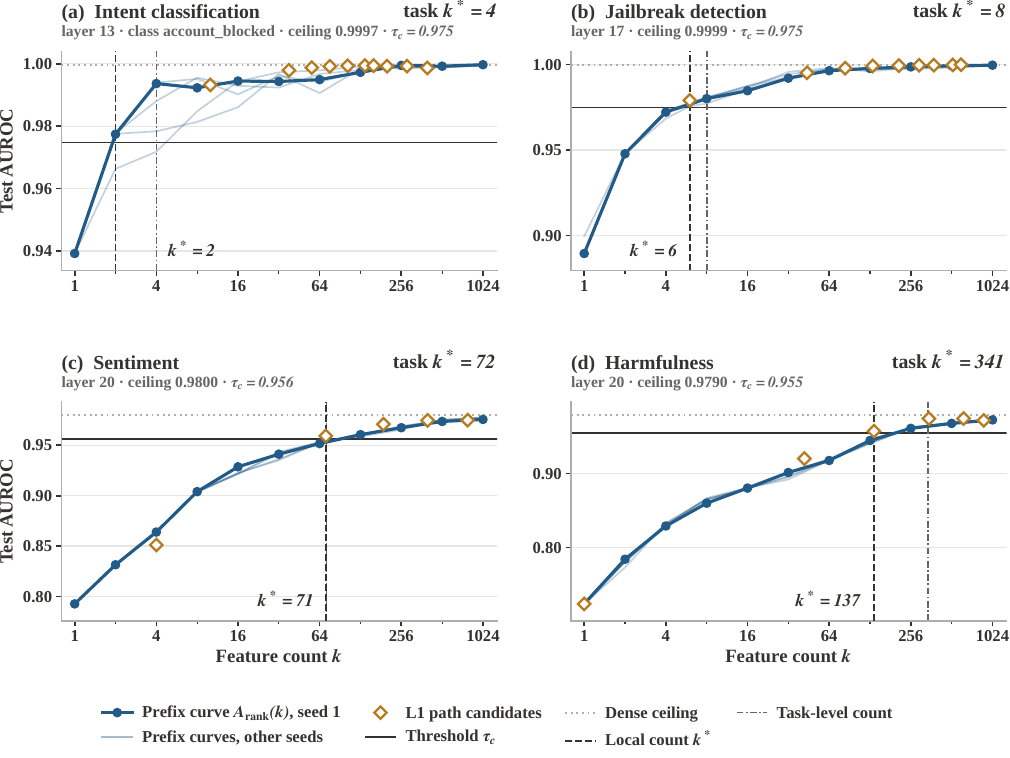}
\caption{\textbf{Reading a local count off one probe cell.}
Each panel shows Gemma-2-9B on one task at the window layer where seed 1 attains its smallest local count.
The dark curve is the prefix curve $A_{\mathrm{rank}}(k)$ of seed 1, the thin curves are the other four seeds at the same layer, and the hollow diamonds are the L1 path candidates of seed 1 placed at their support sizes, with supports above 1024 lying outside the axis.
The solid horizontal line marks the threshold $\tau_c$ of Equation~\ref{eq:threshold}, the dotted horizontal line marks the dense ceiling $C_c$ over all window layers and seeds, and the dashed vertical line marks the local count of Equation~\ref{eq:layer-reading}, the smallest qualifying size on either path.
The gray vertical line and each title give the task-level count of Table~\ref{tab:gold}, the median over five seeds after neighbor confirmation (Appendix~\ref{app:aggregation}).
(a) Intent classification, class account\_blocked, layer 13, ceiling 0.9997, $\tau_c=0.975$.
The prefix curve qualifies at $k^*=2$, layer 17 ties at 2, and the task-level count of 4 is the median over 20 class decisions.
(b) Jailbreak detection, layer 17, ceiling 0.9999, $\tau_c=0.975$.
An L1 support of 6 qualifies before the prefix grid point 8, and neighbor confirmation with layer 20 returns 8.
(c) Sentiment, layer 20, ceiling 0.980, $\tau_c=0.956$.
An L1 support of 71 qualifies before the prefix grid point 128, layers 17 and 24 need 128 for this seed, and the task-level median over seeds is 72.
(d) Harmfulness, layer 20, ceiling 0.979, $\tau_c=0.955$.
An L1 support of 137 qualifies, while layers 17 and 24 need 341 and 342, so neighbor confirmation returns 341.}
\label{fig:app-kcurves}
\end{figure}

\textbf{Question.}
The definition in Equation~\ref{eq:layer-reading} needs a finite candidate family that contains small qualifying subsets whenever they are easy to find.
Exhaustive search over subsets is infeasible for $d$ in the thousands, so we combine two complementary paths.
The first path ranks features and refits probes on prefixes of the ranking, following sparse probing~\citep{gurnee2023finding}.
The second path reads supports off the L1 regularization path.
Algorithm~\ref{alg:search} states the procedure, and this section defines its components and derives its properties.
Figure~\ref{fig:app-kcurves} shows the recorded curves of both paths and the resulting local count for four tasks on Gemma-2-9B.

\textbf{Ranking statistic.}
We rank the standardized training features by the absolute two-sample Welch $t$-statistic~\citep{welch1947generalization} between the positive and negative examples of the class decision,
\begin{equation}
    t_j=\frac{\lvert \bar z_{1j}-\bar z_{0j}\rvert}{\sqrt{s_{1j}^2/n_1 + s_{0j}^2/n_0 + \varepsilon_0}},
    \qquad \varepsilon_0=10^{-12},
    \label{eq:tstat}
\end{equation}
where $\bar z_{yj}$ and $s^2_{yj}$ are the mean and the population variance of feature $j$ among training examples with label $y$, and $n_y$ is their count.
The constant $\varepsilon_0$ only guards against division by zero for constant features.
Let $\pi$ be the permutation that sorts features by decreasing $t_j$, and let $\pi_{1:k}$ be its first $k$ entries.
The statistic scores each feature on its own, so the ranking does not favor features that are useful only in combination.
The L1 path selects features jointly and therefore covers that case.

\textbf{Prefix grid.}
The tested prefix sizes are the powers of two up to the representation width,
\begin{equation}
    K=\{2^i : 0\le i\le 10,\ 2^i\le d\},
    \label{eq:grid}
\end{equation}
so $K$ runs from 1 to 1024 for every model with $d\ge1024$ and stops at 512 for Gemma-3-270M with $d=640$.
The candidate family is the union of the prefix sets and the nonempty L1 supports,
\begin{equation}
    \mathcal{S}^{(r)}_{c,\ell}=\bigl\{\pi_{1:k}:k\in K\bigr\}\ \cup\ \bigl\{T(\gamma):\gamma\in\Gamma_1,\ T(\gamma)\neq\emptyset\bigr\},
    \label{eq:family}
\end{equation}
and it contains at most $|K|+|\Gamma_1|=21$ subsets per class decision, layer, and seed.

\begin{algorithm}[t]
\AppAlgorithmStyle
\caption{Support search for one class decision at one layer and one seed}
\label{alg:search}
\begin{algorithmic}[1]
\Require standardized training features $Z_{\mathrm{tr}}\in\mathbb{R}^{n\times d}$ with labels $y$, test features $Z_{\mathrm{te}}$ with labels $y_{\mathrm{te}}$, threshold $\tau_c$ from Equation~\ref{eq:threshold}, prefix grid $K$ from Equation~\ref{eq:grid}, L1 strengths $\Gamma_1$ from Equation~\ref{eq:l1-grid}
\Ensure local count $k^*_{c,\ell,r}$ and its support $S^*$
\State $\pi \gets$ feature indices sorted by decreasing $t_j$ of Equation~\ref{eq:tstat}
\For{$k\in K$ in increasing order} \Comment{ranked-prefix path}
    \State $A_{\mathrm{rank}}(k)\gets\textsc{FitScore}(Z_{\mathrm{tr}},y,Z_{\mathrm{te}},\pi_{1:k})$
\EndFor
\State $k_{\mathrm{rank}}\gets\min\{k\in K : A_{\mathrm{rank}}(k)\ge\tau_c\}$ \Comment{$\infty$ if the set is empty}
\State $k_{\ell_1}\gets\infty$
\For{$\gamma\in\Gamma_1$} \Comment{L1 path}
    \State $(w,b)\gets$ L1-penalized fit of Equation~\ref{eq:probe} on all $d$ features with strength $\gamma$
    \State $T\gets\{j : w_j\neq 0\}$
    \If{$T\neq\emptyset$ \textbf{and} $\mathrm{AUROC}(Z_{\mathrm{te}}w+b,\ y_{\mathrm{te}})\ge\tau_c$}
        \State $k_{\ell_1}\gets\min(k_{\ell_1},|T|)$ and record $T$ if it attains the minimum
    \EndIf
\EndFor
\State $k^*_{c,\ell,r}\gets\min(k_{\mathrm{rank}},k_{\ell_1})$
\State $S^*\gets$ the prefix or L1 support attaining $k^*_{c,\ell,r}$
\State \Return $k^*_{c,\ell,r}$, $S^*$
\end{algorithmic}
\end{algorithm}

\textbf{Cost.}
For one class decision, layer, and seed, Algorithm~\ref{alg:search} fits one dense probe and up to eleven prefix probes, each with four validation fits and one refit, plus ten L1 fits.
This amounts to at most $5\,(1+|K|)+|\Gamma_1|=70$ logistic fits per class decision, layer, and seed.
The dense probe supplies the ceiling of Equation~\ref{eq:threshold}, so we record the whole curve once and apply the threshold afterwards.
The threshold therefore never influences which probes we fit.

\textbf{Logical status of a count.}
A finite local count certifies that a qualifying subset of that size exists among the candidates, and $\infty$ records that no candidate qualified.
Neither statement certifies that no smaller subset exists, because the candidate family of Equation~\ref{eq:family} is finite.
Lemma~\ref{lem:basic} in Section~\ref{app:aggregation} proves this and the other elementary properties of the count, and the rest of this section establishes the one property that is not elementary, invariance under changes of units.

\begin{appproposition}[invariance to coordinate-wise affine maps]
\label{prop:affine}
Let $x'_j=a_jx_j+b_j$ with $a_j\neq0$ for every residual feature $j$, applied to the training and test splits alike.
In exact arithmetic, and with ties in the ranking broken identically, Algorithm~\ref{alg:search} returns the same local count and the same support on $x'$ as on $x$, provided each L1-penalized fit of Equation~\ref{eq:probe} has a unique solution.
\end{appproposition}

\begin{appproof}
A coordinate that is constant on the training split standardizes to zero there under both $x$ and $x'$, so both penalties give it a coefficient of exactly zero in every fit and it plays no further role, and we consider the remaining coordinates with $\sigma_j>0$.
For these, Equation~\ref{eq:standardize} gives $\mu'_j=a_j\mu_j+b_j$ and $\sigma'_j=\lvert a_j\rvert\sigma_j$, so it maps $x'_j$ to $\operatorname{sign}(a_j)\,z_j$.
The standardized features therefore change only by coordinate sign flips, which we write as $z'=Dz$ with $D=\operatorname{diag}(\operatorname{sign}(a_1),\dots,\operatorname{sign}(a_d))$.
A sign flip leaves $\lvert\bar z_{1j}-\bar z_{0j}\rvert$ and both class variances unchanged, hence every $t_j$ and the ranking $\pi$.
For any subset $S$, the map $(w,b)\mapsto(D_Sw,b)$ is a bijection between parameter vectors on $z$ and on $z'$ that preserves every margin $w^\top z_{i,S}+b$ and both penalties $\lVert w\rVert_2$ and $\lVert w\rVert_1$.
The objective of Equation~\ref{eq:probe} therefore takes the same value at corresponding points, and the unique minimizer on $z'$ is the image of the minimizer on $z$.
The decision function on every training, validation, and test point is therefore unchanged, and so are every AUROC, the validation choice of Equation~\ref{eq:gamma-select}, and the support of every L1 fit, because $D_S$ preserves which coefficients are zero.
\end{appproof}

Proposition~\ref{prop:affine} states exactly which changes of basis leave $k^*$ unchanged.
Rescaling, shifting, or negating a coordinate does not move the count, while mixing coordinates can.
The measurement therefore concerns the model's own residual coordinates, and the random-rotation control of Table~\ref{tab:rotation} measures how much a basis change moves it.
For SAE features, which Section~\ref{app:sae} scales without centering, the invariance holds for pure rescalings with $b_j=0$.

\textbf{Design rationale.}
The two paths cover each other's blind spots.
A univariate ranking misses features that help only in combination, and the L1 path selects features jointly.
An L1 path can skip sizes when several features enter at one strength, and the prefix grid tests every power of two.
Recording the complete curve before applying the threshold prevents the threshold from steering which probes are fitted.

\textbf{Remarks.}
We make four further observations.
First, the local count is a search result, not a certified minimum.
A finite value states that a qualifying subset of that size exists among the candidates, and $\infty$ states that no candidate qualified.
Second, the prefixes are nested, $\pi_{1:k}\subset\pi_{1:2k}$, but the prefix curve $A_{\mathrm{rank}}(k)$ need not increase in $k$, because each prefix probe selects its own regularization strength.
Algorithm~\ref{alg:search} takes the smallest qualifying $k$ rather than requiring every larger prefix to qualify.
Third, the grid of Equation~\ref{eq:grid} has a resolution of one octave above $k=1$, and the L1 path supplies sizes between grid points, which is why we combine the two paths.
Fourth, the comparison with $\tau_c$ in Equation~\ref{eq:layer-reading} reads test AUROCs, so $k^*$ summarizes the test curve over candidate sizes rather than a subset selected on held-out data.
Section~\ref{app:finite-sample} quantifies the consequences of this choice.

\textbf{Takeaway for practitioners.}
A sparse-probe count is an upper bound attached to a search procedure, so a study should report the candidate family together with the count.
Combining a univariate ranking with an L1 path costs 70 fits per class decision, layer, and seed, and removing the L1 path changes the task order only slightly on our models (median $\rho=0.99$ in Table~\ref{tab:estimator}), so the ranking path alone is a reasonable budget option when only task order matters.

\subsubsection{Layer Windows, Ceiling, Neighbor Confirmation, and Aggregation}
\label{app:aggregation}

\textbf{Question.}
A local count describes one class decision at one layer and one seed.
The paper reports one number per task and model, so this section specifies how local counts combine, and it shows that the combination inherits the properties of Section~\ref{app:search}.
The measurement runs in two stages.
A preliminary sweep locates the layers at which a task is most compactly readable, and a five-seed measurement on a window of those layers produces the reported numbers.
Algorithm~\ref{alg:aggregate} states the second stage.

\textbf{Preliminary sweep.}
For every model we first run Algorithm~\ref{alg:search} with one seed on the full training split at every layer of the grid in Table~\ref{tab:models}, for all three representations of Section~\ref{app:models}.
In this sweep the ceiling of a class decision is the largest test AUROC of any probe fitted for it, across representations, layers, and search paths, and the threshold follows Equation~\ref{eq:threshold}.
The best layer of a task in a representation is the layer with the smallest median local count across class decisions, with ties resolved toward the larger dense AUROC.
The sweep serves three purposes.
It places the window of the second stage, it supplies the SAE supports of Section~\ref{app:sae}, and it provides the layer-resolved curves behind the depth analyses of the paper.

\begin{figure}[t]
\centering
\includegraphics[width=\linewidth]{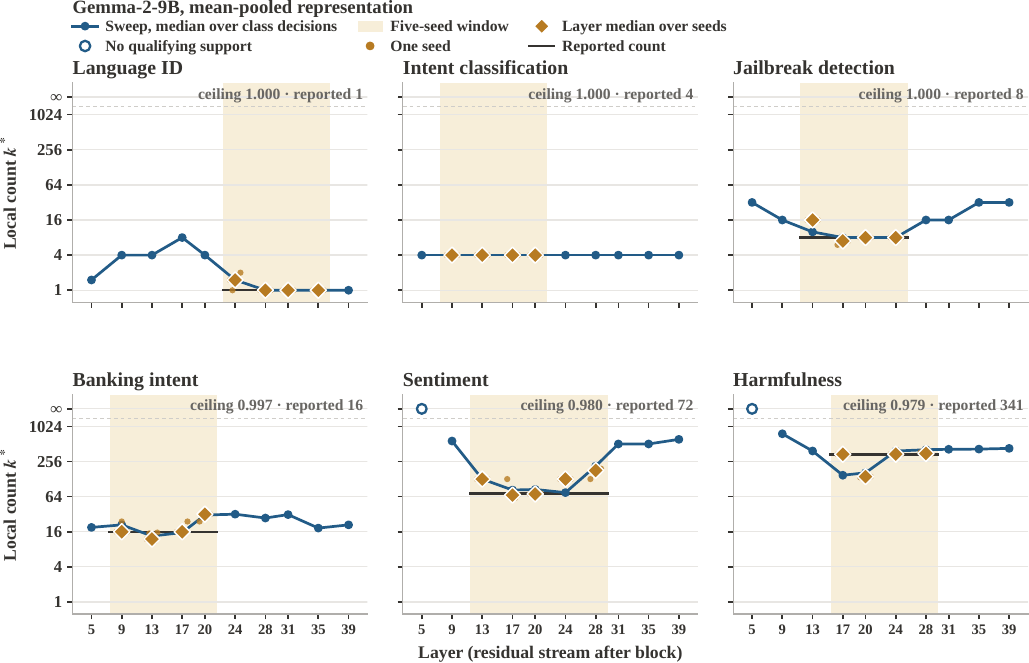}
\caption{\textbf{Local counts across depth on Gemma-2-9B and the five-seed window inside them.}
Each panel shows one task in the mean-pooled representation.
The blue line traces the preliminary single-seed sweep on the full training split over the ten grid layers, giving at each layer the median local count over class decisions under the sweep's ceiling for each decision.
An open circle at the $\infty$ tick marks a layer where no candidate qualified, which happens at layer 5 for sentiment and for harmfulness detection.
The shaded band covers the layers of the five-seed measurement, which are 24 to 35 for language identification, 9 to 20 for intent classification and banking intent, 13 to 24 for jailbreak detection, 17 to 28 for harmfulness detection, and the five-layer follow-up 13 to 28 for sentiment.
Inside the band, small gold dots give the median local count over class decisions for each seed under the common ceiling of the window, gold diamonds give the median of the five seeds at each layer, and the black line gives the reported count after neighbor confirmation and the medians over class decisions and seeds.
The reported counts read 1, 4, 8, 16, 72, and 341, and the median dense ceilings over class decisions read 1.000, 1.000, 1.000, 0.997, 0.980, and 0.979, in panel order.
The five seeds spread by at most one count at the window layers of language identification, intent classification, and jailbreak detection, and they spread from 61 to 128 at layer 17 for sentiment.
Neighbor confirmation keeps single-layer minima out of the reported count, as at layer 17 for jailbreak detection, where the seed median of 7 sits below the reported 8, and at layer 20 for harmfulness detection, where the seed median of 141 sits below the reported 341.
The sweep and the five-seed medians differ most for harmfulness detection at window layers 17 and 20, where the single-seed sweep reads 149 and 164 and the five-seed medians read 341 and 141.}
\label{fig:app-layers}
\end{figure}

\textbf{Layer window.}
The five-seed measurement probes a window of four consecutive grid layers per task--model cell.
The window contains the sweep's best layer in the mean-pooled representation and brackets it, so that the best layer has a grid neighbor on each side wherever the grid allows.
Four layers give Equation~\ref{eq:nc} three adjacent pairs.
Figure~\ref{fig:app-layers} shows, for six tasks on Gemma-2-9B, how the sweep profile, the window, and the five seed values relate.
We hold the window size fixed at four for every cell, because the minimum over more layer pairs tends to be smaller, and cells with different layer counts would not be comparable.
The four-layer window covers 240 of the 266 core task--model cells.
The remaining 26 cells come from follow-up measurements under the same subsample, representation, and number of seeds, on three to five grid layers around the located minimum, which are adjacent grid entries in 18 of these cells and more widely spaced grid entries in five, and in three cells on the full grid.
These cells are sentiment on Gemma-2-9B, Gemma-3-4B, Llama-3.1-8B, Mistral-7B, and OLMo-2-7B in both base and instruction-tuned versions, harmfulness detection on Llama-3.1-8B, Qwen3-8B, Gemma-2-9B-IT, Gemma-3-270M-IT, Llama-3.1-8B-Instruct, and both versions of Mistral-7B and OLMo-2-7B, mathematical-step correctness on both versions of Gemma-2-9B, Gemma-3-4B, and Gemma-3-12B, and CLINC150 on Gemma-2-9B-IT.

\textbf{Seeds and subsamples.}
The five-seed measurement uses seeds $R=\{1,\dots,5\}$ in every cell except harmfulness detection on OLMo-2-7B, which uses seeds 0 to 4.
Each seed draws a stratified subsample of the training split that keeps 85\% of each class, rounded down, and it also seeds the validation portion of Section~\ref{app:probes}.
The test split stays fixed across seeds, so seed variation reflects the training data and the fit, not the evaluation set.
Every probe in Algorithm~\ref{alg:search}, including the dense probe that supplies the ceiling, is fitted on the subsample of its seed, and only the mean-pooled representation of Equation~\ref{eq:pool} enters this second stage.

\textbf{Common ceiling.}
The ceiling $C_c$ of Equation~\ref{eq:threshold} is the largest dense-probe AUROC across the window layers and all five seeds, and the threshold $\tau_c$ derived from it applies to every layer and seed of the window.
This common reference prevents a layer or seed with a weak dense probe from receiving an easier threshold.
As a consequence, a layer whose own dense AUROC lies below $\tau_c$ reads $\infty$, which the aggregation treats as a legitimate value.
By Lemma~\ref{lem:basic}(ii), the common ceiling yields counts at least as large as a layer-specific ceiling would, since it can only raise each threshold.

\textbf{Neighbor confirmation.}
For each class decision and seed, we form the larger of the two local counts in every pair of consecutive window layers and keep the smallest of these pairwise maxima,
\begin{equation}
    \tilde k_{c,r}=\min_{1\le j<m}\ \max\bigl(k^*_{c,\ell_j,r},\,k^*_{c,\ell_{j+1},r}\bigr),
    \qquad \ell_1<\dots<\ell_m,
    \label{eq:nc}
\end{equation}
where $\ell_1,\dots,\ell_m$ are the window layers in order.
A size therefore counts only when two consecutive layers both reach it, which removes minima that appear at a single isolated layer.
When only one layer is available, $\tilde k_{c,r}$ is that layer's local count.
For example, local counts of $32$, $8$, $16$, and $\infty$ on four window layers give pairwise maxima of $32$, $16$, and $\infty$, so $\tilde k_{c,r}=16$ and the isolated count of $8$ does not enter the result.
Consecutive entries of the probed grid are often several blocks apart, so neighbor confirmation asks for stability across a span of depth rather than across adjacent blocks.

\textbf{Medians over extended values.}
We use one median convention throughout.
For values $u_1\le\dots\le u_N$ in $\overline{\mathbb{N}}$,
\begin{equation}
    \operatorname{med}(u_1,\dots,u_N)=
    \begin{cases}
        u_{(N+1)/2}, & N\ \text{odd},\\[2pt]
        \tfrac12\bigl(u_{N/2}+u_{N/2+1}\bigr), & N\ \text{even},
    \end{cases}
    \qquad \tfrac12(u+\infty)=\infty.
    \label{eq:median}
\end{equation}
A median over an odd number of values is therefore infinite exactly when more than half of them are infinite, and a median over an even number is infinite exactly when at least half are.
An even number of finite values can produce a non-integer median, which we round for display.

\textbf{Task-level estimator.}
Within each seed, the task value is the median of the confirmed counts across class decisions, and the reported cell is the median of these task values across seeds,
\begin{equation}
    K_r=\operatorname{med}_{c\in\mathcal{C}}\ \tilde k_{c,r},
    \qquad
    k^*=\operatorname{med}_{r\in R}\ K_r,
    \qquad
    \bigl[\min_{r}K_r,\ \max_{r}K_r\bigr],
    \qquad
    C=\operatorname{med}_{c\in\mathcal{C}}\ C_c.
    \label{eq:task-level}
\end{equation}
The third expression is the seed range, and the fourth is the reported task ceiling.
The median over class decisions makes a multiclass task's count describe a typical class rather than its hardest class, and per-class counts appear where an analysis needs them.

\textbf{Properties.}
The following lemma collects the elementary properties of the local, confirmed, and task-level counts that the paper relies on.
Each holds for the recorded curves of one class decision.

\begin{applemma}[basic properties of the count]
\label{lem:basic}
Let the fitted candidates be fixed.
\begin{enumerate}
\item[(i)] Let $k_{\min}(\tau_c)$ be the smallest $|S|$ over subsets $S\subseteq[d]$ for which some linear classifier supported on $S$ reaches AUROC $\tau_c$ on $\mathcal{D}_{\mathrm{te}}$. Then $k^*_{c,\ell,r}\ge k_{\min}(\tau_c)$.
\item[(ii)] Assume $C_c>0.5$, which holds for every measured class decision in the task matrix since the smallest ceiling is 0.62. Let $k^*(\alpha)$ denote the local count when the factor $0.95$ in Equation~\ref{eq:threshold} is replaced by $\alpha\in(0,1]$. Then $k^*(\alpha)$ is non-decreasing in $\alpha$.
\item[(iii)] Let $k_{(1)}\le\dots\le k_{(m)}$ be the ordered local counts of one class decision and seed over $m\ge2$ window layers. Then the confirmed count satisfies $k_{(2)}\le\tilde k_{c,r}\le k_{(m)}$.
\item[(iv)] The map from the array of local counts $(k^*_{c,\ell,r})_{c,\ell,r}$ to the task-level $k^*$ of Equation~\ref{eq:task-level} is non-decreasing in every coordinate. Consequently, raising the factor $\alpha$ cannot lower the confirmed count, the per-seed task value, or the task-level $k^*$.
\end{enumerate}
\end{applemma}

\begin{appproof}
For (i), if $k^*_{c,\ell,r}=\infty$ the claim is immediate.
Otherwise a candidate $S^*$ attains the minimum in Equation~\ref{eq:layer-reading} together with a linear classifier supported on $S^*$, namely the refitted prefix probe or the L1 fit, whose test AUROC reaches $\tau_c$.
This classifier belongs to the set over which $k_{\min}(\tau_c)$ minimizes, so $|S^*|\ge k_{\min}(\tau_c)$.
For (ii), the threshold $0.5+\alpha(C_c-0.5)$ increases with $\alpha$ by the assumption $C_c>0.5$.
A higher threshold shrinks the set of qualifying candidates in Equation~\ref{eq:layer-reading}, and the minimum over a smaller set cannot decrease.
For (iii), every pairwise maximum in Equation~\ref{eq:nc} is at least as large as both of its arguments, and two distinct layers contribute these arguments, so every pairwise maximum is at least $k_{(2)}$, and the minimum over pairs inherits this bound.
The upper bound holds because every pairwise maximum is at most $k_{(m)}$.
For (iv), the functions $\max$, $\min$, and $\operatorname{med}$ on $\overline{\mathbb{N}}$ are each non-decreasing in every argument, since increasing one argument cannot move any order statistic down and the average in Equation~\ref{eq:median} is non-decreasing in both terms.
Equations~\ref{eq:nc} and~\ref{eq:task-level} compose these functions, and a composition of coordinatewise non-decreasing maps is coordinatewise non-decreasing.
The consequence follows from (ii).
\end{appproof}

Three readings of the lemma matter for the paper.
By (i), a finite count certifies that a readout of that size exists and never certifies that no smaller readout exists.
By (ii) and (iv), the threshold variants of Table~\ref{tab:estimator} move every count in a known direction, and those variants test whether the task order survives the move, which the lemma does not decide.
By (iii), the confirmed count never falls below the second-smallest local count, so a single anomalous layer cannot determine it.

\begin{algorithm}[t]
\AppAlgorithmStyle
\caption{Task-level $k^*$ with seed range from the five-seed layer window}
\label{alg:aggregate}
\begin{algorithmic}[1]
\Require window layers $\ell_1<\dots<\ell_m$, seeds $R=\{1,\dots,5\}$, class decisions $\mathcal{C}$, dense AUROCs $A^{(r)}_{c,\ell}([d])$ and recorded prefix and L1 curves for every $(c,\ell,r)$
\Ensure task $k^*$ with seed range, task ceiling
\For{$c\in\mathcal{C}$}
    \State $C_c\gets\max_{\ell,r}A^{(r)}_{c,\ell}([d])$ and $\tau_c\gets0.5+0.95\,(C_c-0.5)$ \Comment{Equation~\ref{eq:threshold}}
    \For{$r\in R$}
        \For{$j=1,\dots,m$}
            \State $k^*_{c,\ell_j,r}\gets$ Algorithm~\ref{alg:search} applied to the recorded curves of $(c,\ell_j,r)$ with threshold $\tau_c$
        \EndFor
        \State $\tilde k_{c,r}\gets\min_{j<m}\max(k^*_{c,\ell_j,r},k^*_{c,\ell_{j+1},r})$ \Comment{Equation~\ref{eq:nc}}
    \EndFor
\EndFor
\For{$r\in R$}
    \State $K_r\gets\operatorname{med}_{c\in\mathcal{C}}\tilde k_{c,r}$ \Comment{Equation~\ref{eq:median}, $\infty$ above every finite value}
\EndFor
\State \Return $\operatorname{med}_{r}K_r$, $[\min_r K_r,\max_r K_r]$, $\operatorname{med}_{c}C_c$
\end{algorithmic}
\end{algorithm}

\textbf{Design rationale.}
Each aggregation step rules out one way of obtaining a small count by chance.
The common ceiling prevents a layer or seed with a weak dense probe from receiving an easier threshold.
Neighbor confirmation prevents a single fortunate layer from setting the count, which Lemma~\ref{lem:basic}(iii) makes exact.
The fixed window of four layers prevents cells with more layers from receiving smaller counts through a minimum over more pairs.
The fixed test split across seeds isolates training-data variation in the seed range, and the medians over class decisions and seeds prevent one extreme class or one extreme subsample from setting the task value.

\textbf{Readability flag.}
We report counts for all task--model pairs and flag task ceilings at or below 0.90 in every table.
Analyses that require reliable readability restrict to pairs above this value, as stated where used.
The flag is a reporting convention and does not alter any count.
A low ceiling makes the threshold easy to reach in absolute terms, so a small count at a low ceiling can describe a weak readout, which is why we do not interpret such counts as task sizes.

\textbf{Takeaway for practitioners.}
A count at a single layer or with a layer-specific ceiling can be made small by the choice of layer alone.
Reporting counts with a common ceiling, neighbor confirmation, and a fixed window size makes counts comparable across models, and the best-single-layer variant of Table~\ref{tab:estimator} shows how much the ordering moves when these safeguards are dropped.

\subsubsection{Worked Example on One Cell}
\label{app:worked}

\textbf{Question.}
This section traces the procedure of Sections~\ref{app:search} and~\ref{app:aggregation} through one task--model cell with its actual numbers, so that a reader can check every step of Algorithm~\ref{alg:aggregate} against Table~\ref{tab:worked}.
The cell is harmfulness detection on Gemma-2-9B, which has one class decision, a four-layer window, and a reported count of 341 with seed range 332 to 344.

\textbf{Data.}
The window layers are 17, 20, 24, and 28, and the seeds are 1 to 5.
The fixed test split holds 2000 prompts, of which 1039 are harmful.
Each of the 20 layer and seed combinations has one dense probe, eleven prefix probes, and ten L1 fits, as in Algorithm~\ref{alg:search}.

\begin{table}[t]
\centering\AppTableStyle
\caption{\textbf{Every local count behind one reported cell.}
Harmfulness detection on Gemma-2-9B.
The left block gives the dense-probe test AUROC at each window layer and seed, and the right block gives the local count of Equation~\ref{eq:layer-reading} at the common threshold.
The last row applies neighbor confirmation of Equation~\ref{eq:nc} within each seed.
Every local count in this cell comes from the L1 path.}
\label{tab:worked}
\begin{tabular}{@{}lrrrrrrrrrr@{}}
\toprule
\rowcolor{AppGold} & \multicolumn{5}{c}{\AppHead{Dense AUROC by seed}} & \multicolumn{5}{c}{\AppHead{Local count by seed}} \\
\rowcolor{AppGold} \AppHead{Layer} & \AppHead{1} & \AppHead{2} & \AppHead{3} & \AppHead{4} & \AppHead{5} & \AppHead{1} & \AppHead{2} & \AppHead{3} & \AppHead{4} & \AppHead{5} \\
\midrule
17 & 0.9748 & 0.9751 & 0.9763 & 0.9759 & 0.9750 & 341 & 333 & 361 & 351 & 339 \\
20 & 0.9787 & 0.9779 & 0.9782 & 0.9790 & 0.9788 & 137 & 137 & 141 & 142 & 141 \\
24 & 0.9768 & 0.9756 & 0.9776 & 0.9769 & 0.9772 & 342 & 332 & 342 & 344 & 340 \\
28 & 0.9719 & 0.9709 & 0.9720 & 0.9715 & 0.9724 & 369 & 369 & 341 & 352 & 347 \\
\midrule
Confirmed & & & & & & 341 & 332 & 342 & 344 & 339 \\
\bottomrule
\end{tabular}
\end{table}

\textbf{Ceiling and threshold.}
The largest dense AUROC over the 20 combinations is $C_c=0.9790$, reached at layer 20 with seed 4, so Equation~\ref{eq:threshold} gives $\tau_c=0.5+0.95\times0.4790=0.9551$.
The same threshold applies to all 20 combinations, including layer 28, whose own dense AUROCs of 0.9709 to 0.9724 lie 0.007 to 0.008 below the ceiling.
The gap between the ceiling and the threshold is 0.024 AUROC, and Section~\ref{app:finite-sample} shows that the test-split standard error of this ceiling is far smaller than that gap.

\textbf{Local counts.}
At layer 20 the L1 path reaches $\tau_c$ with 137 to 142 features across the five seeds, and at layers 17, 24, and 28 it needs 332 to 369.
The ranked-prefix path never qualifies with fewer features than the L1 path at any layer or seed of this cell, so every local count in Table~\ref{tab:worked} is an L1 support size, which is why the counts are not powers of two.

\textbf{Neighbor confirmation.}
For seed 1, the three pairwise maxima of Equation~\ref{eq:nc} are 341, 342, and 369, namely $\max(341,137)$, $\max(137,342)$, and $\max(342,369)$, and their minimum is 341.
The other seeds give 332, 342, 344, and 339.
The task-level count is the median of these five confirmed counts, 341, with seed range 332 to 344, which matches the reported cell.

\textbf{What the example shows.}
Layer 20 reads the task with fewer than half the features of its neighbors in every seed, so its smaller count is a stable property of that layer and not seed noise.
Neighbor confirmation nevertheless reports 341, because the protocol counts a size only when two adjacent grid layers both reach it, and no neighbor of layer 20 reaches 142.
The best-single-layer variant of Table~\ref{tab:estimator} would report 141 for this cell.
The two readings answer different questions.
The confirmed count describes the size that the task needs across a span of depth, and the single-layer count describes the narrowest layer.
Lemma~\ref{lem:basic}(iii) states the general form of this behavior, since the confirmed count can never fall below the second-smallest local count.
Across the 19 models, the two readings order the 14 core tasks alike at a median Spearman correlation of 0.98 in Table~\ref{tab:estimator}, so the choice between them moves absolute counts and not the task order.

\subsubsection{Finite-Sample Behavior of the Count}
\label{app:finite-sample}

\textbf{Question.}
The count reads test AUROCs twice, once through the ceiling and once through the comparison with the threshold.
A reader can therefore ask how far the reported count can move away from the count that the same candidates would receive on the population of texts from which the test split is drawn.
This section answers the question in two steps.
A distribution-free bound states the worst case, and a data-dependent standard error measures the actual precision of every ceiling in the task matrix.

\textbf{Assumption A (sampling of the test split).}
Conditional on the training data, the positive test texts $\mathcal{P}_c$ are independent draws from one class-conditional distribution, the negative test texts $\mathcal{N}_c$ are independent draws from another, and the two samples are independent of each other.

Assumption~A holds by construction for the native test splits, and it holds approximately for the grouped hold-outs of Section~\ref{app:tasks}, which sample problems or templates rather than single texts.
Every candidate of Equation~\ref{eq:family}, every dense probe, and every decision function depend on the training data only, so conditional on the training data the scorers are fixed functions.
Let $\mathcal{G}_c$ collect all scorers evaluated for class decision $c$, namely the dense probes, the prefix probes, and the L1 fits over all window layers and seeds, and let $M=\lvert\mathcal{G}_c\rvert$.
For a scorer $g$, the population AUROC is
\begin{equation}
    \bar A(g)=\Pr\bigl[g(X_1)>g(X_0)\bigr]+\tfrac12\Pr\bigl[g(X_1)=g(X_0)\bigr],
    \label{eq:pop-auroc}
\end{equation}
where $X_1$ and $X_0$ are independent draws from the positive and negative class-conditional distributions, and Equation~\ref{eq:auroc} gives its empirical counterpart $\hat A(g)$ on the test split.

\begin{applemma}[uniform deviation of test AUROCs]
\label{lem:uniform}
Under Assumption~A, for every $\delta\in(0,1)$, with probability at least $1-\delta$ over the draw of the test split of class decision $c$,
\begin{equation}
    \max_{g\in\mathcal{G}_c}\bigl\lvert\hat A(g)-\bar A(g)\bigr\rvert\le\varepsilon_\delta,
    \qquad
    \varepsilon_\delta=\sqrt{\frac12\Bigl(\frac{1}{n^{\mathrm{te}}_1}+\frac{1}{n^{\mathrm{te}}_0}\Bigr)\log\frac{2M}{\delta}}.
    \label{eq:eps}
\end{equation}
\end{applemma}

\begin{appproof}
Fix $g$.
The statistic $\hat A(g)$ is a function of $n_1^{\mathrm{te}}+n_0^{\mathrm{te}}$ independent texts, and its expectation is $\bar A(g)$ because each summand of Equation~\ref{eq:auroc} has expectation $\bar A(g)$.
Replacing one positive text changes at most $n_0^{\mathrm{te}}$ of the $n_1^{\mathrm{te}}n_0^{\mathrm{te}}$ summands, each by at most $1$, so it changes $\hat A(g)$ by at most $1/n_1^{\mathrm{te}}$.
Replacing one negative text likewise changes $\hat A(g)$ by at most $1/n_0^{\mathrm{te}}$.
The bounded-differences inequality~\citep{mcdiarmid1989method} then gives, for every $\varepsilon>0$,
\begin{equation*}
    \Pr\bigl[\lvert\hat A(g)-\bar A(g)\rvert\ge\varepsilon\bigr]\le2\exp\Bigl(-\frac{2\varepsilon^2}{n_1^{\mathrm{te}}(1/n_1^{\mathrm{te}})^2+n_0^{\mathrm{te}}(1/n_0^{\mathrm{te}})^2}\Bigr)=2\exp\Bigl(-\frac{2\varepsilon^2}{1/n_1^{\mathrm{te}}+1/n_0^{\mathrm{te}}}\Bigr).
\end{equation*}
A union bound over the $M$ scorers and the choice $\varepsilon=\varepsilon_\delta$ bound the probability of the complementary event by $\delta$.
\end{appproof}

\begin{appproposition}[the count is sandwiched by population counts]
\label{prop:sandwich}
Let $\varepsilon>0$ and suppose that $\lvert\hat A(g)-\bar A(g)\rvert\le\varepsilon$ for every scorer $g\in\mathcal{G}_c$.
For a threshold $t$, let $\bar k(t)$ denote the local count of Equation~\ref{eq:layer-reading} computed with population AUROCs $\bar A$ in place of $\hat A$ and with the fixed threshold $t$, and let $\bar\tau_c=0.5+0.95\,(\bar C_c-0.5)$, where $\bar C_c$ is the largest population AUROC among the dense probes.
Then every local count of class decision $c$ at every window layer and seed satisfies
\begin{equation}
    \bar k\bigl(\bar\tau_c+1.95\,\varepsilon\bigr)\ \ge\ k^*_{c,\ell,r}\ \ge\ \bar k\bigl(\bar\tau_c-1.95\,\varepsilon\bigr),
    \label{eq:sandwich}
\end{equation}
and by Lemma~\ref{lem:basic}(iv) the same sandwich holds for the confirmed count and, when the premise holds for every $c\in\mathcal{C}$, for the per-seed task value and the task-level $k^*$.
\end{appproposition}

\begin{appproof}
The maximum is $1$-Lipschitz in the sup norm, so the empirical ceiling satisfies $\lvert C_c-\bar C_c\rvert\le\varepsilon$ and hence $\lvert\tau_c-\bar\tau_c\rvert\le0.95\,\varepsilon$.
If a candidate satisfies $\bar A(S)\ge\bar\tau_c+1.95\,\varepsilon$, then $\hat A(S)\ge\bar A(S)-\varepsilon\ge\bar\tau_c+0.95\,\varepsilon\ge\tau_c$, so it qualifies empirically and the empirical minimum is at most its size.
This gives the left inequality.
If a candidate qualifies empirically, then $\bar A(S)\ge\hat A(S)-\varepsilon\ge\tau_c-\varepsilon\ge\bar\tau_c-1.95\,\varepsilon$, so every empirically qualifying candidate also qualifies at the lower population threshold.
This gives the right inequality.
The extension to aggregated counts applies Lemma~\ref{lem:basic}(iv) to the two bounding arrays.
\end{appproof}

\textbf{What the worst case says.}
Lemma~\ref{lem:uniform} supplies a radius $\varepsilon_\delta$ that satisfies the premise of Proposition~\ref{prop:sandwich} with probability $1-\delta$, and Proposition~\ref{prop:sandwich} then places the reported count between the population counts at two thresholds that differ from $\bar\tau_c$ by $1.95\,\varepsilon_\delta$.
The relevant comparison is with the gap $0.05\,(C_c-0.5)$ between the ceiling and the threshold, which is at most 0.025 AUROC.
At $\delta=0.05$ the radius is 0.08 for the largest and most balanced test splits and 0.36 at the median class decision of the task matrix, where the rarer test class holds 40 texts and $M=440$ scorers enter the union bound.
The worst-case radius therefore exceeds the gap in every class decision, so the distribution-free bound establishes the form of the dependence but not a tight range for any individual count.
Two features of Equation~\ref{eq:eps} carry over to the tighter analysis below.
The radius is set by the rarer test class, and the number of scorers enters only through $\log M$.

\textbf{Data-dependent precision.}
The worst case is loose for scorers whose AUROC lies near one, where the variance of Equation~\ref{eq:auroc} is small.
We therefore measure the precision of every ceiling in the task matrix with the standard error of Hanley and McNeil~\citep{hanley1982meaning}, which approximates the sampling variance of an AUROC $A$ from the class counts alone,
\begin{equation}
    \widehat{\mathrm{se}}(A)=\sqrt{\frac{A(1-A)+(n_1^{\mathrm{te}}-1)(Q_1-A^2)+(n_0^{\mathrm{te}}-1)(Q_2-A^2)}{n_1^{\mathrm{te}}\,n_0^{\mathrm{te}}}},
    \label{eq:hanley-mcneil}
\end{equation}
with $Q_1=A/(2-A)$ and $Q_2=2A^2/(1+A)$.
We evaluate Equation~\ref{eq:hanley-mcneil} at the ceiling $C_c$ of every class decision of the 14 core tasks on the 19 models, which gives 1025 class decisions, and we compare $1.96\,\widehat{\mathrm{se}}(C_c)$ with the gap $0.05\,(C_c-0.5)$.
If the gap exceeded this radius, a shift of the ceiling by its plausible sampling error would move the threshold by less than the distance a candidate has to cover, and the count would be insensitive to which test texts were drawn.

\begin{table}[t]
\centering\AppTableStyle
\caption{\textbf{Test-split precision of the 1025 class-decision ceilings.}
The standard error is Equation~\ref{eq:hanley-mcneil} at the ceiling with the test class counts of the class decision, the worst-case radius is Equation~\ref{eq:eps} at $\delta=0.05$, and the gap is $0.05\,(C_c-0.5)$.
Readable class decisions are those with $C_c>0.90$.}
\label{tab:ceiling-precision}
\begin{tabular}{@{}lrrr@{}}
\toprule
\rowcolor{AppGold} \AppHead{Quantity} & \AppHead{Median} & \AppHead{90th percentile} & \AppHead{Maximum} \\
\midrule
Standard error $\widehat{\mathrm{se}}(C_c)$, all 1025 decisions & 0.0015 & 0.0090 & 0.0208 \\
Standard error $\widehat{\mathrm{se}}(C_c)$, 941 readable decisions & 0.0011 & 0.0071 & \\
Worst-case radius $\varepsilon_{0.05}$ & 0.36 & & 0.41 \\
Gap $0.05\,(C_c-0.5)$ & 0.025 & & 0.025 \\
\midrule
Decisions with $1.96\,\widehat{\mathrm{se}}(C_c)$ below the gap, all & \multicolumn{3}{r}{92.1\%} \\
Decisions with $1.96\,\widehat{\mathrm{se}}(C_c)$ below the gap, readable & \multicolumn{3}{r}{99.4\%} \\
Rarer test class, minimum and median size & \multicolumn{3}{r}{30 and 40 texts} \\
\bottomrule
\end{tabular}
\end{table}

\textbf{Result.}
We make two observations on Table~\ref{tab:ceiling-precision}.
First, the median standard error of a ceiling is 0.0015 AUROC, one sixteenth of the median gap of 0.025, and $1.96\,\widehat{\mathrm{se}}(C_c)$ lies below the gap in 99.4\% of the 941 readable class decisions.
Second, the exceptions concentrate where the ceiling is low, since a ceiling of 0.62 has both a small gap of 0.006 and a large standard error, and such cells already carry the readability flag of Section~\ref{app:aggregation}.
The worked example of Section~\ref{app:worked} illustrates the typical readable case, with a ceiling of 0.9790 on a test split of 1039 positives and 961 negatives, a standard error of 0.0032 AUROC, and a gap of 0.024 AUROC.

\textbf{Scope.}
Equation~\ref{eq:hanley-mcneil} is an approximation that treats the decision values as continuous and uses only the class counts, and it describes the ceiling and not the candidate curve, whose points share the same test texts.
Both summaries condition on the training data and therefore describe test-split variation only, while the seed range of Equation~\ref{eq:task-level} describes training-data variation.
The two summaries answer different questions, and we report them separately.

\textbf{Takeaway for practitioners.}
The precision of a sparse-probe count is set by the rarer class of the test split, because both Equation~\ref{eq:eps} and Equation~\ref{eq:hanley-mcneil} scale with $1/n_1^{\mathrm{te}}$ and $1/n_0^{\mathrm{te}}$.
A relative threshold at 95\% of the above-chance ceiling leaves a gap of at most 0.025 AUROC, so a study should check that the standard error of its ceilings is small against that gap, and the check fails first for one-vs-rest decisions with few test positives and for tasks with low ceilings.

\subsubsection{Models and Probed Layers}
\label{app:models}

Table~\ref{tab:models} lists the ten base models, their instruction-tuned counterparts, and the layer grid of the preliminary sweep, from which the five-seed window of Section~\ref{app:aggregation} is drawn.
The Gemma-3 grids combine the layers with published Gemma Scope 2 SAEs and an even sweep over depth, giving 10 to 12 layers from block 1 to the last block.
The Gemma-2-9B grid takes every fourth block from 5 to 39 together with the two Gemma Scope layers 20 and 31, and the Llama, Mistral, OLMo, and Qwen grids take every fourth block from 3.
Each base model and its counterpart share the same grid, so the comparisons of Section~\ref{sec:training} draw their windows from the same layer set.
Every model provides three representations at each probed layer, namely the mean-pooled residual, the last-position residual, and the SAE encoding of the last-position residual where a suite exists.
The main protocol reads the mean-pooled residual, and Section~\ref{app:sae} describes the use of the SAE encoding.

\begin{table}[t]
\AppTableStyle
    \centering
    \caption{\textbf{Models and probed layers.}
    Blocks is the number of transformer blocks and $d$ the residual width.
    Layer $\ell$ denotes the residual stream after block $\ell$, with blocks indexed from 0, and the listed layers form the grid of the preliminary sweep.
    IT marks base models whose instruction-tuned counterpart is also in the study and probed on the same layers.
    The SAE suite and Features columns name the suite used for the SAE representation and give its dictionary size.}
    \label{tab:models}
    \resizebox{\linewidth}{!}{%
    \begin{tabular}{lrrrllr}
        \toprule
\rowcolor{AppGold}
        \AppHead{Base model} & \AppHead{Blocks} & \AppHead{$d$} & \multicolumn{1}{c}{\AppHead{Probed layers}} & \AppHead{IT} & \AppHead{SAE suite} & \AppHead{Features} \\
        \midrule
        Gemma-3-270M & 18 & 640 & 1, 3, 5, 6, 8, 9, 10, 12, 15, 17 & yes & Gemma Scope 2 & 16384 \\
        Gemma-3-1B & 26 & 1152 & 1, 4, 7, 8, 11, 13, 15, 17, 18, 22, 25 & yes & Gemma Scope 2 & 16384 \\
        Gemma-3-4B & 34 & 2560 & 1, 6, 9, 10, 15, 17, 19, 22, 24, 28, 29, 33 & yes & Gemma Scope 2 & 16384 \\
        Gemma-2-9B & 42 & 3584 & 5, 9, 13, 17, 20, 24, 28, 31, 35, 39 & yes & Gemma Scope & 16384 \\
        Gemma-3-12B & 48 & 3840 & 1, 8, 12, 14, 21, 24, 27, 31, 34, 40, 41, 47 & yes & Gemma Scope 2 & 16384 \\
        Gemma-3-27B & 62 & 5376 & 1, 10, 16, 18, 27, 31, 35, 40, 44, 52, 53, 61 & yes & Gemma Scope 2 & 16384 \\
        Llama-3.1-8B & 32 & 4096 & 3, 7, 11, 15, 19, 23, 27, 31 & yes & Llama Scope & 32768 \\
        Mistral-7B-v0.3 & 32 & 4096 & 3, 7, 11, 15, 19, 23, 27, 31 & yes & none & \\
        OLMo-2-7B & 32 & 4096 & 3, 7, 11, 15, 19, 23, 27, 31 & yes & none & \\
        Qwen3-8B & 36 & 4096 & 3, 7, 11, 15, 19, 23, 27, 31, 35 & no & Qwen-Scope & 65536 \\
        \bottomrule
    \end{tabular}}
\end{table}

The base models are Gemma-3 at 270M, 1B, 4B, 12B, and 27B parameters~\citep{team2025gemma}, Gemma-2-9B~\citep{team2024gemma}, Llama-3.1-8B~\citep{grattafiori2024llama}, Mistral-7B-v0.3~\citep{jiang2023mistral7b}, OLMo-2-1124-7B~\citep{olmo20242}, and Qwen3-8B-Base~\citep{yang2025qwen3}.
The instruction-tuned counterparts are the corresponding releases of each lab, and Qwen3-8B has no counterpart in the study.
The 14 core tasks run on all 19 models, and the ten additional tasks of Table~\ref{tab:tasks} run on Gemma-2-9B only, where they serve the label-budget and input-baseline analyses.
The additional tasks therefore do not form a complete 24-task matrix across models.

\subsubsection{SAE Feature Protocol}
\label{app:sae}

\textbf{Question.}
The correspondence analysis of Section~\ref{sec:correspondence} compares individual features across models, which requires a basis whose coordinates are meant to be individually interpretable.
This section specifies the dictionaries and the protocol that selects SAE supports.

\textbf{Encoders.}
Each suite provides one SAE per layer that maps a residual vector $h\in\mathbb{R}^d$ to a sparse feature vector $g\in\mathbb{R}^{d_{\mathrm{sae}}}$ with $d_{\mathrm{sae}}\gg d$.
Gemma Scope~\citep{lieberum2024gemma}, Gemma Scope 2~\citep{mcdougall2025gemmascope2}, and Llama Scope~\citep{he2024llama} use JumpReLU encoders~\citep{rajamanoharan2024jumping}, which keep a pre-activation only above its threshold,
\begin{equation}
    u = W_{\mathrm{enc}}h+b_{\mathrm{enc}},
    \qquad
    g_j = u_j\,\mathbf{1}[u_j>\theta_j],
    \label{eq:jumprelu}
\end{equation}
with $W_{\mathrm{enc}}\in\mathbb{R}^{d_{\mathrm{sae}}\times d}$ and $b_{\mathrm{enc}}\in\mathbb{R}^{d_{\mathrm{sae}}}$, a threshold $\theta_j$ per feature for the two Gemma suites, and a shared scalar threshold for Llama Scope.
Llama Scope stores a dataset-level input normalization and unnormalized decoder columns, and we fold both into $W_{\mathrm{enc}}$ and $b_{\mathrm{enc}}$ so that Equation~\ref{eq:jumprelu} applies to raw activations.
The folded encoder matches the suite's reference implementation to within $10^{-6}$.
Qwen-Scope~\citep{deng2026qwen} uses a TopK encoder~\citep{gao2025scaling},
\begin{equation}
    g=\operatorname{TopK}_{50}\bigl(\operatorname{ReLU}(W_{\mathrm{enc}}h+b_{\mathrm{enc}})\bigr),
    \label{eq:topk}
\end{equation}
which keeps the 50 largest pre-activations of each input and sets the rest to zero.

\textbf{Models.}
We use the 16384-feature Gemma Scope release whose average number of active features is closest to 70, the 16384-feature medium-sparsity tier of Gemma Scope 2, the 32768-feature Llama Scope release, and the 65536-feature Qwen-Scope release.
SAEs exist at every probed layer of the eight base models that have a suite.
For the instruction-tuned models used in the intervention experiments of Section~\ref{sec:behavior}, the Gemma-3 models use the instruction-tuned Gemma Scope 2 releases at every probed layer, and Gemma-2-9B-IT uses the instruction-tuned Gemma Scope releases at layers 9, 20, and 31.
Llama-3.1-8B-Instruct has no instruction-tuned suite of its own, so it uses the base-model Llama Scope SAEs at every probed layer.

\textbf{Procedure.}
We encode the last-position residual of every text and store the result as a sparse matrix.
Probes on SAE features follow Sections~\ref{app:probes} and~\ref{app:search} without change, except that standardization divides by the training standard deviation without centering, and the dense probe uses all $d_{\mathrm{sae}}$ features.
The prefix grid of Equation~\ref{eq:grid} stays at 1 to 1024, so a class decision whose SAE probe needs more than 1024 dictionary features reads $\infty$ unless the L1 path finds a smaller support.

\textbf{Supports for feature correspondence.}
The supports compared in Section~\ref{sec:correspondence} come from the preliminary sweep of Section~\ref{app:aggregation}, which fits every probe on the full training split with one seed over the full grid.
For each class decision, the ceiling is that sweep's ceiling across the three representations, all grid layers, and both search paths.
We choose one layer per task in the SAE representation by neighbor confirmation over class-decision medians, taking the pair of consecutive grid layers with the smallest larger median and the better layer of that pair.
At that layer, the support of each class decision is the smaller of the first qualifying ranked prefix and the smallest qualifying L1 support of Algorithm~\ref{alg:search}, returned as a set of dictionary indices.
Because the ceiling includes the residual representations, an SAE support qualifies only when the SAE probe reaches 95\% of the above-chance AUROC of the best probe in any representation.

\textbf{Scope.}
Class decisions without a qualifying SAE support do not enter the correspondence tests, and Section~\ref{sec:correspondence} reports the resulting coverage.
Counts in the SAE basis and in the residual basis are not interchangeable, and the paper compares counts only within one basis.
Proposition~\ref{prop:affine} holds in the SAE basis for pure rescalings, so SAE counts refer to the dictionary's own features in the same sense that residual counts refer to residual coordinates.

\textbf{Design rationale.}
The SAE ceiling includes the residual representations, so an SAE support must match the best readout available in any representation.
This choice prevents the correspondence analysis from comparing SAE features that read a task much worse than the residual stream does.

\subsubsection{Design Choices and Their Measured Effects}
\label{app:design}

\textbf{Question.}
Each setting in Table~\ref{tab:protocol-glance} is a choice, and a reader can reasonably ask whether a different choice would change the conclusions.
Table~\ref{tab:design-choices} states the reason for each choice and the control that measures its effect.
The controls separate two kinds of dependence.
Some choices move every count by construction, and the relevant question is whether the task order survives.
Other choices change what the measurement reads, and the relevant question is how far the counts move.

\textbf{Estimator of task order.}
For the estimator variants, the task value of each model is the median over seeds of the per-seed task values, and infinite values share the largest rank.
The agreement between a variant and the main protocol on one model is the Spearman correlation~\citep{spearman1987proof} of the task-level counts over the tasks that both protocols define,
\begin{equation}
    \rho=\frac{\sum_{t}\bigl(\mathrm{rk}_t-\overline{\mathrm{rk}}\bigr)\bigl(\mathrm{rk}'_t-\overline{\mathrm{rk}'}\bigr)}{\sqrt{\sum_{t}\bigl(\mathrm{rk}_t-\overline{\mathrm{rk}}\bigr)^2\sum_{t}\bigl(\mathrm{rk}'_t-\overline{\mathrm{rk}'}\bigr)^2}},
    \label{eq:spearman-estimator}
\end{equation}
where $\mathrm{rk}_t$ and $\mathrm{rk}'_t$ are the average ranks of task $t$ under the variant and the main protocol.
We skip a model when fewer than six tasks are defined under both protocols, and Table~\ref{tab:estimator} reports the median and range of $\rho$ over the retained models for each variant.

\textbf{Null.}
If the estimator choices determined the conclusions, we would expect the task order of the 14 core tasks to reshuffle under each variant and $\rho$ to fall well below $1$.

\textbf{Result.}
We make two observations.
First, the estimator choices move absolute counts but leave the task order nearly intact, which supports the cross-model comparisons of Section~\ref{sec:agreement}.
Second, the representation choices move absolute counts substantially, which is why the paper compares counts only within one basis, one input format, and one set of texts.

\begin{table}[t]
\centering\AppTableStyle
\caption{\textbf{Why each design choice was made, and what changing it does.}
Task order is the Spearman correlation of Equation~\ref{eq:spearman-estimator} between task-level $k^*$ under a variant and under the main protocol, as a median over the 19 models.
Appendix~\ref{app:protocol-controls} reports every control in full.}
\label{tab:design-choices}
\begin{tabular}{@{}p{0.17\linewidth}p{0.40\linewidth}p{0.36\linewidth}@{}}
\toprule
\rowcolor{AppGold} \AppHead{Choice} & \AppHead{Reason} & \AppHead{Measured effect of changing it} \\
\midrule
Relative threshold at 95\% & A fraction of the above-chance ceiling applies the same relative requirement to tasks with different ceilings & 90\% and 97.5\% keep the task order at 0.97 (Table~\ref{tab:estimator}) \\
Two search paths & The ranked path has one-octave resolution and scores features one at a time, and the L1 path selects them jointly & The ranked path alone keeps the order at 0.99 (Table~\ref{tab:estimator}) \\
Neighbor confirmation & A single layer can reach a small count by chance on a fixed test set & The best single layer keeps the order at 0.98 (Table~\ref{tab:estimator}), and Table~\ref{tab:worked} shows a cell where the two readings differ by a factor of 2.4 \\
Linear probe & $k^*$ measures linear readout, which the dense ceiling also measures & A one-hidden-layer MLP gives the same count on 5 of 7 tasks and one grid step apart on 2 (Table~\ref{tab:probefamily}) \\
Model coordinates & The residual stream is the basis the model computes in, and Proposition~\ref{prop:affine} fixes the invariances & Random rotations raise counts by factors of 1.6 and 2.0 on two models, and by at most 20 in one cell (Table~\ref{tab:rotation}) \\
Original wording & The pools are fixed across models, so wording is held constant & Paraphrased test inputs raise counts by 1.5 to 10.8 on six cells (Table~\ref{tab:paraphrase}) \\
Class-balanced loss & One-vs-rest decisions have positive rates near 1\% & Not varied, and held identical for every model and task \\
Shuffled labels & A chance reference for the search itself & The best AUROC at any $k$ stays at or below 0.577 on seven tasks of Gemma-2-9B (Table~\ref{tab:probefamily}) \\
\bottomrule
\end{tabular}
\end{table}

\textbf{Confounds and the elements that rule them out.}
The paragraphs headed Design rationale in this appendix organize the same design from the opposite direction, by naming each alternative explanation of a small or a large count together with the design element that removes it and the evidence that the removal works.
They cover data leakage and surface cues in the task pools, the common ceiling, neighbor confirmation, the fixed window, seed variation, test-split reuse, label noise, joint feature selection, units, basis, and the probe family of the linear readout.

\textbf{Takeaway for practitioners.}
The estimator choices move absolute counts but not the task order, so task rankings obtained with this protocol transfer across reasonable estimator settings.
Absolute counts move with the basis, the input format, and the wording of the inputs, so counts from two studies are comparable only when these three are the same.

\subsubsection{Selection Effects and Statistical Units}
\label{app:selection}

\textbf{Test-set use.}
Training data determine standardization, feature ranking, and the validation choice of regularization.
The reported count nevertheless reads the test curve to identify qualifying supports and layers, and the common ceiling is itself a maximum over test AUROCs.
The count is therefore a descriptive measurement on a fixed evaluation set, and Proposition~\ref{prop:sandwich} states how far it can lie from the corresponding population count for a given deviation of the test AUROCs.
Two selection effects act in opposite directions.
Taking the ceiling as a maximum over the 20 dense fits of a four-layer window raises the threshold, since $\mathbb{E}\max_g\hat A(g)\ge\max_g\bar A(g)$, which pushes counts up.
Taking the minimum over candidates and layer pairs favors a fortunate fit, since $\mathbb{E}\min_g\hat A(g)\le\min_g\bar A(g)$ for any fixed collection, which pushes counts down.
We do not claim that the two effects cancel.
Table~\ref{tab:ceiling-precision} shows that the sampling error of the ceiling is small against the threshold gap in 99.4\% of readable class decisions, neighbor confirmation requires two layers to agree, and the seed range reports how far the count moves when the training data change.
The test splits hold 472 to 3000 texts, and the shuffled-label control in Table~\ref{tab:design-choices} shows that the search does not produce high AUROCs from label noise.

\textbf{Statistical units.}
Table~\ref{tab:uncertainty-units} states the unit and the meaning of every uncertainty summary and missingness convention in the paper.
Seed ranges, bootstrap intervals, and held-out prediction errors summarize different sources of variation, and we never combine them.

\begin{table}[t]
\centering\AppTableStyle
\caption{\textbf{Uncertainty and missingness conventions.} Each quantity is interpreted at its actual unit of analysis, and the paper never combines quantities with different units.}
\label{tab:uncertainty-units}
\begin{tabular}{@{}p{0.25\linewidth}p{0.30\linewidth}p{0.37\linewidth}@{}}
\toprule
\rowcolor{AppGold} \AppHead{Reported quantity} & \AppHead{Unit} & \AppHead{Meaning} \\
\midrule
Count seed range & Training subsample and fit seed & Minimum and maximum on the same evaluation pool, not a confidence interval \\
Ceiling standard error & Test texts of one class decision & Equation~\ref{eq:hanley-mcneil} at the ceiling, conditional on the training data, with the worst case in Lemma~\ref{lem:uniform} \\
Cross-model $\rho$ & Model pair over shared tasks & Descriptive rank agreement, with pairs that share models \\
Matching pass rate & Class decision and model pair & Fraction above a resampled-null threshold, with tests that share data and models \\
Slope interval & Resampled task block & Percentile interval with all student rows repeated per sampled task \\
Held-out error & Student and task residual & Empirical error after holding out each complete task \\
Pruning $q$ & Model, method, and sparsity & Exact-permutation diagnostic with Benjamini--Hochberg adjustment~\citep{benjamini1995controlling} within the stated family \\
$k^*=\infty$ & Search at its stated aggregation & No qualifying tested support, which does not prove unreadability \\
Low ceiling & Dense-probe task reading & Count not interpreted as a reliably readable task size \\
$N^*\leq n_1$ & Learning curve & Left-censored requirement, excluded from the exponent fit \\
\bottomrule
\end{tabular}
\end{table}

\subsubsection{Estimator Variants Across Experiments}
\label{app:protocol-registry}

Later experiments reuse the probe fitting of Section~\ref{app:probes} but adapt the representation, the ceiling, or the aggregation to their own question.
Table~\ref{tab:protocol-registry} lists these variants so that a common symbol does not suggest interchangeable readings.
We compare counts only within one row of the table.
In every row, neighboring layers are adjacent entries in the probed layer set of the cell, not necessarily consecutive transformer blocks.
Lemma~\ref{lem:basic}(i) and (ii) hold in every row, because each variant still takes a minimum over a finite candidate family against a threshold, and Lemma~\ref{lem:basic}(iv) holds in every row whose aggregation composes minima, maxima, and medians.

\begin{table}[t]
\centering\AppTableStyle
\caption{\textbf{Which estimator belongs to which experiment.} The main estimator is defined in Algorithms~\ref{alg:search} and~\ref{alg:aggregate}, and the appendix in the last column specifies each variant.}
\label{tab:protocol-registry}
\begin{tabular}{@{}p{0.18\linewidth}p{0.24\linewidth}p{0.34\linewidth}p{0.16\linewidth}@{}}
\toprule
\rowcolor{AppGold} \AppHead{Experiment} & \AppHead{Representation} & \AppHead{Ceiling and aggregation} & \AppHead{Details} \\
\midrule
Main counts and input-format factorial & Mean-pooled residual coordinates & Common dense maximum per class decision, neighbor confirmation, median over class decisions and then seeds & \ref{app:aggregation}, \ref{app:factorial} \\
SAE correspondence supports & Last-position SAE features & Qualifying support at the selected SAE layer, with a sweep ceiling that may come from other representations & \ref{app:sae}, \ref{app:matching} \\
Controlled synthetic models & Query-position residual or its principal components & Layer-specific dense reference, minimum across readable layers & \ref{app:toy-results} \\
Continual learning & Mean-pooled residual coordinates & Dense maximum per class decision and neighbor-confirmed count, compared with the same chain's baseline & \ref{app:continual} \\
Pruning & Mean-pooled residual coordinates & Median over class decisions per layer, then maximum over layers, compared with the main count & \ref{app:pruning} \\
Label requirements & Student test AUROC & Median at the actual training size, first threshold crossing on a logarithmic budget grid & \ref{app:label-count} \\
\bottomrule
\end{tabular}
\end{table}

\subsubsection{Software and Compute}
\label{app:compute}

Activation extraction uses the Hugging Face Transformers implementations of all models and runs on single GPUs.
Probe fitting runs on CPU, because each fit is a small convex problem and the cost lies in their number rather than their size.
At most 70 fits per class decision, layer, and seed give at most 1400 fits per class decision in the five-seed measurement of a four-layer window.
The main protocol draws its randomness from the fixed seed that builds every task pool and from the five measurement seeds, which fix the training subsample and the validation portion.
The only other source of variation is the L1 solver, whose coordinate order is not seeded and whose cap of 2000 iterations can leave an approximate solution, so an L1 support can differ between runs when the problem has several minimizers or the solver stops before convergence.

\subsubsection{Scope of the Measurement}
\label{app:scope}

The ceiling and $k^*$ describe what linear probes of the stated family can read from a frozen representation under the stated protocol.
Five consequences follow.
First, by Lemma~\ref{lem:basic}(i) a finite $k^*$ upper-bounds the smallest linear support that reaches the threshold, and $\infty$ records that the search found none, not that none exists.
Second, the ceiling is a linear-probe reference and not an upper bound on nonlinear predictors or on what fine-tuning could reach.
Third, $k^*$ depends on the feature basis, the pooling position, the input format, the wording of the inputs, the data, the threshold, and the search, and Section~\ref{app:design} reports how much each tested change moves it.
Fourth, Section~\ref{app:finite-sample} describes variation over test texts under Assumption~A and not variation over task definitions or data sources.
Fifth, a feature that predicts a label need not affect the model's behavior, and the intervention experiments of Section~\ref{sec:behavior} address that question separately.
The paper therefore compares counts only within a specified protocol and treats agreement across models as agreement in measured linear readouts rather than as evidence of shared computation.

\subsubsection{Recommendations for Measuring Readout Sparsity}
\label{app:recommendations}

The protocol and its controls suggest a reporting standard for studies that measure how many features a linear readout needs.
Each item below follows from a result stated in this appendix.
\begin{enumerate}
    \item Report the dense ceiling and the sparse count together for every cell, since two tasks with nearly equal ceilings can differ in count by a factor of 64 (Section~\ref{app:notation}).
    \item Define the threshold relative to the ceiling above chance, and report the task order under at least one alternative threshold (Lemma~\ref{lem:basic}(ii) and Table~\ref{tab:estimator}).
    \item Take the dense ceiling as a common reference over all layers and seeds that enter the count, so that no layer or seed receives an easier threshold (Section~\ref{app:aggregation}).
    \item Require agreement of two neighboring layers before a size counts, and hold the number of probed layers fixed across cells (Lemma~\ref{lem:basic}(iii) and Table~\ref{tab:worked}).
    \item Combine a univariate ranking with a joint selection method, and state the candidate family, since the count is an upper bound attached to that family (Lemma~\ref{lem:basic}(i)).
    \item Repeat the measurement over training subsamples on a fixed test split, report the seed range, and keep it separate from test-split uncertainty (Section~\ref{app:finite-sample}).
    \item Require a minimum number of test positives for every class decision, and check the standard error of each dense ceiling against the gap between ceiling and threshold, since the size of the rarer test class sets the precision of every AUROC (Table~\ref{tab:ceiling-precision}).
    \item Flag low ceilings and do not interpret the counts of such cells as task sizes (Section~\ref{app:aggregation}).
    \item Compare counts only within one basis, one input format, and one set of texts, and test the basis with a random rotation (Proposition~\ref{prop:affine} and Table~\ref{tab:rotation}).
    \item Remove surface cues from the task pools and verify their removal with a cue-only classifier, as the length-matched jailbreak pool does (Section~\ref{app:tasks}).
\end{enumerate}

%% file: appendix/appendix_cross_model.tex
\subsection{Details of the Cross-Model Agreement Experiments}
\label{app:agreement}

This appendix gives the procedures, estimators, controls, and per-cell readings behind Section~\ref{sec:agreement}.
Every experiment reuses the measurement protocol of Appendix~\ref{app:measurement} unless a paragraph states a change.
The appendix follows the order of the main text.
Section~\ref{app:ordering} compares counts and task orders across models and tests whether the text alone predicts the shared order, and Section~\ref{app:protocol-controls} measures how far each protocol choice moves them.
Section~\ref{app:inputbase} compares the models with a bag-of-words reading of the raw text.
Sections~\ref{app:matching} and~\ref{app:dictionary} test whether selected SAE features correspond across models and compose within a model, and Section~\ref{app:interventions} tests whether selected features matter for the model's own behavior.
Table~\ref{tab:agreement-claims} maps every claim of Section~\ref{sec:agreement} to its evidence.

\begin{table}[h]
\centering\AppTableStyle
\caption{\textbf{Claims of Section~\ref{sec:agreement} and the evidence behind each.}
The last column names the statistic and the reference value against which the main text reads each claim.}
\label{tab:agreement-claims}
\begin{tabular}{@{}p{0.30\linewidth}p{0.22\linewidth}p{0.40\linewidth}@{}}
\toprule
\rowcolor{AppGold} \AppHead{Claim} & \AppHead{Evidence} & \AppHead{Statistic and reference} \\
\midrule
Counts differ across tasks but agree within a task & \ref{app:ordering}, Figure~\ref{fig:cross-model-story}a, Table~\ref{tab:gold} & Cross-model spread ratio of Equation~\ref{eq:spread}, below 3.2 on 8 of 9 readable tasks \\
Models rank the tasks alike & \ref{app:ordering}, Table~\ref{tab:ordering} & Spearman $\rho$ of Equation~\ref{eq:rank-agreement}, median 0.93 over 30 cross-lab pairs, against a within-model permutation null with 99th percentile 0.21 \\
The order survives protocol changes & \ref{app:protocol-controls}, Tables~\ref{tab:protocol} to~\ref{tab:estimator} & Order correlation with the main protocol, 0.97 to 0.99 for estimator variants and 0.61 to 0.72 for rotations \\
Absolute counts depend on basis and wording & \ref{app:protocol-controls}, Tables~\ref{tab:rotation} and~\ref{tab:paraphrase} & Multiplicative factors of 1.6 and 2.0 under rotation and 1.5 to 10.8 under paraphrase, with ceilings within 0.03 \\
The agreement is not a shared text-derived order & \ref{app:ordering}, Table~\ref{tab:text-control} & Median cross-lab partial $\rho$ of Equation~\ref{eq:partial-order}, 0.86 to 0.92 under six text-side controls, against 0.93 without control \\
Selected SAE features correspond across models & \ref{app:matching}, Tables~\ref{tab:identity} to~\ref{tab:scale}, Figure~\ref{fig:pair-matching} & Pass rate of Equation~\ref{eq:pass-rate} against a nominal 1\% under the label-correlation-matched null \\
Selected language features drive reply language & \ref{app:interventions}, Tables~\ref{tab:langid-arm} and~\ref{tab:dose} & Gap of Equation~\ref{eq:reply-gap} with 95\% bootstrap intervals against frequency-matched and equal-norm controls \\
Selected coordinates are sufficient but not exclusive & \ref{app:interventions}, Table~\ref{tab:pinning} & Dense AUROC after pinning 256 coordinates, within 0.0011 of the untouched probe \\
\bottomrule
\end{tabular}
\end{table}

\textbf{Design rationale.}
Every experiment in this appendix answers a specific alternative explanation of a main-text finding, and its paragraph headed Design rationale names that alternative, the design element that rules it out, and the evidence that it does.
The design elements of these controls share one principle.
Each control holds fixed everything except the factor under suspicion, so that any change in the reading can be attributed to that factor alone.

\textbf{Notation.}
We write $\mathcal M$ for a set of models and $\mathcal T$ for a set of tasks.
For a model $m$ and a task $t$, $k^*_{t,m}$ denotes the task-level count of Algorithm~\ref{alg:aggregate} and $C_{t,m}$ the task ceiling, namely the median of the class-decision ceilings $C_c$ of Equation~\ref{eq:threshold}.
A task is readable in model $m$ when $C_{t,m}>0.90$, the readability flag of Appendix~\ref{app:aggregation}.
Every count in this appendix is unrounded unless a caption says otherwise.

\subsubsection{Task Ordering Across Models}
\label{app:ordering}

\textbf{Question.}
This section asks whether two models that differ in lab, scale, and training data assign similar counts to the same task, and whether they order the tasks alike.

\textbf{Data and models.}
The comparison uses the 14 core tasks of Table~\ref{tab:tasks} on the ten base models of Table~\ref{tab:models}, and a second pass adds the nine instruction-tuned models.
Every model reads the same text pools, so any difference between two models comes from the models and not from the data.
The ten base models come from five labs and form $\binom{10}{2}=45$ pairs.
Six of the ten base models are Gemma models, so 15 pairs share a lab and 30 cross labs.
The 19 models of the second pass form 171 pairs, of which 69 share a lab and 102 cross labs.

\textbf{Estimators.}
We summarize agreement within a task by the cross-model spread ratio
\begin{equation}
    \sigma_t=\frac{\max_{m\in\mathcal M}k^*_{t,m}}{\min_{m\in\mathcal M}k^*_{t,m}},
    \label{eq:spread}
\end{equation}
computed over the models in which task $t$ is readable.
A ratio of 1 means identical counts, and a ratio of 2 means that the widest model needs one more step of the power-of-two grid of Equation~\ref{eq:grid} than the narrowest.
We summarize agreement across tasks by the Spearman correlation~\citep{spearman1987proof} between the count vectors of two models $A$ and $B$ over a shared task set $\mathcal T_{AB}$,
\begin{equation}
    \rho(A,B)=\frac{\sum_{t\in\mathcal T_{AB}}\bigl(r^A_t-\bar r^A\bigr)\bigl(r^B_t-\bar r^B\bigr)}{\sqrt{\sum_{t}\bigl(r^A_t-\bar r^A\bigr)^2}\sqrt{\sum_{t}\bigl(r^B_t-\bar r^B\bigr)^2}},
    \label{eq:rank-agreement}
\end{equation}
where $r^A_t$ is the rank of $k^*_{t,A}$ among the counts of model $A$ on $\mathcal T_{AB}$, tied values receive the average of their ranks, and $\bar r^A$ is the mean of these ranks over the shared task set.

\textbf{Treatment of infinite counts.}
A cell with $k^*=\infty$ has no finite size, and the task set $\mathcal T_{AB}$ admits three definitions.
The \emph{all} treatment keeps every task and gives every infinite cell the largest tied rank of its model, which matches how Figure~\ref{fig:task-organization}b places such cells.
The \emph{finite} treatment drops a task when either model reads infinite on it.
The \emph{gated} treatment additionally drops a task when either model's ceiling falls below 0.90, which leaves only cells where a count is interpretable.
The main-text summary of Section~\ref{sec:agreement} uses a fourth and stricter set, namely the nine tasks readable in all ten base models, so that every pair compares the same nine tasks.

\textbf{Design rationale.}
Two summaries answer two different doubts.
The spread ratio asks whether counts agree in absolute size within a task, and the rank correlation asks whether models agree on which tasks are narrow, which survives any uniform rescaling of a model's counts.
Infinite and low-ceiling cells carry no interpretable size, and the three treatments bracket how we rank them, from keeping every cell to keeping only interpretable ones.
Splitting pairs by lab tests whether the agreement comes from shared training recipes, which would show as a same-lab advantage in the medians of Table~\ref{tab:ordering}.
The permutation test addresses the alternative that nine tasks with ties agree by chance, and the text-level control addresses the alternative that every model reads one order off the text, which would leave no agreement once the text-side order is removed.

\textbf{Remark (invariance of the order statistic).}
Let $g$ be any strictly increasing map on $[1,\infty]$ with $g(\infty)=\infty$, and replace every count $k^*_{t,m}$ by $g(k^*_{t,m})$ in every model.
Then every correlation $\rho(A,B)$ of Equation~\ref{eq:rank-agreement} is unchanged under each of the three treatments, because a strictly increasing map preserves the order of any two counts and preserves equality, so it leaves every rank $r^A_t$ of every task unchanged, including the average ranks of ties and the tied maximal rank of infinite cells, and Equation~\ref{eq:rank-agreement} depends on the counts only through these ranks.

The invariance of the order statistic has three consequences for the analysis.
First, reporting $\log k^*$ instead of $k^*$ changes no correlation.
Second, an estimator change that multiplies every count of a model by a common factor leaves that model's order correlation with the main protocol at exactly 1, so the estimator controls of Section~\ref{app:protocol-controls} measure task-specific distortion and not uniform rescaling.
Third, rounding is not strictly increasing because it can merge distinct values, which is why every correlation in this paper uses unrounded counts.

\textbf{Permutation test.}
The null hypothesis is that the ten models order the nine tasks independently of one another.
We shuffle the task labels within every model, which keeps each model's ties, recompute the median of Equation~\ref{eq:rank-agreement} over the 30 cross-lab pairs, and repeat this $10^5$ times with seed 0.
A shuffled model enters all nine of its pairs, so the null distribution carries the dependence between pairs that share a model.
The test rejects only independence, since two instruments could agree through a shared text-derived order, which the text-level control below measures.

\textbf{Count ranges and ceilings.}
Across the ten base models, language identification has a median $k^*$ of 1, tool selection and routing require one or two features, and harmfulness detection ranges from 153 to 391.
Among the nine tasks readable in every model, $\sigma_t$ is at most 3.2, except for PII detection, which ranges from 1 to 8.
Below the readability flag, mathematical correctness ranges from 8 to 981 among the seven models with finite readings.
Ceilings differ by at most 0.05 across models on every readable task.
Table~\ref{tab:gold} lists the five tasks outside the common readable subset.

\begin{table}[!ht]
\centering\AppTableStyle
\caption{\textbf{Tasks outside the common high-ceiling subset.} Each cell shows the unrounded $k^*$ followed by the dense-probe ceiling in small type. Muted counts have a ceiling $C\leq0.90$ and are not interpreted as reliable task sizes, and $\infty$ means that no tested support qualifies. Together with Figure~\ref{fig:cross-model-story}a, these rows cover all 14 core tasks on the ten base models.}
\label{tab:gold}
\begin{tabular}{@{}lrrrrr@{}}
\toprule
\rowcolor{AppGold} \AppHead{Model} & \AppHead{Fact checking} & \AppHead{Relevance} & \AppHead{Math answer} & \AppHead{Math step} & \AppHead{Code bug} \\
\midrule
Gemma 3 270M & \textcolor{AppMuted}{235}\,{\scriptsize\textcolor{AppMuted}{0.840}} & \textcolor{AppMuted}{303}\,{\scriptsize\textcolor{AppMuted}{0.636}} & \textcolor{AppMuted}{$\infty$}\,{\scriptsize\textcolor{AppMuted}{0.646}} & \textcolor{AppMuted}{94}\,{\scriptsize\textcolor{AppMuted}{0.751}} & \textcolor{AppMuted}{$\infty$}\,{\scriptsize\textcolor{AppMuted}{0.622}} \\
Gemma 3 1B & \textcolor{AppMuted}{358}\,{\scriptsize\textcolor{AppMuted}{0.879}} & \textcolor{AppMuted}{$\infty$}\,{\scriptsize\textcolor{AppMuted}{0.647}} & \textcolor{AppMuted}{$\infty$}\,{\scriptsize\textcolor{AppMuted}{0.634}} & \textcolor{AppMuted}{140}\,{\scriptsize\textcolor{AppMuted}{0.756}} & \textcolor{AppMuted}{$\infty$}\,{\scriptsize\textcolor{AppMuted}{0.643}} \\
Gemma 3 4B & \textcolor{AppMuted}{206}\,{\scriptsize\textcolor{AppMuted}{0.888}} & \textcolor{AppMuted}{128}\,{\scriptsize\textcolor{AppMuted}{0.639}} & \textcolor{AppMuted}{256}\,{\scriptsize\textcolor{AppMuted}{0.654}} & \textcolor{AppMuted}{223}\,{\scriptsize\textcolor{AppMuted}{0.763}} & \textcolor{AppMuted}{$\infty$}\,{\scriptsize\textcolor{AppMuted}{0.656}} \\
Gemma 2 9B & 253\,{\scriptsize\textcolor{AppMuted}{0.910}} & \textcolor{AppMuted}{128}\,{\scriptsize\textcolor{AppMuted}{0.649}} & \textcolor{AppMuted}{144}\,{\scriptsize\textcolor{AppMuted}{0.661}} & \textcolor{AppMuted}{295}\,{\scriptsize\textcolor{AppMuted}{0.782}} & \textcolor{AppMuted}{937}\,{\scriptsize\textcolor{AppMuted}{0.650}} \\
Gemma 3 12B & 222\,{\scriptsize\textcolor{AppMuted}{0.913}} & \textcolor{AppMuted}{128}\,{\scriptsize\textcolor{AppMuted}{0.645}} & \textcolor{AppMuted}{128}\,{\scriptsize\textcolor{AppMuted}{0.666}} & \textcolor{AppMuted}{255}\,{\scriptsize\textcolor{AppMuted}{0.781}} & \textcolor{AppMuted}{$\infty$}\,{\scriptsize\textcolor{AppMuted}{0.670}} \\
Gemma 3 27B & \textcolor{AppMuted}{239}\,{\scriptsize\textcolor{AppMuted}{0.887}} & \textcolor{AppMuted}{128}\,{\scriptsize\textcolor{AppMuted}{0.644}} & \textcolor{AppMuted}{8}\,{\scriptsize\textcolor{AppMuted}{0.675}} & \textcolor{AppMuted}{128}\,{\scriptsize\textcolor{AppMuted}{0.789}} & \textcolor{AppMuted}{$\infty$}\,{\scriptsize\textcolor{AppMuted}{0.660}} \\
Llama 3.1 8B & 644\,{\scriptsize\textcolor{AppMuted}{0.902}} & \textcolor{AppMuted}{256}\,{\scriptsize\textcolor{AppMuted}{0.638}} & \textcolor{AppMuted}{512}\,{\scriptsize\textcolor{AppMuted}{0.639}} & \textcolor{AppMuted}{266}\,{\scriptsize\textcolor{AppMuted}{0.770}} & \textcolor{AppMuted}{$\infty$}\,{\scriptsize\textcolor{AppMuted}{0.655}} \\
Mistral 7B & 255\,{\scriptsize\textcolor{AppMuted}{0.903}} & \textcolor{AppMuted}{256}\,{\scriptsize\textcolor{AppMuted}{0.641}} & \textcolor{AppMuted}{981}\,{\scriptsize\textcolor{AppMuted}{0.637}} & \textcolor{AppMuted}{70}\,{\scriptsize\textcolor{AppMuted}{0.753}} & \textcolor{AppMuted}{$\infty$}\,{\scriptsize\textcolor{AppMuted}{0.657}} \\
OLMo 2 7B & \textcolor{AppMuted}{680}\,{\scriptsize\textcolor{AppMuted}{0.890}} & \textcolor{AppMuted}{128}\,{\scriptsize\textcolor{AppMuted}{0.627}} & \textcolor{AppMuted}{$\infty$}\,{\scriptsize\textcolor{AppMuted}{0.653}} & \textcolor{AppMuted}{257}\,{\scriptsize\textcolor{AppMuted}{0.767}} & \textcolor{AppMuted}{1024}\,{\scriptsize\textcolor{AppMuted}{0.645}} \\
Qwen 3 8B & \textcolor{AppMuted}{241}\,{\scriptsize\textcolor{AppMuted}{0.893}} & \textcolor{AppMuted}{512}\,{\scriptsize\textcolor{AppMuted}{0.651}} & \textcolor{AppMuted}{176}\,{\scriptsize\textcolor{AppMuted}{0.713}} & \textcolor{AppMuted}{344}\,{\scriptsize\textcolor{AppMuted}{0.804}} & \textcolor{AppMuted}{$\infty$}\,{\scriptsize\textcolor{AppMuted}{0.663}} \\
\bottomrule
\end{tabular}
\end{table}

\textbf{Rounding in Figure~\ref{fig:cross-model-story}a.}
A cell's $k^*$ is a median over class decisions and then over seeds, so a cell with an even number of contributing values can land halfway between two grid points.
Six of the 90 cells in Figure~\ref{fig:cross-model-story}a do.
Tool selection reads 1.5 on Gemma-3-12B, Mistral-7B, and OLMo-2-7B, intent classification reads 4.5 on Qwen3-8B, and banking intent reads 15.5 on Gemma-3-4B and 17.5 on Llama-3.1-8B.
The figure rounds these to the nearest integer and resolves a tie to the even value, so tool selection displays 2 and intent classification displays 4.
Every correlation in this paper uses the unrounded values, as the invariance of the order statistic requires.

\textbf{Result for the main-text summary.}
For the nine readable tasks, Equation~\ref{eq:rank-agreement} over the 30 cross-lab pairs gives a median of 0.9294 and a range of 0.7812 to 1.0000.
The permutation null has a median of $-0.004$, a 99th percentile of 0.21, and a maximum of 0.42, so no permutation reaches the observed median and $p<10^{-5}$.
The 15 same-lab pairs and all 45 pairs give the same result, at null 99th percentiles of 0.31 and 0.18.
This summary uses the same nine tasks for every pair, and the pairwise treatments below form a separate summary over all 14 tasks.

\begin{table}[t]
\AppTableStyle
\centering\small
\caption{\textbf{Spearman correlation between two models' $k^*$ orderings over the 14 core tasks.}
Each entry is the median over pairs with the minimum and maximum in brackets, and $n$ is the number of pairs.
The three treatments differ only in how they handle cells with no finite $k^*$ or a ceiling below 0.90.}
\label{tab:ordering}
\resizebox{\linewidth}{!}{%
\begin{tabular}{llll}
\toprule
\rowcolor{AppGold}
\AppHead{Model set and treatment} & \AppHead{Same lab} & \AppHead{Different labs} & \AppHead{All pairs} \\
\midrule
Ten base models, all & $0.886$ [$0.807$, $0.967$], $n=15$ & $0.911$ [$0.818$, $0.969$], $n=30$ & $0.907$ [$0.807$, $0.969$], $n=45$ \\
Ten base models, finite & $0.926$ [$0.846$, $0.981$], $n=15$ & $0.904$ [$0.771$, $0.981$], $n=30$ & $0.906$ [$0.771$, $0.981$], $n=45$ \\
Ten base models, gated & $0.931$ [$0.897$, $1.000$], $n=15$ & $0.929$ [$0.817$, $1.000$], $n=30$ & $0.931$ [$0.817$, $1.000$], $n=45$ \\
\midrule
All 19 models, all & $0.892$ [$0.750$, $0.993$], $n=69$ & $0.905$ [$0.767$, $0.969$], $n=102$ & $0.901$ [$0.750$, $0.993$], $n=171$ \\
All 19 models, finite & $0.889$ [$0.718$, $0.991$], $n=69$ & $0.890$ [$0.750$, $0.981$], $n=102$ & $0.890$ [$0.718$, $0.991$], $n=171$ \\
All 19 models, gated & $0.929$ [$0.785$, $1.000$], $n=69$ & $0.915$ [$0.797$, $1.000$], $n=102$ & $0.920$ [$0.785$, $1.000$], $n=171$ \\
\bottomrule
\end{tabular}}
\end{table}

\textbf{Result for the pairwise treatments.}
We make two observations on Table~\ref{tab:ordering}.
First, the three treatments agree to within 0.05 in every column, so the stability of the order does not depend on how we handle unreadable cells.
Second, the same-lab and cross-lab ranges overlap substantially, and the medians show no consistent same-lab advantage across treatments.
Adding the nine instruction-tuned models leaves every median between 0.889 and 0.929.

\textbf{Text-level control.}
We run the probing pipeline of Appendix~\ref{app:inputbase} on TF-IDF features of the raw text and read, for each of the nine tasks, the prefix score $s(k)$ of Equation~\ref{eq:prefix-score} at $k=4,8,16,64,256$ and the text-only ceiling.
With any one of these six readings as the control $z$, we compute for every pair of models $(A,B)$ the partial rank correlation of their two $k^*$ orders,
\begin{equation}
    \rho_{AB\cdot z}=\frac{\rho_{AB}-\rho_{Az}\,\rho_{Bz}}{\sqrt{\left(1-\rho_{Az}^2\right)\left(1-\rho_{Bz}^2\right)}},
    \label{eq:partial-order}
\end{equation}
where each $\rho$ is the Spearman correlation of Equation~\ref{eq:rank-agreement} between two of the three task orders over the nine tasks (Table~\ref{tab:text-control}).
Each reading predicts part of every model's order, with a median correlation of $-0.51$ to $-0.71$ across the ten models, because tasks that the text decides with few terms also admit sparse readouts.
After control, the median cross-lab partial correlation is 0.86 to 0.92, against 0.93 without control, and its lowest value of 0.860 comes with $s(256)$ as the control.
The agreement thus rests mostly on the part of the order that the text does not predict, such as the ranking of the seven tasks whose $s(16)$ lies between 0.88 and 1.00.

\begin{table}[!ht]
\centering\AppTableStyle
\caption{\textbf{Text-side readings of the nine readable tasks and the cross-model order they leave unexplained.} Columns $s(k)$ give the prefix score of Equation~\ref{eq:prefix-score} on TF-IDF features of the raw text, the fraction of the text-only ceiling reached within $k$ terms, and the text ceiling is that representation's dense-probe AUROC. Rows are ordered by the median $k^*$ over the ten base models, rounded for display. The first summary row is the median over models of the Spearman correlation between each model's $k^*$ and the column, and the second is the median over the 30 cross-lab pairs of the partial rank correlation of Equation~\ref{eq:partial-order} with the column as the control, against 0.929 without control.}
\label{tab:text-control}
\begin{tabular}{@{}lrrrrrrr@{}}
\toprule
\rowcolor{AppGold} \AppHead{Task} & \AppHead{$s(4)$} & \AppHead{$s(8)$} & \AppHead{$s(16)$} & \AppHead{$s(64)$} & \AppHead{$s(256)$} & \AppHead{Text ceiling} & \AppHead{Median $k^*$} \\
\midrule
Language ID & 0.911 & 0.940 & 0.967 & 0.996 & 1.000 & 1.000 & 1 \\
PII & 0.789 & 0.842 & 0.884 & 0.994 & 0.998 & 1.000 & 1 \\
Tool selection & 0.995 & 1.000 & 1.000 & 1.000 & 1.000 & 1.000 & 1 \\
Routing & 0.780 & 0.826 & 0.899 & 0.969 & 0.985 & 0.995 & 2 \\
Intent (CLINC150) & 1.000 & 1.000 & 1.000 & 1.000 & 1.000 & 1.000 & 4 \\
Jailbreak & 0.679 & 0.832 & 0.890 & 0.938 & 0.973 & 0.997 & 12 \\
Banking77 & 0.843 & 0.969 & 0.969 & 0.998 & 1.000 & 1.000 & 17 \\
Sentiment & 0.191 & 0.479 & 0.530 & 0.724 & 0.838 & 0.867 & 163 \\
Harmfulness & 0.420 & 0.559 & 0.637 & 0.776 & 0.884 & 0.883 & 310 \\
\midrule Alone, median $\rho$ with $k^*$ & $-0.64$ & $-0.55$ & $-0.51$ & $-0.59$ & $-0.71$ & $-0.67$ & --- \\
Controlled cross-lab $\rho$, median & 0.877 & 0.898 & 0.917 & 0.891 & 0.860 & 0.877 & --- \\
\bottomrule
\end{tabular}
\end{table}

\textbf{Scope.}
The correlations of Table~\ref{tab:ordering} are measured over a fixed population of model pairs.
They establish that the tested models order the tasks alike under one measurement protocol, and Section~\ref{app:protocol-controls} measures how far that protocol can move the order.
The text-level control uses one text representation, and a richer text-only baseline could explain more of the order.

\textbf{Finding.}
Readout sparsity orders tasks alike across models and labs.
On the nine readable tasks the median cross-lab Spearman correlation is 0.93 and the lowest is 0.78, and same-lab pairs show no advantage in any treatment.
Controlling for text-side readings leaves the median at 0.86 to 0.92.
Sparse probing has so far characterized one model at a time~\citep{gurnee2023finding}, and the task order it yields is largely shared across ten base models from five labs.

\textbf{Implication.}
For ranking tasks by readout sparsity, one reference model can stand in for models of other labs, up to the residual disagreement that the lowest pair quantifies.
Absolute counts carry no such guarantee, as the protocol controls of the next section show.

\subsubsection{Protocol and Estimator Controls}
\label{app:protocol-controls}

\textbf{Question.}
The count of a task in one model depends on the representation basis, the input format, the wording of the inputs, and four estimator choices.
This section asks, for each of these, how far the choice moves the absolute count, the ceiling, and the task order.
Table~\ref{tab:protocol} summarizes the answers, and the paragraphs below give each control in full.

\begin{table}[t]
\AppTableStyle
\centering\small
\caption{\textbf{What each protocol choice moves.}
The upper block changes the representation or the inputs and the lower block changes the estimator.
Effects on $k^*$ are multiplicative factors, effects on the ceiling are absolute AUROC changes, and effects on task order are Spearman correlations against the main protocol, computed over 22 cells for the rotation and over the 14 tasks for the estimator variants.
Two tasks do not define an order, so the chat-template row reports none.}
\label{tab:protocol}
\resizebox{\linewidth}{!}{%
\begin{tabular}{lllll}
\toprule
\rowcolor{AppGold}
\AppHead{Change} & \AppHead{Measured on} & \AppHead{$k^*$} & \AppHead{Ceiling} & \AppHead{Task order} \\
\midrule
Feature basis, random rotation & 2 models, 22 cells & $\times1.6$ and $\times2.0$, at most $\times20$ & $\leq0.008$ & $0.72$ and $0.61$ \\
Input format, chat template & 1 model, 2 tasks & $\div1.2$ to $\div5.7$ & $\leq0.02$ & not measured \\
Wording, paraphrase & 1 model, 6 cells & $\times1.5$ to $\times10.8$ & $\leq0.03$ & kept at task level \\
\midrule
Threshold, 90\% or 97.5\% of ceiling & 19 models & moves by construction & unaffected & $0.97$ and $0.97$ \\
Search, ranked path only & 19 models & moves by construction & unaffected & $0.99$ \\
Layers, best single layer & 19 models & moves by construction & not reported & $0.98$ \\
Probe, MLP for linear & 1 model, 7 tasks & equal on 5 of 7, one grid step on 2 & $\leq0.003$ & not measured \\
\bottomrule
\end{tabular}}
\end{table}

\paragraph{Random rotations of the feature basis.}
\textbf{Question.}
A small $k^*$ has two possible sources.
The model may place task information in few residual coordinates, or a strong discriminative direction may be readable from few coordinates of any basis.
The rotation control separates the two.

\textbf{Procedure.}
For each rotation $r\in\{0,1,2\}$ we draw a $d\times d$ matrix with independent standard normal entries from a generator seeded with $1000+r$ and take the orthogonal factor $Q_r$ of its QR decomposition.
We multiply the training and test activations of the task's neighbor-confirmed sweep layer by $Q_r$, then rerun the full search of Algorithm~\ref{alg:search}, including standardization, with one seed on the full training split.
The unrotated reference uses the same layer, seed, and split, so its counts differ from the five-seed values of Table~\ref{tab:gold}.
The control covers 22 cells on each of Gemma-2-9B and Gemma-3-4B, namely the 20 class decisions of intent classification together with the binary tasks harmfulness detection and mathematical-step correctness.

\textbf{Remark (a rotation preserves linear readability).}
Let $Q\in\mathbb R^{d\times d}$ be invertible.
The set of affine decision functions $\{x\mapsto w^\top x+b : w\in\mathbb R^d,\ b\in\mathbb R\}$ equals the set $\{x\mapsto w'^\top Qx+b : w'\in\mathbb R^d,\ b\in\mathbb R\}$, since the choice $w'=Q^{-\top}w$ gives $w'^\top Qx=w^\top x$ and the choice $w=Q^\top w'$ gives the converse.
The largest test AUROC that any dense linear probe attains is therefore the same on $x$ and on $Qx$, because AUROC depends only on the values of the decision function on the test points.

This remark concerns the hypothesis class and not the fitted probe, because per-coordinate standardization and the penalty of Equation~\ref{eq:probe} are not rotation invariant.
A change in the fitted ceiling therefore measures the fitting procedure, and a change in $k^*$ measures how the discriminative direction distributes over coordinates.
Proposition~\ref{prop:affine} gives the complementary statement that coordinate-wise affine maps such as standardization change neither quantity.

\textbf{Design rationale.}
An orthogonal map is the one change that provably preserves linear readability while destroying the coordinate system, so any change in $k^*$ under it isolates the role of the model's basis.
Fixing the 20-fold criterion before the run prevents an after-the-fact threshold from turning a modest effect into a claim of a privileged basis.

\textbf{Null prediction and criterion.}
If the model's coordinates carried no special role, a random rotation would leave the counts unchanged.
If the task information sat on a few coordinates of the model's basis, a rotation would spread it and raise the counts sharply.
As a heuristic, a unit direction concentrated on a few coordinates acquires small components on nearly all coordinates after a random rotation, which is the mechanism behind the second prediction.
Before running the control we fixed a 20-fold rise as the criterion for a privileged basis.

\textbf{Estimator.}
For a cell $c$ with unrotated count $k_c$ and rotated counts $k_{c,r}$, the cell factor is $\phi_c=\operatorname{median}_r k_{c,r}/k_c$.
We summarize a model by the median of $\phi_c$ over its 22 cells and test whether the cells rise more often than they fall with a two-sided sign test,
\begin{equation}
    p_{\mathrm{sign}}=\min\Bigl\{1,\ 2\sum_{i=0}^{\min(n_+,n_-)}\binom{n}{i}2^{-n}\Bigr\},
    \qquad n=n_++n_-,
    \label{eq:sign-test}
\end{equation}
where $n_+$ and $n_-$ count the cells whose factor $\phi_c$ exceeds or falls below 1, and the cells with $\phi_c=1$, whose median rotated count equals the original, drop out.

\begin{table}[t]
\AppTableStyle
\centering\small
\caption{\textbf{Effect of a random rotation on $k^*$.}
Entries are medians over the three rotations, and the intent row is additionally a median over the 20 class decisions.
The factor column is the median over cells of each cell's own ratio, which is why it does not equal the ratio of the two preceding columns.}
\label{tab:rotation}
\begin{tabular}{llrrr}
\toprule
\rowcolor{AppGold}
\AppHead{Model} & \AppHead{Task} & \AppHead{Original $k^*$} & \AppHead{Rotated $k^*$} & \AppHead{Factor} \\
\midrule
Gemma-2-9B & Intent, 20 class decisions & 4 & 8 & $1.44$ \\
           & Step check & 68 & 244 & $3.59$ \\
           & Harmfulness & 150 & 251 & $1.67$ \\
\midrule
Gemma-3-4B & Intent, 20 class decisions & 4 & 12 & $2.38$ \\
           & Step check & 128 & 214 & $1.67$ \\
           & Harmfulness & 139 & 142 & $1.02$ \\
\bottomrule
\end{tabular}
\end{table}

\textbf{Result.}
Rotation raises $k^*$ by a median factor of 1.59 on Gemma-2-9B, with an interquartile range of 1.00 to 2.00 and a maximum of 4.0.
There, 14 of the 22 cells rise while 3 fall and 5 stay put, and Equation~\ref{eq:sign-test} gives $p=0.013$.
On Gemma-3-4B the median factor is 2.00, the interquartile range is 1.68 to 4.00, the maximum is 20.0, and 18 cells rise against 1 falling and 3 unchanged, giving $p=8\times10^{-5}$.
The rise is real and an order of magnitude short of the criterion we had fixed, and we report it against that criterion.
Taking the median over the three rotations in each cell, ceilings fall by at most 0.0002 on Gemma-2-9B and 0.0078 on Gemma-3-4B, and the worst single rotation lowers a ceiling by 0.0032 and 0.0131, as the remark on rotations leads one to expect.
Rank correlations between rotated and original counts over the 22 cells are 0.720 with $p=1.6\times10^{-4}$ and 0.606 with $p=0.0028$.
In both models the rotated intent decisions, at 8 and 12 features, stay far below the two wide tasks, at 142 to 251.
The two wide tasks exchange places on Gemma-3-4B, where step checking rises by 1.67 and harmfulness by 1.02, so the size of the rise has no consistent relation to encoding width.
The rise stays modest because the largest rotated count is 251 out of $d=3584$, so a few hundred random mixtures still span a discriminative direction that lies in a high-variance subspace.

\textbf{Scope.}
The basis contributes a factor of one to four to the typical cell and up to twenty in the most affected cell, and it leaves the separation between narrow and wide tasks intact.
The result applies to one seed, one layer per task, and three rotations on two models.

\paragraph{Chat template.}
Section~\ref{sec:training} reports the four-cell design that crosses base and instruction-tuned weights of Gemma-2-9B with plain and chat-formatted inputs on harmfulness detection and fact checking.
Appendix~\ref{app:factorial} gives the conditional effects, the interaction formula, the seed ranges, and the common-layer controls, and Table~\ref{tab:protocol} summarizes only their implication for measurement sensitivity.
The design covers two tasks, so Table~\ref{tab:protocol} reports no task order for this control.

\paragraph{Paraphrase.}
\textbf{Question.}
This control asks whether a count describes a task itself or the particular wording of the test texts on which the count was originally measured.

\textbf{Procedure.}
We paraphrase a seeded subsample of at most 1500 texts from each task's test split with Mistral-7B-Instruct-v0.3 at temperature 0.7, discard refusals and overly short generations, and retain about 99\% of the items.
Probes train on the original training activations and evaluate on the original and the paraphrased test sets, with Gemma-2-9B as the instrument.
The evaluation sets hold 1499 intent texts, 472 jailbreak prompts, 855 sentiment sentences, and 1498 harmfulness prompts, and the original arm evaluates on test sets of the same sizes.
The control uses one seed, the full training split, and neighbor confirmation over all ten grid layers.
A second run paraphrases the training split as well, capped near 3000 texts per task, with 2988, 1900, 2991, and 2997 training texts, and adds an original-text baseline with the same training sizes.

\textbf{Estimator.}
Write $k^*_{a\to b}$ for the count when probes train on distribution $a$ and test on distribution $b$, with $o$ the original texts, $o'$ the size-matched original subset, and $p$ the paraphrases.
The expansion factor of Table~\ref{tab:paraphrase} then decomposes exactly into three ratios,
\begin{equation}
    \frac{k^*_{o\to p}}{k^*_{o\to o}}
    =\underbrace{\frac{k^*_{o'\to o'}}{k^*_{o\to o}}}_{\text{sample size}}
    \cdot\underbrace{\frac{k^*_{p\to p}}{k^*_{o'\to o'}}}_{\text{intrinsic width}}
    \cdot\underbrace{\frac{k^*_{o\to p}}{k^*_{p\to p}}}_{\text{distribution shift}},
    \label{eq:paraphrase-decomp}
\end{equation}
which holds as an identity because the intermediate counts cancel.
The first ratio isolates the smaller training set, the second measures how much wider the paraphrased distribution is when read on its own terms, and the third measures the extra cost of reading across the two distributions.

\textbf{Design rationale.}
A single cross-distribution comparison confounds the three effects of Equation~\ref{eq:paraphrase-decomp}, and the four train and test combinations separate them.
The size-matched original arm removes the smaller paraphrased training set as an explanation, and the paraphrased-to-paraphrased arm measures the width of the new distribution without any transfer across distributions.

\textbf{Null prediction.}
If a count described only the task, paraphrasing would leave all three ratios near 1.

\begin{table}[t]
\AppTableStyle
\centering\small
\caption{\textbf{Effect of paraphrasing the test inputs.}
Each entry is a ceiling followed by $k^*$, with Gemma-2-9B as the instrument.
The factor is the ratio of the two $k^*$ values.}
\label{tab:paraphrase}
\begin{tabular}{lccr}
\toprule
\rowcolor{AppGold}
\AppHead{Task} & \AppHead{Original} & \AppHead{Paraphrased} & \AppHead{Factor} \\
\midrule
Intent, account blocked & $1.000$ / 2 & $1.000$ / 8 & $4.0$ \\
Intent, smart home & $1.000$ / 4 & $1.000$ / 6 & $1.5$ \\
Intent, weather & $1.000$ / 4 & $0.990$ / 16 & $4.0$ \\
Jailbreak & $1.000$ / 8 & $0.982$ / 86 & $10.8$ \\
Sentiment & $0.980$ / 85 & $0.970$ / 208 & $2.4$ \\
Harmfulness & $0.977$ / 166 & $0.947$ / 389 & $2.3$ \\
\bottomrule
\end{tabular}
\end{table}

\begin{table}[t]
\AppTableStyle
\centering\small
\caption{\textbf{Decomposition of the paraphrase expansion by Equation~\ref{eq:paraphrase-decomp}.}
The count columns give $k^*$ for each train and test combination, and the factor columns multiply to the factor of Table~\ref{tab:paraphrase}.}
\label{tab:paraphrase-decomp}
\resizebox{\linewidth}{!}{%
\begin{tabular}{lrrrrrrr}
\toprule
\rowcolor{AppGold}
\AppHead{Task} & \AppHead{$k^*_{o\to o}$} & \AppHead{$k^*_{o'\to o'}$} & \AppHead{$k^*_{p\to p}$} & \AppHead{$k^*_{o\to p}$} & \AppHead{Sample size} & \AppHead{Intrinsic width} & \AppHead{Distribution shift} \\
\midrule
Intent, account blocked & 2 & 2 & 8 & 8 & $1.0$ & $4.0$ & $1.0$ \\
Intent, smart home & 4 & 8 & 8 & 6 & $2.0$ & $1.0$ & $0.8$ \\
Intent, weather & 4 & 8 & 15 & 16 & $2.0$ & $1.9$ & $1.1$ \\
Jailbreak & 8 & 8 & 45 & 86 & $1.0$ & $5.6$ & $1.9$ \\
Sentiment & 85 & 86 & 128 & 208 & $1.0$ & $1.5$ & $1.6$ \\
Harmfulness & 166 & 222 & 264 & 389 & $1.3$ & $1.2$ & $1.5$ \\
\bottomrule
\end{tabular}}
\end{table}

\textbf{Result.}
We make two observations on Tables~\ref{tab:paraphrase} and~\ref{tab:paraphrase-decomp}.
First, the paraphrased distribution is intrinsically wider on five of the six cells, by 1.2 to 5.6, and reading across distributions adds a further 1.5 to 1.9 on the three binary tasks and 0.8 to 1.1 on the intents.
Second, jailbreak detection expands the most in both parts, which fits the reading that jailbreak detection partly rests on surface wording.
The sample-size ratio stays at one grid step or less in every cell.
Ceilings stay at or above 0.947 everywhere, and the order of the four tasks holds in all four train and test combinations, while within the intent task two class decisions exchange places and ties at 8 appear.

\textbf{Scope.}
The result holds for one paraphrasing model, one paraphrase seed, and one instrument, Gemma-2-9B.
It shows that absolute counts belong to a fixed set of texts, which is why the paper compares counts across models only on identical text pools.

\paragraph{Probe family and probe capacity.}
\textbf{Question.}
Two standard objections apply to any sparse-probe count such as $k^*$.
A nonlinear probe might read the same class decision from fewer features, which would make $k^*$ a property of linear readability rather than of the representation, and a finite count might reflect probe capacity rather than signal in the activations.

\textbf{Procedure.}
We reuse the same feature ranking, the same grid of $k$, and the same validation split, and replace the linear probe with a one-hidden-layer MLP in one arm and permute the training labels in the other.
The MLP has 32 or 128 hidden units and an $\ell_2$ penalty of $10^{-3}$ or $10^{-1}$, chosen on the validation split at every $k$ exactly as the linear probe chooses $\gamma$, and it trains with early stopping.
The control uses Gemma-2-9B, one seed, the full training split, and one layer per task.

\textbf{Design rationale.}
Both arms reuse the ranking, the grid, and the validation split of the linear arm, so a difference can come only from the probe family or from the labels.

\textbf{Null prediction.}
If the search could manufacture a finite count from noise, the permuted-label arm would reach AUROCs near the thresholds of the real tasks.
If linearity of the probe limited the count, the MLP arm would reach the same threshold with fewer features.

\begin{table}[t]
\AppTableStyle
\centering\small
\caption{\textbf{Probe-family and capacity controls on Gemma-2-9B.}
The MLP column reports $k^*$ at the linear arm's threshold.
The last column is the highest test AUROC reached at any $k$ on the grid after permuting the training labels.
This control uses one seed, the full training split, and one layer per task, so its absolute values differ from the five-seed values of Table~\ref{tab:gold}.}
\label{tab:probefamily}
\begin{tabular}{lrrrrrr}
\toprule
\rowcolor{AppGold}
\AppHead{Task} & \AppHead{Layer} & \AppHead{Linear ceiling} & \AppHead{Linear $k^*$} & \AppHead{MLP ceiling} & \AppHead{MLP $k^*$} & \AppHead{Shuffled labels} \\
\midrule
Language id & 35 & $1.0000$ & 1 & $1.0000$ & 1 & $0.535$ \\
Intent & 5 & $1.0000$ & 4 & $0.9999$ & 4 & $0.569$ \\
Jailbreak & 17 & $0.9999$ & 8 & $0.9983$ & 8 & $0.552$ \\
Banking intent & 20 & $0.9999$ & 8 & $0.9990$ & 8 & $0.577$ \\
Sentiment & 20 & $0.9798$ & 128 & $0.9825$ & 128 & $0.521$ \\
Fact check & 24 & $0.9093$ & 256 & $0.9094$ & 512 & $0.504$ \\
Harmfulness & 20 & $0.9785$ & 256 & $0.9808$ & 128 & $0.507$ \\
\bottomrule
\end{tabular}
\end{table}

\textbf{Result.}
Five of the seven tasks return the same count under both probe families.
Harmfulness improves by one grid step and fact checking worsens by one, and each grid step doubles $k$, so neither move exceeds the resolution of the search.
Shuffled labels reach at most 0.577 anywhere on the grid, far below every threshold of the real tasks, so the search does not manufacture a finite $k^*$.
Both arms share their data and their selection rule, which makes the comparison internal to this table.

\textbf{Scope.}
The comparison covers one model, one layer per task, and two MLP widths, and it omits the seed averaging, the subsampling, and the neighbor confirmation of the main protocol.

\paragraph{Estimator variants.}
\textbf{Procedure.}
We vary one estimator choice at a time and correlate the resulting task order against the main protocol by Equation~\ref{eq:rank-agreement}, separately in each of the 19 models.
The threshold variants replace the factor 0.95 of Equation~\ref{eq:threshold} by 0.90 or 0.975.
The search variant drops the L1 path of Algorithm~\ref{alg:search}.
The layer variant replaces the neighbor-confirmed count of Equation~\ref{eq:nc} by the smallest local count at any single layer.
Infinite counts take the tied maximal rank.

\textbf{Null prediction.}
By Lemma~\ref{lem:basic}(ii), the threshold variants move every count in a known direction, and by the invariance of the order statistic a uniform move leaves the order intact.
The correlations therefore test whether the moves are uniform enough across tasks to preserve the order.

\begin{table}[t]
\AppTableStyle
\centering\small
\caption{\textbf{Task order under estimator variants.}
Each entry is a Spearman correlation against the main protocol over the 14 core tasks, summarized over the 19 models.}
\label{tab:estimator}
\begin{tabular}{lrr}
\toprule
\rowcolor{AppGold}
\AppHead{Variant} & \AppHead{Median} & \AppHead{Range} \\
\midrule
Threshold at 90\% of the ceiling & $0.97$ & $0.93$ to $0.99$ \\
Threshold at 97.5\% of the ceiling & $0.97$ & $0.91$ to $0.99$ \\
Ranked-feature path only, no L1 path & $0.99$ & $0.91$ to $1.00$ \\
Best single layer, no neighbor confirmation & $0.98$ & $0.93$ to $0.99$ \\
\bottomrule
\end{tabular}
\end{table}

\textbf{Result.}
Every variant keeps the task order at a median of 0.97 or more over the 19 models, and the lowest single model reaches 0.91.
The estimator choices therefore move absolute counts but leave the task order nearly intact, while the representation and input changes of the upper block of Table~\ref{tab:protocol} move absolute counts by larger factors of 1.5 to 20.

\textbf{Finding.}
The controls separate two kinds of dependence on the protocol.
Estimator choices keep the task order at a median correlation of 0.97 or more.
Changes of basis and wording move absolute counts by factors of 1.5 to 20, while ceilings move by at most 0.03 and the order of the tested cells holds.
The count is therefore a property of a task read in a fixed basis on a fixed set of texts, and its rank among tasks is the part that transfers across models.

\textbf{Implication.}
A reported $k^*$ should state its basis, its input format, and its text pool, and comparisons of absolute counts should hold all three fixed.
Among the tested factors, wording and basis move counts the most, so they deserve the first control in any new study.

\subsubsection{Bag-of-Words Input Baseline}
\label{app:inputbase}

\textbf{Question.}
This baseline asks whether a sparse readout merely restates that a task is easy to decide from its raw text alone, with no learned representation.

\textbf{Procedure.}
We run the entire probing pipeline on TF-IDF features of the raw text over all 24 tasks, with unigrams and bigrams, sublinear term frequency, and a vocabulary capped at 30000 terms.
The baseline uses the same dense probe, the same prefix grid, the same L1 path, and the same threshold as the model side, and Gemma-2-9B serves as the model-side instrument.

\textbf{Design rationale.}
Running the identical pipeline on the raw text gives the model a reference instrument with no learned representation.
A task whose model-side reading merely reflected how easily its text decides it would show the same ceiling and the same prefix curve on both instruments.

\textbf{Estimators.}
Let $A_t(k)$ denote the best test AUROC at $k$ features for task $t$ in one representation, and let $C_t$ be that representation's own ceiling.
Two crossing points summarize a curve,
\begin{equation}
    k^{\mathrm{first}}_t=\min\{k:A_t(k)\ge\tau_t\},
    \qquad
    k^{\mathrm{stable}}_t=\min\{k:A_t(k')\ge\tau_t\ \text{for all}\ k'\ge k\},
    \label{eq:crossings}
\end{equation}
and they coincide exactly when the curve never falls back below $\tau_t$ after its first crossing.
The robust statistic is the prefix score, the running maximum of the normalized AUROC,
\begin{equation}
    \bar s_t(k)=\max_{k'\le k}\ \frac{A_t(k')-0.5}{C_t-0.5},
    \label{eq:prefix-score}
\end{equation}
which is non-decreasing in $k$ by construction and moves only when a new point exceeds all earlier points on the curve.
We compare the two representations at a fixed $k$ of 64 or 256 with a two-sided Wilcoxon signed-rank test~\citep{wilcoxon1945individual} over the 24 tasks on the per-task differences of prefix scores $\delta_t=\bar s^{\mathrm{model}}_t(k)-\bar s^{\mathrm{text}}_t(k)$, dropping tasks with $\delta_t=0$.

\textbf{The ceiling axis.}
Input-side and model-side ceilings correlate at $\rho=0.950$ with $p=1.3\times10^{-12}$ over the 24 tasks.
The model exceeds the input baseline by a median of 0.026 AUROC.
The largest gains are $+0.153$ on hate speech, $+0.133$ on relevance, and $+0.130$ on fact checking, and the only negative value is $-0.033$ on emotion.
Ceilings move by at most 0.0024 between independent runs, so this comparison rests on a stable statistic.
Readability is therefore mostly a property of the text, and the ceiling axis measures what the text decides at least as much as what the model contributes.

\textbf{Why we report no input-side $k^*$.}
TF-IDF accuracy is flat and non-monotone near the threshold, so $k^{\mathrm{first}}_t$ is decided by differences of order 0.001.
In 14 of the 24 tasks, $k^{\mathrm{first}}_t\neq k^{\mathrm{stable}}_t$.
Rerunning the same pipeline with a different task list moves fact checking from 5591 features to 512 while its ceiling moves from 0.7885 to 0.7870, and moves harmfulness from 5705 to 8233 at an unchanged ceiling.
Fact checking crosses at 512 by $+0.0012$ and falls back by $-0.0066$ at 1024.
Every per-task ratio between input-side and model-side counts therefore rests on a crossing point that the data do not identify, and we report no such ratio.

\begin{table}[t]
\AppTableStyle
\centering\small
\caption{\textbf{Fraction of its own ceiling each representation reaches within $k$ features.}
Entries are medians over the 24 tasks of the prefix score of Equation~\ref{eq:prefix-score}, the model is Gemma-2-9B, and the $p$ value comes from a two-sided Wilcoxon signed-rank test over tasks.}
\label{tab:prefix}
\begin{tabular}{lrrrr}
\toprule
\rowcolor{AppGold}
\AppHead{Features} & \AppHead{Bag of words} & \AppHead{Model} & \AppHead{Model higher} & \AppHead{Wilcoxon $p$} \\
\midrule
64 & $0.919$ & $0.991$ & 16 of 24 & $0.029$ \\
256 & $0.969$ & $0.998$ & 13 of 24 & $0.099$ \\
\bottomrule
\end{tabular}
\end{table}

\textbf{Result.}
The model reaches more of its own ceiling within 64 features, and the difference is no longer significant at 256.
The comparison favors the model in one respect and the text in another.
The model selects from 3584 residual coordinates while the baseline selects from up to 30000 terms, so a larger pool of candidates serves the baseline.
Each residual coordinate, however, already aggregates the whole input, while each term reads one surface pattern.
The table therefore describes the two representations under a common search and does not isolate the model's contribution.

\textbf{Scope.}
The same bag-of-words readings serve as the text-level control of the cross-model task order in Appendix~\ref{app:ordering}, where they predict part of each model's order and leave the median cross-lab partial correlation at 0.86 to 0.92 against 0.93 without control.

\textbf{Finding.}
Dense readability is mostly a property of the text.
Input-side and model-side ceilings correlate at 0.950, and the model adds a median of 0.026 AUROC.
The model's advantage lies in sparsity, since within 64 features it reaches 0.991 of its own ceiling against 0.919 for the text.

\textbf{Implication.}
A probing study should report a text-only baseline under the same pipeline, since a high ceiling alone says little about what the model contributes.
On flat, non-monotone curves such as those of the TF-IDF baseline, the first crossing point of a threshold is not identified, and the running-maximum prefix score of Equation~\ref{eq:prefix-score} is the statistic to report instead.

\subsubsection{Cross-Model Feature Matching}
\label{app:matching}

\textbf{Question.}
Equal counts in two models leave open whether the two models use corresponding features.
This section asks whether the SAE features that each model selects for a class decision vary alike across individual inputs, beyond what their shared label explains.

\begin{table}[t]
\AppTableStyle
\centering\small
\caption{\textbf{Where feature correspondence can be tested and what it shows.}
The upper block reports the three tasks the matching test covers.
Matching runs over the eight base models with an SAE suite, and one test is one model pair on one class decision.
The dictionary columns report the median over three models of a held-out class decision's recovery and of its same-size random-dictionary null.
The double dagger marks tool selection, whose four class decisions give a leave-one-out dictionary of a median 10 features, so its recovery reflects the power of the protocol rather than a negative result.
The lower block lists the remaining core tasks with the number of the eight models that hold a support.
Sections~\ref{app:matching} and~\ref{app:dictionary} give the per-pair and per-model readings.}
\label{tab:identity}
\resizebox{\linewidth}{!}{%
\begin{tabular}{lrrrrrr}
\toprule
\rowcolor{AppGold}
 & \multicolumn{4}{c}{\AppHead{Cross-model matching}} & \multicolumn{2}{c}{\AppHead{Dictionary}} \\
\cmidrule(lr){2-5}\cmidrule(lr){6-7}
\rowcolor{AppGold} \AppHead{Task} & \AppHead{Models} & \AppHead{Tests} & \AppHead{Match} & \AppHead{Median $z$} & \AppHead{Recovery} & \AppHead{Null} \\
\midrule
Intent classification & 8 & 447 & 56.2\% & $+3.08$ & $0.984$ & $0.630$ \\
Language identification & 8 & 280 & 48.9\% & $+3.71$ & $0.996$ & $0.022$ \\
Tool selection & 8 & 105 & 34.3\% & $+1.68$ & $0.533^{\ddagger}$ & $0.020$ \\
\midrule
\multicolumn{7}{l}{\emph{Core tasks the matching test cannot cover}} \\
Routing & 7 & 21 & \multicolumn{4}{l}{supports of 1 to 2 features, every test degenerate} \\
PII detection & 3 & 3 & \multicolumn{4}{l}{too few models with a support} \\
Jailbreak detection & 2 & 1 & \multicolumn{4}{l}{too few models with a support} \\
Banking intent, harmfulness, sentiment & 1 & 0 & \multicolumn{4}{l}{one model each, no pair} \\
Relevance, fact check, code bug, math, step check & 0 & 0 & \multicolumn{4}{l}{no SAE support in any model} \\
\bottomrule
\end{tabular}}
\end{table}

\textbf{Data and models.}
Matching runs on the eight base models with an available SAE suite, namely Gemma-3-270M, Gemma-3-1B, Gemma-3-4B, Gemma-2-9B, Gemma-3-12B, Gemma-3-27B, Llama-3.1-8B, and Qwen3-8B.
They give 28 pairs, of which 13 cross labs and 15 pair two Gemma models.
The shared inputs are the tasks' test pools, which contain 3000 texts each for intent classification and language identification and 840 texts for tool selection.

\textbf{Supports.}
Supports come from the SAE protocol of Appendix~\ref{app:sae}, read at the last token.
A class decision holds a support only when some subset of SAE features within the search reaches 95\% of that decision's best probe in any representation at the task's neighbor-confirmed SAE layer.
A class decision enters a pair only when both models hold a support for it.
The number of tests per pair is therefore the size of that intersection, which ranges from 11 to 19 on intent classification.

\textbf{Estimator.}
For a model pair $(A,B)$ and class decision $c$ with binary labels $y\in\{0,1\}^n$, let $X^A\in\mathbb R^{n\times k_A}$ and $X^B\in\mathbb R^{n\times k_B}$ contain the selected features on the $n$ shared test inputs, and let $n_v$ be the number of inputs with label $v$.
Shared label responses can by themselves induce activation agreement between two models, so we compare response profiles within each class.
We remove each feature's mean within each class and scale by the resulting standard deviation,
\begin{equation}
 \widetilde X^A_{ij}=X^A_{ij}-\frac{1}{n_{y_i}}\sum_{u:y_u=y_i}X^A_{uj},
 \qquad P^A_{ij}=\frac{\widetilde X^A_{ij}}{\operatorname{sd}(\widetilde X^A_{:j})},
 \qquad R=\frac{|(P^A)^\top P^B|}{n},
 \label{eq:matching-profiles}
\end{equation}
where $\operatorname{sd}$ uses the denominator $n$, the absolute value acts entrywise, and a zero standard deviation is replaced by one so that a constant centered column stays at zero.
With $q=\min(k_A,k_B)$ and $\mathcal M_q$ the one-to-one matchings of cardinality $q$ between the columns of $P^A$ and $P^B$, the statistic is
\begin{equation}
 T(A,B;c)=\max_{M\in\mathcal M_q}\frac1q\sum_{(i,j)\in M}R_{ij}.
 \label{eq:matching-score}
\end{equation}
The Hungarian algorithm~\citep{kuhn1955hungarian} solves this rectangular assignment exactly.

\begin{applemma}[what the matching statistic ignores]
\label{lem:matching}
The following hold for Equations~\ref{eq:matching-profiles} and~\ref{eq:matching-score}.
\begin{enumerate}
\item For any functions $f_j:\{0,1\}\to\mathbb R$, replacing $X^A_{ij}$ by $X^A_{ij}+f_j(y_i)$ leaves $P^A$ unchanged.
\item For any nonzero scalars $a_j$, replacing $X^A_{:j}$ by $a_jX^A_{:j}$ leaves the $j$th row of $R$ unchanged.
\item Every entry of $R$ lies in $[0,1]$, so $T(A,B;c)\in[0,1]$, and $T(A,B;c)=T(B,A;c)$.
\end{enumerate}
\end{applemma}

\begin{appproof}
For the first claim, the added term is constant within each class, so the within-class centering of Equation~\ref{eq:matching-profiles} removes it exactly.
For the second claim, scaling by $a_j$ scales the centered column and its standard deviation by $a_j$ and $|a_j|$, so $P^A_{:j}$ changes at most in sign, and the absolute value removes the sign.
For the third claim, every nonconstant column of $P^A$ has mean zero and $\lVert P^A_{:j}\rVert_2^2=n$, so $R_{ij}$ is an absolute Pearson correlation and the Cauchy--Schwarz inequality bounds it by 1.
A constant column gives $R_{ij}=0$.
Transposing $R$ maps matchings of $(A,B)$ onto matchings of $(B,A)$ with the same value.
\end{appproof}

Lemma~\ref{lem:matching} states precisely what a high matching score can and cannot reflect.
The first claim shows that agreement in class means, however large, contributes nothing, so a high score requires agreement in how responses vary among inputs of the same class.
The second claim shows that the statistic ignores the scale and the sign of each selected feature, so a high score does not imply identical activation signs or identical computations in the two models.

\textbf{Design rationale.}
Two models that both read a label correlate on every input merely because both respond to that label.
The statistic removes this source by construction, as Lemma~\ref{lem:matching} shows, and the null removes the next source, namely that any features correlated with the label might agree within class as well.
Drawing the null from unselected features of the same model keeps the dictionary, the layer, and the inputs fixed, so the comparison isolates the selection itself.

\textbf{Null.}
Two selected sets can agree within class merely because both contain features correlated with the label, so the null must match label correlation.
For each selected feature $j$ of model $B$, let $\lambda_j=|\operatorname{corr}(X^B_{:j},y)|$ be its absolute point-biserial correlation with the label on the shared inputs.
We form a candidate list $\mathcal N_j$ from the 50 nonconstant, unselected features of $B$ whose label correlations lie closest to $\lambda_j$.
Each null draw $b$ samples one replacement uniformly from each list $\mathcal N_j$, independently across $j$, and recomputes Equation~\ref{eq:matching-score} with model $A$'s profiles unchanged.
Independent sampling can draw the same replacement for two selected features, so a null set is a collection with multiplicity.
The construction preserves the number of columns and approximately preserves their label correlations.
With 1000 draws $T^{(1)},\dots,T^{(1000)}$, we define the exceedance indicator and the standardized score,
\begin{equation}
 I_{ABc}=\mathbf1\bigl\{T(A,B;c)>Q_{0.99}(T^{(1)},\ldots,T^{(1000)})\bigr\},\qquad
 z_{ABc}=\frac{T(A,B;c)-\overline T_{\rm null}}{\operatorname{sd}(T_{\rm null})+10^{-12}},
 \label{eq:matching-null}
\end{equation}
where $Q_{0.99}$ is the empirical 0.99 quantile with linear interpolation and the exceedance is strict.

\textbf{Null prediction.}
If the selected features of $B$ corresponded to those of $A$ no better than label-matched unselected features do, $T(A,B;c)$ would be exchangeable with the null draws up to the approximation in the matching of label correlations.
The pass rate would then sit near the nominal 1\%.
This 1\% is a nominal tail expectation for the resampling construction and not a separately calibrated error rate.

\begin{algorithm}[t]
\AppAlgorithmStyle
\caption{Cross-model correspondence with a label-correlation-matched null}
\label{alg:matching}
\begin{algorithmic}[1]
\Require ordered models $A,B$, selected supports, aligned test inputs and binary labels
\State Verify that the label vectors agree and retain class decisions with supports on both sides
\For{each retained class decision $c$}
 \State Center and scale the selected profiles and compute $T(A,B;c)$ by Equation~\ref{eq:matching-score}
 \State Form the 50-feature candidate lists $\mathcal N_j$ from model $B$'s nonconstant unselected features
 \State Seed the generator with the run seed and the CRC32 checksum of the ordered pair and the class name
 \For{$b=1,\ldots,1000$}
  \State Sample one replacement per selected $B$ feature independently, keeping multiplicities
  \State Recenter, rescale, and solve the rectangular assignment to obtain $T^{(b)}$
 \EndFor
 \State Record $I_{ABc}$, $z_{ABc}$, and both support sizes by Equation~\ref{eq:matching-null}
\EndFor
\end{algorithmic}
\end{algorithm}

\textbf{Aggregation.}
For a task with eligible comparisons $\mathcal E$, each one model pair on one class decision with supports on both sides, the pooled pass rate and the pair-level pass rate are
\begin{equation}
    \widehat p=\frac{1}{|\mathcal E|}\sum_{(A,B,c)\in\mathcal E}I_{ABc},
    \qquad
    \widehat p_{AB}=\frac{1}{|\mathcal C_{AB}|}\sum_{c\in\mathcal C_{AB}}I_{ABc},
    \label{eq:pass-rate}
\end{equation}
where $\mathcal C_{AB}$ is the set of class decisions eligible for the pair $(A,B)$.
The pooled rate weights every test equally, and a median of $\widehat p_{AB}$ over pairs weights every pair equally.
The order of the model list fixes which side supplies replacements, so reversing that order can change the null and the reported rates even though the observed score is symmetric by Lemma~\ref{lem:matching}.
Seeding each pair and class decision separately ensures that adding a model does not change the random draws of existing pairs.

\textbf{Dependence.}
Tests that share a model or a class decision share data, so the tests in $\mathcal E$ are not independent.
We therefore read $\widehat p$ as a descriptive rate against the nominal 1\% and do not convert it into a single $p$ value.
The test labels enter both the centering and the null construction, so the procedure describes correspondence on the shared evaluation pool.

\begin{figure}[t]
\centering
\includegraphics[width=\linewidth]{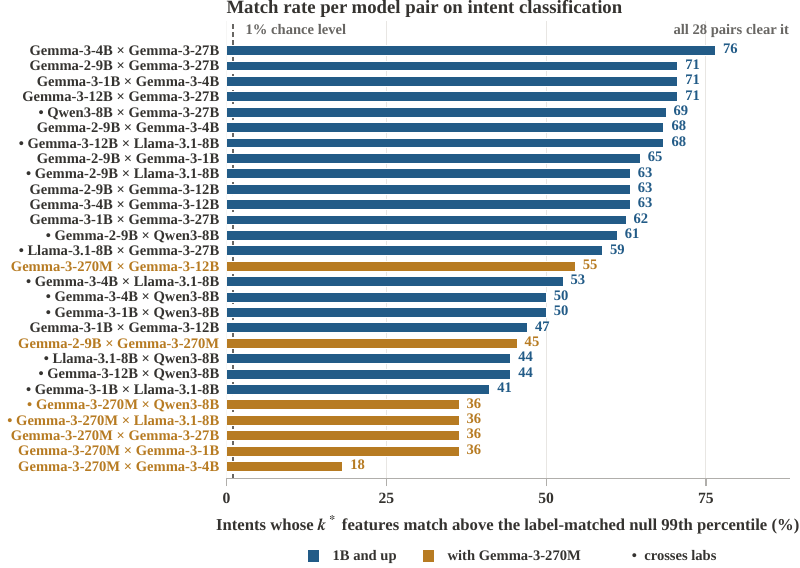}
\caption{\textbf{Match rate for every model pair on intent classification.}
Each bar is the fraction of that pair's shared intents whose one-to-one feature match exceeds the label-correlation-matched null's 99th percentile.
The dashed line marks the 1\% rate the null allows, and every pair clears it.
Dots mark cross-lab pairs.
Pairs containing Gemma-3-270M occupy most of the lower half.}
\label{fig:pair-matching}
\end{figure}

\textbf{Result by task.}
Intent classification exceeds the null in 56.2\% of 447 tests at a median $z$ of $+3.08$, with 84\% of all $z$ scores positive.
Language identification reaches 48.9\% of 280 at $+3.71$, and tool selection reaches 34.3\% of 105 at $+1.68$.
The 13 cross-lab pairs match at median rates of 0.50, 0.30, and 0.50 on the three tasks, with ranges of 0.36 to 0.69, 0.20 to 0.50, and 0.00 to 0.75, so one cross-lab pair matches on no tool-selection decision.
The strongest cross-lab pair on intent classification is Qwen3-8B with Gemma-3-27B at a match rate of 0.69, and the highest median $z$ of the whole table is $+4.66$ on Gemma-3-12B with Llama-3.1-8B at a rate of 0.68.
Figure~\ref{fig:pair-matching} lists all 28 pairs.

\textbf{Degenerate comparisons.}
Language identification needs one further split, because at least one support has size 1 in 245 of its 280 tests, and both supports have size 1 in 189 of them.
With $q=1$ the statistic reduces to one absolute correlation when both supports have size 1, and to the largest absolute correlation between the single feature and the other side's features otherwise, and in either case it must still exceed the null threshold to pass.
Seven of the ten language decisions have single-feature supports in most eligible model pairs.
Its 35 tests with larger supports reach 60.0\% at a median $z$ of $+4.92$, and the 66 such tests of tool selection reach 30.3\% at $+1.48$.

\begin{figure}[t]
\centering
\includegraphics[width=\linewidth]{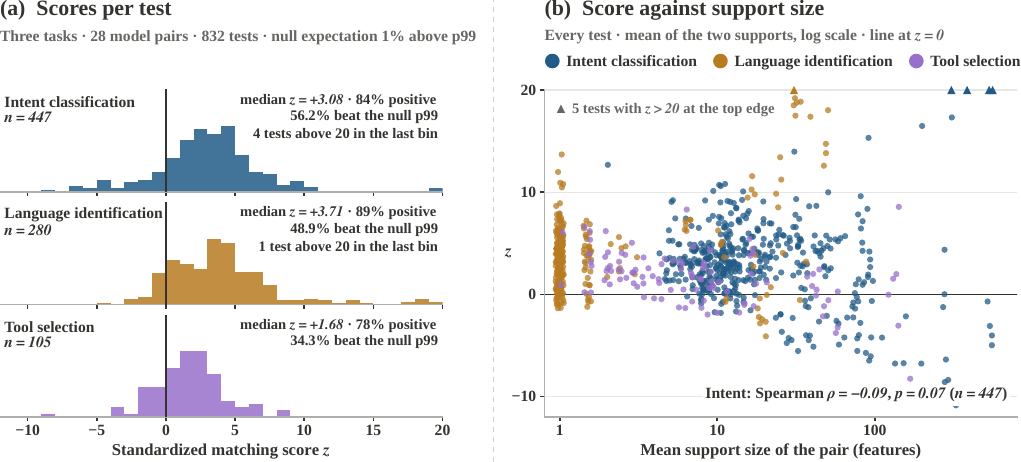}
\caption{\textbf{Cross-model SAE feature correspondence, test by test.}
Each test pairs two of the eight base models on one class decision and compares the observed one-to-one feature match with a null of equally sized feature sets matched in label correlation.
(a) Histograms of the standardized score $z$ over the 447 intent classification tests, the 280 language identification tests, and the 105 tool selection tests.
The median $z$ is $+3.08$, $+3.71$, and $+1.68$, positive scores make up 84\%, 89\%, and 78\% of each task, and 56.2\%, 48.9\%, and 34.3\% of tests exceed the null 99th percentile where the null allows 1\%.
Four intent tests and one language test with $z$ above 20 fall in the last bin.
(b) The same 832 tests plotted against the mean of the two support sizes on a logarithmic axis.
Tests with equal support sizes are jittered horizontally, and the five tests with $z$ above 20 appear as triangles at the top edge.
Over the 447 intent tests the Spearman correlation between support size and $z$ is $-0.09$ with $p=0.07$.}
\label{fig:app-matching}
\end{figure}

\textbf{Result by class decision.}
Match rates in Table~\ref{tab:perintent} vary far more across the 19 intents, from 0.14 to 1.00, than across the 28 pairs, from 0.18 to 0.76.
Figure~\ref{fig:app-matching} shows the distribution of the standardized scores of all 832 tests and their relation to the mean support size.

\begin{table}[t]
\AppTableStyle
\centering\small
\caption{\textbf{Per-intent match rates on intent classification.}
The match rate is the fraction of the pairs holding a support for that intent whose observed match exceeds the null's 99th percentile.
Support is the median over models of the two supports' mean size in features.}
\label{tab:perintent}
\resizebox{\linewidth}{!}{%
\begin{tabular}{lrrr@{\hskip 1.5em}lrrr}
\toprule
\rowcolor{AppGold}
\AppHead{Intent} & \AppHead{Match} & \AppHead{Median $z$} & \AppHead{Support} & \AppHead{Intent} & \AppHead{Match} & \AppHead{Median $z$} & \AppHead{Support} \\
\midrule
smart\_home & $1.00$ & $+5.56$ & 38 & update\_playlist & $0.46$ & $+2.52$ & 8 \\
ingredients\_list & $0.96$ & $+5.19$ & 12 & spelling & $0.40$ & $+1.59$ & 72 \\
sync\_device & $0.93$ & $+4.23$ & 12 & weather & $0.39$ & $+2.26$ & 10 \\
traffic & $0.79$ & $+5.05$ & 16 & yes & $0.33$ & $-1.94$ & 50 \\
account\_blocked & $0.75$ & $+4.55$ & 14 & confirm\_reservation & $0.33$ & $-5.73$ & 155 \\
goodbye & $0.71$ & $+3.69$ & 28 & change\_volume & $0.33$ & $-1.14$ & 63 \\
credit\_limit & $0.71$ & $+3.25$ & 8 & order\_status & $0.32$ & $+1.12$ & 10 \\
income & $0.68$ & $+4.51$ & 31 & reset\_settings & $0.18$ & $+1.87$ & 11 \\
find\_phone & $0.64$ & $+4.04$ & 8 & roll\_dice & $0.14$ & $+1.28$ & 12 \\
tell\_joke & $0.62$ & $+3.56$ & 10 & & & & \\
\bottomrule
\end{tabular}}
\end{table}

Seven intents match in at least 71\% of their pairs and six in at most 34\%.
Every intent whose median support reaches 50 features matches in at most 40\% of pairs.
A small support does not guarantee a match, since roll\_dice, reset\_settings, and order\_status have supports of 10 to 12 and match in 14 to 32\% of pairs.
The rank correlation between support size and $z$ over all 447 tests is only $-0.09$ with $p=0.07$, so support size predicts the extremes rather than the whole table.

\paragraph{Subsampling control for large supports.}
\textbf{Question.}
One-to-one matching may become conservative as supports grow, since the average of Equation~\ref{eq:matching-score} runs over more matched pairs and a large assignment has more ways to fall short of the null.
This control asks whether the low scores of large supports reflect that property of the statistic or an absence of correspondence.

\textbf{Procedure.}
We cut the second model's support to 20 features at random, redraw five times, keep the first model's support whole, and draw the null at the reduced size.

\textbf{Null prediction.}
If large supports failed only because of the conservativeness of the statistic, reduction to 20 features would restore their pass rates to those of the three converging intents.

\begin{table}[t]
\AppTableStyle
\centering\small
\caption{\textbf{Median $z$ before and after reducing every support to 20 features.}
The upper group matches at full size and the lower group, the four intents with supports of 50 or more, does not.}
\label{tab:subsample}
\begin{tabular}{lrrr}
\toprule
\rowcolor{AppGold}
\AppHead{Intent} & \AppHead{Support} & \AppHead{Full-size $z$} & \AppHead{Reduced $z$} \\
\midrule
smart\_home & 38 & $+5.56$ & $+3.1$ \\
income & 31 & $+4.51$ & $+3.8$ \\
goodbye & 28 & $+3.69$ & $+2.7$ \\
\midrule
spelling & 72 & $+1.59$ & $-0.6$ \\
change\_volume & 63 & $-1.14$ & $-0.7$ \\
yes & 50 & $-1.94$ & $-1.4$ \\
confirm\_reservation & 155 & $-5.73$ & $-1.4$ \\
\bottomrule
\end{tabular}
\end{table}

\textbf{Result.}
We make two observations on Table~\ref{tab:subsample}.
First, the three converging intents still exceed the null in 57\% of their reduced tests, so reduction preserves a real signal.
Second, the four intents with large supports reach only 12 to 27\%, so their low rate is not an artifact of support size alone, although that rate still exceeds the nominal 1\% and is consistent with partial correspondence.
The deep negative scores do disappear, so their magnitude measures the conservativeness of the statistic on large sets and carries no evidence against correspondence.

\paragraph{Model scale.}
We group the 28 pairs by the smallest model they contain and take the median of $\widehat p_{AB}$ in each group, separately for each of the three tasks.

\begin{table}[t]
\AppTableStyle
\centering\small
\caption{\textbf{Median match rate by the scale of the pair.}
The three groups partition the 28 pairs.
A pair enters a task's column only when that task has at least one test for it.}
\label{tab:scale}
\begin{tabular}{lrrrr}
\toprule
\rowcolor{AppGold}
\AppHead{Group} & \AppHead{Pairs} & \AppHead{Intent} & \AppHead{Language id} & \AppHead{Tool selection} \\
\midrule
Contains the 270M model & 7 & $0.36$ & $0.40$ & $0.33$ \\
Contains the 1B model & 6 & $0.56$ & $0.65$ & $0.25$ \\
Both models at 4B or above & 15 & $0.63$ & $0.40$ & $0.50$ \\
\bottomrule
\end{tabular}
\end{table}

Only intent classification rises with model scale, from 0.36 to 0.63, and the ten pairs among models of 7B and above give the same median of 0.63.
Language identification peaks in the middle group, at 0.65, tool selection falls and then rises, and the three groups overlap in their ranges, so these readings support no scale threshold and no general scale trend.

\paragraph{Coverage.}
Only intent classification, language identification, and tool selection provide at least 100 tests each among the 14 core tasks.
The remaining tasks lack enough SAE supports or comparisons with more than one feature on each side.
Routing has one class decision with supports of one or two features, so its 21 pairwise tests are single, mostly degenerate comparisons, and we report its coverage without a match rate.
Sentiment returns zero testable cells rather than a negative result.
Only Qwen3-8B records a support for it, with 32 features, while the other seven models place their neighbor-confirmed SAE layer at layer 1, 3, or 5 and reach no qualifying subset there.
The five tasks listed last in Table~\ref{tab:identity} hold an SAE support on none of the eight models, and their residual-basis ceilings of 0.62 to 0.89 in Table~\ref{tab:gold} show that the models read them poorly in every basis we have.
The correspondence test therefore applies to readable tasks with SAE supports in at least two models.

\textbf{Finding.}
Selected SAE features vary alike across models among inputs of the same class.
After class means are removed and against a null matched in label correlation, 56.2\% of intent tests pass where the null allows 1\%, and the 13 cross-lab pairs pass at a median rate of 0.50.
Correspondence varies more across class decisions, from 0.14 to 1.00, than across model pairs, from 0.18 to 0.76.
Earlier comparisons of units or features across networks match them by activation correlation on shared inputs~\citep{li2015convergent,lan2024quantifying}, and such correlations include the agreement that a shared label induces.
The present test removes that agreement exactly, by Lemma~\ref{lem:matching}, before any feature is matched.

\textbf{Implication.}
Cross-model feature matching should remove class means and compare against a label-correlation-matched null, otherwise two models that merely read the same label appear to share features.
Agreement under this test concerns measured responses on shared inputs and does not by itself establish a shared computation across the two models.

\subsubsection{Leave-One-Out Dictionary Test}
\label{app:dictionary}

\textbf{Question.}
Within one model, the class decisions of a multiclass task each have their own SAE support.
This test asks whether those supports draw on a shared family dictionary or whether each class decision carries private features that no other class decision uses.

\textbf{Procedure.}
For a class decision $i$ with support $S_i$, we take the union of the other class decisions' supports as a dictionary and refit $i$'s probe on those features alone,
\begin{equation}
    U_{-i}=\bigcup_{j\neq i}S_j,
    \qquad
    \mathrm{rec}_i=\frac{A_i(U_{-i})-0.5}{C_i-0.5},
    \qquad
    \mathrm{own}_i=\frac{|S_i\cap U_{-i}|}{|S_i|},
    \label{eq:loo-dictionary}
\end{equation}
where $A_i(U)$ is the test AUROC of the probe of Section~\ref{app:probes} restricted to $U$ and $C_i$ is the ceiling of class decision $i$.
A recovery of 1 means that a class decision that contributed nothing to the dictionary is still fully expressible in it.
Own overlap separates composition from plain inclusion, since a class decision whose support already lies in the dictionary recovers trivially.
The null replaces $U_{-i}$ by a random set of $|U_{-i}|$ SAE features, drawn with five seeds.
The width of the expression is the smallest prefix of the dictionary, ranked by the $t$-statistic of Equation~\ref{eq:tstat} within it, that clears the threshold $\tau_i$.
The test uses the SAE layers 28 of Gemma-2-9B, 23 of Llama-3.1-8B, and 23 of Qwen3-8B.

\textbf{Design rationale.}
Leaving the class decision out of its own dictionary prevents trivial recovery, and own overlap measures how much trivial recovery remains.
The random dictionary of equal size separates composition from the generic power of many features.

\textbf{Criteria and null prediction.}
Before running the test we fixed two criteria, a median recovery of at least 0.9 and recovery above the random null.
If each class decision relied on private features, recovery from $U_{-i}$ would stay near the random null once own overlap is low.

\begin{table}[t]
\AppTableStyle
\centering\small
\caption{\textbf{Leave-one-out dictionary recovery.}
Recovery, the null, and own overlap are medians over class decisions, and the minimum is the lowest recovery of any class decision.}
\label{tab:dictionary}
\resizebox{\linewidth}{!}{%
\begin{tabular}{llrrrrrrr}
\toprule
\rowcolor{AppGold}
\AppHead{Task} & \AppHead{Model} & \AppHead{Decisions} & \AppHead{Dictionary} & \AppHead{$k^*$} & \AppHead{Recovery} & \AppHead{Minimum} & \AppHead{Null} & \AppHead{Own overlap} \\
\midrule
Intent & Llama-3.1-8B & 20 & 835 & 17 & $0.984$ & $0.790$ & $0.630$ & $0.16$ \\
       & Gemma-2-9B & 19 & 499 & 12 & $0.981$ & $0.897$ & $0.766$ & $0.11$ \\
       & Qwen3-8B & 18 & 1428 & 22 & $0.989$ & $0.889$ & $0.562$ & $0.24$ \\
\midrule
Language id & Llama-3.1-8B & 10 & 15 & 1 & $0.999$ & $0.784$ & $0.012$ & $0.00$ \\
            & Gemma-2-9B & 10 & 33 & 1 & $0.990$ & $0.955$ & $0.079$ & $0.00$ \\
            & Qwen3-8B & 10 & 39 & 1 & $0.996$ & $0.654$ & $0.022$ & $0.08$ \\
\midrule
Tool selection & Llama-3.1-8B & 4 & 3 & 1 & $0.533$ & $0.331$ & $0.020$ & $0.00$ \\
               & Gemma-2-9B & 4 & 22 & 4 & $0.347$ & $0.247$ & $0.091$ & $0.00$ \\
               & Qwen3-8B & 4 & 10 & 3 & $0.548$ & $0.461$ & $0.001$ & $0.00$ \\
\bottomrule
\end{tabular}}
\end{table}

\textbf{Result.}
On intent classification, 18 of 20, 17 of 19, and 17 of 18 class decisions reach a recovery of 0.9 on Llama-3.1-8B, Gemma-2-9B, and Qwen3-8B.
All 57 class decisions across the three models exceed their own random control, with Wilcoxon $p$ values of $1.9\times10^{-6}$, $3.8\times10^{-6}$, and $7.6\times10^{-6}$ in the same order.
Own overlap stays between 0.11 and 0.24, so 76 to 89\% of a held-out support lies outside the dictionary and is nonetheless expressible within it.
Language identification reproduces the result on three models at an own overlap of at most 0.08.

\textbf{Composability and minimality.}
Among the class decisions for which some prefix of the dictionary clears the threshold, which are 11 of 20, 14 of 19, and 14 of 18 on the three models, reading a held-out intent from the dictionary takes a median of 64 to 128 features, while the per-model medians of the intents' own shortest supports are 12 to 22, and the median over class decisions of the per-decision ratio is 3.8 to 5.2.
A shared dictionary can express every member of the family, and each member also has a shorter private support.
The quantity $k^*$ measures the second of these.

\textbf{Scope.}
The test has power only when a task has enough class decisions to build a large dictionary.
Tool selection has four class decisions, so its leave-one-out dictionary holds a median of 3 to 22 features and no class decision recovers fully.
The number of class decisions sets the dictionary size and the dictionary size sets the power of the test, so this reading describes the protocol on tasks with few class decisions and not a negative result.
The test also needs a basis in which random subsets are insufficient.
Run on residual coordinates it returns a recovery of 0.997 against a random control of 0.994, because a few hundred of the 3584 coordinates already read an intent, and the comparison between dictionary and random control loses all resolution.

\textbf{Finding.}
The class decisions of a family compose from a shared family dictionary and also keep short private supports.
A held-out intent recovers a median of 0.98 of its performance from the other intents' features on each of three models with an own overlap of at most 0.24, yet its own support is 3.8 to 5.2 times shorter than the dictionary readout.

\textbf{Implication.}
A shared family dictionary and a minimal per-decision readout are distinct quantities, and the count $k^*$ measures only the second of the two.

\subsubsection{Behavioral Interventions}
\label{app:interventions}

\textbf{Question.}
A feature that predicts a label need not affect what the model does.
This section asks whether removing a selected support changes the behavior that the class decision describes, and whether equally large edits in other directions leave the behavior intact.

\paragraph{Editing the residual stream.}
Every SAE intervention edits the output of one transformer block at every token position, for the prompt and for each generated token.
For a removed set $J$ of SAE features, a scale $\alpha>0$, and a residual vector $h$, the treatment is
\begin{equation}
    h\ \mapsto\ h-\alpha\,\Delta_J(h),
    \qquad
    \Delta_J(h)=\sum_{j\in J}g_j(h)\,W_{\mathrm{dec},j},
    \label{eq:sae-ablation}
\end{equation}
where $g_j(h)$ is the JumpReLU activation of Equation~\ref{eq:jumprelu} recomputed from the current residual vector and $W_{\mathrm{dec},j}$ is the decoder column of feature $j$.
At $\alpha=1$ the edit removes exactly the reconstruction that the features of $J$ contribute and leaves the reconstruction of every other feature and the reconstruction error in place.
Frequency-matched controls replace $J$ by unselected features drawn without replacement, one per removed feature, from its 50 nearest neighbors in activation frequency on the training encodings, with three independent draws.
The equal-norm control subtracts a fixed random unit direction $u$ scaled at each position to the size of the treatment,
\begin{equation}
    h\ \mapsto\ h-\alpha\,\lVert\Delta_J(h)\rVert_2\,u,
    \label{eq:equal-norm}
\end{equation}
so it matches the size of the treatment edit at every token position and differs from it only in direction.
Generation is greedy through the model's chat template with at most 48 new tokens.
Every condition uses 200 prompts per class from the task's held-out split.

\paragraph{Reply language.}
\textbf{Estimator.}
The decision variable is a one-vs-rest language identification class decision for a language $L$, and the behavior is the script of the model's reply.
We read
\begin{equation}
    \mathrm{gap}_L = P(\text{reply in } L \mid \text{input in } L) - P(\text{reply in } L \mid \text{input not in } L),
    \label{eq:reply-gap}
\end{equation}
and estimate each probability by its empirical frequency over the 200 prompts of the corresponding class.
The 95\% interval of the gap is the 2.5 to 97.5 percentile range of 10000 parametric bootstrap replicates~\citep{efron1992bootstrap}, each of which redraws the two class frequencies independently from binomial distributions with the observed rates and 200 trials.
A treatment separates from a control when the two intervals do not overlap.
A reply counts as being in $L$ when it has at least 8 alphabetic characters and at least 0.3 of them fall in $L$'s script.

\textbf{Design choices.}
The gap needs no second language classifier as long as the script identifies the language, so we keep only script-unique languages.
This excludes Arabic and Urdu, which share a script, and the four Latin-script languages of the pool.
A gap near zero means that the target-language reply probabilities are similar under the two input conditions, and it does not by itself imply that the model can no longer produce the language.
The layer comes from the protocol's own neighbor confirmation, giving layer 31 of Gemma-2-9B-IT at 74\% of depth, layer 40 of Gemma-3-27B-IT at 65\%, and layer 19 of Llama-3.1-8B-Instruct at 59\%.
On Gemma-2-9B-IT the removed set equals the full protocol's $k^*$ support at that layer for all four languages.

\textbf{Design rationale.}
The two kinds of control answer different doubts.
Frequency-matched features ask whether removing any equally active features disrupts the reply, and the equal-norm direction asks whether an edit of equal size in an arbitrary direction does.
The gap of Equation~\ref{eq:reply-gap} compares inputs in $L$ with inputs in other languages, so it measures the adaptation of reply language to input language rather than the ability to produce $L$ at all.

\textbf{Null prediction.}
If the selected features only correlated with the input language and did not drive the reply language, removing them would move the gap of Equation~\ref{eq:reply-gap} no more than removing frequency-matched features or subtracting an equal-norm random direction does.

\begin{table}[t]
\AppTableStyle
\centering\small
\caption{\textbf{Ablating a language's $k^*$ support at the original magnitude.}
Baseline and treatment are gap values under the script readout of Equation~\ref{eq:reply-gap}.
The corrected column restates the treatment after discounting replies in a same-script neighbor language, for the cells where we recorded that diagnostic.
The last column reports whether the treatment's 95\% bootstrap interval separates from all four controls, three frequency-matched and one equal-norm, under the script readout.}
\label{tab:langid-arm}
\resizebox{\linewidth}{!}{%
\begin{tabular}{llrrrrrl}
\toprule
\rowcolor{AppGold}
\AppHead{Model} & \AppHead{Language} & \AppHead{$k^*$} & \AppHead{Baseline} & \AppHead{Treatment} & \AppHead{Collapse} & \AppHead{Corrected treatment} & \AppHead{Separates} \\
\midrule
Gemma-2-9B-IT & Japanese & 1 & $+0.990$ & $+0.015$ & 98\% & not needed & yes \\
              & Greek & 4 & $+0.890$ & $+0.000$ & 100\% & not needed & yes \\
              & Hindi & 4 & $+0.880$ & $+0.005$ & 99\% & not needed & yes \\
              & Bulgarian & 1 & $+0.915$ & $+0.760$ & 17\% & $+0.560$ & yes \\
\midrule
Gemma-3-27B-IT & Japanese & 1 & $+1.000$ & $+0.960$ & 4\% & not needed & yes \\
               & Greek & 1 & $+1.000$ & $+1.000$ & 0\% & not needed & no \\
               & Hindi & 1 & $+0.995$ & $+0.990$ & 1\% & not needed & no \\
               & Bulgarian & 1 & $+1.000$ & $+0.850$ & 15\% & $+0.105$ & yes \\
\midrule
Llama-3.1-8B-Instruct & Japanese & 1 & $+0.995$ & $+0.965$ & 3\% & $+0.840$ & no \\
                      & Greek & 1 & $+1.000$ & $+0.990$ & 1\% & not needed & no \\
                      & Hindi & 1 & $+0.975$ & $+0.950$ & 3\% & not needed & no \\
                      & Bulgarian & 1 & $+0.995$ & $+0.905$ & 9\% & $+0.585$ & yes \\
\bottomrule
\end{tabular}}
\end{table}

\textbf{Same-script replies.}
Script detection cannot see a switch to a language that shares the script, which inflates negatives.
Ablating the Bulgarian support on Gemma-3-27B-IT makes the model reply in Russian, so the Cyrillic readout still scores the reply as Bulgarian.
We keep the main readout unchanged for comparability across runs and record two diagnostics beside it.
The first is the rate of Cyrillic letters used in Russian but not in Bulgarian, namely Unicode \texttt{U+044B}, \texttt{U+044D}, and \texttt{U+0451}, in Bulgarian replies.
The second is the rate of kana-free replies for Japanese inputs.
Greek needs no correction because its script is unique to it.
Devanagari is shared with Marathi and Nepali, and the Hindi readout applies without that correction.
The correction does not affect the Gemma-2-9B-IT positives, where the model switches to fluent English in Latin script and the readout scores it correctly.
Sample replies after each ablation remain coherent, so the intervention redirects the reply language rather than breaking the model's generation.

\paragraph{Dose response.}
\textbf{Question.}
Scaling the SAE edit of Equation~\ref{eq:sae-ablation} by $\alpha$ separates removing this direction from removing enough of anything, because at each scale the treatment and its equal-norm control of Equation~\ref{eq:equal-norm} differ only in direction and not in size.

\begin{table}[t]
\AppTableStyle
\centering\small
\caption{\textbf{Corrected in-language reply rates under scaled ablation.}
Each treatment column has an equal-norm random direction at the same scale, reported in the last column in the same order.
The $16\times$ column enters no conclusion because its noise control also destroys the behavior.}
\label{tab:dose}
\resizebox{\linewidth}{!}{%
\begin{tabular}{llrrrrrl}
\toprule
\rowcolor{AppGold}
\AppHead{Model} & \AppHead{Language} & \AppHead{Baseline} & \AppHead{$\times1$} & \AppHead{$\times2$} & \AppHead{$\times4$} & \AppHead{$\times16$} & \AppHead{Equal-norm noise} \\
\midrule
Gemma-3-27B-IT & Japanese & $1.000$ & $0.960$ & $0.000$ & $0.000$ & $0.000$ & $1.000$, $1.000$, $0.960$, $0.010$ \\
               & Greek & $1.000$ & $1.000$ & $0.870$ & $0.000$ & $0.000$ & $1.000$, $1.000$, $1.000$, $0.090$ \\
               & Hindi & $0.995$ & $0.990$ & $0.575$ & $0.240$ & $0.000$ & $0.995$, $0.995$, $0.995$, $0.160$ \\
               & Bulgarian & $1.000$ & $0.105$ & $0.000$ & $0.000$ & $0.000$ & $1.000$, $1.000$, $1.000$, $0.210$ \\
\midrule
 & & & $\times0.25$ & $\times0.5$ & $\times1$ & & \\
Gemma-2-9B-IT & Japanese & $0.990$ & $0.990$ & $0.940$ & $0.000$ & & $0.990$, $0.990$, $0.990$ \\
              & Greek & $0.890$ & $0.825$ & $0.715$ & $0.000$ & & $0.885$, $0.890$, $0.885$ \\
              & Hindi & $0.880$ & $0.825$ & $0.705$ & $0.005$ & & $0.885$, $0.885$, $0.880$ \\
              & Bulgarian & $0.915$ & $0.905$ & $0.880$ & $0.560$ & & $0.915$, $0.915$, $0.905$ \\
\bottomrule
\end{tabular}}
\end{table}

\textbf{Result.}
Counting every cell of Table~\ref{tab:dose} that has a matched noise control gives 28.
Four of them sit at the 16-fold scale, where the noise control itself collapses, and we set them aside.
Of the remaining 24, the treatment falls more than 0.10 below its noise control in 15, and the two move together in 9.
The original-magnitude column of the 27B model reproduces exactly across two independent runs.

\textbf{Why the edit size explains the 27B readings.}
Collapse correlates with the absolute size of the edit at $\rho=+0.73$ with $p=0.001$ over 16 cells, so the amount removed is a live explanation that the dose curve has to settle.
The supports at layer 40 of Gemma-3-27B-IT carry 0.24 to 0.59\% of the residual norm, against 1.96 to 6.73\% at layer 31 of Gemma-2-9B-IT, so the same ablation removes between 4 and 28 times less there depending on the language.
At two or four times that magnitude all four languages collapse while 11 of the 12 matched noise controls do not move, which shows that the direction rather than the magnitude carries the effect.
SAE reconstruction cosines also differ across these layers, at 0.674 for Gemma-2-9B-IT layer 31, 0.352 for Gemma-3-27B-IT layer 40, and 0.729 for Llama-3.1-8B-Instruct layer 19, and a poorly reconstructing layer weakens the link between an SAE support and the direction the model uses.
Llama-3.1-8B-Instruct is read through SAEs trained on the base model, which adds a second reason for its smaller effects at the original magnitude.
Three of the removed features on Gemma-2-9B-IT also fire on inputs in other languages, so a support is not always exclusive to its own class decision.

\paragraph{Refusal.}
\textbf{Estimator.}
The second behavior uses the harmfulness task with 200 held-out prompts per class.
A reply counts as a refusal when it contains one of 22 fixed phrases such as ``I can't'', ``I'm sorry'', or ``as an AI''.
We read the gap between the refusal rate on harmful prompts and the refusal rate on benign prompts, with the same bootstrap interval as Equation~\ref{eq:reply-gap}.
The harmful refusal rate alone would miss one of the two ways in which an edit can erase the distinction between the two classes.
OLMo-2-7B-IT takes the second way, where harmful refusal rises from 0.705 to 0.840 while benign refusal rises from 0.075 to 0.585, which only the gap registers.

\textbf{Interventions.}
The SAE arm removes the 64 features with the largest two-sample $t$-statistics between harmful and benign prompts on the training encodings at layer 20 of Gemma-2-9B-IT, with the edit of Equation~\ref{eq:sae-ablation} and the same controls as the reply-language arm.
The LEACE arm~\citep{belrose2023leace} applies to the residual vector $h$ the rank-one affine edit
\begin{equation}
    h\ \mapsto\ h-\bigl((h-\mu)^\top b\bigr)\,a,
    \label{eq:leace-edit}
\end{equation}
with $\mu$, $a$, and $b$ fitted on the training activations of the intervened layer so that the label becomes linearly unreadable there, and applies it at every position.
Its controls are norm-matched random rank-one directions in configurations 2, 3, 5, and 6, and shuffled-label edits in configurations 4 and 7, the latter once at their natural size and once rescaled to the Frobenius change of the treatment.
Configurations differ in the layer and in whether the edit is fitted on the activation of the last token or the mean of the last three tokens, which matters because the edit applies at every position.

\begin{table}[t]
\AppTableStyle
\centering\small
\caption{\textbf{All seven fitted refusal configurations.}
Control collapse is the mean reduction in gap across that arm's random controls.
Configurations vary in edit type, layer, and activation pooling.
A configuration is separated when its treated interval overlaps no control interval.}
\label{tab:refusal-arms}
\resizebox{\linewidth}{!}{%
\begin{tabular}{rlllrrl}
\toprule
\rowcolor{AppGold}
 & \AppHead{Model} & \AppHead{Intervention} & \AppHead{Fitting} & \AppHead{Treated gap} & \AppHead{Control collapse} & \AppHead{Separation} \\
\midrule
1 & Gemma-2-9B-IT & SAE top-64 ablation & layer 20 & $+0.275$ & $-0.053$ & separated \\
2 & Gemma-2-9B-IT & Rank-1 LEACE & default & $+0.450$ & $+0.015$ & overlaps \\
3 & Gemma-2-9B-IT & Rank-1 LEACE & layer 20, last token & $+0.420$ & $-0.057$ & separated \\
4 & Gemma-2-9B-IT & Rank-1 LEACE & layer 20, last 3 tokens & $+0.415$ & $+0.003$ & separated \\
5 & OLMo-2-7B-IT & Rank-1 LEACE & default & $+0.205$ & $+0.007$ & separated \\
6 & OLMo-2-7B-IT & Rank-1 LEACE & layer 15, last token & $+0.255$ & $+0.262$ & overlaps \\
7 & OLMo-2-7B-IT & Rank-1 LEACE & layer 15, last 3 tokens & $+0.255$ & $+0.012$ & separated \\
\bottomrule
\end{tabular}}
\end{table}

\textbf{Result.}
The untreated gap is $+0.565$ on Gemma-2-9B-IT and $+0.630$ on OLMo-2-7B-IT.
Configuration 1 has frequency-matched controls at $+0.600$, $+0.625$, and $+0.630$ and an equal-norm control at $+0.550$, and its treated interval of $[0.195, 0.355]$ overlaps none of them.
Configuration 2 produces the smallest collapse of the seven, and its interval is too wide to separate.
Configuration 6 overlaps on its control side rather than its treatment side, since one of its three random controls collapses further than the treatment, and the same layer and intervention fitted on three tokens returns controls at $+0.615$, $+0.620$, and $+0.620$.
We report one configuration per model and intervention in the main text and list all seven here, because the count of fitted configurations belongs beside the ones we report.

\textbf{Two controls on the size of the edit.}
On Gemma-2-9B-IT a shuffled-label edit rescaled to the same Frobenius change as the LEACE edit leaves the gap between $+0.535$ and $+0.605$, so an equally large edit that carries no label information does not move the behavior.
On OLMo-2-7B-IT the LEACE edit is smaller than its random-direction controls, at 0.184 against 0.277 to 0.294, and collapses the gap further, so a larger edit is neither necessary nor sufficient.
Generations after the LEACE edit remain fluent and become evasive rather than degenerate.

\paragraph{Sufficient and nonexclusive readouts.}
\textbf{Definitions.}
Call a coordinate set $S$ sufficient at level $a$ when a probe restricted to $S$ reaches test AUROC $a$.
Pinning $S$ replaces every coordinate in $S$ by its training mean on both splits, and we then refit a dense probe on all $d$ coordinates.
Call $S$ exclusive when pinning it drives the refitted dense AUROC toward chance.

\textbf{Remark (pinning removes only what the complement cannot restore).}
Let $\bar S=[d]\setminus S$.
The refitted dense probe after pinning $S$ can realize every decision function that a probe on $\bar S$ alone can realize, because after pinning every coordinate in $S$ is constant across inputs, so a weight on it only shifts the intercept, and setting those weights to zero recovers every linear function of $\bar S$.
Its best attainable test AUROC is therefore at least that of the best probe on $\bar S$.

This remark predicts that pinning a sufficient set leaves dense readability intact whenever the complement is itself sufficient, which is the question the experiment answers.

\textbf{Procedure.}
We rank the mean-pooled residual coordinates of the harmfulness task by $|t|$ of Equation~\ref{eq:tstat} on the training split, at layer 24 of Gemma-2-9B-IT, layer 15 of OLMo-2-7B-IT, and layer 11 of Mistral-7B-Instruct.
For $n\in\{16,51,64,128,256\}$ we pin the top $n$ coordinates on both splits and refit a dense probe, and we also fit a probe on the top $n$ coordinates alone.
This experiment uses a standardized logistic probe with inverse strength 1, no class weighting, and the full training split, so its values are comparable within Table~\ref{tab:pinning} and not with the main protocol.

\textbf{Design rationale.}
Refitting after pinning asks the strongest version of the necessity question, namely whether any linear readout survives the removal, rather than whether one particular probe breaks.

\begin{table}[t]
\AppTableStyle
\centering\small
\caption{\textbf{Sufficiency without exclusivity on harmfulness detection.}
Each cell gives the test AUROC of a probe on the top $n$ coordinates alone, followed by the refitted dense AUROC after pinning those coordinates to their training means.
The last row gives the dense AUROC with nothing pinned.}
\label{tab:pinning}
\begin{tabular}{rccc}
\toprule
\rowcolor{AppGold}
\AppHead{$n$} & \AppHead{Gemma-2-9B-IT} & \AppHead{OLMo-2-7B-IT} & \AppHead{Mistral-7B-Instruct} \\
\midrule
16 & $0.9341$ / $0.9826$ & $0.9462$ / $0.9880$ & $0.8836$ / $0.9779$ \\
51 & $0.9646$ / $0.9825$ & $0.9674$ / $0.9880$ & $0.9276$ / $0.9773$ \\
64 & $0.9676$ / $0.9827$ & $0.9733$ / $0.9879$ & $0.9310$ / $0.9772$ \\
128 & $0.9752$ / $0.9824$ & $0.9817$ / $0.9875$ & $0.9490$ / $0.9768$ \\
256 & $0.9783$ / $0.9822$ & $0.9824$ / $0.9873$ & $0.9631$ / $0.9768$ \\
\midrule
None pinned & $0.9825$ & $0.9878$ & $0.9778$ \\
\bottomrule
\end{tabular}
\end{table}

\textbf{Result.}
The top 256 coordinates alone reach 0.963 to 0.982 across the three models, and pinning them leaves the refitted dense AUROC within 0.0011 of the untouched probe in every model.
Pinning the same coordinates during generation also changes no behavior at any $n$ up to 256 in the three families.
Both the selected coordinates and their complements are therefore sufficient, and the residual representation is highly redundant.
The count $k^*$ names a minimally sufficient readout and not a necessary set.
This also explains why the rank-one LEACE edit of Equation~\ref{eq:leace-edit} changes behavior where coordinate pinning does not, since LEACE removes a direction that spans all coordinates and drives linear readability to chance, from 0.95 to 0.46 and from 0.988 to 0.494.
We do not report iterative nullspace projection as an encoding width, because erasing 64 directions that way only lowers AUROC from 0.98 to between 0.80 and 0.83 in the three families and measures the inefficiency of that method rather than a property of the representation.

\textbf{Finding.}
Selected features matter for behavior in the direction the probe identifies.
On Gemma-2-9B-IT, removing one to four features collapses the reply-language gap for Japanese, Greek, and Hindi from 0.88 to 0.99 to at most 0.015, while matched controls leave it near baseline.
The dose sweep shows that the direction and not the size of the edit carries the effect.
The residual coordinates, by contrast, form sufficient but nonexclusive readouts, since pinning the top 256 coordinates leaves dense readability within 0.0011 of its untouched value.

\textbf{Implication.}
A test of necessity in a redundant basis needs a direction-level erasure or a feature-level ablation with matched controls, since coordinate pinning cannot remove information that the complement also carries.
Reporting every fitted configuration beside the reported one, as Table~\ref{tab:refusal-arms} does, lets a reader judge how selective the reported effects are.

\paragraph{Scope of the interventions.}
The refusal arm covers one task and reads refusal from a phrase list.
Mistral-7B-Instruct enters the pinning experiments but not the SAE and LEACE refusal arms, because its baseline refusal rate of 0.110 leaves little refusal to collapse.
Both arms intervene inside one model at a time.
They establish that a model's own behavior depends on its selected features, and they make no claim that a matched pair of supports across two models implements one computation.

\subsubsection{Summary of Findings and Open Questions}
\label{app:agreement-summary}

The experiments of this appendix support five statements, each under the protocol of Appendix~\ref{app:measurement} and on the ten base models and nine instruction-tuned models stated above.
\begin{enumerate}
\item Readout sparsity orders tasks alike across the ten base models, at a median cross-lab Spearman correlation of 0.93 on nine readable tasks that remains 0.86 to 0.92 after controlling for any one of six text-side readings, and this order survives changes of threshold, search path, and layer rule at median correlations of 0.97 to 0.99.
\item Absolute counts depend on the basis and on the wording of the texts by factors of 1.5 to 20, while ceilings move by at most 0.03, so only counts measured in one basis on one text pool are comparable.
\item Dense ceilings are largely fixed by the text, at a correlation of 0.950 with a bag-of-words instrument, and the models' contribution shows in sparsity at small feature counts.
\item Selected SAE features correspond across models beyond class means and label correlation on three tasks, at pass rates of 34.3 to 56.2\% against a nominal 1\%.
\item Selected language features drive reply-language adaptation in Gemma-2-9B-IT, and selected residual coordinates are sufficient but not necessary for harmfulness readability.
\end{enumerate}

Four questions remain open under the present evidence.
First, text-side readings predict part of each model's task order but not the agreement between models, and whether a richer text-only baseline would explain more of that agreement remains open.
Second, correspondence varies strongly across class decisions, and support size explains its extremes but not its middle range.
Third, the correspondence test compares responses and not mechanisms, so whether corresponding features implement one computation across models requires interventions in both models at once.
Fourth, the correspondence and intervention results cover the tasks for which SAE supports exist, and the reconstruction quality of the SAE at the intervened layer, 0.35 to 0.73 in cosine, bounds how directly an SAE support reflects the direction the model uses.

%% file: appendix/appendix_training.tex
\subsection{Details of the Training Experiments}
\label{app:training}

Section~\ref{sec:training} asks how training changes the sparsity of task readouts.
This appendix answers three questions in turn.
First, which tasks change their counts between a released base model and its instruction-tuned counterpart under the respective deployment formats.
Second, how model weights and input format contribute to those changes and whether the two contributions interact.
Third, whether supervision for a target produces a sparse readout of that target in a controlled setting where the answer to every query is known.
The real-model comparisons in Sections~\ref{app:training-pairs} and~\ref{app:factorial} reuse the estimator of Appendix~\ref{app:measurement} without change.
The controlled synthetic experiments use their own readout protocol, which Section~\ref{app:toy-results} states and contrasts with the main estimator.
Table~\ref{tab:training-claims} maps every claim of Section~\ref{sec:training} to the evidence and the statistic that support it.

\begin{table}[ht]
\centering\AppTableStyle
\caption{\textbf{Claims of Section~\ref{sec:training} and their evidence.}
Each row names the claim, the place in this appendix that reports it, and the statistic that carries it.
Seed ranges in the last column are minima and maxima over five seed-level task counts, not confidence intervals.}
\label{tab:training-claims}
\begin{tabular}{@{}p{0.34\linewidth}p{0.22\linewidth}p{0.36\linewidth}@{}}
\toprule
\rowcolor{AppGold} \AppHead{Claim} & \AppHead{Evidence} & \AppHead{Supporting statistic} \\
\midrule
Deployment reductions are task-selective on Gemma-2-9B & Table~\ref{tab:app-9b-tasks} & Harmfulness 341 to 47 and fact checking 253 to 40, while five tasks with $k^*\le8$ keep a ratio of 1.00 \\
Harmfulness reductions depend on the model & Tables~\ref{tab:app-selectivity} and~\ref{tab:app-harm-scale} & Ratios from 1.09 on Gemma-3-270M to 8.53 on Gemma-3-12B, with non-overlapping seed ranges in all six Gemma pairs \\
Weights and format both reduce counts & Table~\ref{tab:app-factorial} & Every single-factor change lowers the median count on both tasks \\
Harmfulness is approximately separable & Equation~\ref{eq:factorial-interaction} & Interaction ratio $I\approx1.08$ \\
Fact checking needs both changes & Equation~\ref{eq:factorial-interaction} & Interaction ratio $I\approx4.87$ \\
Factorial medians do not depend on the layer window & Table~\ref{tab:app-factorial} & Shared-layer medians reproduce all eight cells \\
Readable supervised targets admit one-feature readouts & Table~\ref{tab:app-toy} & $k^{\mathrm{syn}}=1$ in 20 of 20 models for seven families in both bases \\
Readable controls need tens of features & Table~\ref{tab:app-toy} & Family medians 29 to 60.5 in the raw basis and 16 to 64 in the PCA basis \\
Sharing inputs with a target does not narrow a control & Section~\ref{app:toy-results} & Overlapping control median 60.5 in the raw basis and 64 in the PCA basis \\
Four- and eight-factor parity is not learned & Tables~\ref{tab:app-toy} and~\ref{tab:app-toy-trajectory} & No model passes the readability gate, and answer accuracy stays at chance after 150,000 updates \\
\bottomrule
\end{tabular}
\end{table}

Each experiment in Sections~\ref{app:training-pairs} to~\ref{app:toy-results} also states its design rationale, meaning the alternative explanation it rules out, the design element that rules it out, and the evidence that the element works.

\textbf{Notation.}
We write $k^*_{t,m,s}$ for the task-level count of Algorithm~\ref{alg:aggregate} on task $t$, model pair $m$, and setting $s$.
The setting is a pair of weights and input format, with $s=\mathrm{base/plain}$ for the base checkpoint on raw text and $s=\mathrm{IT/chat}$ for the instruction-tuned checkpoint on its chat template.
Every count in Sections~\ref{app:training-pairs} and~\ref{app:factorial} uses the ceiling $C_c$ and threshold $\tau_c$ of Equation~\ref{eq:threshold}, the local counts of Equation~\ref{eq:layer-reading}, and the confirmed counts $\tilde k_{c,r}$ of Equation~\ref{eq:nc}.
We call the synthetic readout of Section~\ref{app:toy-results} $k^{\mathrm{syn}}$ to keep it apart from these counts.

\subsubsection{Paired Deployment Comparisons}
\label{app:training-pairs}

\textbf{Question.}
Does instruction-tuned deployment change the count of every task by a similar factor, or does it act selectively on some tasks and leave others unchanged?

\textbf{Models.}
The nine pairs comprise the six Gemma sizes and the Llama, Mistral, and OLMo base and instruction-tuned pairs listed in Table~\ref{tab:models}.
Qwen contributes a base checkpoint to the main matrix but no matched instruction-tuned checkpoint to this analysis.
Each pair shares its layer grid, so both members of a pair draw their windows from the same layers.
No model is trained for this comparison, which compares released checkpoints together with their deployment formats.

\textbf{Data and gate.}
Both members of a pair read the same 14 task pools of Table~\ref{tab:tasks}.
Within each pair, a task enters only if both ceilings exceed 0.90 and both counts are finite.
We write $\mathcal T_m$ for the tasks that pass this gate, and the gate leaves nine or ten tasks per pair, as the Tasks column of Table~\ref{tab:app-selectivity} lists, rather than one common subset across pairs.

\textbf{Design rationale.}
A count can fall after instruction tuning because the task becomes less readable, because every readout shrinks by a common factor, or because the deployment change acts on some tasks more than others.
The gate requires a ceiling above 0.90 on both sides, so a reported ratio compares two readouts of a reliably readable task rather than a readout with a lost signal.
Both members of a pair read identical task pools on the same layer grid, so data and depth sampling cannot differ between them.
The ratio is computed per task within one pair, so a common rescaling would appear as equal ratios across tasks, which the per-task table tests directly.

\textbf{Estimator.}
For a model pair $m$ and a task $t\in\mathcal T_m$, the deployment reduction and its Spearman rank association with the base count over the gated tasks are
\begin{equation}
 R_{t,m}=\frac{k^*_{t,m,\mathrm{base/plain}}}{k^*_{t,m,\mathrm{IT/chat}}},
 \qquad
 \rho_m=\operatorname{Spearman}_{t\in\mathcal T_m}\!\left(k^*_{t,m,\mathrm{base/plain}},\,R_{t,m}\right).
 \label{eq:deployment-ratio}
\end{equation}
A ratio above one means that the deployed instruction-tuned setting admits a smaller readout than the base setting on plain text.
The counts are medians over five probe seeds, and a seed range is the minimum and maximum of the five seed-level task counts.

\textbf{Null reference.}
If deployment rescaled every count by one common factor, every ratio $R_{t,m}$ would equal that factor and the task profile would keep its shape.
The statistic $\rho_m$ does not test this null directly, because the base count enters both of its arguments in Equation~\ref{eq:deployment-ratio}.
The following proposition makes this coupling precise under an independence assumption.

\begin{appproposition}[coupling of the selectivity statistic]
\label{prop:coupling}
Let $X=\log k^*_{\mathrm{base/plain}}$ and $Y=\log k^*_{\mathrm{IT/chat}}$ be random over tasks with finite variances $\sigma_X^2>0$ and $\sigma_Y^2$, and let $X$ and $Y$ be independent.
Then the Pearson correlation of $X$ with $\log R=X-Y$ equals $\sigma_X/\sqrt{\sigma_X^2+\sigma_Y^2}>0$.
\end{appproposition}

\begin{appproof}
Independence gives $\operatorname{Cov}(X,X-Y)=\operatorname{Var}(X)=\sigma_X^2$ and $\operatorname{Var}(X-Y)=\sigma_X^2+\sigma_Y^2$.
Dividing the covariance by the standard deviations $\sigma_X$ and $\sqrt{\sigma_X^2+\sigma_Y^2}$ gives the stated value.
\end{appproof}

Proposition~\ref{prop:coupling} shows that a positive association between base counts and ratios arises even when the instruction-tuned counts carry no information about the base counts.
A positive $\rho_m$ is therefore a descriptive summary of the selectivity pattern and not an independent test of it.
The evidence for selectivity rests instead on the per-task counts and their seed ranges in Table~\ref{tab:app-9b-tasks}, where the ratio is exactly 1.00 on five tasks and far above one on two.
The nine correlations $\rho_m$ also share their tasks across the model pairs, so they are dependent summaries.

\begin{table}[!ht]
\centering\AppTableStyle
\caption{\textbf{Selectivity across all nine base/IT pairs.} Rank correlations compare base counts with base/IT count ratios over tasks that pass the ceiling gate on both sides. Harmfulness columns show the deployment comparison, which combines weights and input format. Ratios are descriptive and share the base count with the predictor, as Proposition~\ref{prop:coupling} shows.}
\label{tab:app-selectivity}
\begin{tabular}{@{}lrrrrr@{}}
\toprule
\rowcolor{AppGold} \AppHead{Model} & \AppHead{Tasks} & \AppHead{$\rho$} & \AppHead{$k^*_{\rm base}$} & \AppHead{$k^*_{\rm IT}$} & \AppHead{Ratio} \\
\midrule
Gemma 3 270M & 9 & 0.626 & 174 & 159 & 1.09 \\
Gemma 3 1B & 9 & 0.814 & 234 & 91 & 2.57 \\
Gemma 3 4B & 9 & 0.995 & 293 & 64 & 4.58 \\
Gemma 2 9B & 10 & 0.826 & 341 & 47 & 7.26 \\
Gemma 3 12B & 10 & 0.760 & 324 & 38 & 8.53 \\
Gemma 3 27B & 9 & 0.597 & 374 & 50 & 7.48 \\
Llama 3.1 8B & 10 & 0.604 & 153 & 128 & 1.20 \\
Mistral 7B & 10 & 0.869 & 391 & 128 & 3.05 \\
OLMo 2 7B & 9 & 0.321 & 380 & 57 & 6.67 \\
\bottomrule
\end{tabular}
\end{table}

\textbf{Result on Gemma-2-9B.}
Table~\ref{tab:app-9b-tasks} lists all ten tasks that pass the gate on Gemma-2-9B.
The five tasks with base counts of at most 8 keep the same count after deployment, and their seed ranges overlap.
The reduction concentrates on the widest readouts, with harmfulness falling from 341 to 47 and fact checking from 253 to 40.
Sentiment is the one wide readout that barely moves, with a ratio of 1.12.
We report this case as measured and do not assign it a cause.

\begin{table}[t]
\centering\AppTableStyle
\caption{\textbf{Every gated task on Gemma-2-9B.}
Counts are medians over five seeds with the seed range in brackets where the five seed-level counts differ.
The ratio is the base count divided by the instruction-tuned count, as in Equation~\ref{eq:deployment-ratio}, and 1.00 marks an unchanged count.}
\label{tab:app-9b-tasks}
\begin{tabular}{@{}lrrr@{}}
\toprule
\rowcolor{AppGold} \AppHead{Task} & \AppHead{Base / plain} & \AppHead{IT / chat} & \AppHead{Ratio} \\
\midrule
Harmfulness detection & 341 [332, 344] & 47 [44, 48] & 7.26 \\
Fact checking & 253 [233, 256] & 40 [39, 51] & 6.33 \\
Intent (Banking77) & 16 [15.5, 16] & 4 [4, 6] & 4.00 \\
Intent (CLINC150) & 4 [3, 4] & 2 & 2.00 \\
Sentiment & 72 [68, 128] & 64 [60, 64] & 1.12 \\
Tool selection & 1 [1, 1.5] & 1 & 1.00 \\
Routing & 2 & 2 & 1.00 \\
PII detection & 4 & 4 & 1.00 \\
Language identification & 1 & 1 & 1.00 \\
Jailbreak detection & 8 & 8 & 1.00 \\
\bottomrule
\end{tabular}
\end{table}

\textbf{Result across scale.}
Table~\ref{tab:app-harm-scale} follows harmfulness detection across the six Gemma sizes under one protocol, one task pool, and one gate.
The ratio grows from 1.1 at 270M parameters to 8.5 at 12B and stays at 7.5 at 27B.
The seed ranges of the base and instruction-tuned counts do not overlap in any of the six pairs.
Outside the Gemma family, the harmfulness ratio is 1.20 on Llama-3.1-8B, 3.05 on Mistral-7B, and 6.67 on OLMo-2-7B, so the size of the reduction also depends on the lab.

\begin{table}[t]
\centering\AppTableStyle
\caption{\textbf{Harmfulness detection across Gemma sizes.}
Counts are medians over five seeds, and brackets give the seed range.
The ratios repeat Table~\ref{tab:app-selectivity} at one decimal.}
\label{tab:app-harm-scale}
\begin{tabular}{@{}lrrr@{}}
\toprule
\rowcolor{AppGold} \AppHead{Model} & \AppHead{Base / plain} & \AppHead{IT / chat} & \AppHead{Ratio} \\
\midrule
Gemma-3-270M & 174 [162, 179] & 159 [154, 161] & 1.1 \\
Gemma-3-1B & 234 [231, 243] & 91 [91, 92] & 2.6 \\
Gemma-3-4B & 293 [256, 310] & 64 [43, 64] & 4.6 \\
Gemma-2-9B & 341 [332, 344] & 47 [44, 48] & 7.3 \\
Gemma-3-12B & 324 [308, 329] & 38 [37, 40] & 8.5 \\
Gemma-3-27B & 374 [365, 389] & 50 [48, 54] & 7.5 \\
\bottomrule
\end{tabular}
\end{table}

\textbf{Seed interpretation.}
Each count uses five 85\% training subsamples, the shared ceiling, and neighbor confirmation from Appendix~\ref{app:aggregation}.
The seed ranges summarize variation across seed-level task medians on a fixed test split.
Non-overlap of two ranges indicates that no seed reverses the base and instruction-tuned comparison, and it is not a calibrated significance test.

\textbf{Scope.}
The comparison changes weights and input format together, as deployment does.
It describes released checkpoints and does not isolate any stage of their post-training recipes.
Section~\ref{app:factorial} separates the two changes on one model, Gemma-2-9B, in a factorial design.

\textbf{Finding.}
Instruction-tuned deployment reshapes the sparsity profile across tasks rather than rescaling it.
On Gemma-2-9B, the five tasks with base counts of at most 8 keep ratios of exactly 1.00, while harmfulness and fact checking shrink by factors of 7.26 and 6.33 with both ceilings above 0.90.
Within one task and one model family under one protocol, the harmfulness reduction rises from 1.1 at 270M parameters to 8.5 at 12B and stays at 7.5 at 27B.

\textbf{Implication.}
A single summary of how instruction tuning changes representations can hide opposite behavior across tasks, so base and instruction-tuned comparisons should be reported per task.
The size of the harmfulness reduction also depends on the model family and scale, so a result on one pair does not transfer to another pair without measurement.

\subsubsection{Weights and Format as a Factorial Design}
\label{app:factorial}

\textbf{Question.}
Does the deployment reduction of Section~\ref{app:training-pairs} on Gemma-2-9B come from the instruction-tuned weights, from the chat format, or from their combination?

\textbf{Conditions.}
Let $w\in\{0,1\}$ indicate base and instruction-tuned weights of Gemma-2-9B, and let $f\in\{0,1\}$ indicate plain text and chat format.
The four conditions cross the two factors.
Both chat conditions render the input through the chat template of the Gemma-2-9B-IT tokenizer, followed by its generation prompt, so the base checkpoint receives the template of its instruction-tuned counterpart.
Both plain conditions receive the raw text with the tokenizer's default special tokens, as in Appendix~\ref{app:inputs}.
All four conditions share the task pools, mean pooling over attended positions, the 768-token limit, the five seeds, and the probe settings of Appendix~\ref{app:measurement}.

\textbf{Tasks.}
The design measures harmfulness detection and fact checking, the two tasks with the largest deployment reductions in Table~\ref{tab:app-9b-tasks}.
Intent classification on CLINC150 serves as the floor control of the design, because its counts of 2 or 4 lie near the bottom of the size grid.

\textbf{Design rationale.}
Deployment changes two things at once, the weights and the input format, so a base/IT difference alone cannot say which of them moves the count.
Crossing the two factors turns one comparison into four cells, and the interaction ratio of Equation~\ref{eq:factorial-interaction} then separates additive from joint action on the log-count scale.
Both chat cells use one template string, so the format factor means the same thing under either weight setting.
Three further elements close the remaining alternatives.
The shared-layer recomputation rules out a shift of the located window, the per-cell ceilings rule out a change of readability, and the floor control shows where the grid resolution makes ratios unreliable.

\textbf{Estimator.}
Write the four task-level counts as $k_{wf}$.
The conditional reduction factors of the weights and of the format, each at a fixed level of the other, and the deployment reduction, are
\begin{equation}
 R_W(f)=\frac{k_{0f}}{k_{1f}},\qquad
 R_F(w)=\frac{k_{w0}}{k_{w1}},\qquad
 R_{\mathrm{deploy}}=\frac{k_{00}}{k_{11}}.
 \label{eq:factorial-effects}
\end{equation}
The deployment ratio factors along either path through the design,
\begin{equation}
 R_{\mathrm{deploy}}=R_W(0)\,R_F(1)=R_F(0)\,R_W(1).
 \label{eq:factorial-paths}
\end{equation}
This identity holds whether or not the factors interact, so it carries no evidence about independence.
To quantify interaction on the multiplicative scale, we define
\begin{equation}
 I=\frac{R_W(1)}{R_W(0)}=\frac{R_F(1)}{R_F(0)}
   =\frac{k_{01}k_{10}}{k_{00}k_{11}},\qquad
 \delta=\log I.
 \label{eq:factorial-interaction}
\end{equation}
The value $I=1$ means that each factor reduces the count by the same factor under either level of the other, so the solo reductions multiply to the deployment reduction.

\begin{appproposition}[saturated log-linear decomposition]
\label{prop:loglinear}
Let the four task-level counts $k_{wf}$ be finite and positive.
Then there are unique real coefficients $\mu,\alpha,\beta,\delta$ with
\begin{equation}
 \log k_{wf}=\mu-\alpha\,w-\beta\,f-\delta\,wf\qquad\text{for all } w,f\in\{0,1\},
 \label{eq:loglinear}
\end{equation}
namely $\mu=\log k_{00}$, $\alpha=\log R_W(0)$, $\beta=\log R_F(0)$, and $\delta=\log I$.
Moreover, the deployment reduction factors into the two solo reductions and the interaction,
\begin{equation}
 R_{\mathrm{deploy}}=R_W(0)\,R_F(0)\,I.
 \label{eq:deploy-product}
\end{equation}
\end{appproposition}

\begin{appproof}
Equation~\ref{eq:loglinear} imposes four linear equations on four unknowns.
Setting $w=f=0$ gives $\mu=\log k_{00}$.
Setting $(w,f)=(1,0)$ gives $\alpha=\log k_{00}-\log k_{10}=\log R_W(0)$, and $(w,f)=(0,1)$ gives $\beta=\log R_F(0)$ in the same way.
Setting $w=f=1$ gives $\log k_{11}=\mu-\alpha-\beta-\delta$, which rearranges to $\delta=\log k_{01}+\log k_{10}-\log k_{00}-\log k_{11}=\log I$.
The four equations have a unique solution because their coefficient matrix is lower triangular with a nonzero diagonal.
For Equation~\ref{eq:deploy-product}, the product $R_W(0)R_F(0)I$ equals $(k_{00}/k_{10})(k_{00}/k_{01})(k_{01}k_{10})/(k_{00}k_{11})=k_{00}/k_{11}$.
\end{appproof}

The sign convention makes $\alpha$ and $\beta$ positive when the solo changes reduce the count.
With this convention, a positive $\delta$ means that the joint change reduces the count by more than the product of the solo reductions, which we call complementarity.
The decomposition is saturated, so it describes the four medians exactly and has no residual from which to estimate uncertainty.

\textbf{Null prediction.}
If the two factors acted separably on the log-count scale, we would expect $I=1$, and the deployment reduction would equal the product of the two solo reductions.
If the reduction came from the weights alone, we would expect $R_F(0)=R_F(1)=1$.

\begin{table}[!ht]
\centering\AppTableStyle
\caption{\textbf{Seed ranges and ceilings for the four-cell design.} Brackets give the minimum and maximum over five seed-level task counts, not confidence intervals. Restricting to the shared layers reproduces every median, with layers $17,20,24,28$ for harmfulness and $20,24,28$ for fact checking.}
\label{tab:app-factorial}
\begin{tabular}{@{}llrrr@{}}
\toprule
\rowcolor{AppGold} \AppHead{Task} & \AppHead{Weights / input} & \AppHead{$k^*$} & \AppHead{Seed range} & \AppHead{Ceiling} \\
\midrule
Harmfulness & Base / plain & 341 & [332, 344] & 0.979 \\
Harmfulness & Base / chat & 135 & [130, 141] & 0.981 \\
Harmfulness & IT / plain & 128 & [128, 128] & 0.984 \\
Harmfulness & IT / chat & 47 & [44, 48] & 0.986 \\
Fact checking & Base / plain & 253 & [233, 256] & 0.910 \\
Fact checking & Base / chat & 217 & [212, 222] & 0.909 \\
Fact checking & IT / plain & 227 & [215, 239] & 0.915 \\
Fact checking & IT / chat & 40 & [39, 51] & 0.927 \\
\bottomrule
\end{tabular}
\end{table}

\textbf{Result.}
The observed counts give $I\approx1.08$ for harmfulness and $I\approx4.87$ for fact checking.
Harmfulness therefore admits an approximately separable description, in which the format alone takes the count from 341 to 135 and the weights alone take it from 341 to 128.
Fact checking instead shows strong complementarity, since each solo change leaves the count between 217 and 227, while the joint change takes it to 40.
Neither task supports the reading that the weights alone carry the reduction, because the format alone lowers both counts.
The ceilings stay between 0.979 and 0.986 on harmfulness and between 0.909 and 0.927 on fact checking, so the count changes do not come from changes in linear readability.
These ratios are arithmetic contrasts of the eight measured medians in Table~\ref{tab:app-factorial} and not fitted causal effects with estimated uncertainty.

\textbf{Layer control.}
The four conditions draw their windows from the same layer grid, but the located windows can differ across conditions.
Restricting all four conditions to their shared layers, which are 17, 20, 24, and 28 for harmfulness and 20, 24, and 28 for fact checking, reproduces every median in Table~\ref{tab:app-factorial}.
The factorial contrasts therefore do not come from a shift of the window.

\textbf{Floor control.}
The intent control has counts of 4, 2, 2, and 2 on the available grids and 4, 4, 4, and 2 on the shared layers 13, 17, and 20.
Counts at the bottom of the size grid change by whole octaves, so a single grid step changes the ratio by a factor of two.
This resolution leaves too little room for a stable multiplicative decomposition, and the control does not carry the factorial conclusion.

\textbf{Scope.}
The design uses one model family and one chat template, so it establishes that weights and format jointly shape readout sparsity on Gemma-2-9B and does not estimate a universal template effect.
The mean over attended positions includes template tokens in the chat conditions, so the format factor combines a change of context with a change of pooled positions.

\textbf{Finding.}
Weights and input format both reduce readout counts on Gemma-2-9B, and the way they combine depends on the task.
Harmfulness is close to separable with $I\approx1.08$, while fact checking needs both changes together with $I\approx4.87$, so either change alone leaves its count between 217 and 227 near the base count of 253 and the joint change takes it to 40.

\textbf{Implication.}
A base and instruction-tuned comparison run on different input formats measures the joint effect of weights and format.
Attributing a readout change to training therefore requires holding the chat template fixed.
On fact checking the weight change barely moves the count on plain inputs and reduces it strongly on chat inputs, so the chosen template decides the conclusion.

\subsubsection{Synthetic World and Training Objective}
\label{app:toy-design}

\textbf{Question.}
Does supervision for a target produce a sparse readout of that target, beyond the presence of the target's input information in the representation?
Answering this question requires tasks whose answers are known exactly and whose supervision we control.
We therefore build a synthetic world of 32 binary factors in which every query has a known true answer, and we vary only whether the training objective rewards predicting that answer.

\textbf{Generative model.}
Each example draws a latent factor vector $z\in\{0,1\}^{32}$ with independent fair coordinates, $z_j\sim\operatorname{Bernoulli}(1/2)$.
A uniformly random permutation $\sigma$ of $\{1,\dots,32\}$ orders the observations.
The observation for factor $j$ is a single token $o(j,z_j)$ that names the factor and its value, so the vocabulary holds 64 observation tokens.
A training sequence is
\begin{equation}
 x=\bigl(\langle\mathrm{bos}\rangle,\ o(\sigma_1,z_{\sigma_1}),\dots,o(\sigma_{32},z_{\sigma_{32}}),\ \langle\mathrm{sep}\rangle,\ q_1,a_1,\dots,q_{12},a_{12}\bigr),
 \label{eq:toy-sequence}
\end{equation}
where $q_1,\dots,q_{12}$ are distinct query tokens drawn uniformly without replacement from a pool of 50 and each $a_i$ is one of two answer tokens.
The sequence has $1+32+1+24=58$ tokens, and the vocabulary has $4+64+50=118$ tokens, counting the beginning, separator, and two answer tokens.
A measurement sequence stops at a single query token, which gives a prefix of 35 tokens.

\textbf{Query pool.}
The pool contains 32 factor queries with answers $z_j$ and 18 aggregate queries, namely one target and one control instance for each of nine function families.
For a factor subset $S$, the three function families are threshold, conjunction, and parity,
\begin{equation}
 g_{\mathrm{thr},S}(z)=\mathbf 1\!\left\{\sum_{j\in S}z_j\geq |S|/2\right\},\quad
 g_{\mathrm{and},S}(z)=\prod_{j\in S}z_j,\quad
 g_{\mathrm{xor},S}(z)=\Bigl(\sum_{j\in S}z_j\Bigr)\bmod 2.
 \label{eq:toy-functions}
\end{equation}
Threshold subsets have sizes 2, 4, 8, and 16, conjunction subsets have sizes 2 and 4, and parity subsets have sizes 2, 4, and 8.
The threshold uses at least half of the factors and so includes ties.
Table~\ref{tab:toy-subsets} lists every subset.
The two eight-factor threshold instances share factors 16 to 19 by design, which forms the overlap control of Section~\ref{app:toy-results}.
Other control subsets also intersect target subsets of different families, for example the sixteen-factor threshold control contains both conjunction targets.

\begin{table}[t]
\centering\AppTableStyle
\caption{\textbf{Factor subsets of the 18 aggregate queries.}
Factors are numbered from 0 to 31.
Target and control instances of a family use subsets of equal size.
The eight-factor threshold pair shares factors 16 to 19, which forms the overlap control of Section~\ref{app:toy-results}.}
\label{tab:toy-subsets}
\begin{tabular}{@{}lll@{}}
\toprule
\rowcolor{AppGold} \AppHead{Family} & \AppHead{Target subset} & \AppHead{Control subset} \\
\midrule
thr2 & 0, 1 & 2, 3 \\
thr4 & 4, 5, 6, 7 & 8, 9, 10, 11 \\
thr8 & 12 to 19 & 16 to 23 \\
thr16 & 0 to 15 & 16 to 31 \\
and2 & 24, 25 & 26, 27 \\
and4 & 28, 29, 30, 31 & 20, 21, 22, 23 \\
xor2 & 2, 5 & 9, 14 \\
xor4 & 3, 10, 17, 24 & 6, 13, 20, 27 \\
xor8 & 1, 4, 11, 18, 21, 26, 29, 30 & 0, 7, 8, 15, 22, 23, 25, 28 \\
\bottomrule
\end{tabular}
\end{table}

\begin{applemma}[matched true-label distributions]
\label{lem:matched}
For every family, the true label of the target instance and the true label of the control instance have the same distribution, namely
\begin{equation}
 \Pr[g_{\mathrm{thr},S}=1]=2^{-|S|}\sum_{i\geq |S|/2}\binom{|S|}{i},\qquad
 \Pr[g_{\mathrm{and},S}=1]=2^{-|S|},\qquad
 \Pr[g_{\mathrm{xor},S}=1]=\tfrac12.
 \label{eq:toy-rates}
\end{equation}
\end{applemma}

\begin{appproof}
The sum $\sum_{j\in S}z_j$ of independent fair bits follows a binomial distribution with $|S|$ trials and success probability one half, whatever the identity of the factors in $S$.
Each family's label is a function of this sum alone, so its distribution depends on $S$ only through $|S|$.
Target and control subsets of a family have equal sizes by Table~\ref{tab:toy-subsets}.
The three expressions of Equation~\ref{eq:toy-rates} follow from the binomial probabilities of the sum, and the parity case uses the fact that a binomial count with success probability one half is even with probability one half for every $|S|\ge1$.
\end{appproof}

Lemma~\ref{lem:matched} rules out one alternative explanation of any target and control difference.
The probes of the target and control instances of one family face the same class balance and the same chance level, so a difference in their counts cannot come from the label distribution.

\textbf{Design rationale.}
The design pairs every supervised target with a control that differs from it in one respect only, whether training rewards predicting the aggregate.
Four elements make this contrast clean.
First, control answers are fresh fair bits, so the objective gives no reason to compute the control aggregate, while the control's input factors stay in the sequence and every factor is itself a supervised query that the model must retain up to the query position.
Second, target and control subsets have equal sizes within each family, so Lemma~\ref{lem:matched} equalizes their class balance and chance level.
Third, the eight-factor threshold control shares half of its factors with a trained target, which tests whether training a target narrows readouts of anything computed from its inputs.
Fourth, each observation appears at a uniformly random position, so no factor can be read from a fixed position, and every probe uses fresh sequences, so a probe cannot recognize training examples.
The contrast sits within one model, one layer grid, and one probe protocol, which removes differences between models from every target and control comparison.

\textbf{Implication for experimental design.}
Pairing a supervised target with an independent-answer control over the same kind of inputs converts the question of whether training shapes a readout into a within-model contrast with a known null.
The same construction applies to any setting in which the true answer can be computed but its supervision can be withheld.

\textbf{Target and control supervision.}
Factor queries and target aggregates receive their correct answers during training.
Control aggregates instead receive a fresh fair bit $\xi$ for every occurrence, drawn independently of the factors and of every other answer.
The control's input factors remain in the sequence, and every factor is itself a supervised factor query.
Probes always use the true aggregate value $g_q(z)$ as their label, including for controls.
The design thus holds the input information fixed and varies only whether the objective rewards computing the aggregate.

\textbf{Objective.}
Let $\ell_\theta(x_{<i})\in\mathbb R^{118}$ be the output logits of the model with parameters $\theta$ before position $i$ and let $\mathcal A(x)$ be the 12 answer positions of Equation~\ref{eq:toy-sequence}.
Training minimizes the expected cross-entropy over the full vocabulary of 118 tokens at the 12 answer positions only,
\begin{equation}
 \begin{aligned}
 \mathcal L(\theta)
   &= -\,\mathbb E_{x}\,\frac{1}{12}\sum_{i\in\mathcal A(x)}
      \log \operatorname{softmax}\bigl(\ell_\theta(x_{<i})\bigr)_{x_i}, \\[6pt]
 x_i
   &= \begin{cases}
        \langle a_{g_q(z)}\rangle, & q \text{ a factor or target query}, \\
        \langle a_{\xi}\rangle,  & q \text{ a control query},
      \end{cases}
 \end{aligned}
 \label{eq:toy-loss}
\end{equation}
where $q$ is the query that precedes position $i$ and $\langle a_0\rangle,\langle a_1\rangle$ are the two answer tokens.
Each update estimates the expectation with 256 freshly generated sequences, so no sequence is seen twice.

\begin{appproposition}[what the objective rewards]
\label{prop:objective}
First, the prefix before any answer position determines $z$, so the answer to every factor and target query is a deterministic function of the prefix.
Second, for every control query the conditional distribution of the answer given the prefix is uniform on the two answer tokens, so the expected loss at that position is at least $\log 2$, with equality exactly when the model assigns probability one half to each answer token.
Third, the infimum of $\mathcal L$ over all predictors equals $\tfrac{9}{50}\log 2$, the expected fraction of control positions times $\log 2$.
\end{appproposition}

\begin{appproof}
Each factor appears in exactly one observation token, and that token names its value, so the 32 observations determine $z$.
Every factor and target answer is $g_q(z)$ and is therefore a function of the prefix.
A control answer $\xi$ is independent of the prefix, because it is drawn independently of $z$, of the permutation, of the queries, and of the other answers.
The expected loss at a control position is the cross-entropy between the uniform distribution on the two answer tokens and the model's distribution, which is at least the entropy $\log 2$ and equals it exactly when the model puts probability one half on each answer token, since any mass on other tokens only increases the loss.
At factor and target positions, a predictor that places probability tending to one on the correct token drives the loss toward zero.
Each of the 12 query slots holds one of the 50 queries uniformly at random, and 9 of the 50 are controls, so a fraction $9/50$ of answer positions carries the irreducible loss $\log 2$ in expectation, and the two facts together give the infimum $\tfrac{9}{50}\log 2$.
\end{appproof}

Proposition~\ref{prop:objective} states exactly what the two conditions of the design differ in.
The objective rewards computing $g_q(z)$ at the query position for targets, while for controls every function of the input is equally good.
Proposition~\ref{prop:objective} does not imply a one-coordinate readout, since the transformer output passes through a final layer normalization and a vocabulary projection, and the GRU output through a vocabulary projection.
The single-feature result for targets in Section~\ref{app:toy-results} is therefore an empirical finding and not a consequence of the objective.

\begin{algorithm}[t]
\AppAlgorithmStyle
\caption{Generation and training of one controlled synthetic model}
\label{alg:toy}
\begin{algorithmic}[1]
\Require architecture and width from Table~\ref{tab:toy-settings}, training seed $s$, number of updates $T$
\State Initialize parameters with seed $s$ and the data generator with seed $1000+s$
\For{$t=1,\dots,T$}
 \For{each of 256 sequences}
  \State Draw $z\sim\operatorname{Bernoulli}(1/2)^{32}$ and a uniform permutation $\sigma$
  \State Write $\langle\mathrm{bos}\rangle$, the observations $o(\sigma_j,z_{\sigma_j})$ for $j=1,\dots,32$, and $\langle\mathrm{sep}\rangle$
  \State Draw 12 distinct queries uniformly from the pool of 50
  \For{each query $q$}
   \State Append $q$ and then $\langle a_{g_q(z)}\rangle$ if $q$ is a factor or target, or $\langle a_\xi\rangle$ with a fresh $\xi\sim\operatorname{Bernoulli}(1/2)$ if $q$ is a control
  \EndFor
 \EndFor
 \State Compute Equation~\ref{eq:toy-loss} on the batch and take one AdamW step with the scheduled learning rate
\EndFor
\end{algorithmic}
\end{algorithm}

\begin{algorithm}[t]
\AppAlgorithmStyle
\caption{Readout measurement for one query of one trained model}
\label{alg:toy-measure}
\begin{algorithmic}[1]
\Require trained model, query $q$, measurement seed $777+q$
\State Draw 6,000 fresh examples, each ending at the query token $q$ with no answer token
\State Label each example with the true value $g_q(z)$, also when $q$ is a control
\State Record the residual stream, or the GRU hidden state, at the query position after the embedding and after every block or recurrent layer
\State Use the first 4,000 examples for training and the last 2,000 for testing
\State Fit the probe suite of Appendix~\ref{app:search} with probe seed 0 on the raw coordinates
\State Repeat the fit on a PCA basis fitted to the training examples only
\State Read $k^{\mathrm{syn}}_q$ from Equation~\ref{eq:toy-reading}
\end{algorithmic}
\end{algorithm}

\begin{table}[t]
\centering\AppTableStyle
\caption{\textbf{Architecture and optimization of the controlled synthetic models.} The main 20-model grid uses these settings. Section~\ref{app:toy-results} covers the auxiliary width, duration, and trajectory runs.}
\label{tab:toy-settings}
\begin{tabular}{@{}lp{0.63\linewidth}@{}}
\toprule
\rowcolor{AppGold} \AppHead{Component} & \AppHead{Setting} \\
\midrule
Transformer grid & Depth $\{2,4\}$, width $\{64,128,256\}$, seeds $\{0,1,2\}$, giving 18 models \\
Transformer block & Pre-layer normalization, 4 attention heads, causal mask, GELU MLP of width $4d$, learned positions, no dropout \\
Transformer output & Final layer normalization and an untied vocabulary projection without bias \\
Initialization & Transformer linear and embedding weights normal with standard deviation 0.02 and zero biases, with attention input projections, layer normalization, and the GRU at their library defaults \\
Recurrent grid & Two-layer GRU, width 256, seeds $\{0,1\}$, giving 2 models \\
Optimizer & AdamW~\citep{loshchilov2017decoupled}, learning rate $3\times10^{-4}$, weight decay 0.01, batch size 256 \\
Schedule & 500-step linear warmup multiplied by cosine decay, 30,000 updates \\
Sequence & 58 training tokens, 35-token measurement prefix ending at the query \\
Probe data & 6,000 fresh examples per query, the first 4,000 for training and the last 2,000 for testing \\
PCA & Fit on probe training features only, retaining $\min(256,d)$ components \\
Probe grid & $1,2,4,\ldots,256$, capped at the feature width, with the shared L1 path \\
\bottomrule
\end{tabular}
\end{table}

\subsubsection{Synthetic Readout, Coverage, and Controls}
\label{app:toy-results}

\textbf{Question.}
In models trained on the objective of Equation~\ref{eq:toy-loss}, how many coordinates does a readout of each aggregate need, and does the answer differ between targets and controls?

\textbf{Representation.}
The readout uses the residual stream at the query position, where the model must produce the answer.
Layer zero denotes the embedding output and layer $\ell\ge1$ the output of block or recurrent layer $\ell$, which differs from the block indexing of Appendix~\ref{app:inputs}.
The transformer residual is read before the final layer normalization, and the GRU readout uses the hidden state of each recurrent layer.
The raw basis uses the model's own coordinates.
The PCA basis centers the training features and rotates them onto their principal axes, and since every main-grid width is at most 256, it keeps all $d$ components and is a full-rank orthogonal change of basis.

\textbf{Estimator.}
Let $C_{q\ell}$ be the dense AUROC for query $q$ at layer $\ell$, and let $\widetilde A_{q\ell}(k)$ be the median AUROC over the ranked-prefix and L1 candidates of size $k$ at that layer.
The synthetic readout is
\begin{equation}
 k^{\mathrm{syn}}_q=\min_{\ell:\,C_{q\ell}>0.90}\ \min\left\{k:
 \widetilde A_{q\ell}(k)\geq0.5+0.95\,(C_{q\ell}-0.5)\right\}.
 \label{eq:toy-reading}
\end{equation}
A query whose dense AUROC never exceeds 0.90 at any layer is recorded as unreadable, which is distinct from a large finite count.
Table~\ref{tab:toy-vs-main} states every difference from the main estimator.

\begin{table}[t]
\centering\AppTableStyle
\caption{\textbf{Synthetic readout versus the main estimator.}
The two estimators share standardization, feature ranking, the probe objective, the regularization grid, and the L1 path of Appendix~\ref{app:search}.}
\label{tab:toy-vs-main}
\begin{tabular}{@{}p{0.22\linewidth}p{0.34\linewidth}p{0.36\linewidth}@{}}
\toprule
\rowcolor{AppGold} \AppHead{Component} & \AppHead{Main estimator} & \AppHead{Synthetic readout} \\
\midrule
Position & Mean over attended positions & Query position \\
Seeds & Five probe seeds on 85\% subsamples & One probe seed on all 4,000 training examples \\
Ceiling & Common maximum over window layers and seeds & Layer-specific dense AUROC \\
Candidates of one size & Any qualifying candidate & Median AUROC over candidates of that size \\
Layers & Neighbor confirmation over a four-layer window & Minimum over layers that pass the gate \\
Gate & Ceiling below 0.90 flagged & Layers at or below 0.90 excluded \\
\bottomrule
\end{tabular}
\end{table}

The median over candidates of one size makes a size qualify only when its typical candidate reaches the threshold.
The minimum over layers without neighbor confirmation is the less conservative choice, and the target counts of one sit at the floor of the grid, where neighbor confirmation cannot lower a count.
Variation across the 20 models comes from training seeds, widths, depths, and architectures rather than from probe seeds, since every model uses one probe seed.

\textbf{Design rationale.}
The readout position, the gate, and the candidate rule each address one alternative reading.
The query position is where the model must produce the answer, and the answer token is excluded from the prefix, so the probe reads what the model has computed before the answer appears and not the answer itself.
The gate separates a readout that is absent from a readout that is wide, so an aggregate the model never computes cannot masquerade as a large count.
The median over candidates of one size prevents a single fortunate candidate from setting a small count.
The PCA basis is a data-dependent rotation of the full space, so agreement between the two bases shows that the target and control separation does not come from alignment with the model's own coordinates.

\textbf{Null predictions.}
If supervision played no role in readout sparsity, target and control instances of one family would show the same distribution of counts, since Lemma~\ref{lem:matched} matches their labels and their inputs carry the same information.
If a control's count simply reflected the number of factors it depends on, the count would grow in proportion to $|S|$ across families.
If training a target narrowed the readout of any aggregate computed over its inputs, the overlapping eight-factor threshold control would read narrower than the other readable controls.

\begin{table}[!ht]
\centering\AppTableStyle
\caption{\textbf{Target and control readouts in the complete main toy grid.} Each entry is the median count over readable models, followed by readable coverage. A dash indicates no model above the ceiling gate, not a large finite count. Raw and PCA bases are separate measurements.}
\label{tab:app-toy}
\begin{tabular}{@{}lrrrr@{}}
\toprule
\rowcolor{AppGold} \AppHead{Family} & \AppHead{Raw target} & \AppHead{Raw control} & \AppHead{PCA target} & \AppHead{PCA control} \\
\midrule
thr2 & 1 (20/20) & 37 (20/20) & 1 (20/20) & 31.5 (20/20) \\
thr4 & 1 (20/20) & 53.5 (20/20) & 1 (20/20) & 48 (20/20) \\
thr8 & 1 (20/20) & 60.5 (20/20) & 1 (20/20) & 64 (20/20) \\
thr16 & 1 (20/20) & 52.5 (20/20) & 1 (20/20) & 32 (20/20) \\
and2 & 1 (20/20) & 30 (20/20) & 1 (20/20) & 30 (20/20) \\
and4 & 1 (20/20) & 29 (20/20) & 1 (20/20) & 16 (20/20) \\
xor2 & 1 (20/20) & --- (0/20) & 1 (20/20) & --- (0/20) \\
xor4 & --- (0/20) & --- (0/20) & --- (0/20) & --- (0/20) \\
xor8 & --- (0/20) & --- (0/20) & --- (0/20) & --- (0/20) \\
\bottomrule
\end{tabular}
\end{table}

\begin{figure}[t]
\centering
\includegraphics[width=\linewidth]{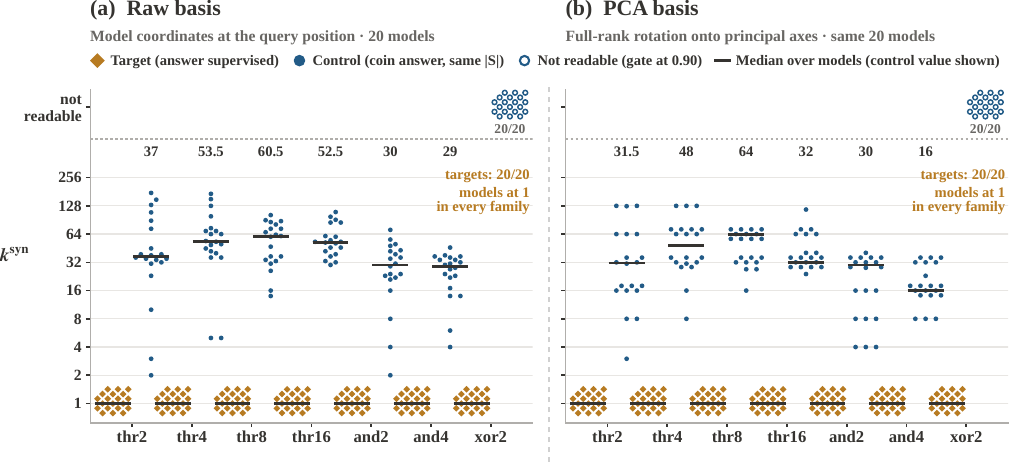}
\caption{\textbf{Synthetic readout $k^{\mathrm{syn}}$ for every target and control in the 20 main-grid models.}
Each marker is one model.
Within each query family, gold diamonds show the supervised target, blue circles show the control that shares the factor count but answers a coin, and black bars mark the median over models.
(a) Raw basis.
Every target reads at $k^{\mathrm{syn}}=1$ in all 20 models, and the control medians are 37, 53.5, 60.5, 52.5, 30, and 29 for thr2, thr4, thr8, thr16, and2, and and4.
(b) PCA basis.
Every target again reads at 1 in all 20 models, and the control medians are 31.5, 48, 64, 32, 30, and 16.
The xor2 control passes the readability gate in none of the 20 models in either basis and appears as open markers in the band at the top.
No control reaches a count of one in any model.}
\label{fig:app-synthetic}
\end{figure}

\textbf{Result.}
Every readable target in the seven learned families has $k^{\mathrm{syn}}=1$ in all 20 main models and in both bases.
Figure~\ref{fig:app-synthetic} shows every one of these counts.
Readable controls span family medians of 29 to 60.5 in the raw basis and 16 to 64 in the PCA basis, and no control reaches a count of one in any model.
The target and control distributions therefore do not overlap, which rejects the first null prediction.
Four- and eight-factor parity targets fail the readability gate, as do the parity controls, and we keep them apart from finite wide readouts.

\textbf{Idealized factor coordinates.}
The second null prediction has an exact form in an idealized representation whose 32 coordinates are the binary factors themselves.

\begin{appproposition}[exact recovery needs every factor]
\label{prop:recovery}
Let the representation be $z\in\{0,1\}^{32}$ with independent fair coordinates, let $|S|\ge2$, and let $g$ be $g_{\mathrm{thr},S}$ or $g_{\mathrm{and},S}$.
A linear score on a coordinate set $T$ reaches population AUROC one for $g$ only if $S\subseteq T$, and $\sum_{j\in S}z_j$ reaches it.
\end{appproposition}

\begin{appproof}
Let $T$ miss some $j\in S$.
For the threshold, fix the other coordinates of $S$ so that exactly $\lceil|S|/2\rceil-1$ of them equal one, which is possible because $0\le\lceil|S|/2\rceil-1\le|S|-1$.
The label then equals $z_j$.
For the conjunction, fix the other coordinates of $S$ at one, and the label again equals $z_j$.
Fix every coordinate outside $S$ arbitrarily.
The two inputs that differ only in $z_j$ have positive probability, opposite labels, and equal scores, since the score ignores coordinate $j$.
A tie between a positive and a negative contributes one half rather than one to the AUROC, so the AUROC falls strictly below one.
For the converse, both labels are nondecreasing threshold functions of $\sum_{j\in S}z_j$, so every positive has a strictly larger sum than every negative.
\end{appproof}

Proposition~\ref{prop:recovery} concerns exact recovery, while $k^{\mathrm{syn}}$ asks for 95\% of the above-chance ceiling, so it motivates the second null prediction without implying it.
The measurements do not follow this idealized picture.
Across the 120 readable controls of the threshold and conjunction families, the Spearman correlation~\citep{spearman1987proof} between $|S|$ and the count is 0.289 in the raw basis and 0.347 in the PCA basis, with log-log slopes of 0.31 and 0.35 rather than one.
The two-sided $p$-values are 0.0014 and 0.0001, but the 120 observations share models and families, so these values describe an association and do not test independent units.
Control counts thus depend on the number of input factors only weakly, and the trained models do not store controls as sums over separate factor coordinates.

\textbf{Overlapping inputs.}
The target threshold over factors 12 to 19 and the control threshold over factors 16 to 23 share four of eight factors.
The control's median count is 60.5 in the raw basis and 64 in the PCA basis, with a raw range of 14 to 102 over the 20 models.
The other readable controls have a median of 38 in the raw basis and 32 in the PCA basis.
A two-sided Mann--Whitney test~\citep{mann1947test} comparing the overlapping control with the other controls gives $p=0.099$ in the raw basis and $p=0.005$ in the PCA basis, with the overlapping control reading wider.
Training the target therefore does not narrow the readout of an untrained aggregate over half of its inputs, which rejects the third null prediction.
The wider reading in the PCA basis, with $p=0.005$, rests on one pair of instances, and we do not interpret its direction further.

\textbf{Width-limited and longer runs.}
Six additional two-layer transformers of widths 16 and 32, three two-layer width-128 transformers with 1,500 updates, and two four-layer transformers of widths 128 and 256 trained for 150,000 updates serve as capacity and duration diagnostics.
They are excluded from the 20-model main grid.
At width 16, six or seven of the nine targets read at $k^{\mathrm{syn}}=1$ and the remaining two or three are unreadable, and at width 32 the split is seven and two.
Limited width therefore removes some readouts entirely, and every target that stays readable keeps a count of one.
In the 150,000-step runs, answer accuracy on a fixed evaluation set of 512 held-out sequences, logged every 2,000 updates, stays between 0.459 and 0.535 for four-factor parity and between 0.471 and 0.529 for eight-factor parity across all 75 logged evaluations, against a chance level of one half, while two-factor parity and the eight-factor threshold end at 1.000 and never fall below 0.959 after the first 2,000 updates.
The dense ceilings of the parity targets stay between 0.500 and 0.521 over the two runs, and those of the parity controls reach at most 0.527.
After these runs, the overlapping eight-factor threshold control reads 26 at width 128 and 51 at width 256, in the range of the main grid, so five times more training does not narrow an unsupervised readout.
These results concern the tested architectures and optimization and do not establish that the parity targets are unlearnable.

\textbf{Training trajectory.}
Table~\ref{tab:app-toy-trajectory} reports the query-position counts at all six saved checkpoints of the four-layer width-256 transformer with training seed 0.
The readable threshold target already has a one-feature readout at the earliest checkpoint after 500 updates.
The two-factor parity target is unreadable at 500 and 1,000 updates and reads at a count of one from 2,000 updates on.
Neither tracked target shows a wide readout at one checkpoint that narrows at a later one.
This sampling cannot exclude a wider transient readout between checkpoints.

\begin{table}[!ht]
\centering\AppTableStyle
\caption{\textbf{Readouts at saved training checkpoints.} Four-layer width-256 transformer, training seed 0, query-token raw basis. Dashes mark failed readability gates. The checkpoints locate changes only at sampled times, and they cannot exclude a transition between observations.}
\label{tab:app-toy-trajectory}
\begin{tabular}{@{}rrrrrr@{}}
\toprule
\rowcolor{AppGold} \AppHead{Step} & \AppHead{thr8 target} & \AppHead{xor2 target} & \AppHead{xor4 target} & \AppHead{xor8 target} & \AppHead{thr8 control} \\
\midrule
500 & 1 & --- & --- & --- & 69 \\
1000 & 1 & --- & --- & --- & 67 \\
2000 & 1 & 1 & --- & --- & 53 \\
4000 & 1 & 1 & --- & --- & 78 \\
8000 & 1 & 1 & --- & --- & 65 \\
16000 & 1 & 1 & --- & --- & 83 \\
\bottomrule
\end{tabular}
\end{table}

\textbf{Readout sparsity and information.}
A single-feature readout of a target might suggest that training discards information.
The minimum expected cross-entropy of any decoder of a target $Y$ from a representation $V$ is $H(Y\mid V)=H(Y)-I(Y;V)$, which depends on $V$ only through the information it retains.
Mutual information is invariant to invertible maps of $V$, while readout sparsity depends on the basis.
The next proposition makes the second half exact.

\begin{appproposition}[information is basis-free and readout sparsity is not]
\label{prop:basis}
Let $Y\sim\operatorname{Bernoulli}(1/2)$, let $U=(2Y-1,\,N)$ with $N\sim\mathcal N(0,\sigma^2)$ independent of $Y$, and let $V=QU$ with the orthogonal map $Q$ whose rows are $(1,1)/\sqrt2$ and $(1,-1)/\sqrt2$.
Define the population count of a representation as the smallest number of its coordinates on which some linear score reaches the threshold $0.5+0.95\,(C-0.5)$, where $C$ is the best population AUROC of a linear score on all coordinates.
Then $I(Y;V)=I(Y;U)$, the population count of $U$ is one, and the population count of $V$ is two whenever $\sigma>\sqrt2/\Phi^{-1}(0.975)$, where $\Phi$ is the standard normal distribution function.
\end{appproposition}

\begin{appproof}
The map $Q$ is invertible, so $V$ and $U$ generate the same information about $Y$ and $I(Y;V)=I(Y;U)$.
The first coordinate of $U$ equals $2Y-1$ and separates the classes perfectly, so $C=1$ for both representations, since $V$ determines $U$ linearly.
The threshold is then $0.975$, and the population count of $U$ is one.
Each coordinate of $V$ equals $(2Y-1\pm N)/\sqrt2$.
A linear score on one coordinate either preserves or reverses the order of that coordinate, and the reversal yields an AUROC of at most one half, so the best single-coordinate AUROC is the probability that a positive exceeds an independent negative.
The difference of a positive and a negative value equals $(2+N_1\mp N_2)/\sqrt2$ with independent copies $N_1,N_2$, which is normal with mean $\sqrt2$ and variance $\sigma^2$, since $N_2$ and $-N_2$ have the same distribution.
The AUROC of either coordinate is therefore $\Phi(\sqrt2/\sigma)$, which falls below $0.975$ exactly when $\sigma>\sqrt2/\Phi^{-1}(0.975)$.
The sum of the two coordinates of $V$ equals $\sqrt2\,(2Y-1)$ and reaches AUROC one, so the population count of $V$ is two.
\end{appproof}

Proposition~\ref{prop:basis} shows that a basis change can move a readout count while leaving the retained information untouched.
The agreement of the raw and PCA bases in Table~\ref{tab:app-toy} is therefore an empirical finding about these two bases rather than a consequence of the definition, and it does not establish invariance to arbitrary changes of basis.
The single-feature readouts characterize how the target is accessible under the tested bases and probe protocol.
They do not establish compression of information or a gradual narrowing of the readout during training.

\textbf{Scope.}
The synthetic experiments establish the target and control separation for threshold, conjunction, and two-factor parity functions of independent binary factors, in transformers and GRUs of width 64 to 256 trained on the objective of Equation~\ref{eq:toy-loss}.
The absolute control counts depend on the basis, which Table~\ref{tab:app-toy} measures for the raw and PCA bases.
The experiments do not model the continuous and correlated features of natural language, and they do not explain the intermediate counts between one and several hundred that Section~\ref{sec:agreement} reports for pretrained models.

\textbf{Finding.}
In 20 of 20 trained models and in both tested bases, every readable supervised aggregate admits a one-feature readout, while readable controls over inputs of the same size need family medians of 29 to 60.5 features in the raw basis.
Sharing half of its factors with a trained target does not narrow a control, which reads 60.5 and 64 in the raw and PCA bases.
Threshold targets over 2, 4, 8, and 16 factors all read at one feature in both bases, so the number of input dependencies does not set the readout size of a learned target in these models.

\textbf{Implication.}
In this controlled setting, a sparse readout marks the variables that the objective rewarded computing, and linear readability alone marks only that the information is present.
A dense probe with a high ceiling therefore cannot distinguish a computed target from a readable consequence of its inputs, while the pair of ceiling and count can.
The basis construction of Proposition~\ref{prop:basis} adds a caution for interpretation, since a count describes access in a chosen basis and a small count does not by itself indicate that information has been discarded.

\subsubsection{Summary of Findings and Open Questions}
\label{app:training-summary}

The training experiments support four findings, each under the conditions stated above.
\begin{enumerate}
\item Instruction-tuned deployment changes readout counts selectively across tasks, leaving counts of at most 8 unchanged on Gemma-2-9B and reducing the two widest readable readouts, fact checking and harmfulness, by factors of 6.33 and 7.26 (Table~\ref{tab:app-9b-tasks}).
\item The harmfulness reduction depends on the model, rising from 1.1 to 8.5 across Gemma sizes and ranging from 1.20 to 6.67 across the other three labs (Tables~\ref{tab:app-selectivity} and~\ref{tab:app-harm-scale}).
\item Weights and input format both contribute, with an interaction ratio of about 1.08 on harmfulness and about 4.87 on fact checking (Equation~\ref{eq:factorial-interaction}).
\item In controlled synthetic models, supervision concentrates each learned target into a one-feature readout, while matched readable controls need tens of features, and sharing half of its inputs with a trained target does not narrow a control (Table~\ref{tab:app-toy}).
\end{enumerate}

The following questions remain open and are stated as testable conditions.
\begin{enumerate}
\item Whether the saturated decomposition yields similar interaction ratios on model families other than Gemma-2-9B and on chat templates other than its own.
\item Which stage of post-training produces the selective reduction, which requires checkpoints between the base and the released instruction-tuned model.
\item Why sentiment keeps a ratio of 1.12 on Gemma-2-9B despite a base count of 72, unlike the other wide readouts on that model, harmfulness and fact checking.
\item Whether four- and eight-factor parity becomes learnable under other architectures or optimizers, and whether it then also reads at one feature at the query position.
\item Which mechanism produces the intermediate counts of pretrained models, which the synthetic setting, with counts of one or of tens, does not reproduce.
\end{enumerate}

%% file: appendix/appendix_applications.tex
\subsection{Details of the Learning and Maintenance Experiments}
\label{app:applications}

This appendix specifies the experiments of Section~\ref{sec:applications} at the level needed to recompute every number in that section.
The section makes three predictive claims and two intervention claims, and each claim rests on a distinct estimator with its own unit of analysis.
Table~\ref{tab:app-claims} maps every claim to the subsection that defines its estimator, the display that reports it, and the statistic that carries it.
The task pools and the probe family come from Appendix~\ref{app:measurement}, and every reference count $k^*_t$ and ceiling $C_t$ below is the task-level output of Algorithm~\ref{alg:aggregate} on Gemma-2-9B unless a subsection states otherwise.
We keep three procedures apart throughout.
Initial learning adapts a pretrained student, recovery adapts a backbone that sequential training has already changed, and frozen probing fits new linear probes without changing any backbone weight.

\begin{table}[!ht]
\centering\AppTableStyle
\caption{\textbf{Claim-to-evidence map for Section~\ref{sec:applications}.}
Each row names a main-text claim, the subsection that defines its estimator, the display that reports it, and the statistic that carries it.
The last two rows are intervention claims whose tested rules did not outperform their uniform and random baselines, and we report them with the same level of detail as the three predictive claims.}
\label{tab:app-claims}
\begin{tabular}{@{}p{0.29\linewidth}p{0.14\linewidth}p{0.20\linewidth}p{0.29\linewidth}@{}}
\toprule
\rowcolor{AppGold} \AppHead{Claim} & \AppHead{Estimator} & \AppHead{Display} & \AppHead{Supporting statistic} \\
\midrule
Reference $k^*$ anticipates student label requirements & \ref{app:label-count} & Eq.~\ref{eq:label-budget}, Table~\ref{tab:app-nstar} & Shared exponent 0.65 on 109 uncensored observations \\
The association carries task-level uncertainty & \ref{app:label-inference} & Table~\ref{tab:app-inference} & Task-block interval $[0.44,0.91]$, held-out factor 5.3 \\
The association remains under transferability and text controls & \ref{app:learning-controls} & Table~\ref{tab:app-partial-controls} & Partial $\rho$ of $k^*$ of $-0.67$ to $-0.79$ given LogME or the H-score and $-0.55$ to $-0.67$ given the bag-of-words ceiling \\
Saved heads fail while linear readability persists & \ref{app:continual} & Eq.~\ref{eq:forgetting-channels} & Head losses of 0.47 to 0.68 over all 18 chain seeds against ceiling losses of at most 0.029 \\
Recovery needs fewer labels than initial learning & \ref{app:continual} & Tables~\ref{tab:recovery} and~\ref{tab:app-cl-curves} & Labels to a common target, 101 to 244 for recovery against 194 to 602 for initial learning \\
Wider readouts lose more readability under pruning & \ref{app:pruning} & Table~\ref{tab:app-pruning-ranks} & $\rho\geq0.886$ with $q<0.05$ in 30 of 32 conditions \\
Allocation by predicted $N^*$ does not consistently save labels & \ref{app:allocation} & Table~\ref{tab:app-allocation} & Median ratio 1.03 to uniform, cheaper in 6 of 18 settings \\
Readout-guided protection does not beat random protection & \ref{app:pruning-recovery} & Table~\ref{tab:app-protection} & Selected minus random differences within 0.010 \\
\bottomrule
\end{tabular}
\end{table}

Each experiment rules out an alternative explanation, and every subsection below names that alternative and the design element that addresses it in a paragraph headed Design rationale.

\textbf{Shared notation.}
We index students by $m$, tasks by $t$, class decisions by $c$, and realized training-set sizes by $n$.
For a student $m$ adapted on task $t$ with $n$ labels, $A^{\mathrm{stu}}_{tmc}(n,r)$ denotes the test AUROC of the adapted model for class decision $c$ and training seed $r$, evaluated on the fixed test split of Table~\ref{tab:tasks}.
The reference quantities $k^*_t$ and $C_t$ carry no student index, since they come from the frozen Gemma-2-9B measurement.
All logarithms of label counts and feature counts are base 10.

\subsubsection{Construction of Figure~\ref{fig:applications-story}}
\label{app:app-figure}

\textbf{Panel a.}
The panel plots the 109 uncensored student and task observations over 16 tasks and eight students that enter the shared-exponent fit of Equation~\ref{eq:label-budget}, with raw label requirements on the vertical axis and the fitted student-specific intercepts drawn as parallel lines with the shared slope.
The student abbreviations G3, L3.2, and Q3 denote Gemma 3, Llama 3.2, and Qwen 3.

\textbf{Panel b.}
The panel uses harmfulness detection at the end of the harmfulness, sentiment, and jailbreak chain in each of four models, and its bars show medians over three paired chain seeds.
Saved-head losses are relative to the checkpoint immediately after harmfulness training, whereas probe losses are relative to the pretrained checkpoint, following Equation~\ref{eq:forgetting-channels}.
Negative losses denote improvements, and we retain them.
The right half compares, on the shared grid of 16 to 1024 labels, the labels that initial learning and recovery need to reach one target AUROC per model, with horizontal whiskers spanning the three chain seeds of recovery, and Table~\ref{tab:recovery} gives the numerical values.
The two losses in the panel do not form an additive decomposition of forgetting, for the reason stated after Equation~\ref{eq:forgetting-channels}.

\textbf{Panel c.}
The panel shows the six-task Wanda experiment on Gemma-3-1B using Equation~\ref{eq:pruning-loss}, with the unpruned reference at zero sparsity and pre-pruning counts annotated beside the curves.
Its expanded vertical scale resolves the small pruning losses of the narrow-readout tasks.

\subsubsection{Student Adaptation and Label Sources}
\label{app:student-training}

\textbf{Question.}
The label-budget analysis needs, for every student and task, a learning curve that relates the number of labels to held-out AUROC.
This subsection specifies how we produce those curves.

\textbf{Models.}
The eight students are Gemma-3-270M, Gemma-3-1B, Gemma-3-4B, Llama-3.2-1B, Llama-3.2-3B, Qwen3-0.6B-Base, Qwen3-1.7B-Base, and OLMo-2-0425-1B.
Each run starts from the pretrained checkpoint with a newly initialized binary classification head and newly initialized adapters.
The head reads the final non-padding position of the last layer, following the sequence-classification convention of each architecture.
LoRA~\citep{hu2021lora} adapters with the settings of Table~\ref{tab:adaptation-settings} attach to every attention and MLP projection that exists in the architecture, namely the query, key, value, and output projections and the gate, up, and down projections.

\textbf{Data.}
Each class decision defines a binary problem on the task pool of Table~\ref{tab:tasks}.
Up to three class decisions per multiclass task enter the adaptation sweep, so these learning curves need not cover every decision that enters the reference $k^*_t$.
For a requested budget $n$ and a training split with $n_{\mathrm{tr}}$ texts of which a fraction $\pi$ is positive, we write $\bar n=\min\{n,n_{\mathrm{tr}}\}$ and draw without replacement
\begin{equation}
    n_+ = \max\bigl\{1,\ \min\{\operatorname{round}(\bar n\pi),\ n_{\mathrm{tr}}\pi,\ \bar n-1\}\bigr\},
    \qquad
    n_- = \min\{\bar n-n_+,\ n_{\mathrm{tr}}(1-\pi)\},
    \label{eq:stratified-draw}
\end{equation}
positive and negative texts, so every subset keeps the pool's class ratio and contains both classes.
A requested budget above the pool size maps to the full pool, and we collapse requested budgets that map to the same realized size.
Budgets and seeds are independent runs, not successive extensions of one adapter, and every run initializes its head and adapters from the same seeded draw, with the pretrained weights and the random state restored from an in-process per-seed cache so that repeated runs of one seed start from identical initial states.

\textbf{Label sources.}
Main runs train on ground-truth labels.
Hard-teacher runs replace the ground-truth training labels with the predictions of a frozen Gemma-2-9B probe, and soft-teacher runs train on the same probe's predicted positive-class probabilities $p_i$ with the loss
\begin{equation}
 \mathcal L_{\rm soft}(\theta)=-\frac1n\sum_{i=1}^{n}\Bigl[p_i\log q_\theta(1\mid x_i)
 +(1-p_i)\log q_\theta(0\mid x_i)\Bigr],
 \label{eq:soft-label-loss}
\end{equation}
where $q_\theta(\cdot\mid x)$ is the student's softmax output.
Stratified subsampling for soft-label runs uses the hard labels $\mathbf1[p_i>0.5]$.
Before training, we verify that the teacher labels align text by text with the student's training pool.
Every run evaluates against the original ground-truth test labels of the fixed test split, so all three label sources share one evaluation target.

\textbf{Procedure.}
Table~\ref{tab:adaptation-settings} lists the settings of the three adaptation procedures.
All procedures use AdamW~\citep{loshchilov2017decoupled} with a constant learning rate, batch size 16, and no validation-based early stopping.
The number of epochs adapts to the realized training size,
\begin{equation}
    E(n)=\max\Bigl\{3,\ \min\bigl\{50,\ \lfloor 3000/n\rfloor\bigr\}\Bigr\},
    \label{eq:epochs}
\end{equation}
so a run on 16 labels takes 50 epochs and a run on 4096 labels takes three.
Trainable parameters are stored in float32, and the forward pass uses bfloat16 autocasting on GPU.
Students receive plain text without a chat template, truncated at the token limits of Table~\ref{tab:adaptation-settings}.

\begin{table}[!ht]
\centering\AppTableStyle
\caption{\textbf{Adaptation settings of the three procedures.}
Sequential updates train all backbone weights, whereas initial learning and recovery train rank-16 LoRA adapters and a head.
The recovery label grid of four budgets from 16 to 1024 is coarser than the eight-point initial-learning grid, and Section~\ref{app:continual} states how this difference enters the comparison.}
\label{tab:adaptation-settings}
\begin{tabular}{@{}lp{0.25\linewidth}p{0.25\linewidth}p{0.22\linewidth}@{}}
\toprule
\rowcolor{AppGold} \AppHead{Setting} & \AppHead{Initial learning} & \AppHead{Sequential updates} & \AppHead{Recovery} \\
\midrule
Trainable weights & LoRA and head & All backbone weights and head & LoRA and new head \\
Learning rate & $2\times10^{-4}$ & $10^{-5}$ & $2\times10^{-4}$ \\
LoRA & $r=16$, $\alpha=32$, dropout 0.05 & Not used & Same as initial learning \\
Batch size & 16 & 16 & 16 \\
Token limit & 768, and 512 for the Gemma-3 students on the 14 core tasks & 512 & 512 \\
Label grid & 16, 32, 64, 256, 512, 1024, 2048, 4096 & Up to 4096 per step & 16, 64, 256, 1024 \\
Seeds & $0,1,2$ & $0,1,2$ per chain & $0,1,2$ per final backbone \\
\bottomrule
\end{tabular}
\end{table}

\textbf{Design rationale.}
Three choices keep the learning curves comparable across budgets.
First, every budget is an independent run from the pretrained checkpoint, so points on one curve do not share an adapter and a poor early run cannot propagate to later budgets.
Second, curves are keyed by the realized training size, so a budget above the pool size never appears at a label count the student did not receive.
Third, stratified sampling by Equation~\ref{eq:stratified-draw} holds the class ratio fixed, so a change in AUROC along a curve reflects the number of labels and not a change in prevalence.

\textbf{Scope.}
Initial learning truncates inputs at 768 tokens for the five non-Gemma students and for the Gemma-3 students on the ten additional tasks, and at 512 tokens for the Gemma-3 students on the 14 core tasks, whose runs predate the change of the default limit.
Sequential updates and recovery truncate at 512 tokens.
The recovery comparison of Section~\ref{app:continual} therefore uses one limit on both sides for Gemma-3-1B and Gemma-3-4B, and for Llama-3.1-8B and Qwen3-8B it compares two complete procedures that also differ in this limit, so it does not isolate the change of backbone state alone.
These limits are separate from the frozen-probe limits of Appendix~\ref{app:inputs}.

\subsubsection{Label Requirements, Censoring, and the Complete Fit Panel}
\label{app:label-count}

\textbf{Question.}
Do tasks with wider readouts in the frozen reference model require more labels for students to learn?
Answering this requires a label requirement that is comparable across tasks with different attainable AUROC, a treatment of curves that saturate before their first point, and a regression that separates student-specific offsets from the shared dependence on $k^*_t$.

\begin{appdefinition}[learning curve and label requirement]
\label{def:nstar}
Let $n_1<\dots<n_J$ be the realized training sizes observed for student $m$ on task $t$.
The learning curve is the median over the training seeds and the evaluated class decisions at each of these realized training sizes,
\begin{equation}
    \bar A_{tm}(n_j)=\operatorname{median}_{c,\,r}\,A^{\mathrm{stu}}_{tmc}(n_j,r),
    \label{eq:learning-curve}
\end{equation}
keyed by the realized size rather than by a requested budget that may exceed the pool.
The label threshold applies the normalization of Equation~\ref{eq:threshold} to the curve's own endpoint,
\begin{equation}
 \tau_{tm}=0.5+0.95\left[\bar A_{tm}(n_J)-0.5\right].
 \label{eq:nstar-threshold}
\end{equation}
If $\bar A_{tm}(n_1)\geq\tau_{tm}$, the requirement is left-censored, and we record only $N^*_{tm}\leq n_1$.
Otherwise let $j$ be the first index with $\bar A_{tm}(n_{j+1})\geq\tau_{tm}$, and define the weight $w$ and the requirement by
\begin{equation}
 w=\frac{\tau_{tm}-\bar A_{tm}(n_j)}{\bar A_{tm}(n_{j+1})-\bar A_{tm}(n_j)},\qquad
 \log_{10}N^*_{tm}=(1-w)\log_{10}n_j+w\log_{10}n_{j+1}.
 \label{eq:nstar-interpolation}
\end{equation}
Equivalently, $N^*_{tm}=n_j^{\,1-w}\,n_{j+1}^{\,w}$ is the geometric interpolation between the two bracketing sizes.
The reader takes the first crossing on the observed curve and imposes no monotonicity.
\end{appdefinition}

\textbf{Remark (existence and bracketing).}
If $\bar A_{tm}(n_J)>0.5$, then $N^*_{tm}$ is either left-censored or finite with $n_j< N^*_{tm}\leq n_{j+1}$ for the first crossing bracket $(n_j,n_{j+1})$.
The threshold satisfies $\tau_{tm}=\bar A_{tm}(n_J)-0.05\,[\bar A_{tm}(n_J)-0.5]<\bar A_{tm}(n_J)$, so the endpoint itself reaches $\tau_{tm}$ and a first crossing exists.
If the crossing occurs at $n_1$, the value is left-censored.
Otherwise $\bar A_{tm}(n_j)<\tau_{tm}\leq\bar A_{tm}(n_{j+1})$ by the choice of $j$, which gives $0<w\leq1$ in Equation~\ref{eq:nstar-interpolation}, and the geometric interpolation then lies in the half-open bracket $(n_j,n_{j+1}]$.

Every uncensored requirement is therefore finite and lies within one grid step of an observed crossing.
The endpoint $\bar A_{tm}(n_J)$ is an empirical reference, so extending the grid to larger budgets can change both $\tau_{tm}$ and $N^*_{tm}$.
The requirement therefore measures how quickly a curve approaches its own largest-budget performance, not the number of labels needed for a fixed absolute AUROC.

\textbf{Remark (reference point).}
Reading $N^*$ against the running maximum of each learning curve instead of its endpoint changes 31 of the 109 requirements and moves the shared exponent from 0.65 $[0.44,0.91]$ to 0.63 $[0.40,0.90]$.
One requirement changes by an order of magnitude, CLINC150 on Gemma-3-1B from 28 to 295, where an early spike raises the reference.
The two references trade one sensitivity for another, and we report the endpoint reference as primary and this variant as a sensitivity analysis.
On the shared four-point grid of Section~\ref{app:continual} every recovery and initial-learning curve is monotone in the label count, so the two references coincide there.

\textbf{Censoring.}
Left-censored observations carry only an upper bound $N^*_{tm}\leq16$, and a least-squares fit cannot use a bound as a response value.
We exclude them from the regression and display them in Table~\ref{tab:app-nstar}, so a reader can see which tasks saturate early.
Eighteen of the 19 censored cells occur at $k^*_t\leq2$, and every censored requirement lies below every included one, since an uncensored requirement exceeds 16 by construction.
The exclusion therefore removes information from the low end of the $k^*$ range rather than adding points at an imputed value.

\textbf{Design rationale.}
The relative threshold of Equation~\ref{eq:nstar-threshold} removes differences in attainable AUROC between tasks and students, so a hard task with a low endpoint does not receive a large requirement merely because it never reaches a fixed absolute target.
The ceiling gate $C_t>0.90$ admits only tasks whose reference count describes a reliably readable readout, as defined in Appendix~\ref{app:aggregation}.
Student intercepts absorb every student-level offset, which Proposition~\ref{prop:pooled-slope} below makes precise.

\textbf{Estimator.}
We relate the requirement of task $t$ in student $m$ to the reference count by
\begin{equation}
    \log_{10} N^*_{t,m}
    = a_m + b\log_{10} k^*_t + e_{t,m},
    \label{eq:label-budget}
\end{equation}
where $a_m$ is a student-specific intercept, $b$ is the shared exponent, and $e_{t,m}$ is a residual.
The panel $\mathcal P$ contains every pair with $C_t>0.90$, $0<k^*_t<\infty$, and a finite uncensored $N^*_{tm}$, which yields 109 observations across eight students and 16 tasks.
We estimate $(a_1,\dots,a_8,b)$ by ordinary least squares on the design matrix whose first column holds $\log_{10}k^*_t$ and whose remaining eight columns are student indicators, without a separate global intercept.

\begin{appproposition}[the shared exponent pools within-student slopes]
\label{prop:pooled-slope}
Let $x_{tm}=\log_{10}k^*_t$ and $y_{tm}=\log_{10}N^*_{tm}$, let $\mathcal T_m$ be the tasks of student $m$ in $\mathcal P$, and let $\bar x_m,\bar y_m$ be their means over $\mathcal T_m$.
Write $S^{xx}_m=\sum_{t\in\mathcal T_m}(x_{tm}-\bar x_m)^2$ for the within-student sum of squares, write $S^{xy}_m=\sum_{t\in\mathcal T_m}(x_{tm}-\bar x_m)(y_{tm}-\bar y_m)$ for the within-student cross product, and assume $\sum_m S^{xx}_m>0$.
Then the shared least-squares exponent of Equation~\ref{eq:label-budget} fitted on the panel $\mathcal P$ is
\begin{equation}
    \widehat b=\frac{\sum_m S^{xy}_m}{\sum_m S^{xx}_m}
    =\sum_{m:\,S^{xx}_m>0}\frac{S^{xx}_m}{\sum_{m'}S^{xx}_{m'}}\ \widehat b_m,
    \qquad
    \widehat b_m=\frac{S^{xy}_m}{S^{xx}_m},
    \label{eq:pooled-slope}
\end{equation}
where $\widehat b_m$ is the slope of a separate regression for student $m$, and $\widehat a_m=\bar y_m-\widehat b\,\bar x_m$.
\end{appproposition}

\begin{appproof}
For fixed $b$, the objective $\sum_m\sum_{t\in\mathcal T_m}(y_{tm}-a_m-bx_{tm})^2$ separates over students, and each $a_m$ minimizes its own block at $a_m=\bar y_m-b\bar x_m$.
Substituting this minimizer leaves $\sum_m\sum_{t}\bigl[(y_{tm}-\bar y_m)-b(x_{tm}-\bar x_m)\bigr]^2$, a convex quadratic in $b$ whose stationarity condition is $\sum_m S^{xy}_m=b\sum_m S^{xx}_m$.
The second equality in Equation~\ref{eq:pooled-slope} follows by writing $S^{xy}_m=S^{xx}_m\widehat b_m$ for every student with $S^{xx}_m>0$.
\end{appproof}

Proposition~\ref{prop:pooled-slope} connects the two fits we report.
The shared exponent is a convex combination of the separate student slopes of Table~\ref{tab:app-student-slopes}, weighted by how widely each student's tasks spread in $\log k^*$.
The shared exponent therefore lies between the smallest and the largest separate slope, 0.454 and 0.700.
The fit identifies the exponent only from variation of $k^*_t$ across tasks within each student, so a student-level offset such as model size or family enters $a_m$ and never $b$.

\begin{algorithm}[t]
\AppAlgorithmStyle
\caption{Construct the label-requirement panel and fit the shared exponent}
\label{alg:nstar-panel}
\begin{algorithmic}[1]
\Require adapted-student test AUROCs with realized training sizes, reference counts $k^*_t$ and ceilings $C_t$ from Algorithm~\ref{alg:aggregate}
\For{each student $m$ and task $t$}
 \State Group AUROCs by realized training size and form $\bar A_{tm}$ by Equation~\ref{eq:learning-curve}
 \State Compute $N^*_{tm}$ and its censoring flag by Equations~\ref{eq:nstar-threshold} and~\ref{eq:nstar-interpolation}
 \If{$C_t>0.90$, $0<k^*_t<\infty$, and $N^*_{tm}$ is finite and uncensored}
  \State Add $(t,m,\log_{10}k^*_t,\log_{10}N^*_{tm})$ to the panel $\mathcal P$
 \EndIf
\EndFor
\State Fit Equation~\ref{eq:label-budget} by least squares with a shared slope and student-specific intercepts
\end{algorithmic}
\end{algorithm}

\begin{table}[!ht]
\centering\AppTableStyle
\caption{\textbf{Label requirements underlying the shared-exponent fit.} Columns identify the eight students (G for Gemma 3, L for Llama 3.2, Q for Qwen 3, O for OLMo 2). Counts are rounded for display, and $\leq16$ denotes a left-censored requirement that the regression excludes. Rows contain all 16 contributing tasks, and the reference model is Gemma 2 9B.}
\label{tab:app-nstar}
\begin{tabular}{@{}lrrrrrrrrr@{}}
\toprule
\rowcolor{AppGold} \AppHead{Task} & \AppHead{$k^*$} & \AppHead{G270M} & \AppHead{G1B} & \AppHead{G4B} & \AppHead{L1B} & \AppHead{L3B} & \AppHead{Q0.6B} & \AppHead{Q1.7B} & \AppHead{O1B} \\
\midrule
Language ID & 1 & $\leq 16$ & $\leq 16$ & $\leq 16$ & $\leq 16$ & $\leq 16$ & $\leq 16$ & $\leq 16$ & 29 \\
Tool selection & 1 & $\leq 16$ & 25 & 21 & $\leq 16$ & $\leq 16$ & 19 & $\leq 16$ & 44 \\
Routing & 2 & $\leq 16$ & $\leq 16$ & $\leq 16$ & 24 & $\leq 16$ & $\leq 16$ & $\leq 16$ & $\leq 16$ \\
Intent (CLINC150) & 4 & 316 & 28 & 35 & 164 & 387 & 239 & 24 & 282 \\
PII & 4 & 58 & 80 & 55 & 73 & 58 & 54 & 63 & 90 \\
Jailbreak & 8 & 59 & 61 & 41 & 62 & 52 & 30 & 29 & 94 \\
MASSIVE & 8 & 407 & 379 & 288 & 245 & 241 & 237 & 384 & 351 \\
Subjectivity & 8 & 50 & 17 & 27 & 21 & $\leq 16$ & 38 & 47 & 37 \\
Yelp & 8 & 57 & 28 & 26 & 28 & 21 & 31 & 32 & 30 \\
Banking77 & 16 & 240 & 232 & 187 & 239 & 262 & 241 & 233 & 339 \\
IMDB & 16 & 368 & 78 & 30 & 61 & 54 & 145 & 52 & 74 \\
AG News & 24 & 120 & 54 & 108 & 57 & 103 & 132 & 103 & 164 \\
Newsgroups & 24 & 1653 & 1798 & 1301 & 1612 & 1216 & 1193 & 1245 & 1796 \\
Sentiment & 72 & 351 & 215 & 211 & 248 & 203 & 223 & 248 & 240 \\
Fact checking & 253 & 2596 & 1181 & 822 & 1028 & 739 & 1665 & 1255 & 2702 \\
Harmfulness & 341 & 1467 & 864 & 478 & 714 & 572 & 988 & 440 & 891 \\
\bottomrule
\end{tabular}
\end{table}
\begin{table}[!ht]
\centering\AppTableStyle
\caption{\textbf{Student-specific descriptive fits.} Each slope comes from a separate regression on the uncensored tasks of that student. By Proposition~\ref{prop:pooled-slope} the shared exponent of Equation~\ref{eq:label-budget} is a weighted average of these slopes, and the shared fit remains the primary analysis.}
\label{tab:app-student-slopes}
\begin{tabular}{@{}lrr@{}}
\toprule
\rowcolor{AppGold} \AppHead{Student} & \AppHead{Tasks} & \AppHead{Separate slope} \\
\midrule
G270M & 13 & 0.694 \\
G1B & 14 & 0.696 \\
G4B & 14 & 0.641 \\
L1B & 14 & 0.640 \\
L3B & 12 & 0.454 \\
Q0.6B & 14 & 0.700 \\
Q1.7B & 13 & 0.685 \\
O1B & 15 & 0.629 \\
\bottomrule
\end{tabular}
\end{table}

\begin{figure}[t]
\centering
\includegraphics[width=\linewidth]{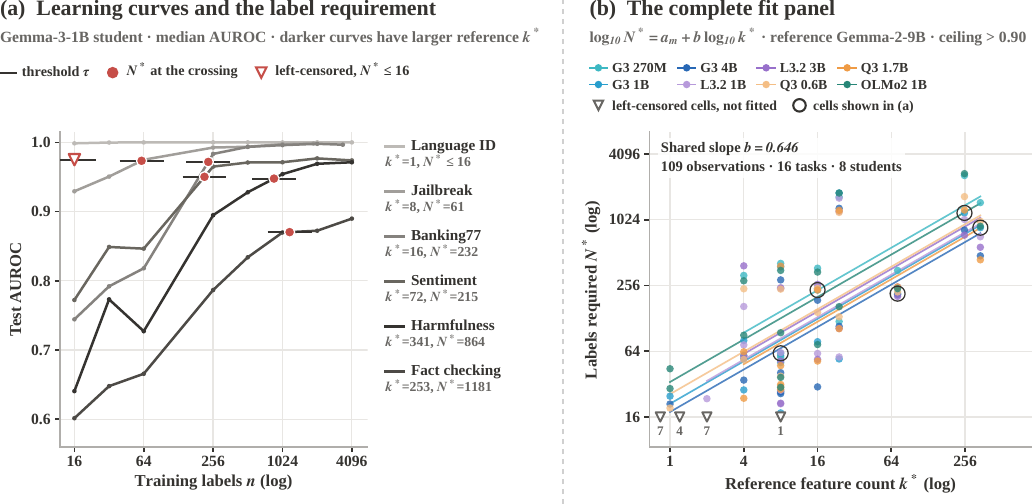}
\caption{\textbf{Learning curves behind the label requirement and the complete fit panel.}
Panel (a) shows the Gemma 3 1B student on six tasks whose reference counts in Gemma 2 9B are $k^*=1$ (language identification), 8 (jailbreak), 16 (Banking77), 72 (sentiment), 253 (fact checking), and 341 (harmfulness).
Each curve is the median test AUROC over seeds and class decisions at every realized training size from 16 to 4096 labels.
The short horizontal tick marks the threshold $\tau=0.5+0.95\,[\bar A(n_J)-0.5]$ of Equation~\ref{eq:nstar-threshold}, and the dot marks $N^*$ from the geometric interpolation of Equation~\ref{eq:nstar-interpolation}.
The curves give $N^*=61$ for jailbreak, 232 for Banking77, 215 for sentiment, 1181 for fact checking, and 864 for harmfulness.
Language identification already exceeds its threshold at 16 labels, so its requirement is left-censored at $N^*\leq16$ and appears as an open triangle.
Panel (b) plots all 109 uncensored observations over 16 tasks and 8 students with one color per student and the fitted lines of Equation~\ref{eq:label-budget}, which share the slope $b=0.646$ and differ only in their student intercepts.
Open triangles at $N^*=16$ mark the 19 left-censored cells, with the number of cells below each triangle (7 for language identification, 4 for tool selection, 7 for routing, and 1 for subjectivity), and the regression excludes them.
Rings mark the five uncensored cells of panel (a).}
\label{fig:app-learning}
\end{figure}

\textbf{Result.}
Table~\ref{tab:app-nstar} lists the contributing tasks together with the left-censored cells that the regression excludes, and the resulting panel contains 109 observations.
Figure~\ref{fig:app-learning} shows the learning curves of one student with their thresholds and crossings, and the full 109-observation panel with the fitted parallel lines.
Seven of the eight separate slopes lie between 0.629 and 0.700, which shows that the shared exponent does not average over students with opposite trends.
Llama-3.2-3B has the flattest relationship at 0.454, and we keep it in the primary fit.
Its slope is sensitive to the CLINC150 observation, whose curve crosses its threshold close to the smallest budget in several other students.

\textbf{Scope.}
The fit uses 16 of the 24 tasks, namely those with $C_t>0.90$ and at least one uncensored requirement.
The exponent describes how requirements scale across tasks within this range of $k^*_t$, from 1 to 341, and it makes no claim about extrapolation beyond it.

\textbf{Finding.}
Across eight students from four families, the label requirement grows as a power of the frozen Gemma-2-9B reference model's count over 16 tasks, with a shared exponent of 0.65 that seven of the eight separate student fits reproduce to within 0.06.

\textbf{Implication.}
A single measurement of $k^*$ on one reference model supplies a relative ranking of annotation needs across tasks before any student is trained.

\subsubsection{Task-Block Uncertainty and Held-Out Prediction}
\label{app:label-inference}

\textbf{Question.}
How precisely do the 109 observations determine the shared exponent, and how accurately does the fit predict the requirement of a task it has not seen?

\textbf{Unit of resampling.}
Observations from different students share tasks and reference counts, so the 109 rows are not independent draws.
We resample whole task blocks, which is a block bootstrap~\citep{efron1992bootstrap} with the task as the block.
Let $\mathcal B_t=\{(t,m)\in\mathcal P\}$ be the block of task $t$ and $T=16$ the number of tasks.
Each bootstrap replicate draws task identities $t^{(1)},\dots,t^{(T)}$ uniformly with replacement and fits Equation~\ref{eq:label-budget} to the concatenation $\mathcal B_{t^{(1)}}\cup\dots\cup\mathcal B_{t^{(T)}}$ taken as a multiset, so a task drawn twice contributes its block twice.
We use 10,000 replicates with seed 0 and skip replicates with fewer than three distinct tasks, since the slope is not estimable from fewer.
The reported interval is the 2.5th to 97.5th percentile of the retained slopes $\widehat b^{(1)},\dots,\widehat b^{(B)}$,
\begin{equation}
    \mathrm{CI}_{0.95}(b)=\bigl[\,Q_{0.025}(\{\widehat b^{(i)}\}),\ Q_{0.975}(\{\widehat b^{(i)}\})\,\bigr].
    \label{eq:task-bootstrap}
\end{equation}

\textbf{Held-out error.}
For each task $t$, we refit Equation~\ref{eq:label-budget} on $\mathcal P\setminus\mathcal B_t$ and predict every row of $\mathcal B_t$.
Writing $\widehat y^{(-t)}_{tm}$ for this leave-one-task-out prediction, the error summary is
\begin{equation}
 e_{0.9}=Q_{0.9}\!\left(\left\{\bigl|\log_{10}N^*_{tm}-\widehat y^{(-t)}_{tm}\bigr|\right\}_{(t,m)\in\mathcal P}\right),
 \qquad F_{0.9}=10^{e_{0.9}}.
 \label{eq:heldout-error}
\end{equation}

\textbf{Remark (multiplicative reading of the held-out error).}
For any observation with $\bigl|\log_{10}N^*_{tm}-\widehat y^{(-t)}_{tm}\bigr|\leq e_{0.9}$, exponentiating the two inequalities with base 10 shows that the prediction $\widehat N^{(-t)}_{tm}=10^{\widehat y^{(-t)}_{tm}}$ satisfies $\widehat N^{(-t)}_{tm}/F_{0.9}\leq N^*_{tm}\leq F_{0.9}\,\widehat N^{(-t)}_{tm}$, and by the definition of the empirical 0.9 quantile at least 90\% of the observations in $\mathcal P$ satisfy the premise.

By this remark, the reported factor of 5.3 means that at least nine of ten held-out requirements fall within a factor of 5.3 of their leave-one-task-out prediction in either direction.

\begin{algorithm}[t]
\AppAlgorithmStyle
\caption{Task-block bootstrap and leave-one-task-out prediction}
\label{alg:task-bootstrap}
\begin{algorithmic}[1]
\Require panel $\mathcal P$ from Algorithm~\ref{alg:nstar-panel} with task blocks $\mathcal B_t$
\State Fit the shared slope and student intercepts by least squares
\For{$i=1,\ldots,10000$}
 \State Draw $T$ task identities with replacement
 \If{at least three distinct tasks are drawn}
  \State Concatenate their blocks, retaining repeated blocks
  \State Refit Equation~\ref{eq:label-budget} and retain its slope $\widehat b^{(i)}$
 \EndIf
\EndFor
\State Report the 2.5th and 97.5th percentiles of the retained slopes
\For{each task $t$}
 \State Fit on $\mathcal P\setminus\mathcal B_t$
 \State Predict every row of $\mathcal B_t$ and retain its absolute log error
\EndFor
\State Report the pooled 90th-percentile error $e_{0.9}$ and the factor $10^{e_{0.9}}$
\end{algorithmic}
\end{algorithm}

\begin{table}[!ht]
\centering\AppTableStyle
\caption{\textbf{Shared fit and task-level uncertainty.} The percentile interval uses 10,000 task-block bootstrap draws with multiplicities retained. Held-out error leaves whole task blocks out.}
\label{tab:app-inference}
\begin{tabular}{@{}lr@{}}
\toprule
\rowcolor{AppGold} \AppHead{Quantity} & \AppHead{Value} \\
\midrule
Observations / tasks / students & 109 / 16 / 8 \\
Shared exponent & 0.645848 \\
95\% task-block interval & [0.438945, 0.907697] \\
90th-percentile held-out log error & 0.727170 \\
Corresponding multiplicative error & 5.335 \\
\bottomrule
\end{tabular}
\end{table}

\textbf{Design rationale.}
Resampling rows would treat eight observations of one task as eight independent pieces of evidence, although they share one reference count and one set of texts.
The task block is the unit that varies when a new task is added, so it is the unit we resample.
Leave-one-task-out prediction answers the practical question of how well the fit anticipates the requirement of a task it has not seen, which an in-sample residual cannot answer.

\textbf{Null and result.}
If requirements did not depend on the reference count, the task-block interval would cover zero.
The interval $[0.44,0.91]$ excludes zero and also excludes one, so requirements grow with $k^*_t$ but more slowly than in proportion to it.
The held-out factor of 5.3 states the precision of individual predictions and is the relevant quantity for any use of the fit as a budget planner.

\textbf{Scope.}
The held-out error is a pooled quantile over observations, so tasks with more eligible students contribute more errors.
It is neither a calibrated conformal interval nor a coverage guarantee for arbitrary unseen tasks.
The bootstrap conditions on the observed task collection, the measured reference counts, and the fitted student curves, and it does not propagate the seed variation of $k^*_t$ reported in Appendix~\ref{app:aggregation}.
Retaining multiplicities is essential for a valid block bootstrap, since collapsing a draw to its distinct tasks changes the sampling distribution of the replicate.

\textbf{Finding.}
The exponent interval excludes both zero and one, so requirements grow with the count but sublinearly.
Individual forecasts carry a held-out factor of 5.3.

\textbf{Implication.}
The fit supports planning at the level of orders of magnitude and task rankings, and a practitioner should budget with the factor of 5.3 in mind rather than with the point prediction alone.

\subsubsection{Incremental Predictive Value and Teacher Controls}
\label{app:learning-controls}

\textbf{Question.}
Does $k^*_t$ carry information about student performance beyond transferability scores and difficulty proxies, and does the teacher's internal count matter once its labels are fixed?

\textbf{Estimator.}
The fixed-budget analysis relates $k^*_t$ to student AUROC at 256 labels after accounting for a control variable.
Let $r_x$ and $r_y$ be average-rank vectors over tasks, and let $Z$ contain an intercept and the ranked controls.
The partial rank correlation is
\begin{equation}
 e_x=r_x-ZZ^+r_x,\quad e_y=r_y-ZZ^+r_y,\qquad
 \rho_{xy\cdot Z}=\operatorname{Corr}(e_x,e_y),
 \label{eq:partial-rank}
\end{equation}
where $Z^+$ is the Moore–Penrose pseudoinverse, so $ZZ^+$ projects onto the column space of $Z$.
The residuals $e_x$ and $e_y$ are therefore the parts of the two rank vectors that the controls do not explain linearly.
The permutation diagnostic permutes $e_x$ 10,000 times with seed 0 and reports the fraction of permutations whose absolute correlation reaches the observed one.
This diagnostic tests association under the stated residual construction and is not a general test of conditional independence.

\textbf{Controls.}
The controls are LogME~\citep{you2021logme}, the H-score~\citep{bao2019information}, label entropy, the bag-of-words input ceiling of Appendix~\ref{app:inputbase}, and the student base-model ceiling where available.
With ten high-ceiling tasks, we add controls one at a time, and multi-control fits serve only as sensitivity analyses because they leave little residual variation.
The counts in this analysis come from the preliminary single-seed sweep of Appendix~\ref{app:aggregation}, not from the five-seed counts of the primary exponent fit, and the analysis reads the requested 256-label runs, a budget below the pool size of all ten tasks.
Table~\ref{tab:app-partial-controls} therefore reports a separate control analysis rather than a second fit of Equation~\ref{eq:label-budget}.

\begin{table}[!ht]
\centering\AppTableStyle
\caption{\textbf{Fixed-budget incremental validity on ten high-ceiling tasks.} Partial rank correlations with student AUROC at 256 labels, where $x\mid z$ denotes predictor $x$ with control $z$ and bag of words denotes the text-only ceiling of Appendix~\ref{app:inputbase}. The residual-permutation $p$ values are not adjusted for multiplicity. These controls use the single-seed counts of the preliminary sweep and the requested 256-label runs, as specified in Appendix~\ref{app:learning-controls}, which also reports the $N^*$-level control.}
\label{tab:app-partial-controls}
\begin{tabular}{@{}llrr@{}}
\toprule
\rowcolor{AppGold} \AppHead{Student} & \AppHead{Predictor / control} & \AppHead{$\rho$} & \AppHead{$p$} \\
\midrule
Gemma 3 270M & $k^*\mid$ LogME & -0.668 & 0.0343 \\
Gemma 3 270M & LogME $\mid k^*$ & +0.140 & 0.7071 \\
Gemma 3 270M & $k^*\mid$ H-score & -0.708 & 0.0228 \\
Gemma 3 270M & H-score $\mid k^*$ & +0.813 & 0.0028 \\
Gemma 3 270M & $k^*\mid$ bag of words & -0.547 & 0.0957 \\
Gemma 3 270M & Bag of words $\mid k^*$ & +0.551 & 0.0970 \\
Gemma 3 1B & $k^*\mid$ LogME & -0.763 & 0.0101 \\
Gemma 3 1B & LogME $\mid k^*$ & +0.012 & 0.9798 \\
Gemma 3 1B & $k^*\mid$ H-score & -0.786 & 0.0075 \\
Gemma 3 1B & H-score $\mid k^*$ & +0.736 & 0.0168 \\
Gemma 3 1B & $k^*\mid$ bag of words & -0.670 & 0.0316 \\
Gemma 3 1B & Bag of words $\mid k^*$ & +0.479 & 0.1603 \\
Gemma 3 4B & $k^*\mid$ LogME & -0.716 & 0.0221 \\
Gemma 3 4B & LogME $\mid k^*$ & +0.128 & 0.7348 \\
Gemma 3 4B & $k^*\mid$ H-score & -0.736 & 0.0153 \\
Gemma 3 4B & H-score $\mid k^*$ & +0.547 & 0.1080 \\
Gemma 3 4B & $k^*\mid$ bag of words & -0.649 & 0.0444 \\
Gemma 3 4B & Bag of words $\mid k^*$ & +0.406 & 0.2409 \\
\bottomrule
\end{tabular}
\end{table}

\textbf{Design rationale.}
Rank-based partial correlations test whether $k^*_t$ adds information beyond each transferability score without assuming linear relations with AUROC.
The teacher swap uses a natural experiment in which two teachers disagree in their internal counts but agree in their labels, which separates the path through labels from the path through representation sparsity.

\textbf{Result.}
Given LogME, $k^*_t$ keeps a partial correlation between $-0.668$ and $-0.763$ with student AUROC in all three students, whereas LogME given $k^*_t$ falls to between $+0.012$ and $+0.140$.
Given the H-score, both predictors keep substantial partial correlations, so the H-score and $k^*_t$ carry complementary information about student AUROC at 256 labels.
Given the bag-of-words ceiling of Appendix~\ref{app:inputbase}, $k^*_t$ keeps a partial correlation between $-0.547$ and $-0.670$, with residual-permutation $p$ values of 0.032 to 0.096, although the two predictors correlate at $-0.80$ over the ten tasks.
The same control on the 109 observations of Equation~\ref{eq:label-budget} leaves a partial rank correlation of $+0.51$ between $k^*_t$ and $N^*$ after removing each student's mean, with $p<10^{-4}$, and adding the bag-of-words ceiling as a covariate moves the shared exponent from 0.65 to 0.55 while $R^2$ changes from 0.525 to 0.528.
On the ten tasks of the fixed-budget analysis the bag-of-words ceiling exceeds 0.98 on all but fact checking, sentiment, and harmfulness, which limits this control there.

\begin{table}[!ht]
\centering\AppTableStyle
\caption{\textbf{What each learning control tests.} Distinct controls address distinct alternative explanations, and no single control identifies the cause of the relation between counts and requirements.}
\label{tab:learning-control-map}
\begin{tabular}{@{}p{0.23\linewidth}p{0.33\linewidth}p{0.36\linewidth}@{}}
\toprule
\rowcolor{AppGold} \AppHead{Control} & \AppHead{Comparison} & \AppHead{Interpretation} \\
\midrule
Reference-model replacement & Refit using another LM's measured counts & Instrument sensitivity on the same student curves \\
LogME and H-score & Partial rank correlations at fixed budget & Incremental or complementary predictive information \\
Difficulty proxies & Add measured proxies individually & Sensitivity to measured difficulty, not exhaustive adjustment \\
Base and IT hard teachers & Same task, different teacher-probe labels & Label transfer can stay similar despite different teacher counts \\
Soft teachers & Train on teacher probabilities & Probability information beyond hard labels \\
\bottomrule
\end{tabular}
\end{table}

\textbf{Teacher controls.}
For harmfulness and intent classification, probes on base and instruction-tuned Gemma-2-9B produce nearly identical labels despite different internal counts.
Substituting one teacher's hard labels for the other's changes student AUROC by at most 0.005 over the tested budgets, and soft-label curves remain close at larger budgets.
These controls change the teacher's labels and cannot change its representation sparsity alone, so they test the consequences of label substitution rather than a surgical intervention on sparsity.
Our class-mean approximation to the STS transferability score is likewise a simplified variant and not a reproduction of the original procedure.

\textbf{Finding.}
Given LogME, the count keeps a partial correlation of $-0.668$ to $-0.763$ with student AUROC in all three tested students, whereas LogME given the count retains at most $+0.140$.
Given the bag-of-words ceiling, the count keeps $-0.547$ to $-0.670$ at fixed budget and $+0.51$ with $N^*$ on the full panel of 109 observations across eight students.

\textbf{Implication.}
Readout sparsity is not a restatement of a transferability score or of text decidability, and in all three students it complements the H-score rather than replacing it.

\subsubsection{Multi-Task Budget Replay}
\label{app:allocation}

\textbf{Question.}
Under one total budget for several tasks, does allocation in proportion to predicted requirements reach a joint performance target with fewer labels than uniform allocation?

\textbf{Data.}
The replay includes the nine tasks whose Gemma-2-9B ceilings exceed 0.90, namely language identification, tool selection, routing, intent, PII detection, jailbreak detection, banking intent, harmfulness detection, and fact checking.
For each student and task, $a_t(n)$ is the median test accuracy over seeds and class decisions at realized size $n$, linearly interpolated in $\log n$ between measured sizes and held constant beyond the smallest and largest sizes.

\textbf{Estimator.}
For a total budget $B$ and nonnegative weights $w_t$, each policy assigns
\begin{equation}
 b_t(B)=B\frac{w_t}{\sum_u w_u},\qquad
 U(B)=\prod_t a_t\!\left(b_t(B)\right),
 \label{eq:allocation}
\end{equation}
and its cost for a target $U^\star$ is the smallest total budget with $U(B)\geq U^\star$.
We report every policy's cost as a ratio to the uniform policy's cost in the same setting.
The policies differ only in their weights,
\begin{equation}
    w^{\mathrm{uni}}_t=1,\qquad
    w^{\mathrm{pilot}}_t=\max\{10^{-6},\,1-a_t(16)\},\qquad
    w^{k}_t=k^*_t,\qquad
    w^{N}_t=10^{\,\widehat\alpha^{(-t)}}\bigl(k^*_t\bigr)^{\widehat\beta^{(-t)}},
    \label{eq:allocation-weights}
\end{equation}
where $(\widehat\alpha^{(-t)},\widehat\beta^{(-t)})$ is a least-squares fit of $\log_{10}N^*$ on $\log_{10}k^*$ over the other pipeline tasks of the same student, and a student leaves the predicted-requirement policy when any of its tasks has fewer than four other tasks to fit on, a condition that no tested pipeline triggers.
The requirement in this fit comes from accuracy curves with chance level 0.5, not from the AUROC curves of Equation~\ref{eq:label-budget}.

\textbf{Remark (monotone utility and exact search).}
If every curve $a_t$ is nondecreasing and nonnegative and every $w_t>0$, then $U(B)$ is nondecreasing in $B$, because each $b_t(B)$ increases linearly in $B$, log-linear interpolation of a nondecreasing sequence is nondecreasing, and a product of nonnegative nondecreasing functions is nondecreasing.
The set $\{B:U(B)\geq U^\star\}$ is then an interval $[B^\star,10^6]$, and geometric bisection on $[10^2,10^6]$ evaluates $U$ at the geometric midpoint $\sqrt{\ell h}$ of the current bracket $[\ell,h]$, keeps the half that contains $B^\star$, and halves $\log(h/\ell)$ at every step.
After 60 halvings the returned budget lies within a factor of $(10^4)^{2^{-60}}$ above $B^\star$, whenever the target is reachable within the interval and not already reached at its lower end.

The measured curves are unsmoothed and can be locally nonmonotone, in which case the bisection is a numerical policy evaluation rather than a certified minimum.
A replay on the running-maximum envelope of each curve, which satisfies the premise of this remark, serves as a sensitivity analysis.

\textbf{Remark (why the count rule overspends).}
If one fit with exponent $\widehat\beta$ served all tasks, the predicted-requirement weights would equal $w^{N}_t\propto(k^*_t)^{\widehat\beta}$, since the student intercept cancels in Equation~\ref{eq:allocation}.
The raw-count rule $w^k_t=k^*_t$ then corresponds to an exponent of one.
Whenever the fitted exponent lies below one, as the AUROC-based exponent of Equation~\ref{eq:label-budget} does, the raw-count rule concentrates more labels on the widest-readout tasks than their requirements warrant.
The leave-one-task-out fits make this correspondence approximate, since each task receives its own fitted pair.

\textbf{Design rationale.}
The replay evaluates every policy on the same measured learning curves, so differences between policies come from allocation alone and not from new training noise.
Predicted requirements for a task come from fits that exclude that task, so no task's own measured requirement enters its weight.
The pipeline admits only the nine tasks with $C_t>0.90$, and it excludes relevance, mathematical-step correctness, code-bug detection, and mathematical-answer correctness because their Gemma-2-9B reference ceilings fall below this gate.

\textbf{Null.}
Uniform allocation is the null policy, and a rule that carries useful allocation information should reach every joint target in every student at a cost ratio below one.

\begin{table}[!ht]
\centering\AppTableStyle
\caption{\textbf{Budget allocation replay relative to uniform allocation.} Ratios below one use fewer labels under the replay accounting. Only settings in which the uniform, pilot, count, and predicted-requirement policies all reach the target enter the 18-setting summary, and the oracle policy is excluded because its exact requirements are undefined for left-censored tasks.}
\label{tab:app-allocation}
\begin{tabular}{@{}lrrr@{}}
\toprule
\rowcolor{AppGold} \AppHead{Rule} & \AppHead{Median ratio} & \AppHead{Range} & \AppHead{Cheaper} \\
\midrule
Pilot error & 0.66 & [0.56, 0.82] & 18/18 \\
Predicted $N^*$ & 1.03 & [0.65, 1.98] & 6/18 \\
Reference $k^*$ & 4.03 & [2.29, 13.02] & 0/18 \\
\bottomrule
\end{tabular}
\end{table}

\textbf{Result.}
The predicted-requirement rule costs a median of 1.03 times the uniform budget and is cheaper in 6 of 18 settings, so it does not consistently save labels.
The raw-count rule costs between 2.29 and 13.02 times the uniform budget and is cheaper in none of the 18 settings, consistent with the remark above.
The pilot rule costs a median of 0.66 times the uniform budget and is cheaper than uniform allocation in all 18 settings under the stated accounting.

\textbf{Scope.}
We evaluate targets $U^\star\in\{0.50,0.60,0.70\}$, and the 18 settings comprise eight students at each of the first two targets and two students at the third, since the other students have at least one policy that cannot reach the target within $10^6$ labels.
The pilot rule sets its weights from 16-label runs, but the replay neither adds the pilot's own labels and training runs to $B$ nor enforces $b_t\geq16$, so its ratio describes this accounting convention rather than an end-to-end budget.
The product in Equation~\ref{eq:allocation} is a replay utility and not an observed probability of joint success, since we estimate no dependence among task errors.
The result concerns proportional allocation, and other allocation policies that use count information remain untested.
Predicting where a learning curve reaches its criterion differs from locating where additional labels most improve a joint objective, and the proportional policies of Equation~\ref{eq:allocation-weights} use only the former.

\textbf{Finding.}
Allocating labels in proportion to predicted requirements reaches the three joint targets at a median of 1.03 times the uniform budget over 18 settings, and allocating in proportion to raw reference counts costs 2.29 to 13.02 times the uniform budget in the same settings.

\textbf{Implication.}
Forecasting where each learning curve reaches its criterion does not by itself specify where the next label is most valuable to the joint target.
The boundary of usefulness is precise, since proportional rules built on counts or predicted requirements do not save labels in the tested pipelines, while a 16-label pilot does in all 18 settings under the stated accounting.

\subsubsection{Retention, Readability, and Recovery under Continual Learning}
\label{app:continual}

\textbf{Question.}
When sequential training disrupts an earlier task's classification head, does the backbone still carry linearly readable information about that task, and how many labels does recovery need compared with initial learning on the pretrained backbone?

\textbf{Models and data.}
The common chain trains harmfulness detection, then sentiment, then jailbreak detection, with three chain seeds in each of Gemma-3-1B, Gemma-3-4B, Llama-3.1-8B, and Qwen3-8B.
Expanded Gemma-3-1B experiments change the initial task or the order, and six-task chains append banking intent, CLINC150, and language identification, which Llama-3.1-8B also runs with three seeds.
Each step trains one binary class decision on up to 4096 labels.

\textbf{Procedure.}
Each chain fine-tunes all backbone parameters on one task at a time with a fresh classification head, and the backbone after step $s$ initializes step $s+1$.
Old heads are saved and evaluated on each updated backbone without refitting.
Separately, we freeze each updated backbone, extract its representations, and fit new probes for the evaluation tasks by the procedure of Appendix~\ref{app:probes}.
The 4B and 8B backbones train with gradient checkpointing.
Algorithm~\ref{alg:continual} states the full procedure.

\textbf{Estimator.}
For task $t$, let $H_t^{(s)}$ be the saved-head AUROC after step $s$, let $s_t$ be the step at which task $t$ was trained, and let $C_t^{(s)}$ be the dense-probe ceiling on the backbone after step $s$, with $s=0$ denoting the pretrained checkpoint.
The two reported changes use different reference states,
\begin{equation}
 L^{\rm head}_{t,s}=H_t^{(s_t)}-H_t^{(s)},\qquad
 L^{\rm ceiling}_{t,s}=C_t^{(0)}-C_t^{(s)}.
 \label{eq:forgetting-channels}
\end{equation}
The ceiling $C_t^{(s)}$ is the largest dense-probe AUROC over the evaluation layers, and the accompanying count is the neighbor-confirmed count of Equation~\ref{eq:nc}, both computed per class decision.

\textbf{Remark (the two losses are not a decomposition).}
The saved head reads the final non-padding position of the last layer through weights fitted before the update, whereas the ceiling reads mean-pooled representations at the evaluation layers through a probe fitted after it.
Neither predictor bounds the other, so $L^{\rm head}_{t,s}-L^{\rm ceiling}_{t,s}$ has no interpretation as a readout component of forgetting.
Instead, a large $L^{\rm head}$ together with a small $L^{\rm ceiling}$ shows qualitatively that the updated representation still supports an accurate linear readout that the saved head no longer implements.

\begin{algorithm}[t]
\AppAlgorithmStyle
\caption{Separate behavioral retention from frozen readability}
\label{alg:continual}
\begin{algorithmic}[1]
\Require pretrained backbone $\theta_0$, ordered tasks $(t_1,\ldots,t_S)$
\For{$s=1,\ldots,S$}
 \State Initialize a new classification head and train it with all weights of $\theta_{s-1}$ on $t_s$
 \State Save the updated backbone $\theta_s$ and the trained head for $t_s$
 \State Evaluate each available saved head on $\theta_s$ without refitting
 \State Freeze $\theta_s$, extract representations, and fit new probes for the evaluation tasks
 \State Record head AUROC, dense ceiling, and neighbor-confirmed count
\EndFor
\For{each recovery budget and recovery seed}
 \State Start independently from $\theta_S$, and train fresh LoRA adapters and a new head
 \State Evaluate on the original test split of the recovered task
\EndFor
\State Read recovery and initial-learning requirements at one target on the shared grid
\end{algorithmic}
\end{algorithm}

\textbf{Baselines.}
The Gemma-3-1B baseline $C_t^{(0)}$ comes from the pretrained model's single-seed preliminary sweep of Appendix~\ref{app:aggregation}, and the other three models use the first seed of their five-seed measurement.
Baseline and post-update rows pass through the same ceiling and count estimator, so every comparison is internal to one chain and one estimator.
These are within-chain comparisons and do not replace the five-seed task-level counts of Algorithm~\ref{alg:aggregate}.
Probe coverage varies across chains, and in the third seed of the six-task Llama chain three of the six evaluation tasks were not probed, which we treat as missing rather than as retained.

\textbf{Recovery estimator.}
Each final chain backbone receives three independent recovery seeds at each of the budgets $16,64,256,1024$.
We first aggregate the recovery AUROCs into one curve per final backbone, so each chain seed yields one recovery curve.
The initial-learning curve on the same pretrained backbone comes from the eight-point grid $16,32,64,256,512,1024,2048,4096$, and Llama-3.1-8B and Qwen3-8B supply learning curves beyond the eight-student panel of Section~\ref{app:label-count}.
We restrict this curve to the four recovery budgets and set the target $\tau=0.5+0.95\,[\bar A^{\rm init}(1024)-0.5]$, the threshold of Equation~\ref{eq:nstar-threshold} at the shared endpoint.
Both requirements are the first crossing of $\tau$ under the interpolation of Equation~\ref{eq:nstar-interpolation}, so the initial requirement equals its own $N^*$ on the shared grid and the recovery requirement is read against the same AUROC.
Table~\ref{tab:recovery} reports the median and range of the recovery requirements over the three chain seeds, together with recovery $N^*$ against each recovery curve's own endpoint and initial $N^*$ on the full eight-point grid.

\textbf{Design rationale.}
Re-probing the frozen updated backbone separates two explanations of forgetting that a saved-head evaluation alone confounds.
A failing head can reflect a readout that no longer matches the representation, or a representation that no longer carries the information.
Each loss in Equation~\ref{eq:forgetting-channels} uses the reference state that matches its question, and baseline and updated rows pass through one ceiling and count estimator.
Recovery and initial learning run on the same four backbones with the same adapter family, so the comparison isolates the effect of starting from the updated backbone up to the input-length difference stated in Section~\ref{app:student-training}.
Reading both curves on one grid against one target removes two differences that own-endpoint requirements on unequal grids would carry, a higher endpoint and denser interior points on the longer grid.

\textbf{Null.}
If sequential training erased the task information that a linear probe reads, $L^{\rm ceiling}$ would be comparable to $L^{\rm head}$.
If the updated backbone offered no advantage for relearning the disrupted task, recovery and initial-learning requirements would be similar.

\begin{table}[!ht]
\centering\AppTableStyle
\caption{\textbf{Recovering harmfulness detection after the same three-task sequence requires fewer labels than initial learning.} Both curves are read on the recovery grid of 16 to 1024 labels. The target is the AUROC at which initial learning attains 95\% of its gain above chance at 1024 labels, and the first requirement columns give the labels that each procedure needs to reach it. Recovery values give the median and, in brackets, the range over three chain seeds. The last two columns give recovery $N^*$ against its own endpoint and initial $N^*$ on the full grid of 16 to 4096 labels.}
\label{tab:recovery}
\begin{tabular}{@{}lrrrrr@{}}
\toprule
\rowcolor{AppGold} \AppHead{Model} & \AppHead{Target} & \AppHead{Initial} & \AppHead{Recovery} & \AppHead{Recovery, own endpoint} & \AppHead{Initial, 16 to 4096} \\
\midrule
Gemma-3-1B & 0.931 & 602 & \textbf{101} [77, 129] & 182 [133, 208] & 864 \\
Gemma-3-4B & 0.951 & 520 & \textbf{155} [141, 229] & 163 [163, 224] & 478 \\
Llama-3.1-8B & 0.949 & 524 & \textbf{244} [237, 544] & 236 [220, 456] & 497 \\
Qwen3-8B & 0.947 & 194 & \textbf{126} [43, 164] & 176 [61, 200] & 269 \\
\bottomrule
\end{tabular}
\end{table}

\begin{table}[!ht]
\centering\AppTableStyle
\caption{\textbf{Harmfulness recovery curves after the common three-task chain.} We first aggregate the three recovery seeds within each final backbone and then report the median and range over three chain seeds. These endpoints supplement the label requirements in Table~\ref{tab:recovery}.}
\label{tab:app-cl-curves}
\begin{tabular}{@{}lrrr@{}}
\toprule
\rowcolor{AppGold} \AppHead{Model} & \AppHead{Labels} & \AppHead{Median AUROC} & \AppHead{Chain range} \\
\midrule
Gemma 3 1B & 16 & 0.754 & [0.749, 0.862] \\
Gemma 3 1B & 64 & 0.920 & [0.908, 0.926] \\
Gemma 3 1B & 256 & 0.955 & [0.955, 0.965] \\
Gemma 3 1B & 1024 & 0.970 & [0.970, 0.971] \\
Gemma 3 4B & 16 & 0.763 & [0.720, 0.784] \\
Gemma 3 4B & 64 & 0.915 & [0.827, 0.931] \\
Gemma 3 4B & 256 & 0.966 & [0.961, 0.971] \\
Gemma 3 4B & 1024 & 0.977 & [0.972, 0.978] \\
Llama 3.1 8B & 16 & 0.658 & [0.653, 0.669] \\
Llama 3.1 8B & 64 & 0.846 & [0.839, 0.888] \\
Llama 3.1 8B & 256 & 0.951 & [0.927, 0.955] \\
Llama 3.1 8B & 1024 & 0.967 & [0.966, 0.971] \\
Qwen 3 8B & 16 & 0.873 & [0.727, 0.910] \\
Qwen 3 8B & 64 & 0.920 & [0.905, 0.962] \\
Qwen 3 8B & 256 & 0.975 & [0.968, 0.984] \\
Qwen 3 8B & 1024 & 0.984 & [0.980, 0.985] \\
\bottomrule
\end{tabular}
\end{table}

\begin{figure}[t]
    \centering
    \includegraphics[width=\textwidth]{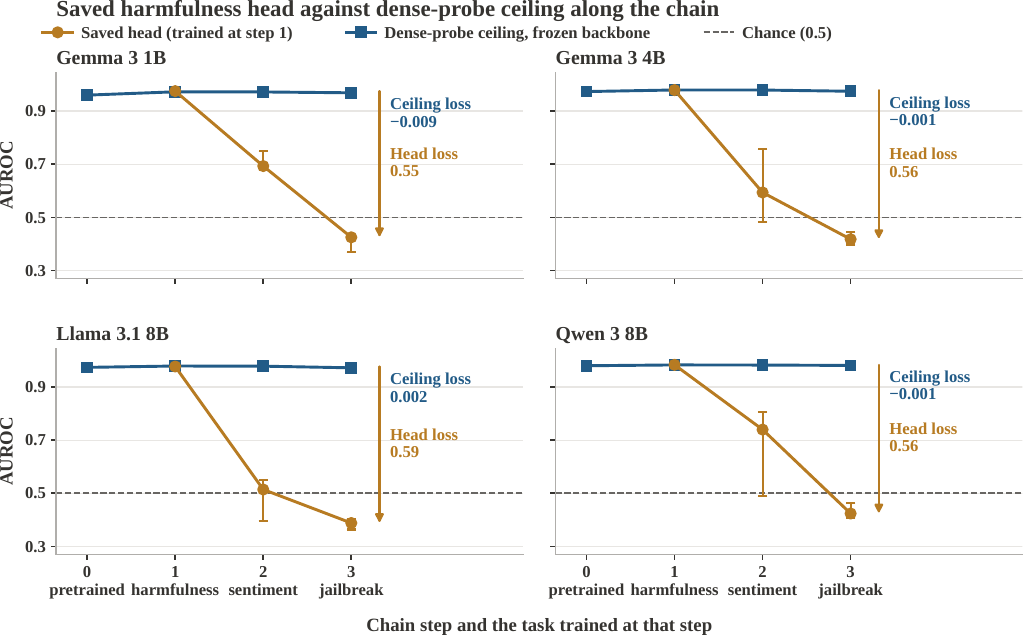}
    \caption{\textbf{Saved heads fail while new probes keep reading harmfulness along the common chain.}
    Each panel follows one model through the chain that trains harmfulness, then sentiment, then jailbreak detection, with three chain seeds.
    Gold circles give the AUROC of the harmfulness head saved after step 1 and evaluated on each later backbone without refitting.
    Blue squares give the dense-probe ceiling of new probes fitted on the frozen backbone after each step, and step 0 denotes the pretrained model.
    Markers show the median over the three chain seeds and whiskers span the seed range.
    The dashed line marks chance.
    The arrow and the labels report the final losses of Equation~\ref{eq:forgetting-channels} from the seed medians.
    Head losses are 0.55, 0.56, 0.59 and 0.56 for Gemma-3-1B, Gemma-3-4B, Llama-3.1-8B and Qwen3-8B, and the ceiling losses are $-0.009$, $-0.001$, $0.002$ and $-0.001$, where a negative value means that the final ceiling exceeds the pretrained ceiling.
    Over every step and seed of these twelve chains the largest harmfulness ceiling loss stays below 0.010.}
    \label{fig:app-continual}
\end{figure}

\textbf{Result.}
Over the twelve common-chain seeds, saved-head losses of 0.52 to 0.61 coexist with harmfulness ceiling losses of at most 0.010, and the six-task chains of Gemma-3-1B and Llama-3.1-8B extend the head losses to 0.47 to 0.68 while their ceiling losses stay at most 0.029.
The harmfulness medians of the main text are 0.55 to 0.59 for the heads against at most 0.009 for the ceilings.
On the shared grid, recovery reaches the common target with a median of 101 to 244 labels against 194 to 602 for initial learning, a reduction of 1.5- to 6.0-fold, and every recovery median lies below its model's initial requirement.
One Llama-3.1-8B chain seed needs 544 labels, above that model's initial 524.
Against each recovery curve's own endpoint the reduction is 2.2- to 3.3-fold in three models, and on Qwen3-8B the two requirements are 176 and 194, because its initial curve falls from 0.978 at 512 labels to 0.971 at 1024 and so lowers its own reference.
On the full eight-point grid the initial requirements are 269 to 864, so the requirement moves with the grid, which is why the comparison holds the grid fixed.
Figure~\ref{fig:app-continual} plots the head and ceiling trajectories of all four models along the chain.
The recovery curves of Table~\ref{tab:app-cl-curves} reach AUROC between 0.967 and 0.984 at 1024 labels, so the recovered heads attain high absolute performance and not only a fast approach to a low endpoint.

\textbf{Scope.}
The ceiling bound depends on chain length.
Three-step chains lose at most 0.015 of the pretrained ceiling, whereas the six-step Llama chain shows a small accumulation after the fourth step that reaches 0.029, which rules out a claim of zero information loss.
The comparison uses the recovery grid, which ends at 1024 labels, so it does not show how the two procedures compare at larger budgets, and Table~\ref{tab:app-cl-curves} supplies the absolute recovery endpoints.
Across six tasks, pre-update $k^*$ and the recovery discount show a positive trend with $\rho=0.77$ and $p=0.072$, which we report as a descriptive trend and not as a prediction law.

\textbf{Finding.}
After sequential training, saved heads lose 0.47 to 0.68 AUROC while dense probes on the same frozen backbones stay within 0.029 of the pretrained ceiling, and recovery reaches a common target with 101 to 244 labels in median against 194 to 602 for initial learning.

\textbf{Implication.}
A failing saved head does not show that the task information is gone.
Re-probing the frozen backbone distinguishes a displaced readout from lost readability at the cost of one linear fit, and when readability survives, a small recovery budget suffices in the tested chains.

\subsubsection{Pruning Operators, Calibration, and Statistical Family}
\label{app:pruning}

\textbf{Question.}
Does the pre-pruning count of a task predict how much linear readability that task loses when the weights of the same model are pruned by magnitude or by Wanda?

\textbf{Models and data.}
We prune Gemma-3-1B, Gemma-2-9B, OLMo-2-7B, and Llama-3.1-8B and probe six tasks, namely language identification, CLINC150, jailbreak detection, banking intent, sentiment, and harmfulness detection, at requested sparsities of 30\%, 50\%, 60\%, 70\%, 80\%, and 90\%.

\textbf{Operators.}
Both methods act on every linear map inside the transformer blocks.
Magnitude pruning treats each weight matrix $W$ with $|W|$ entries separately and zeros the entries whose magnitude does not exceed the $\lfloor s|W|\rfloor$th smallest absolute value in that matrix,
\begin{equation}
    W_{ij}\leftarrow W_{ij}\,\mathbf1\bigl[\,|W_{ij}|>\theta_W\,\bigr],
    \qquad
    \theta_W=\bigl|W\bigr|_{(\lfloor s|W|\rfloor)},
    \label{eq:magnitude-prune}
\end{equation}
where $s$ is the requested sparsity and $|W|_{(k)}$ is the $k$th order statistic of the absolute entries, so ties at the threshold can raise the realized fraction slightly above $s$.
Wanda~\citep{sun2024simple} scores each weight by its magnitude times the norm of its input feature over a calibration set,
\begin{equation}
 \nu_j=\Bigl(\sum_{x,t}u_j(x,t)^2\Bigr)^{1/2},\qquad
 S_{ij}=|W_{ij}|\,\nu_j,
 \label{eq:wanda-score}
\end{equation}
where $u_j(x,t)$ is the $j$th input coordinate of the linear map at position $t$ of calibration text $x$.
Each output row then zeros its $\lfloor s\,d_{\rm in}\rfloor$ lowest-scoring weights, so every row reaches the same sparsity.
The default calibration draws 128 texts with seed 0 from the union of the evaluated tasks' training pools, and a Gemma-2-9B calibration control substitutes Pile text with the same count and sampling rule.

\textbf{Estimator.}
After pruning, we fit dense probes on the pruned backbone for up to three class decisions per task, take the median dense AUROC over the evaluated class decisions at each probed layer, and take the maximum of these medians over the probed layers,
\begin{equation}
 C^{\rm prune}_{t,m,a,s}=\max_\ell\ \operatorname{median}_{c\in\mathcal C_t}
 A^{\rm dense}_{t,c,m,a,s,\ell},\qquad
 D_{t,m,a,s}=C_{t,m}-C^{\rm prune}_{t,m,a,s},
 \label{eq:pruning-loss}
\end{equation}
where $a$ indexes the pruning method and $C_{t,m}$ is the unpruned task ceiling from Algorithm~\ref{alg:aggregate}.
The baseline count $k^*_{t,m}$ comes from the same unpruned measurement.

\textbf{Test statistic.}
Each condition $(m,a,s)$ pairs the six baseline counts with the six ceiling losses and computes the Spearman correlation~\citep{spearman1987proof} $\rho_{\rm obs}$ with average ranks for ties.
The one-sided exact test enumerates all $6!$ relabelings of the losses,
\begin{equation}
 p=\frac{1}{6!}\sum_{\pi\in\mathfrak S_6}
 \mathbf1\!\left\{\operatorname{Corr}(r_{k},\,r_{D}\circ\pi)\geq\rho_{\rm obs}-10^{-12}\right\},
 \label{eq:pruning-permutation}
\end{equation}
where the tolerance $10^{-12}$ guards the comparison against floating-point error.

\begin{applemma}[exactness and resolution of the permutation test]
\label{lem:perm}
Under the null hypothesis that the pairing of counts with losses is exchangeable, $\Pr(p\leq\alpha)\leq\alpha$ for every $\alpha\in[0,1]$.
Without ties, the smallest attainable value is $p=1/6!$, reached exactly when the two rankings agree.
\end{applemma}

\begin{appproof}
Under exchangeability, the observed pairing is a uniformly random element of the orbit $\{r_D\circ\pi:\pi\in\mathfrak S_6\}$, and $p$ is the fraction of the orbit whose statistic reaches the observed one.
The standard argument for randomization tests then gives $\Pr(p\leq\alpha)\leq\alpha$, since at most $\lfloor\alpha\,6!\rfloor$ orbit elements can have a $p$ value of at most $\alpha$.
Without ties, $\rho=1$ holds for exactly one relabeling, so $\rho_{\rm obs}=1$ gives $p=1/6!$ and every other permutation gives a strictly smaller correlation.
\end{appproof}

Lemma~\ref{lem:perm} shows that six tasks give the test a resolution of $1/6!$, so the statistic and its $p$ value take finitely many values.
This discreteness explains the repeated values of $\rho$ and $q$ in Table~\ref{tab:app-pruning-ranks}, and it makes the test conservative rather than liberal at every nominal level.

\textbf{Multiplicity.}
We apply the Benjamini--Hochberg adjustment~\citep{benjamini1995controlling} to the 32 conditions at sparsities of at least 60\%, formed by four models, two methods, and four sparsities.
For sorted values $p_{(1)}\leq\dots\leq p_{(M)}$ with $M=32$,
\begin{equation}
 q_{(i)}=\min\Bigl\{1,\ \min_{j\geq i}\frac{M}{j}\,p_{(j)}\Bigr\}.
 \label{eq:bh-adjustment}
\end{equation}
The 32 conditions share tasks, model weights, and baseline counts and ceilings, so they are not independent replications.
The adjustment is a conventional control of the reported family, and we do not claim a false-discovery guarantee under arbitrary dependence.
Conditions at 30\% and 50\% sparsity are descriptive, since their losses are too small to order reliably.

\begin{table}[!ht]
\centering\AppTableStyle
\caption{\textbf{Correlation of pre-pruning counts with ceiling losses.} Cells give Spearman $\rho$ between the six baseline counts and the unrounded ceiling losses of Equation~\ref{eq:pruning-loss}. At 60\% to 90\% sparsity, the small number beside each correlation gives the Benjamini--Hochberg $q$ over those 32 conditions. Each one-sided test enumerates all $6!=720$ permutations. Low-sparsity correlations are descriptive, and at 30\% sparsity every maximum ceiling loss is below 0.007.}
\label{tab:app-pruning-ranks}
\begin{tabular}{@{}llrrrrrr@{}}
\toprule
\rowcolor{AppGold} \AppHead{Model} & \AppHead{Method} & \AppHead{30\%} & \AppHead{50\%} & \AppHead{60\%} & \AppHead{70\%} & \AppHead{80\%} & \AppHead{90\%} \\
\midrule
Gemma 3 1B & Magnitude & 0.429 & 0.943 & 0.886\,{\scriptsize\textcolor{AppMuted}{0.0178}} & 0.886\,{\scriptsize\textcolor{AppMuted}{0.0178}} & 0.600\,{\scriptsize\textcolor{AppMuted}{0.1208}} & 0.886\,{\scriptsize\textcolor{AppMuted}{0.0178}} \\
Gemma 3 1B & Wanda & 0.943 & 0.600 & 0.886\,{\scriptsize\textcolor{AppMuted}{0.0178}} & 0.886\,{\scriptsize\textcolor{AppMuted}{0.0178}} & 0.600\,{\scriptsize\textcolor{AppMuted}{0.1208}} & 0.886\,{\scriptsize\textcolor{AppMuted}{0.0178}} \\
Gemma 2 9B & Magnitude & 0.257 & 0.943 & 1.000\,{\scriptsize\textcolor{AppMuted}{0.0074}} & 0.943\,{\scriptsize\textcolor{AppMuted}{0.0127}} & 0.943\,{\scriptsize\textcolor{AppMuted}{0.0127}} & 0.886\,{\scriptsize\textcolor{AppMuted}{0.0178}} \\
Gemma 2 9B & Wanda & 0.257 & 0.943 & 1.000\,{\scriptsize\textcolor{AppMuted}{0.0074}} & 0.943\,{\scriptsize\textcolor{AppMuted}{0.0127}} & 0.943\,{\scriptsize\textcolor{AppMuted}{0.0127}} & 0.943\,{\scriptsize\textcolor{AppMuted}{0.0127}} \\
OLMo 2 7B & Magnitude & 0.714 & 0.771 & 0.943\,{\scriptsize\textcolor{AppMuted}{0.0127}} & 0.943\,{\scriptsize\textcolor{AppMuted}{0.0127}} & 0.943\,{\scriptsize\textcolor{AppMuted}{0.0127}} & 0.943\,{\scriptsize\textcolor{AppMuted}{0.0127}} \\
OLMo 2 7B & Wanda & 0.257 & 1.000 & 1.000\,{\scriptsize\textcolor{AppMuted}{0.0074}} & 0.943\,{\scriptsize\textcolor{AppMuted}{0.0127}} & 0.886\,{\scriptsize\textcolor{AppMuted}{0.0178}} & 0.886\,{\scriptsize\textcolor{AppMuted}{0.0178}} \\
Llama 3.1 8B & Magnitude & 0.429 & 1.000 & 0.943\,{\scriptsize\textcolor{AppMuted}{0.0127}} & 0.943\,{\scriptsize\textcolor{AppMuted}{0.0127}} & 0.943\,{\scriptsize\textcolor{AppMuted}{0.0127}} & 0.943\,{\scriptsize\textcolor{AppMuted}{0.0127}} \\
Llama 3.1 8B & Wanda & 0.928 & 0.600 & 1.000\,{\scriptsize\textcolor{AppMuted}{0.0074}} & 1.000\,{\scriptsize\textcolor{AppMuted}{0.0074}} & 0.943\,{\scriptsize\textcolor{AppMuted}{0.0127}} & 1.000\,{\scriptsize\textcolor{AppMuted}{0.0074}} \\
\bottomrule
\end{tabular}
\end{table}

\textbf{Design rationale.}
Two pruning methods with different scoring rules test whether the order of losses is a property of the tasks or of one scoring function.
With six tasks per condition, an asymptotic correlation test would be unreliable, so we enumerate every relabeling and obtain exact $p$ values.
The adjustment family consists of every condition at 60\% sparsity or higher, a cut defined by sparsity and not by the observed correlations.
Magnitude pruning uses no calibration data at all, and the Pile calibration control on Gemma-2-9B replaces the task texts with generic text, so together they test whether the losses arise from calibrating on the evaluated tasks.

\textbf{Null and result.}
If pre-pruning counts carried no information about pruning losses, the correlations would scatter around zero and the $q$ values would rarely fall below 0.05.
Instead, 30 of the 32 conditions at 60\% sparsity or higher reach $\rho\geq0.886$ with $q<0.05$, and the two exceptions are the 80\% conditions on Gemma-3-1B with $\rho=0.600$.
All 48 correlations in the table are positive, including the descriptive low-sparsity cells.
On Gemma-2-9B, Pile calibration changes every Wanda ceiling at 60\% to 90\% sparsity by at most 0.004 and leaves every rank correlation unchanged.
At 70\% sparsity and above, sentiment loses more readability than harmfulness detection in all four models, although it has the smaller count in three of them, so the association separates stable from vulnerable tasks without ordering the wide-readout tasks among themselves.

\textbf{Precision.}
We compute every correlation from unrounded ceiling losses.
Rounding losses to three decimals would create artificial ties among the small losses of the narrow-readout tasks and change their average ranks.
At 30\% sparsity, even a high correlation orders losses smaller than 0.007, so these cells appear for completeness and do not carry the claim.

\textbf{Scope.}
Post-pruning probes cover up to three class decisions per task, whereas the unpruned task ceiling can cover more, and the losses describe this evaluation convention.
They do not constitute a matched five-seed estimate of information loss for every original class decision.

\textbf{Finding.}
At 60\% sparsity or higher, the pre-pruning count correlates with the loss of linear readability at $\rho\geq0.886$ with $q<0.05$ in 30 of 32 conditions, consistently across two pruning methods and four models, and Pile calibration leaves the Gemma-2-9B order unchanged.

\textbf{Implication.}
Tasks with narrow readouts tolerate aggressive pruning, and tasks with wide readouts deserve the closest monitoring when a model is compressed.
The count separates these two groups but does not order losses within the wide-readout group.

\subsubsection{Pruning Recovery and Support Protection}
\label{app:pruning-recovery}

\textbf{Question.}
How much task performance can a small labeled budget of at most 1,024 labels restore after pruning, and does protecting the weights attached to selected readout coordinates preserve readability better than protecting equally many random coordinates?

\textbf{Recovery procedure.}
Each pruned backbone is adapted independently with LoRA and a fresh classification head at budgets $16,64,256,1024$ and seeds $0,1,2$, using the recovery settings of Table~\ref{tab:adaptation-settings}.
Table~\ref{tab:app-prune-recovery} gives the complete Gemma-3-1B Wanda comparison behind the main-text range, with gaps computed before rounding.
The frozen ceiling is a reference for what a linear probe reads, not an upper bound on larger-budget or full-parameter adaptation.

\begin{table}[!ht]
\centering\AppTableStyle
\caption{\textbf{Pruning ceilings and recovery at 1,024 labels.} Gemma 3 1B with Wanda, median LoRA recovery AUROC across three seeds. The gap is the ceiling minus the recovery AUROC, computed before rounding. These are finite-budget observations, not asymptotic limits.}
\label{tab:app-prune-recovery}
\begin{tabular}{@{}lrrrr@{}}
\toprule
\rowcolor{AppGold} \AppHead{Task} & \AppHead{Sparsity} & \AppHead{Ceiling} & \AppHead{Recovery} & \AppHead{Gap} \\
\midrule
Harmfulness & 50\% & 0.929 & 0.910 & 0.020 \\
Sentiment & 50\% & 0.909 & 0.895 & 0.014 \\
Harmfulness & 70\% & 0.914 & 0.876 & 0.038 \\
Sentiment & 70\% & 0.877 & 0.836 & 0.041 \\
Harmfulness & 80\% & 0.905 & 0.862 & 0.042 \\
Sentiment & 80\% & 0.874 & 0.831 & 0.043 \\
Harmfulness & 90\% & 0.908 & 0.873 & 0.035 \\
Sentiment & 90\% & 0.878 & 0.837 & 0.041 \\
\bottomrule
\end{tabular}
\end{table}

\textbf{Additional recovery arms.}
Tables~\ref{tab:app-repair-wanda} and~\ref{tab:app-repair-magnitude} give every available 1,024-label LoRA endpoint, which covers the four pruned models and four tasks for Wanda and Gemma-3-1B alone for magnitude pruning, and they keep arms with different coverage in separate tables rather than merging them into one denominator.
Table~\ref{tab:app-full-recovery} adds a full-parameter control on Gemma-3-1B with Wanda at the same 1,024-label budget.
Its effects relative to LoRA vary by task and sparsity, so full-parameter adaptation is not uniformly better under the respective default learning rates.
These finite-budget endpoints neither establish convergence nor exhaust alternative recovery procedures.

\begin{table}[!ht]
\centering\AppTableStyle
\caption{\textbf{Wanda recovery endpoints across models.} Median test AUROC at 1,024 labels over the available LoRA recovery seeds. A dash denotes an unavailable endpoint. These arms are separate from the single-model Gemma 3 1B ceiling comparison in Table~\ref{tab:app-prune-recovery}.}
\label{tab:app-repair-wanda}
\begin{tabular}{@{}llrrrr@{}}
\toprule
\rowcolor{AppGold} \AppHead{Model} & \AppHead{Task} & \AppHead{50\%} & \AppHead{70\%} & \AppHead{80\%} & \AppHead{90\%} \\
\midrule
Gemma 3 1B & Banking77 & 0.989 & 0.996 & 0.994 & 0.992 \\
Gemma 3 1B & Jailbreak & 0.996 & 0.995 & 0.996 & 0.993 \\
Gemma 3 1B & Sentiment & 0.895 & 0.836 & 0.831 & 0.837 \\
Gemma 3 1B & Harmfulness & 0.910 & 0.876 & 0.862 & 0.873 \\
Gemma 2 9B & Banking77 & 0.993 & 0.994 & 0.987 & 0.991 \\
Gemma 2 9B & Jailbreak & 0.998 & 0.997 & 0.993 & 0.993 \\
Gemma 2 9B & Sentiment & 0.974 & 0.956 & 0.870 & 0.835 \\
Gemma 2 9B & Harmfulness & 0.958 & 0.928 & 0.877 & 0.868 \\
OLMo 2 7B & Banking77 & 0.989 & 0.992 & 0.984 & 0.990 \\
OLMo 2 7B & Jailbreak & 0.998 & 0.998 & 0.996 & 0.990 \\
OLMo 2 7B & Sentiment & 0.980 & 0.969 & 0.902 & 0.840 \\
OLMo 2 7B & Harmfulness & 0.973 & 0.948 & 0.912 & 0.871 \\
Llama 3.1 8B & Banking77 & 0.992 & 0.993 & 0.990 & 0.986 \\
Llama 3.1 8B & Jailbreak & 0.996 & 0.996 & 0.995 & 0.994 \\
Llama 3.1 8B & Sentiment & 0.974 & 0.943 & 0.887 & 0.830 \\
Llama 3.1 8B & Harmfulness & 0.957 & 0.913 & 0.895 & 0.859 \\
\bottomrule
\end{tabular}%
\par
\end{table}
\begin{table}[!ht]
\centering\AppTableStyle
\caption{\textbf{Magnitude recovery endpoints across models.} Median test AUROC at 1,024 labels over the available LoRA recovery seeds. A dash denotes an unavailable endpoint. These arms are separate from the single-model Gemma 3 1B ceiling comparison in Table~\ref{tab:app-prune-recovery}.}
\label{tab:app-repair-magnitude}
\begin{tabular}{@{}llrrrr@{}}
\toprule
\rowcolor{AppGold} \AppHead{Model} & \AppHead{Task} & \AppHead{50\%} & \AppHead{70\%} & \AppHead{80\%} & \AppHead{90\%} \\
\midrule
Gemma 3 1B & Banking77 & 0.988 & 0.940 & 0.887 & 0.882 \\
Gemma 3 1B & Jailbreak & 0.996 & 0.986 & 0.977 & 0.967 \\
Gemma 3 1B & Sentiment & 0.897 & 0.705 & 0.594 & 0.582 \\
Gemma 3 1B & Harmfulness & 0.905 & 0.775 & 0.762 & 0.755 \\
\bottomrule
\end{tabular}
\par
\end{table}
\begin{table}[!ht]
\centering\AppTableStyle
\caption{\textbf{Adapter versus full-parameter recovery.} Gemma 3 1B with Wanda at 1,024 labels, median AUROC across seeds. Full-parameter and LoRA runs use their respective default learning rates, so the comparison is between procedures and not a capacity-only ablation.}
\label{tab:app-full-recovery}
\begin{tabular}{@{}llrr@{}}
\toprule
\rowcolor{AppGold} \AppHead{Task} & \AppHead{Sparsity} & \AppHead{LoRA} & \AppHead{Full parameters} \\
\midrule
Sentiment & 50\% & 0.895 & 0.912 \\
Sentiment & 70\% & 0.836 & 0.835 \\
Sentiment & 80\% & 0.831 & 0.807 \\
Sentiment & 90\% & 0.837 & 0.838 \\
Harmfulness & 50\% & 0.910 & 0.904 \\
Harmfulness & 70\% & 0.876 & 0.881 \\
Harmfulness & 80\% & 0.862 & 0.849 \\
Harmfulness & 90\% & 0.873 & 0.853 \\
\bottomrule
\end{tabular}
\par
\end{table}

\begin{appdefinition}[support protection]
\label{def:protection}
Let $P\subseteq[d]$ be a set of protected residual coordinates.
For a linear map that writes to the residual stream, namely the attention output projection and the MLP down projection, the rows indexed by $P$ stay dense and every other row follows Wanda.
For a linear map that reads from the residual stream, namely the query, key, value, gate, and up projections, the columns indexed by $P$ receive the score $S_{ij}=+\infty$ in Equation~\ref{eq:wanda-score}, and each row still zeros its $\lfloor s\,d_{\rm in}\rfloor$ lowest-scoring weights.
The protected arms differ only in $P$.
The jailbreak-support arm protects the selected jailbreak coordinates, the single-layer arm protects the selected harmfulness and sentiment coordinates of one layer, the selected-union arm protects the harmfulness and sentiment selections over four layers, and the random control protects as many random coordinates per task.
\end{appdefinition}

\textbf{Remark (what protection guarantees).}
The reader-column rule removes protected columns from the pruning set only while each row has at least $\lfloor s\,d_{\rm in}\rfloor$ unprotected columns.
When the quota exceeds that number, the lowest-score selection must include some infinite-score columns, so the rule is a protection heuristic rather than a hard constraint at every requested sparsity.
The writer-row rule leaves $|P|$ rows dense in each writer map, which lowers the realized sparsity
\begin{equation}
    s_{\mathrm{act}}=\frac{\#\{\text{zero weights in all pruned maps}\}}{\#\{\text{weights in all pruned maps}\}}
    \label{eq:actual-sparsity}
\end{equation}
below the requested value, and we report $s_{\mathrm{act}}$ separately in Table~\ref{tab:app-protection} wherever the run log recorded it.
Protecting weights also does not protect the activations of the protected coordinates, since they receive contributions from every unprotected weight upstream.

\begin{table}[!ht]
\centering\AppTableStyle
\caption{\textbf{Complete support-protection comparison.} Gemma 2 9B with Wanda. Protected dimensions count the distinct residual coordinates in $P$, actual sparsity is $s_{\mathrm{act}}$ of Equation~\ref{eq:actual-sparsity}, and a dash marks a value that was not recorded. Task columns give post-pruning dense-probe ceilings, maximized over the evaluation layers. All three requested sparsities and all five arms are shown.}
\label{tab:app-protection}
\begin{tabular}{@{}llrrrr@{}}
\toprule
\rowcolor{AppGold} \AppHead{Requested} & \AppHead{Arm} & \AppHead{Protected dims} & \AppHead{Actual} & \AppHead{Sentiment} & \AppHead{Harmfulness} \\
\midrule
70\% & Ordinary & 0 & --- & 0.947 & 0.948 \\
70\% & Jailbreak support & 48 & --- & 0.944 & 0.947 \\
70\% & Single-layer support & 385 & 0.675 & 0.931 & 0.940 \\
70\% & Selected union & 969 & --- & 0.891 & 0.925 \\
70\% & Random control & 975 & 0.636 & 0.888 & 0.925 \\
80\% & Ordinary & 0 & --- & 0.904 & 0.932 \\
80\% & Jailbreak support & 48 & --- & 0.900 & 0.930 \\
80\% & Single-layer support & 385 & 0.771 & 0.899 & 0.923 \\
80\% & Selected union & 969 & --- & 0.868 & 0.902 \\
80\% & Random control & 975 & 0.727 & 0.866 & 0.906 \\
90\% & Ordinary & 0 & --- & 0.895 & 0.924 \\
90\% & Jailbreak support & 48 & --- & 0.894 & 0.923 \\
90\% & Single-layer support & 385 & 0.868 & 0.855 & 0.901 \\
90\% & Selected union & 969 & --- & 0.848 & 0.898 \\
90\% & Random control & 975 & 0.818 & 0.858 & 0.898 \\
\bottomrule
\end{tabular}
\end{table}

\textbf{Design rationale.}
We fixed the success criteria before running the protection arms.
The random control protects the same per-task numbers of coordinates as the selected union, so any advantage of the selected arm would reflect which coordinates it protects and not how many.
Reporting the realized sparsity separates a genuine protection effect from the trivial effect of pruning fewer weights.

\textbf{Null.}
If selected readout coordinates marked the weights that task readability depends on, the selected-union arm would retain clearly higher ceilings than the random control of nearly equal size.

\textbf{Result.}
The selected-union minus random-control differences are $+0.003$, $+0.002$, and $-0.010$ for sentiment and $0$, $-0.004$, and $0$ for harmfulness at the three requested sparsities, so selected protection does not outperform random protection.
Both union arms also retain lower ceilings than ordinary Wanda at the same requested sparsity.
At matched actual sparsity, the comparison with ordinary pruning interpolates the measured ordinary degradation curve, and it is not an additional directly measured run at each protected arm's exact sparsity.

\textbf{Scope.}
The selected union contains 969 distinct coordinates and the random control 975, since the per-task sets have the same sizes but overlap differently, so the two arms are nearly but not exactly equal in size.
The retained run log lacks the realized sparsity of the selected-union arm, and an earlier summary of the same runs reported values within 0.001 of the random control, so we compare the two union arms at their common requested sparsity.
The selection budgets were 174 coordinates for harmfulness, 227 for sentiment, and 16 for jailbreak detection, which differ from the final Gemma-2-9B counts of 341, 72, and 8.
The protection arm therefore tests this selection rule, not the preservation of the final protocol's supports.
The result bounds this protection rule and leaves other support-aware compression procedures untested.

\textbf{Finding.}
Protecting the weights attached to selected readout coordinates retains ceilings within 0.010 of protecting equally many random coordinates, and both protected arms retain less readability than ordinary Wanda at the same requested sparsity.

\textbf{Implication.}
Coordinates that suffice for a linear readout do not identify the parameters that sustain it, since every protected coordinate still receives contributions from unprotected weights upstream.
Readout measurements diagnose which tasks pruning endangers, and translating them into a protection rule requires a mechanism that the tested rule does not provide.

\subsubsection{Reproducibility of the Application Results}
\label{app:reproduction}

Every result in this appendix is determined by four ingredients, namely the task pools, the adaptation or modification procedure, the random seeds, and the estimator that turns runs into the reported number.
Table~\ref{tab:code-evidence} lists these four ingredients for each result of Section~\ref{sec:applications}, and the three paragraphs below state the shared conventions that the table does not show.

\begin{table}[!ht]
\centering\AppTableStyle
\caption{\textbf{Ingredients needed to reproduce each application result.} Task pools are those of Table~\ref{tab:tasks}, reference counts and ceilings come from Algorithm~\ref{alg:aggregate} on Gemma-2-9B unless a subsection states otherwise, and every run evaluates on the fixed test split of its task.}
\label{tab:code-evidence}
\begin{tabular}{@{}p{0.20\linewidth}p{0.30\linewidth}p{0.22\linewidth}p{0.20\linewidth}@{}}
\toprule
\rowcolor{AppGold} \AppHead{Result} & \AppHead{Procedure} & \AppHead{Randomness} & \AppHead{Estimator} \\
\midrule
Student learning curves & LoRA and head, Table~\ref{tab:adaptation-settings}, Eqs.~\ref{eq:stratified-draw} and~\ref{eq:epochs} & Seeds 0, 1, 2 per budget & Eq.~\ref{eq:learning-curve} \\
Label requirements & Definition~\ref{def:nstar} on realized sizes & Inherited from the curves & Eqs.~\ref{eq:nstar-threshold} and~\ref{eq:nstar-interpolation} \\
Shared exponent & Least squares with student intercepts & None & Eq.~\ref{eq:label-budget}, Proposition~\ref{prop:pooled-slope} \\
Exponent interval & Task-block bootstrap & 10,000 draws, seed 0 & Eq.~\ref{eq:task-bootstrap} \\
Held-out error & Leave one task out & None & Eq.~\ref{eq:heldout-error} \\
Partial correlations & Rank residuals at 256 labels & 10,000 permutations, seed 0 & Eq.~\ref{eq:partial-rank} \\
Allocation replay & Proportional policies & None beyond the curves & Eqs.~\ref{eq:allocation} and~\ref{eq:allocation-weights} \\
Continual learning & Full-parameter chains, frozen probes & Chain seeds 0, 1, 2 & Eq.~\ref{eq:forgetting-channels} \\
Recovery & LoRA and new head from the final backbone & Seeds 0, 1, 2 per backbone & Definition~\ref{def:nstar} \\
Pruning losses & Eqs.~\ref{eq:magnitude-prune} and~\ref{eq:wanda-score} & Calibration draw, seed 0 & Eq.~\ref{eq:pruning-loss} \\
Pruning order & Exact permutation test with adjustment & None & Eqs.~\ref{eq:pruning-permutation} and~\ref{eq:bh-adjustment} \\
Support protection & Definition~\ref{def:protection} & Random coordinate draw & Eq.~\ref{eq:actual-sparsity} \\
\bottomrule
\end{tabular}
\end{table}

\textbf{Deterministic components.}
The least-squares fits, the leave-one-task-out predictions, the allocation replay, the exact permutation tests, and the Benjamini–Hochberg adjustment involve no randomness beyond their inputs, so they return identical values from identical learning curves and probe results.
The task-block bootstrap and the partial-correlation permutations fix their seeds at 0, which makes their intervals and $p$ values exactly repeatable from the same panel.

\textbf{Stochastic components.}
Adaptation runs draw their training subsets by Equation~\ref{eq:stratified-draw} with the run seed and initialize heads and adapters from the same seed.
GPU training in bfloat16 is not bitwise deterministic, so repeated runs can differ in the last digits of AUROC even under identical seeds, and the three seeds per budget quantify the resulting variation.
The continual-learning chains and the pruning calibration draws are seeded in the same way.

\textbf{Reported precision.}
Tables round values for display, whereas every threshold comparison, correlation, and gap uses unrounded values.
Consequently, a displayed gap can differ in its last digit from the difference of the two displayed values in the same row.

\subsubsection{Summary of Findings, Recommendations, and Open Questions}
\label{app:applications-summary}

\textbf{Findings.}
The experiments of this appendix establish three predictive results and two boundaries.
First, the count of a frozen reference model anticipates label requirements in other students, with $N^*\propto(k^*)^{0.65}$ and a held-out factor of 5.3.
Second, sequential training can disable saved heads while linear readability persists, and recovery then needs fewer labels than initial learning.
Third, wider pre-pruning readouts predict larger losses of readability, in 30 of 32 adjusted conditions.
The two boundaries concern intervention, since neither proportional label allocation by counts nor protection of readout-coordinate weights outperformed its baseline.

\textbf{Recommendations.}
The results support three practices under the tested procedures.
Before annotation, measure $k^*$ on a reference model to rank tasks by expected label needs, and treat individual forecasts as accurate to within a factor of about five.
After sequential updates, re-probe the frozen backbone before retraining from scratch, since a failing head can coexist with intact linear readability.
Before compression, flag tasks with wide readouts for evaluation, since they carry the largest losses.

\textbf{Open questions.}
The results leave three questions open.
The source of the relation between reference counts and student requirements remains unidentified, since the teacher controls change labels and not sparsity alone.
An allocation rule that uses counts to choose where marginal labels help a joint objective most, rather than proportional weights, remains untested.
A compression rule that preserves the activations of selected coordinates, rather than their attached weights, would test whether readout supports can guide preservation under pruning at all.